\documentclass[10pt,twocolumn,letterpaper]{article}
\pdftrailerid{}
\usepackage{cvpr}      

\usepackage{multirow}
\usepackage{bm}
\usepackage{array}
\usepackage{algorithm}
\usepackage{algpseudocode}
\usepackage{capt-of}

\newcommand{\SeeErr}{\epsilon_{\mathrm{cov}}}
\newcommand{\SayErr}{\epsilon_{\mathrm{sel}}}
\newcommand{\Acc}{\mathrm{Acc}}
\newcommand{\one}{\mathbf{1}}
\newcommand{\sd}[1]{{$\pm$#1}}
\newcommand{\ci}[2]{\,{\scriptsize[#1,\,#2]}}

\definecolor{cvprblue}{rgb}{0.21,0.49,0.74}
\usepackage[pagebackref,breaklinks,colorlinks,allcolors=cvprblue]{hyperref}
\def\confName{CVPR}
\def\confYear{2027}

\title{Where Does Proposal Selection Fail? Diagnosing Frozen Vision-Language Queries over Detector Proposals}

\author{Bo Ma\\ Auckland University of Technology \and Weiqi Yan\\ Auckland University of Technology \and Jinsong Wu\\ Guilin University of Electronic Technology}

\begin{document}
\maketitle

\begin{abstract}
In modular grounding a frozen detector proposes regions and a
vision-language model (VLM) selects one, at lower cost than letting the VLM
decode a box, but it is unclear where such systems fail. We hold perception
fixed---every trained selector chooses among the same frozen proposals, which
contain the referent for at least $99.8\%$ of RefCOCO/+/g queries---and vary
the selection interface one factor at a time for two VLMs of one family
(Qwen2.5-VL-7B, Qwen3-VL-8B). On RefCOCOg, what the selector reads matters more
than how it is built: the VLM's image-conditioned query encodes the referent's location and beats a text-only query by $0.15$--$0.18$ with every
head tested (by less for InternVL3-8B, whose vision encoder
differs; one head tested). The remaining errors are rankings among a few top candidates; a cross-attention
selector supervised with every valid box reaches the VLM's coordinate decoding on RefCOCOg, and with Qwen2.5-VL on
RefCOCO/+ when also trained on their data. Beyond that distribution it does not: on Ref-L4's Objects365 images, mostly of categories outside
training, every trained head falls to $0.19$--$0.28$ while
Qwen decoding keeps about $0.8$ (InternVL3's drops too), neither a wider detector prompt nor a
class-agnostic proposer closes the gap, and on public reasoning benchmarks Ref-Adv-s and RefBench-PRO full-budget heads trail decoding by $0.19$--$0.37$ although
proposals cover most referents. A trained verifier reduces to an existence
probe on the VLM's query state and does not beat the VLM's yes/no answer on public no-target benchmarks.
\end{abstract}

\section{Introduction}
\label{sec:intro}

\begin{figure*}[t]
\centering
\begin{minipage}[b]{0.385\textwidth}
\centering
\includegraphics[width=\linewidth]{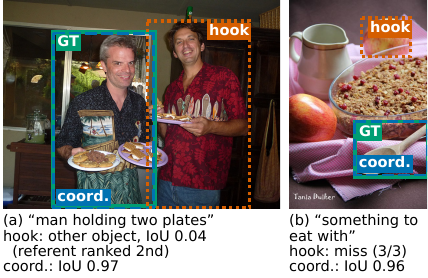}
\end{minipage}\hfill
\begin{minipage}[b]{0.595\textwidth}
\centering
\includegraphics[width=\linewidth]{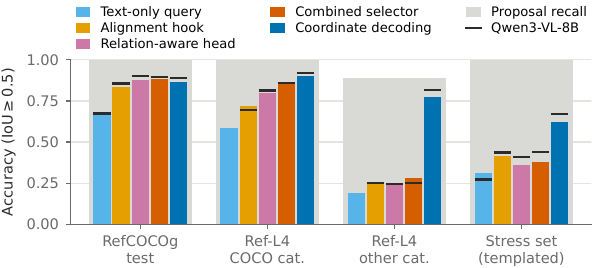}
\end{minipage}
\caption{\textbf{Where selection fails.} Left: the alignment hook's choice
among the frozen detector's proposals (orange, dotted; one seed, $24$k rows) against the
VLM's own coordinate decoding (blue, dashed) and the ground truth (green).
(a)~A ranking error: the referent is ranked second. (b)~A reasoning request,
missed in all three seeds. Right: accuracy of heads trained on all of RefCOCOg
and of coordinate decoding (bars: Qwen2.5-VL-7B; ticks: Qwen3-VL-8B; gray:
proposal recall). The selectors reach decoding on RefCOCOg but not on Ref-L4's Objects365
images or on templated reasoning requests
(values in \cref{tab:full}; Qwen3-VL's text-only query is not split on Ref-L4).}
\label{fig:teaser}
\end{figure*}
Grounding a free-form request---``the person not wearing a helmet'',
``something to cut paper with''---requires two competencies: localizing
candidate objects and selecting the one the request refers to. Two families
of vision-language models (VLMs) combine them differently.
Coordinate-as-text models decode the box inside the language head
\cite{kosmos2,shikra,griffon,qwen25vl,rexomni}; modular designs let a detector
propose regions and let the language model select among them, by retrieval
with pooled token states \cite{groundhog}, by an index token
\cite{chatrex,rexseek,groma}, by attending from proposals to the language model's
hidden states \cite{vgent}, or by reasoning over candidates in text
\cite{rexthinker}. These systems report localization, selection and abstention as one number, so a wrong box cannot be traced to the
detector, the selection step or the decision to answer.

The modular route has practical appeal. One detector pass serves every request
on an image; one forward pass of the VLM replaces autoregressive decoding, so
with our components a proposal pipeline costs $0.41$ of the latency of
coordinate decoding per request ($0.23$ when the detector pass is shared by
the $3.8$ requests per image of RefCOCOg val); boxes
inherit the detector's localization; and a selector can expose a decision to
abstain. What builders of such pipelines need to know is whether selection is
as accurate as the VLM's own grounding, and if not, which part to fix.

We isolate the selection step. A frozen, language-aligned detector
\cite{groundingdino} returns $M{=}300$ proposals with region features; a frozen
VLM (Qwen2.5-VL \cite{qwen25vl}, replicated with Qwen3-VL \cite{qwen3vl}) reads
the image and the request and emits a pooled query embedding; a small trained
\emph{alignment hook} (a region--query matching head) selects a proposal; a
verifier decides whether to answer. Because every trained method selects from
the same proposals---which contain the referent for $99.8$--$100\%$ of
RefCOCO/+/g queries---differences measure the selection interface, not the
detector. Varying one factor at a time yields three findings
(\cref{fig:teaser}), whose core comparisons are repeated for both VLMs on heads
trained with the full RefCOCOg training set (\cref{tab:full}):
\begin{itemize}\setlength\itemsep{1pt}
\item \emph{The query decides, because it localizes.} With any of three heads
fixed, a text-only query loses $0.15$--$0.18$ on RefCOCOg (at $24$k rows the hook's
loss is $0.21$ with Qwen2.5-VL and $0.13$ with InternVL3 \cite{internvl3}, whose
vision encoder differs) against the query the
VLM computes with the image, which linearly encodes much of the referent's location; with that query fixed, pairwise heads of
different form and capacity converge and cross-attention scoring adds a few points (\cref{sec:limits}).
\item \emph{On standard expressions, errors are ranking errors.} The correct
proposal is usually in the top five (\cref{fig:teaser}a); supervision with
every valid box, reading several VLM layers and cross-attention scoring bring a
selector level with coordinate decoding on RefCOCOg. Prompting the VLM to pick
among the top candidates does not help; fine-tuning it to pick, as in
candidate-region selection \cite{crs2025}, does, though on the hook's top five it
ends level with the same VLM fine-tuned to decode boxes on the same rows.
\item \emph{Beyond them, a category boundary.} On Ref-L4 every trained head
collapses on referents outside the COCO categories it was trained on, while
Qwen's coordinate decoding does not (InternVL3's falls too); a detector prompt naming every category or a
class-agnostic proposer leaves it far below decoding, while
training on such referents, not on their caption style, lifts it (hook: $0.24$
to $0.47$ on Objects365 images; \cref{sec:reasoning}). On public reasoning benchmarks
(Ref-Adv-s, RefBench-PRO) every full-budget head trails decoding by $0.19$--$0.37$.
\end{itemize}
For abstention, a verifier on frozen features is an existence probe on the VLM's query state; it adds referent-level abstention to coordinate-snap, but the VLM's yes-minus-no score ranks swapped negatives better and its yes/no answer is better on every public no-target
benchmark
(\cref{sec:verifier}). We contribute a matched-perception protocol for modular
grounding (one metric for every method, with abstention reported as answer rate and
selective accuracy and errors split into perception and selection as in
\cite{hkvlm2026}, \cref{sec:setting}) and the controlled diagnosis above. We claim no state-of-the-art localization: the VLM's untrained
decoding is matched only on RefCOCOg, and on RefCOCO/+ with their data.

\section{Related Work}
\label{sec:related}

\paragraph{Two-stage referring-expression comprehension.}
Selecting among expression-agnostic proposals is the classic two-stage design
\cite{mattnet,reclip,refdetector}, which already compared ground-truth with
detected regions; its ceiling motivated expression-aware proposals \cite{refnms} and
one-stage grounding fusing the request early
\cite{onestagevg,mdetr,transvg,glip}; referring segmentation has been split into perception and cognition \cite{deris2025}. We revisit selection
with a frozen VLM as the query source.

\paragraph{Coordinate-as-text grounding.}
Kosmos-2 \cite{kosmos2}, Shikra \cite{shikra} and Florence-2 \cite{florence2}
emit boxes as location tokens or plain numbers, Ferret \cite{ferret} mixes points with continuous region
features, Griffon \cite{griffon,griffong} spells out dense object locations,
the Qwen-VL family \cite{qwenvl,qwen2vl,qwen25vl,qwen3vl} decodes coordinates
as text (normalized, except absolute pixels in Qwen2.5-VL), and Rex-Omni \cite{rexomni} predicts quantized coordinate tokens.
Frozen VLMs also localize without decoding, through a few attention heads
\cite{lochead}, their attention maps \cite{flmm,mllmsknow} or a learned
positional prompt \cite{pin2024}, and intermediate layers carry region
information \cite{interlayer2024,vlmlocmech}. Contemporaneous work regresses the box from one token state of a fine-tuned VLM \cite{jag},
drafts it with such a head and verifies it with coordinate tokens \cite{locus} (both run in \cref{tab:app-e54}), or fuses several frozen
VLMs' boxes by agreement \cite{forum}. The
frozen VLM's own coordinate decoding is our strongest matched-perception reference.

\paragraph{Proposal-based grounding with multimodal LLMs.}
GROUNDHOG \cite{groundhog} retrieves masks with the averaged hidden state of an
adapted LLM's grounding tokens. ChatRex \cite{chatrex}, RexSeek \cite{rexseek},
Groma \cite{groma} and VLM-FO1 \cite{vlmfo1} insert proposal or region tokens
into a trained LLM that answers with their indices; LLM-wrapper
\cite{llmwrapper} fine-tunes an LLM to pick among candidate boxes listed as
text, candidate-region selection \cite{crs2025} instruction-tunes an MLLM to
answer one of a detector's top candidates or ``none'', and ROD-MLLM
\cite{rodmllm} also rejects absent targets. VGent \cite{vgent} freezes a
reinforcement-learning-tuned Qwen2.5-VL-7B and lets boxes from several
proposers cross-attend to its hidden states of every layer, trained on several
grounding corpora; our combined selector (\cref{sec:method}) is a small instance of that design
trained on RefCOCO-family data alone. Rex-Thinker \cite{rexthinker} reasons in text over
each candidate and may reject all of them. Training-free variants overlay
numbered marks \cite{setofmark,llmoptic}, which can hurt open VLMs \cite{crg}
unless they are trained to read them \cite{somllava}, ask one
true/false question per proposal \cite{tfverify}, or chain agents over
candidates \cite{groundingagent}. PropVG \cite{propvg} and RefFormer
\cite{refformer} avoid an external detector. Published accuracies are not matched-perception results. The alignment hook, the
verifier and the coverage--selection decomposition were introduced in HKVLM
\cite{hkvlm2026}; the matched-perception diagnosis, the category-boundary
analysis, the full-budget and second-VLM replications, the probes on per-token states and
in-VLM selection are new here. The hook is deliberately
the simplest member of the family, so that what these systems add---token-level
access, relations among proposals, selection inside the language model---can be varied singly on identical proposals.

\paragraph{Absent targets and abstention.}
gRefCOCO \cite{grefcoco,grec,grex} adds no-target and multi-target expressions
to the RefCOCO family, and GREC methods predict the number of targets
\cite{hiea2g}; Ref-ZOM \cite{refzom} does the same for referring segmentation;
R-RefCOCO \cite{rrefcoco} builds negatives that include sentences of other
referents, as our swaps do; D$^3$ \cite{d3} annotates descriptions of absence
and finds that REC models rarely reject negatives, and OmniLabel
\cite{omnilabel} adds negative descriptions to language-based detection; and
GroundingME \cite{groundingme} reports that most MLLMs score zero on its
rejection dimension and releases attribute-edited RefCOCOg negatives. Remedies
train the grounder to refuse \cite{counterfactualrec,rcgrpo}, add a
discrimination module \cite{propvg} or verify program steps \cite{viro}; our
verifier is trained on frozen features and acts as a hidden-state probe
\cite{halp}. Contemporaneously, absence is scored from frozen CLIP features
\cite{prove} and black-box VLM groundings are calibrated on held-out data
\cite{surrogatevg}. Recent benchmarks also probe compositional, knowledge-dependent and
reasoning-heavy references with rejection
\cite{finecopsref,refadv,refbenchpro,knowdrrec,openref}. We report the answer
rate, selective accuracy and the area under the risk--coverage curve \cite{geifman2017selective,reliablevqa}; object
hallucination is measured with caption metrics \cite{rohrbach2018object} and
existence probes \cite{pope} and reduced by contrastive decoding \cite{vcd},
and grounding objectives alone do not remove it \cite{objgndhalluc}. Ref-L4
\cite{refl4} pairs long, human-reviewed expressions over $365$ categories with
COCO and Objects365 images and already reports lower accuracy on its
Objects365 categories; open-vocabulary detectors \cite{glip,owlvit,yoloworld}
and grounding heads \cite{ovvg} drop on categories outside their training
vocabulary. We separate prompt and training vocabulary as causes.

\section{Setting and Metrics}
\label{sec:setting}
Given an image $I$ and a request $q$, a system returns a box for the referent
when one exists and abstains otherwise. A perception module supplies proposals
$\mathcal{R}=\{(b_j,e_j)\}_{j=1}^{M}$ with boxes $b_j$ and features $e_j$;
\emph{selection} maps $q$ to one element of $\mathcal{R}$.

\paragraph{One grounding metric.}
\emph{Grounding accuracy} is the fraction of present queries whose returned box
has $\mathrm{IoU}\ge0.5$ with the ground truth, for every method. For methods
that rank the proposals, the accuracy of the top-ranked proposal with abstention
disabled is \emph{selection accuracy}. An abstaining system additionally has an
\emph{answer rate} (fraction of queries answered) and a \emph{selective
accuracy} (accuracy among answered queries). On sets that mix present and
absent referents, \emph{joint accuracy} counts a present query as correct if a
correct box is returned and an absent query if the system abstains; its
\emph{AURC} is the area under the selective risk--coverage curve.
\emph{Recall@$k$} is the share of present queries with a proposal at
$\mathrm{IoU}\ge0.5$ among the top $k$.

\paragraph{Perception versus selection.}
Let $z_n=\one[\max_j\mathrm{IoU}(b_{n,j},b_n^\star)\ge0.5]$ indicate that some
proposal covers referent $n$. Because the returned box is a proposal, a hit
implies $z_n=1$, and selection accuracy factorizes into proposal recall and
conditional selection,
\begin{equation}
\Acc=(1-\SeeErr)(1-\SayErr),\quad
\SeeErr=1-\tfrac{1}{N}\textstyle\sum_n z_n,
\label{eq:decomp}
\end{equation}
where $N$ is the number of present queries and $\SayErr$ the failure rate
among covered referents; the identity separates the detector's share of the error from the selector's.

\paragraph{Terms.}
\emph{Alignment hook}: the cosine matching head of \cref{sec:method}.
\emph{Learned bridge}: the hook trained without the verifier, used to vary the
query source (CLIP, the detector's text tower, the VLM with or without the image).
\emph{Index scorer}: a per-proposal scorer over the query and proposal features
with a learned index embedding, modeled on the index outputs of
\cite{chatrex,groma} but a small head over the pooled query rather than a
trained language model; \emph{audited}: its best variant (no index embedding,
masked uncovered positives, normalized inputs, learning rate swept on RefCOCOg
val). \emph{Relation-aware head}, \emph{token-state probe} and \emph{combined
selector}: the heads of \cref{sec:method}. \emph{Detector-native}: the detector
prompted with the expression, highest-scoring box. \emph{Coordinate decoding}:
the frozen VLM prompted for a box in its documented format (absolute pixels for
Qwen2.5-VL, a $0$--$1000$ grid for Qwen3-VL), used as returned; \emph{coordinate-snap}: that box snapped to the proposal of
maximal IoU. \emph{Swapped} absent referent: an expression paired with another
image. \emph{Stress set}: the templated reasoning requests of
\cref{sec:protocol}. \emph{Late fusion}: the request and the region features meet only in the
selection head, through the VLM's query.
Intervals: $95\%$ bootstrap over queries; \emph{two-level}: seeds, then
queries; \emph{Holm}: multiplicity-adjusted $p$-values.

\section{Selection Interface and Probes}
\label{sec:method}

\begin{figure}[t]
\centering
\includegraphics[width=\columnwidth]{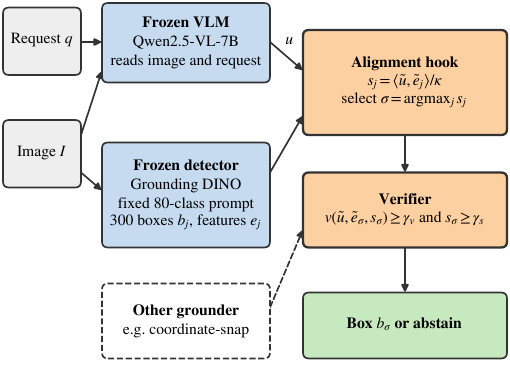}
\caption{\textbf{Matched-perception selection.} A frozen detector, prompted
with one fixed text independent of the request, supplies proposals and region
features; the frozen VLM reads image and request and returns the query $u$.
Only the orange components are trained: the hook selects a proposal and the
verifier decides whether to answer. Dashed: the verifier scores, unchanged,
the proposal returned by another grounder, gating on $v\ge\gamma_v$ alone
(\cref{sec:verifier}).}
\label{fig:arch}
\end{figure}

\paragraph{Frozen components.}
The detector is Grounding DINO \cite{groundingdino} (Swin-B), prompted with one
fixed text listing the $80$ COCO category names, independent of the request; it
returns the $M{=}300$ highest-scoring proposals with $256$-dimensional
vision-language-fused region features. The VLM is Qwen2.5-VL-7B \cite{qwen25vl};
it receives the image and the bare request, with no instruction, and the query $u\in\mathbb{R}^{3584}$ is its final-layer hidden
state at the position that would begin its answer (\cref{fig:arch}). We call
this single vector the pooled query, unlike the per-token states of
the token-state probe.

\paragraph{Alignment hook.}
Two small MLPs map regions and the query into a shared space; the affinity is a
temperature-scaled cosine, $s_j=\langle\tilde u,\tilde e_j\rangle/\kappa$, with
$\ell_2$-normalized projections and a learned temperature $\kappa$. The hook is
trained by contrastive retrieval over the proposals of the image,
$\mathcal{L}_{\mathrm{ret}}=-\one[P\neq\varnothing]\log\sum_{j\in P}\mathrm{softmax}_j(s)$,
where the positive set $P$ is the proposal of maximal IoU with the ground
truth; absent queries and the few present queries that no proposal covers have
$P=\varnothing$ and contribute only to the verifier loss. At inference the hook
selects $\arg\max_j s_j$. With normalized projections any proposal can be ranked
first by some query, as its feature lies outside the convex hull of the others'; the interface itself
rules out no selection.

\paragraph{Verifier.}
A head $v(\tilde u,\tilde e_\sigma,s_\sigma)$ reads the query, the selected
region and their affinity and is trained with binary cross-entropy on
$y=\one[\text{referent present}]$, with swapped absent queries as negatives.
The hook emits a box when $v\ge\gamma_v$ and $s_\sigma\ge\gamma_s$ (its
\emph{veto}); $(\gamma_v,\gamma_s)=(0.3,24)$ is chosen once on RefCOCOg val
($\gamma_s\kappa\approx0.4$ in cosine units at the learned $\kappa\approx0.017$).
Because the verifier reads only a query, a region and their affinity, it can score any grounder's box once snapped to a proposal, gating on
$v\ge\gamma_v$ alone (\cref{sec:verifier}).

\paragraph{Diagnostic probes.}
Each probe changes one factor with perception, training data ($24$k rows,
\cref{sec:protocol}) and evaluation fixed, with three seeds. \emph{Query
source}: the learned bridge on CLIP \cite{clip} text, the detector's text tower
(native and through a trained projection), the VLM without the image, or the
image-conditioned VLM query. \emph{Head form and capacity}: shared dimension
$128$--$512$, width $256$--$1024$, depth $0$--$3$; the index scorer with and
without its index embedding, with a bilinear scorer and with a learning-rate
sweep. \emph{Supervision and readout}: every proposal with
$\mathrm{IoU}\ge0.5$ as a positive; the query read as the mean over expression
tokens or from three depths of the VLM. \emph{Set-level scoring} at a matched
$4.3$M parameters: a relation-aware head in which query-modulated proposal
tokens with box geometry attend to each other under a learned pairwise
geometry bias, optionally with a loss that ranks proposals by IoU; a
cross-attention head in which proposal tokens attend to the pooled query; and
the \emph{token-state probe}, in which they attend to the final-layer states of
the request tokens. The \emph{combined selector} joins token states, two intermediate depths, relations among proposals and the all-positive
IoU-rank loss.

\section{Experimental Protocol}
\label{sec:protocol}
\paragraph{Data.}
We evaluate on RefCOCO, RefCOCO+ \cite{refcoco} and RefCOCOg
\cite{refcocog} (umd split \cite{refcocogumd}) over COCO \cite{coco};
training uses RefCOCOg only, so RefCOCO/+ are held-out expression sets, though
not held-out images (see below). Probes train on the first $24{,}000$ rows of
RefCOCOg train ($16{,}724$ present expressions, $7{,}276$ swapped absent ones);
the core comparisons use the full set ($80{,}512$ and $34{,}505$ rows on
$21{,}899$ images; \cref{tab:full}). Ref-L4 \cite{refl4} ($13{,}420$ validation
queries) adds long expressions and non-COCO categories. A $231$-query stress
set of templated reasoning requests built from COCO val2017 annotations tests
negation, affordance and world knowledge; it is small and templated ($25$
templates over $218$ images, none in any RefCOCO-family split, reviewed by
annotators blind to model outputs), so its intervals are clustered by template (\cref{sec:app-stress}).
Two public reasoning benchmarks complement it: Ref-Adv-s, the public split of
Ref-Adv \cite{refadv} ($1{,}142$ queries, same-category distractors, $40\%$ negated, no image in
RefCOCO-family training data) and RefBench-PRO \cite{refbenchpro} ($5{,}000$ queries in
five perception and reasoning tasks plus $1{,}000$ without a referent). Abstention is tested on POPE \cite{pope}, RefBench-PRO, gRefCOCO \cite{grefcoco} testA and
testB, GroundingME \cite{groundingme} and on RefCOCOg val and test and the stress set
augmented with $30\%$ swapped absent referents, re-encoded with the
new image. Some supplementary analyses use an earlier feature build.

\paragraph{Caveats on the frozen components.}
The Swin-B detector was trained with RefCOCO-family data and the VLM's
grounding corpora are undisclosed, so neither built-in interface is zero-shot on
these splits: low-rank fine-tuning of the VLM in its native box format on
$200$ to $5{,}000$ RefCOCOg expressions moves its RefCOCOg val accuracy only
from $0.865$ to $0.866$--$0.871$ (\cref{sec:app-additional-discussion}). With
proposals from a detector trained without RefCOCO data, recall drops by at most $0.006$ on every whole split but the stress set,
the relation-aware head stays within $0.010$ on the standard splits (the hook
within $0.04$), and the Ref-L4 collapse persists
(\cref{tab:app-second-detector}). Every trained method shares both frozen
components, so comparisons among them stay matched. RefCOCO/+ val images
outside RefCOCOg train are harder for every system: on those of RefCOCO val
the full-budget combined selector reaches $0.861$ against $0.875$ for
coordinate decoding ($-0.014$ [$-0.024$, $-0.004$]; RefCOCO+ $0.784$ against
$0.824$); the heads gain at most $0.013$ more on newly seen training images than
on others, so this reflects image selection, not memorization
(\cref{tab:app-full-leak,tab:app-allsplits}); trained on all three
RefCOCO-family sets it reaches $0.893$ and $0.820$ there. Trained results: three-seed means.

\section{Results}
\label{sec:results}

\subsection{Where selection by alignment stands}
\label{sec:standing}
\Cref{tab:app-claims} indexes the evidence (intervals, paired tests,
per-split and per-bucket tables) behind each claim below.

\begin{table*}[t]
\centering
\footnotesize
\setlength{\tabcolsep}{4.2pt}
\caption{\textbf{Full RefCOCOg training set, two VLMs}: accuracy at
$\mathrm{IoU}\ge0.5$ on present queries, no abstention; trained rows select one
of the same $300$ proposals (mean over three seeds; sample std $\le0.016$, on
Stress $\le0.048$; RefCOCOg val also served for tuning). Heads are trained on all of RefCOCOg train ($115$k rows)
with each VLM's own query; text-only query: the learned bridge on the request
read without the image; pooled query: cross-attention of proposal tokens to
the pooled query. Ref-L4 is also split by image source: COCO images, whose referents are all
COCO categories ($63\%$ of queries), and Objects365 images (other; a third of
their referents map to a COCO class). All three train sets: trained on
RefCOCO, RefCOCO+ and RefCOCOg train, evaluation images removed. RefCOCO/+
cells of the pooled query, token-state probe and combined selector: same-seed
retraining (other cells reproduced within $0.004$ outside Stress). Ref-Adv-s and RB-PRO
(RefBench-PRO, five grounding tasks): public reasoning benchmarks, Qwen2.5-VL
heads only. --: not evaluated. Std,
$\mathrm{IoU}\ge0.75$ and recall@5 where evaluated: \cref{tab:app-full-acc,tab:app-full-iou75}.}
\label{tab:full}
\begin{tabular}{@{}lcccccccccc@{}}
\toprule
& RefCOCO & RefCOCO+ & \multicolumn{2}{c}{RefCOCOg} & \multicolumn{3}{c}{Ref-L4 val} & & & \\
\cmidrule(lr){4-5}\cmidrule(lr){6-8}
& val & val & val & test & all & COCO cat. & other & Stress & Ref-Adv-s & RB-PRO\\
\midrule
Proposal recall & 0.999 & 0.999 & 1.000 & 0.998 & 0.958 & 0.999 & 0.886 & 1.000 & 0.907 & 0.884\\
Detector-native & 0.847 & 0.737 & 0.783 & 0.781 & 0.634 & 0.649 & 0.609 & 0.255 & 0.271 & 0.524\\
\midrule
\multicolumn{11}{@{}l}{\emph{Qwen2.5-VL-7B}}\\
Text-only query & 0.683 & 0.517 & 0.672 & 0.676 & 0.442 & 0.587 & 0.194 & 0.312 & -- & --\\
Alignment hook & 0.834 & 0.760 & 0.843 & 0.837 & 0.549 & 0.718 & 0.259 & 0.413 & 0.308 & 0.297\\
Relation-aware head & 0.872 & 0.802 & 0.877 & 0.878 & 0.595 & 0.798 & 0.247 & 0.358 & 0.317 & 0.290\\
Pooled query, cross-attention & 0.876 & 0.808 & 0.877 & 0.881 & 0.606 & 0.807 & 0.260 & 0.388 & -- & --\\
Token-state probe & 0.886 & 0.815 & 0.871 & 0.878 & 0.627 & 0.834 & 0.272 & 0.355 & 0.326 & 0.313\\
Combined selector & 0.895 & 0.824 & 0.876 & 0.885 & 0.641 & 0.852 & 0.280 & 0.381 & 0.314 & 0.313\\
\quad all three train sets & 0.912 & 0.846 & 0.877 & 0.889 & 0.640 & 0.853 & 0.274 & 0.374 & -- & --\\
Coordinate decoding & 0.893 & 0.838 & 0.865 & 0.866 & 0.853 & 0.900 & 0.774 & 0.623 & 0.520 & 0.662\\
\midrule
\multicolumn{11}{@{}l}{\emph{Qwen3-VL-8B}}\\
Text-only query & 0.669 & 0.508 & 0.671 & 0.674 & 0.469 & -- & -- & 0.273 & -- & --\\
Alignment hook & 0.853 & 0.800 & 0.853 & 0.856 & 0.531 & 0.695 & 0.250 & 0.437 & -- & --\\
Relation-aware head & 0.905 & 0.853 & 0.900 & 0.899 & 0.605 & 0.814 & 0.247 & 0.408 & -- & --\\
Token-state probe & 0.909 & 0.858 & 0.897 & 0.894 & 0.579 & 0.782 & 0.230 & 0.434 & -- & --\\
Combined selector & 0.910 & 0.855 & 0.898 & 0.895 & 0.637 & 0.861 & 0.252 & 0.439 & -- & --\\
Coordinate decoding & 0.918 & 0.868 & 0.886 & 0.890 & 0.880 & 0.918 & 0.815 & 0.671 & -- & --\\
\bottomrule
\end{tabular}
\end{table*}

\cref{tab:full} answers the first question. Proposal recall is at least $0.998$
on the COCO-based splits, so \cref{eq:decomp} attributes the loss to selection:
for the hook trained on $24$k rows $\SayErr$ is $0.21$ on RefCOCO and RefCOCOg,
$0.29$ on RefCOCO+ and $0.62$ on the stress set, whereas on Ref-L4 perception
accounts for $4\%$ of queries. At that budget the
hook is $0.07$--$0.13$ below coordinate-snap on every standard split and not
distinguishable from detector-native selection on RefCOCOg. Trained on the full set, every head gains on
every split outside the stress set, where the relation-aware head, the
token-state probe and the combined selector lose $0.03$, $0.07$ and $0.04$. On RefCOCOg test the
combined selector exceeds the raw box by $0.018$ [$0.012$, $0.025$] with
Qwen2.5-VL and is level with it with Qwen3-VL ($+0.006$ [$0.000$, $0.011$]; on
val $+0.011$ and $+0.013$);
the relation-aware head alone exceeds the raw box by $0.012$ [$0.005$, $0.020$]
and cross-attention to the pooled query reaches $0.881$, so the scoring, not
the combination, carries most of it (two-level intervals). On RefCOCO val the combined selector is level with the raw box,
on RefCOCO+ val $0.014$ behind (Qwen3-VL: $0.008$ and $0.014$ behind); trained on all three RefCOCO-family training sets it leads
on RefCOCO val, RefCOCO+ val and RefCOCOg test by $0.018$, $0.008$ and $0.023$
(intervals above zero; \cref{tab:app-union}), though the added data move
RefCOCOg test by only $+0.005$ [$0.000$, $0.009$] and Ref-L4 not at all. VGent
\cite{vgent} reports $0.01$--$0.035$ more on these splits; as neither these data nor a class-agnostic
proposer (\cref{sec:reasoning}) close that gap, it may lie in what we hold
fixed: its tuned encoder read at every layer, its three proposers used together and its broader corpora. On Ref-L4
and the stress set decoding stays at least $0.21$ ahead of every head.
The findings below hold for both VLMs where \cref{tab:full} has both. On
RefCOCO/+ testA/testB and Ref-L4 test the hook is below the relation-aware
head and both below the snapped box. Selected boxes are not loose: at $\mathrm{IoU}\ge0.75$ the full-budget relation-aware head and
combined selector exceed the snapped box on RefCOCOg test by about $0.04$.

\subsection{What limits selection}
\label{sec:limits}

\begin{table*}[t]
\centering
\footnotesize
\setlength{\tabcolsep}{3.6pt}
\caption{\textbf{Diagnostic probes} (Qwen2.5-VL, $24$k rows; selection accuracy
at $\mathrm{IoU}\ge0.5$; mean$\pm$sample std over three seeds; $^\ast$selected
on RefCOCOg val; --: not evaluated). Test: RefCOCOg test. Left: the query source
with the head fixed (the bridge on the image-conditioned query coincides seed
by seed with the hook trained without the verifier), and the head with the
image-conditioned query fixed. Right: supervision and readout of the hook, and
set-level scoring at a matched $4.3$M parameters (cross-attention rows trained
in one run with a relation-aware head that scores $0.850$ / $0.517$ / $0.380$).}
\label{tab:probes}
\begin{tabular}{@{}lccc@{\hspace{12pt}}lccc@{}}
\toprule
& Test & Ref-L4 & Stress & & Test & Ref-L4 & Stress\\
\midrule
\multicolumn{4}{@{}l}{\emph{Query source (learned bridge)}} & \multicolumn{4}{l}{\emph{Supervision and readout (hook)}}\\
CLIP text & 0.553\sd{.005} & 0.359 & 0.237 & Single positive, final token & 0.793\sd{.007} & 0.464 & 0.384\\
Detector text tower & 0.605\sd{.006} & 0.388 & 0.250 & Every valid box positive & 0.805\sd{.014} & -- & --\\
\quad projected & 0.611\sd{.010} & 0.375 & 0.224 & Mean over expression & 0.721\sd{.008} & -- & --\\
VLM, text only & 0.593\sd{.007} & 0.372 & 0.294 & Three VLM depths & 0.817\sd{.001} & -- & --\\
VLM, image-conditioned & 0.791\sd{.013} & 0.465 & 0.423 & Three depths, every valid box & 0.842\sd{.003} & -- & --\\
\cmidrule(r){1-4}\cmidrule(l){5-8}
\multicolumn{4}{@{}l}{\emph{Head (image-conditioned query)}} & \multicolumn{4}{l}{\emph{Set-level scoring ($4.3$M parameters)}}\\
Alignment hook ($1.1$M) & 0.793\sd{.007} & 0.464 & 0.384 & Hook, single positive & 0.792\sd{.004} & 0.444 & 0.390\\
Index scorer, as proposed & 0.770\sd{.007} & 0.444 & 0.384 & Hook, every valid box & 0.816\sd{.011} & 0.497 & 0.392\\
Index scorer, no index emb. & 0.808\sd{.007} & 0.469 & 0.420 & Relation-aware, every valid box & 0.842\sd{.003} & 0.518 & 0.362\\
Index scorer, audited$^\ast$ & 0.815\sd{.002} & 0.451 & 0.434 & \quad + IoU-rank loss$^\ast$ & 0.849\sd{.001} & 0.522 & 0.388\\
\cmidrule(r){1-4}
Detector-native & 0.781 & 0.634 & 0.255 & Pooled query, cross-attention & 0.854\sd{.001} & 0.502 & 0.356\\
Coordinate-snap & 0.866 & 0.833 & 0.610 & Token states, cross-attention & 0.852\sd{.002} & 0.593 & 0.426\\
& & & & \quad + image tokens & 0.851\sd{.005} & 0.592 & 0.392\\
& & & & \quad + relations & 0.854\sd{.002} & 0.588 & 0.408\\
\bottomrule
\end{tabular}
\end{table*}

\paragraph{The query, and why it matters.}
With the head fixed (\cref{tab:probes}, left), every text-only query source
stays at or below $0.61$ on RefCOCOg test, against $0.79$ for the
image-conditioned VLM query; at the full budget the text-only query loses
$0.15$--$0.18$ on RefCOCOg with the bridge (both VLMs, \cref{tab:full}), the
relation-aware head and the token-state probe, and at $24$k rows it costs the hook
$0.134$ [$0.122$, $0.146$] with InternVL3-8B against $0.208$ with Qwen2.5-VL on
RefCOCOg val (\cref{tab:app-textq-heads,tab:app-ivl3}). The image-conditioned query already
localizes the referent: a linear probe from it to the box explains $68\%$ of the variance of the box center on RefCOCOg val ($6\%$ for the text-only query),
and snapping the probe's box to the nearest proposal selects the referent for
$0.56$ of queries. The hook reads this
location out against the region features ($0.79$); a query computed without the
image cannot carry it, so under late fusion over request-independent proposals
what matters is what the VLM computed with the image. The same detector reaches
$0.78$ when the expression conditions its features (detector-native) but $0.61$
when its text tower supplies the query. With the query fixed, pairwise
per-proposal heads differ little: the index-scorer variants stay within $0.03$
of the hook on RefCOCOg and seven capacity settings span at most $0.025$; scoring proposals by
cross-attention with every valid box as a positive adds $0.06$ over the
single-positive hook on RefCOCOg test, $0.02$--$0.04$ of it at matched
supervision and size, with or without relations among proposals
(\cref{tab:probes}, right). On Ref-L4 the query matters less ($0.06$--$0.13$) and the head more (with the text-only query the token-state
probe leads the bridge by $0.105$).

\paragraph{Ranking, not retrieval.}
On RefCOCOg the correct proposal is ranked first for $0.79$ of queries but is
among the top five for $0.96$ ($0.97$ after suppressing near-duplicates,
\cref{tab:invlm}): the scores contain the answer and
misorder it among a few top candidates. Of the hook's errors on RefCOCOg
val ($0.21$ of queries), four fifths keep the correct proposal in the top five,
split evenly between a box that overlaps the referent
($0.1\le\mathrm{IoU}<0.5$) and a different object; errors grow with the number
of same-category distractors ($0.19$ with one, $0.27$ with three or more). Changes aimed at that ordering pay off
(\cref{tab:probes}, right): every valid box as a positive adds $0.012$ and
three VLM depths $0.024$ (significant after correction only for the former at
matched capacity), whereas averaging over expression
tokens loses $0.07$. Set-level scoring adds $0.02$--$0.04$ over a hook with the
same parameters and supervision on every standard split, at both budgets with Qwen2.5-VL and at the full budget
with Qwen3-VL ($+0.023$ to $+0.035$), with every seed pair positive, but not
on the stress set; the
gain does not come from relations among proposals: removing the pairwise
geometry bias changes accuracy by at most $0.004$ outside the stress set, and
cross-attention to the pooled query alone matches it. Trained on the full set and
combined in one selector, these changes reach coordinate decoding on RefCOCOg
(\cref{tab:full}).

\begin{table}[t]
\centering
\footnotesize
\setlength{\tabcolsep}{3pt}
\caption{\textbf{Selecting inside the VLM} (Qwen2.5-VL; accuracy at
$\mathrm{IoU}\ge0.5$; hook: seed $42$). The VLM chooses among five proposals (near-duplicates
suppressed) ranked by the hook or by the detector prompted with the
expression, drawn as numbered marks, listed as box coordinates, or judged one
at a time. Fine-tuned: one epoch of low-rank adaptation \cite{hu2022lora} on
the $24$k training rows, to answer the box list with a candidate's number (or
``none'' on swapped rows) or to decode a box in the native format (present
rows); three seeds, std $\le0.006$ on RefCOCOg, except the detector row (one
seed). Ref-L4: fixed $3{,}000$-query subset (hook seed: $0.451$; all of Ref-L4:
$0.452$). --: not evaluated.}
\label{tab:invlm}
\begin{tabular}{@{}lcccc@{}}
\toprule
& \multicolumn{2}{c}{RefCOCOg} & &\\
\cmidrule(lr){2-3}
& val & test & Ref-L4 & Stress\\
\midrule
Hook, top-1 & 0.787 & 0.788 & 0.451 & 0.390\\
Correct box in hook's top 5 & 0.973 & 0.973 & 0.757 & 0.736\\
Correct box in detector's top 5 & 0.980 & -- & 0.898 & --\\
\midrule
VLM over hook's top 5: marks & 0.587 & -- & 0.503 & 0.411\\
\quad box list & 0.760 & 0.762 & 0.576 & 0.524\\
\quad true/false per box & 0.735 & -- & 0.587 & 0.489\\
VLM over detector's top 5: marks & 0.479 & -- & 0.485 & 0.429\\
\midrule
Fine-tuned, box list, hook's top 5 & 0.878 & 0.882 & 0.688 & 0.528\\
\quad detector's top 5 & 0.892 & -- & 0.763 & --\\
Fine-tuned, native box & 0.871 & 0.875 & -- & --\\
\midrule
Coordinate decoding & 0.865 & 0.866 & 0.838 & 0.623\\
\bottomrule
\end{tabular}
\end{table}

\paragraph{Selecting inside the VLM.}
If ordering is the bottleneck, the VLM might order the top candidates itself. Given the hook's five highest-ranked proposals, which
contain a correct box for $0.97$ of RefCOCOg queries, none of three zero-shot
prompts does (\cref{tab:invlm}): the best, a list of box coordinates, stays below
the hook ($0.760$ against $0.787$), marks drawn on the image fare worse, as
reported for open VLMs \cite{crg}, and Qwen3-VL behaves alike with the box list. Fine-tuning, as in candidate-region
selection \cite{crs2025}, changes this: one epoch on the hook's $24$k training
rows, answering with a candidate's number, brings the same VLM to
$0.882$ on RefCOCOg test, above the hook and its own decoding ($+0.016$
[$0.007$, $0.024$]; three seeds). The same epoch in the native box format brings
decoding to $0.875$, not distinguishable ($+0.007$ [$-0.001$, $0.015$];
\cref{tab:app-lora-matched}): training adds $0.12$ to choosing from the list
but under $0.01$ to decoding, so it teaches ordering without making
selection among the hook's candidates beat equally trained decoding. On Ref-L4 and the stress set
it gains $0.24$ and $0.14$ over the hook, below decoding; on the expression-prompted detector's top five it reaches $0.892$ on val, above
decoding (one seed).

\subsection{Ref-L4 and the reasoning probes}
\label{sec:reasoning}
On Ref-L4 the hook's top five contain the correct proposal for only $0.76$ of
queries, below decoding's $0.84$ (\cref{tab:invlm}), so reordering them cannot
close the gap; on the stress set ($0.74$ against $0.62$) it could. The relation-aware head gains $0.06$
on Ref-L4 over the hook ($0.02$ at matched parameters and supervision) but
still trails coordinate-snap by $0.31$; on the stress set it gains nothing over the hook ($+0.004$ [$-0.036$, $+0.046$] at
$24$k, clustered by template). Decoding leads
the hook there by $0.62$ to $0.38$, also on each of the $14$ templates with at
least ten queries, and affordance requests stay below $0.40$ for every trained
interface.

\paragraph{Pooled versus token-level query.}
Is the pooled query the limit, as pooled states of generative VLMs are weak
retrieval embeddings \cite{realign2025}? Proposal tokens cross-attending to the request tokens' final-layer states, at the relation head's parameter count, test this (\cref{tab:probes}, right). On Ref-L4 token states
help at this budget: $0.593$ against at most $0.522$ for pooled-query heads and $+0.08$ over a same-run relation-aware head (paired
test, Holm $p\le0.006$), while RefCOCOg is unchanged and image tokens or
relations add nothing. On the stress set they do not ($0.426$ against
$0.423$ and $0.434$ for the best pooled-query heads). The advantage
shrinks with data: at the full budget the pooled control reaches $0.606$ and the
token-state probe $0.627$, it loses $0.07$ on the stress set, and with Qwen3-VL it trails the relation-aware head on Ref-L4 ($0.579$ against $0.605$; \cref{tab:full}).

\paragraph{On Ref-L4 trained heads meet a category boundary.}
On Ref-L4 detector-native selection ($0.634$) leads every trained head except
the full-budget combined selector ($0.641$ and $0.637$). Splitting it by image source---COCO images, whose referents all belong to the
$80$ COCO classes of the proposal prompt and of the training data,
and Objects365 images (``other'')---locates most of the reversal
(\cref{tab:full}). On COCO images ($63\%$ of queries)
the full-budget relation and cross-attention heads reach $0.78$--$0.86$ and the
combined selector comes within $0.06$ of decoding;
on Objects365 images every trained head stays at $0.19$--$0.28$ while the
expression-prompted detector keeps $0.61$ and decoding $0.77$--$0.82$, although proposals still cover $0.89$ of these referents, whose boxes are also
smaller. A third of them map to a COCO class, and these are selected about
twice as often as the rest (hook at $24$k: $0.38$ against $0.17$), a drop $0.14$
larger than that of decoding. In a regression on vocabulary, image source and box size the trained heads'
category term stays at $+0.84$ to $+0.94$ log-odds under every size
specification; the image-source term depends on it. Contemporaneous work attributes
a similar drop of RefCOCO-fine-tuned VLMs to target size \cite{locus}. On these images released Jag and LOCUS v1.0 reach $0.36$--$0.48$; size is their largest term, and a smaller category
term remains ($+0.33$ to $+0.53$; \cref{tab:app-e54-size}). LOCUS v1.1, also trained on small targets, reaches $0.52$--$0.60$. Prompting the detector with every Objects365
name lifts recall on referents without a COCO counterpart to $0.95$
but the hook only from $0.17$ to $0.27$. A class-agnostic proposer without a category prompt (UPN
\cite{chatrex}) raises their recall from $0.89$ to $0.97$, yet the retrained heads stay at $0.17$--$0.29$
there (\cref{tab:app-upn}). Training on half of Ref-L4's Objects365 referents
lifts the other half from $0.24$ to $0.47$ (hook, $24$k; on referents without a
COCO counterpart from $0.17$ to $0.42$, $0.50$ with the extended prompt); as many
COCO-image Ref-L4 rows, same caption style but no new categories, add nothing
there ($-0.008$ [$-0.018$, $0.002$], hook), and $0.05$--$0.09$ of the gain is
specific to the new categories, the rest reaching COCO-mappable referents on
Objects365 images too (\cref{tab:app-e31,tab:app-style}). A RefCOCOg-trained selector thus does not transfer
beyond the COCO categories although the frozen Qwen VLMs do, and $30$k Flickr30k Entities
phrases \cite{flickr30kentities} on non-person, non-COCO objects lift
Objects365 referents by only $0.06$--$0.08$, mostly regardless of naming
(\cref{tab:app-broadvocab}). With InternVL3 the hook ($0.58$ to $0.21$) and decoding ($0.92$ to $0.56$) both drop.

\paragraph{Public benchmarks.}
On Ref-Adv-s, whose own prompt reproduces the published Qwen2.5-VL score
here, and on RefBench-PRO a proposal covers $91\%$ and $88\%$ of referents, yet every full-budget head reaches only $0.31$--$0.33$ and $0.29$--$0.31$, against
$0.52$ and $0.66$ for decoding (\cref{tab:full}). Negation is not the obstacle (heads $0.33$--$0.35$ with it, $0.29$--$0.31$
without); relation and commonsense requests are ($0.155$--$0.244$ against
$0.607$--$0.662$), and on RefBench-PRO even the expression-prompted detector
($0.52$) beats every head (\cref{tab:app-refadv,tab:app-prbench}).
On GroundingME \cite{groundingme} (present referents) decoding reaches
$0.24$ ($0.35$ with Qwen3-VL) and every trained selector $0.05$--$0.07$,
although proposals cover $79\%$ of referents (\cref{tab:app-gme-diag}): not a
pipeline fault---they reach $0.81$--$0.86$ on COCO images---but a
selection failure (first covering proposal at median rank $16$--$31$ of $300$), as the query barely
steers it: another request's query costs the hooks $0.01$--$0.02$, against
$0.50$--$0.55$ on RefCOCOg val.

\subsection{What the verifier contributes}
\label{sec:verifier}
A swapped expression describes a specific object of another image, so the
question is whether this image contains that referent, not a category.
Attached unchanged to coordinate-snap, the verifier gives joint accuracy
$0.888$ on RefCOCOg test with $30\%$ swapped absent referents, against $0.877$
for a threshold on the VLM's yes-probability tuned on val ($+0.011$
[$0.009$, $0.014$], paired sign-flip test), $0.841$ for its plain yes/no answer,
$0.788$ for the hook's own veto and $0.606$ for always answering
(\cref{tab:app-detach}). It leads only at its operating point, on its own kind of negative. As a ranking signal the VLM's
yes-minus-no score is better (AURC $0.134$ against $0.148$ on val); its yes/no answer rejects more same-category hard negatives ($0.91$ against
$0.87$); and on every public no-target benchmark that answer wins: on POPE,
gRefCOCO testA and testB (Pr@(F1$=$1) $0.51$ and $0.52$ against $0.48$, also after retraining on gRefCOCO negatives), GroundingME's attribute-edited negatives
($0.87$--$0.91$ rejected against $0.33$--$0.38$) and RefBench-PRO (joint accuracy $0.52$ against $0.48$). Its signal lies in the query: the projected query alone, or a logistic
probe on the VLM's raw state, matches it; the region alone does not.
Qwen3-VL gives the same ordering. Decoding also refuses on its own:
abstaining when no box parses gives joint accuracy $0.66$ (Qwen2.5-VL, snapped
box, $17\%$ of swapped requests refused) and $0.82$ (Qwen3-VL, raw box, $66\%$),
below either model's yes/no answer. On the stress set
no gate is reliable; the verifier is level with the tuned yes-probability.

\section{Limitations}
\label{sec:limitations}
All conclusions are for two detectors of one family (one without RefCOCO data),
a class-agnostic proposer at the $24$k budget
and two VLMs of one family with undisclosed grounding corpora, plus InternVL3 (language model from Qwen2.5;
hook and decoding only, $24$k rows). The main heads saw only COCO-category referents. The templated reasoning
probes are small, a third of their swaps satisfied by the new image; Ref-Adv-s and
RefBench-PRO used Qwen2.5-VL heads only. The VLMs fine-tuned to select or decode (\cref{tab:invlm}) saw one epoch of $24$k rows.
No VGent-class system is public; its remaining lead is inferred.

\section{Conclusion}
\label{sec:conclusion}
Proposal selection over a frozen VLM's query fails in three places: in what the query saw, in ordering a few top candidates, and beyond the
training distribution (new categories, reasoning requests). The first two are fixed on in-distribution
expressions by reading the query from the VLM with the image, supervising with
every valid box and scoring proposals by cross-attention, at $0.41$ of decoding's latency. The third is not fixed by a better head, a
wider detector prompt or a class-agnostic proposer; training on the target distribution
narrows it. Beyond RefCOCO-family expressions the VLM's own box is better
when its latency is acceptable, and on public no-target benchmarks its yes/no
answer abstains better than a trained verifier.

\section*{LLM Usage Statement}
LLM-based assistants (Anthropic Claude models, used through the Claude Code
agent) were used extensively in this work, often carrying out multi-step tasks
autonomously under the authors' direction. Ideation and experiments: proposing
and planning experiments, and acting as simulated reviewers whose critiques
shaped which analyses were run; the authors decided what to pursue. Code and
figures: writing and debugging the experiment, analysis and figure code, and
launching and monitoring runs. Data: generating the rule-based candidate queries
of the stress set and writing the data-preparation code; annotators who are not
authors filtered the candidates. Related work: finding and summarizing literature.
Text: drafting and revising the paper and the supplementary material, including
grammar and wording. Checking: cross-checking reported numbers against the
result files. No reported number was generated by an LLM, other than the outputs
of the models under study. The authors reviewed all content and take full
responsibility for it.

\clearpage
\renewcommand{\sd}[1]{{\scriptsize$\pm$#1}}
\renewcommand{\thesection}{S\arabic{section}}
\renewcommand{\thetable}{S\arabic{table}}
\renewcommand{\thefigure}{S\arabic{figure}}
\renewcommand{\theequation}{S\arabic{equation}}
\setcounter{section}{0}\setcounter{table}{0}\setcounter{figure}{0}\setcounter{equation}{0}
\maketitlesupplementary
\begin{center}\small Bo Ma and Weiqi Yan\\Auckland University of Technology\\Jinsong Wu\\Guilin University of Electronic Technology\end{center}

This supplementary material accompanies the main paper. It gives the formal
objectives, the accounting identities, the results that the main paper
summarizes in text (reasoning-stress categories, POPE, training budget),
extended results and ablations (Ref-L4 vocabulary, a class-agnostic proposer,
the prompt-versus-training vocabulary test and a caption-style control, margins,
in-VLM selection including the fine-tuned selector and its matched control),
an error taxonomy of the hook, the full evaluation of the verifier as a
detachable abstention layer (including an untouched test split, hard negatives,
its input ablation and the query-location probe), the construction and an
audit of the reasoning-stress set, the public reasoning and rejection
benchmarks, the full training set and training on all three RefCOCO-family sets, the second VLM, a third VLM and a second proposal source,
an audit of image overlap between training and evaluation splits, the
statistical protocol, the feature builds, and the reproducibility
configuration. Notation and metrics follow the main
paper: \emph{selection accuracy} is the accuracy of the top-ranked proposal at
$\mathrm{IoU}\ge0.5$ with abstention disabled, \emph{answer rate} (the fraction
of rows answered) and \emph{selective accuracy} describe an abstaining system, and
\emph{instance-exact selection} is the fraction of present referents whose
top-ranked proposal is the proposal of maximal IoU with the annotated box,
provided that this IoU is at least $0.5$; it is therefore never above selection
accuracy, and it is the correctness criterion of the veto-threshold selection
(\cref{sec:app-thresholds}). The main paper calls the frozen VLM's box snapped to the proposals
\emph{coordinate-snap} and the unsnapped box \emph{coordinate decoding}.

\paragraph{Where each claim is supported.}
\Cref{tab:app-claims} maps each finding of the main paper to its main-paper
table and to the supplementary evidence.

\begin{table*}[t]
\centering
\footnotesize
\setlength{\tabcolsep}{4pt}
\caption{Claim index. Main-paper tables are numbered as in the main paper.}
\label{tab:app-claims}
\begin{tabular}{@{}>{\raggedright\arraybackslash}p{0.43\textwidth}>{\raggedright\arraybackslash}p{0.12\textwidth}>{\raggedright\arraybackslash}p{0.39\textwidth}@{}}
\toprule
Claim & Main paper & Supplement\\
\midrule
The query source decides accuracy under late fusion, because the image-conditioned query encodes the referent's location; head form and capacity matter less than the query on the RefCOCO family (on RefCOCOg test the image-conditioned query adds $0.15$--$0.17$ with each of three heads, a set-level head $0.02$--$0.06$ over the hook; on Ref-L4 the head matters as much) & Tables 1, 2; Fig.~1 & \cref{tab:app-query-location,tab:app-capacity,tab:app-full-cap,tab:app-textq-heads,tab:app-grounding24k,tab:app-matched-hook-q3,tab:app-intervals,tab:app-tests}\\
Remaining errors are ranking errors (correct proposal usually in the top five) & Tables 2, 3; Fig.~1 & \cref{fig:app-topk,tab:app-errors,tab:app-coverage}\\
Cross-attention scoring, token states, multi-layer readout and multi-positive supervision reach coordinate decoding on RefCOCOg at the full budget, and on RefCOCO/+ when trained on all three RefCOCO-family sets (not explained by the smaller RefCOCOg part of that union); VGent's remaining lead is attributed by elimination, also against its class-agnostic proposer & Tables 1, 2; Fig.~1 & \cref{tab:app-grounding24k,sec:app-full,tab:app-full-acc,tab:app-full-iou75,tab:app-allsplits,tab:app-union,tab:app-union-diffs,tab:app-published}\\
Prompting the VLM to choose does not help on RefCOCOg; fine-tuning it to choose adds $0.12$ but ends level with fine-tuning it to decode on the same rows & Table 3 & \cref{tab:app-invlm,tab:app-invlm-splits,tab:app-e32,tab:app-lora-matched,tab:app-lora-diffs}\\
Ref-L4: collapse outside the trained categories, not lifted by a wider detector prompt or a class-agnostic proposer; training on the new categories helps, not through caption style & Table 1; Fig.~1 & \cref{tab:app-refl4buckets,tab:app-refl4source,tab:app-extvocab,tab:app-e31,tab:app-upn,tab:app-style}\\
Released Jag and LOCUS weights on every column of Table 1: on Objects365 images far below decoding, with a smaller category term than our heads' with size controlled & -- & \cref{tab:app-e54,tab:app-e54-size}\\
Second detector without RefCOCO data leaves the findings unchanged & -- & \cref{tab:app-second-detector}\\
A third VLM (InternVL3-8B, hook and decoding at $24$k): the query finding transfers ($0.134$), the decoding side of the category contrast does not (its decoding also falls outside the COCO categories) & -- & \cref{tab:app-ivl3}\\
Reasoning probes and public benchmarks (Ref-Adv-s, RefBench-PRO, GroundingME): trained selectors far below decoding although proposals cover most referents; the GroundingME collapse is on the selection side & Table 1; Fig.~1 & \cref{tab:app-stress,tab:app-stress-templates,tab:app-refadv,tab:app-prbench,tab:app-public,tab:app-gme-diag}\\
Verifier vs.\ the VLM's own signals; prior and hard negatives; RefBench-PRO rejection & -- & \cref{tab:app-detach,tab:app-full-abst,tab:app-prior,tab:app-hardneg,tab:app-abst-intervals,tab:app-public-reject}\\
The verifier's signal lies in the query; the query encodes the referent's location & -- & \cref{tab:app-verifier-inputs,tab:app-query-location}\\
Feature builds and provenance & -- & \cref{tab:app-builds}\\
\bottomrule
\end{tabular}
\end{table*}

\section{Formal Selection Objective}
Let $\mathcal{R}=\{(b_j,e_j)\}_{j=1}^{M}$ be the frozen detector's proposals
and $u$ the query embedding of one request (every experiment has one referent
or none per request). With projections $\tilde{e}_j=f_\theta(e_j)$,
$\tilde{u}=g_\phi(u)$ and affinity
$s_{j}=\langle\tilde{u},\tilde{e}_j\rangle/\kappa$, selection is trained as
contrastive retrieval of one proposal.

\paragraph{Retrieval.}
The positive set $P$ of a present request holds the matched proposal, the
proposal of maximal IoU with the ground-truth box; $P$ is empty for an absent
request and for a present request that no proposal covers at
$\mathrm{IoU}\ge0.5$ ($0.0$--$0.2\%$ of requests at $M{=}300$). Over a batch,
\begin{equation}
\mathcal{L}_{\mathrm{ret}} = -\sum_{n:\,P_n\neq\varnothing}\log
\frac{\sum_{j\in P_n}\exp(s_{nj})}{\sum_{j=1}^{M}\exp(s_{nj})}.
\end{equation}
Requests with $P_n=\varnothing$ enter only the verifier objective. The
multi-positive variant of the supervision probes puts every proposal with
$\mathrm{IoU}\ge0.5$ into $P_n$; the main configuration uses the single
matched proposal. The hook has no box-regression output: the emitted box is
always a frozen proposal.

\section{Faithfulness Objective and Veto}
The verification head reads the projected query, the projected region of
maximal affinity $\sigma=\arg\max_j s_j$ and their affinity,
$v(\tilde{u},\tilde{e}_{\sigma},s_{\sigma})\in[0,1]$, with the same choice of
$\sigma$ during training as at inference, and is trained with
\begin{align}
\mathcal{L}_{\mathrm{faith}}&=\sum_{n}\mathrm{BCE}\big(v_n,y_n\big),\\
y_n&=\one[\text{request }n\text{ has a referent}].\nonumber
\end{align}
A present request receives $y_n{=}1$ whether or not its selected proposal is
correct; the verifier is therefore an existence detector conditioned on the
best-scoring region, not a correctness estimator for the selection. Absent
training requests are constructed by pairing an expression with a different
image (random swap, rate $0.3$); Section~\ref{sec:app-labels} compares the
existence label with a correctness label. At inference a request emits
$b_\sigma$ iff $v\ge\gamma_v$ and $s_\sigma\ge\gamma_s$; otherwise it abstains.
The total loss is
$\mathcal{L}=\mathcal{L}_{\mathrm{ret}}+\lambda_{\mathrm{f}}\mathcal{L}_{\mathrm{faith}}$
with $\lambda_{\mathrm{f}}=1$; gradients of both terms reach the projections.

\section{Accounting Identities}
\label{sec:app-diagnostics}

Let the detector produce proposals $\{(b_{n,j},e_{n,j})\}_{j=1}^{M}$ for input
$n$, let $b_n^\star$ be the ground-truth referent box when it exists, and let
$y_n^\star\in\{0,1\}$ denote existence. Restrict to present queries
($y_n^\star{=}1$, $n=1,\dots,N_+$) and fix the IoU threshold $\tau$.

\paragraph{Definitions.}
The referent is covered by the proposal set when
$z_n=\mathbf{1}[\max_j\mathrm{IoU}(b_{n,j},b_n^\star)\ge\tau]$;
$\SeeErr=1-\frac{1}{N_+}\sum_n z_n$ depends only on the frozen detector and
$M$. With $\hat{b}_n$ the selected proposal and
$h_n=\mathbf{1}[\mathrm{IoU}(\hat{b}_n,b_n^\star)\ge\tau]$ the hit indicator,
$\SayErr=\sum_n z_n(1-h_n)/\sum_n z_n$ is the failure rate given a covering proposal.
False-positive and false-negative existence rates are
$\epsilon_{\mathrm{hall}}=\sum_n\mathbf{1}[y_n^\star{=}0\wedge\hat{y}_n{=}1]/\sum_n\mathbf{1}[y_n^\star{=}0]$
and
$\epsilon_{\mathrm{miss}}=\sum_n\mathbf{1}[y_n^\star{=}1\wedge\hat{y}_n{=}0]/\sum_n\mathbf{1}[y_n^\star{=}1]$.

\paragraph{Proposition 1.}
Because $\hat{b}_n$ is one of the $M$ proposals,
$\mathrm{IoU}(\hat{b}_n,b_n^\star)\le\max_j\mathrm{IoU}(b_{n,j},b_n^\star)$, so
$h_n=1$ implies $z_n=1$ and $h_n=z_nh_n$ for every $n$. Hence
\begin{align*}
\sum_{n} h_n &=\sum_n z_n-\sum_n z_n(1-h_n)\\
&=\Big(\sum_n z_n\Big)\big(1-\SayErr\big),
\end{align*}
and dividing by $N_+$ gives $\Acc=(1-\SeeErr)(1-\SayErr)$. The only
assumption is that the emitted box is drawn from the proposal set, which is
how every hook-level accuracy in this paper is measured.

\paragraph{Numerical check.}
At $M{=}300$ proposal recall is $1.000$ (RefCOCOg val), $0.998$ (RefCOCOg test),
$0.999$ (RefCOCO val) and $0.999$ (RefCOCO+ val)
(Table~\ref{tab:app-coverage}). With the three-seed selection accuracies of the
main paper the identity gives $\SayErr(0.5)=0.213$ (RefCOCOg val), $0.205$
(RefCOCOg test), $0.210$ (RefCOCO val) and $0.292$ (RefCOCO+ val).

\paragraph{Margin bound and selection ceiling.}
Fix a query and write $s_j=\langle\tilde u,\tilde e_j\rangle/\kappa$ with
$\|\tilde u\|=\|\tilde e_j\|=1$. For any two proposals,
\begin{equation}
s_i-s_j=\tfrac{1}{\kappa}\langle\tilde u,\tilde e_i-\tilde e_j\rangle
\le\tfrac{1}{\kappa}\|\tilde e_i-\tilde e_j\|
\label{eq:app-margin}
\end{equation}
by Cauchy--Schwarz, with equality when $\tilde u$ is parallel to
$\tilde e_i-\tilde e_j$. The score margin between the referent's proposal and
a distractor is therefore bounded by their distance in the projected region
space, uniformly over queries. For selectability, let
$P\subseteq\{1,\dots,M\}$ be the correct proposals of a covered referent and
$D$ its complement. Some query ranks a member of $P$ first iff there exist
$p\in P$ and $u$ with $\langle u,\tilde e_p\rangle>\max_{j\in D}\langle
u,\tilde e_j\rangle$, i.e.\ iff $\tilde e_p$ is strictly separable from
$\{\tilde e_j\}_{j\in D}$, and the largest margin it can attain,
\begin{equation}
\mu_p=\max_{\|u\|\le1}\ \min_{j\in D}\ \langle u,\tilde e_p-\tilde e_j\rangle,
\label{eq:app-achievable}
\end{equation}
equals the Euclidean distance from $\tilde e_p$ to the convex hull of the
distractor features (linear separation duality).

Separability itself is not informative here: the projections are
$\ell_2$-normalized, every $\tilde e_j$ lies on the unit sphere, and every
point of a sphere is an extreme point of any set of points on it, so
$\mu_p>0$ always and the interface can express \emph{any} selection. We
verify this numerically ($\mu_p>0$ for every covered referent of every split)
and report it only as a check. What the geometry does constrain is the
\emph{width} of the cone of query directions that select correctly, measured
by $\mu_p$, and how much of it the emitted query attains: writing
$p$ for the proposal of maximal IoU with the referent, the realized margin is
$\rho=s_p-\max_{j\in D}s_j$, with $\rho>0$ exactly when that proposal
outranks every incorrect proposal, which implies a correct selection (another
correct proposal may still be ranked above $p$), and $\rho\le\mu_p/\kappa$ by
\cref{eq:app-achievable}. The ratio
$\kappa\rho/\mu_p\in(-\infty,1]$ is the share of the available room the
trained interface (query and projections together) uses. We compute $\mu_p$ with a batched Frank--Wolfe
iteration that returns both a separating direction (a certified lower bound
on $\mu_p$) and the hull iterate (an upper bound).

\paragraph{Corollary 1.}
With gate $g_n\in\{0,1\}$, the abstaining system counts a present query as
correct only if it is both answered and a hit:
$\Acc_{\mathrm{veto}}=\frac{1}{N_+}\sum_n g_nh_n
=\Acc-\frac{1}{N_+}\sum_n(1-g_n)h_n\le\Acc$, with equality iff
$(1-g_n)h_n=0$ for all $n$. The answer rate on present queries is $\frac{1}{N_+}\sum_n g_n$ and
selective accuracy is $\sum_n g_nh_n/\sum_n g_n$. On a positive-only split
abstention can therefore only lower accuracy; its benefit is measured on
queries without a referent, where the same gate converts false emissions into
correct abstentions.

\section{Results Summarized in the Main Paper}
\label{sec:app-main-results}

\paragraph{Grounding at the $24$k-row budget.}
\Cref{tab:app-grounding24k} gives the matched-perception comparison at the
probes' training budget, with the learned bridge and the index scorer that
Table 2 of the main paper varies. At this budget the hook is $0.07$--$0.13$
below coordinate-snap on every standard split and not distinguishable from
detector-native selection on RefCOCOg; the relation-aware head shown is the
variant with the IoU-rank loss, selected among five on RefCOCOg val
(\cref{tab:app-tuning}).

\begin{table*}[t]
\centering
\footnotesize
\setlength{\tabcolsep}{5pt}
\caption{\textbf{Grounding under matched perception at the $24$k-row budget}
(Qwen2.5-VL): accuracy at $\mathrm{IoU}\ge0.5$ on present queries, no
abstention. Trained rows select one of the same $300$ proposals
(mean$\pm$sample std, three seeds; $^\ast$selected on RefCOCOg val);
detector-native: raw box; coordinate-snap: the VLM's box snapped to the
proposals. The learned bridge here and the CLIP row of Table 2 of the main
paper are separate runs of the same recipe. Intervals: \cref{tab:app-intervals}.}
\label{tab:app-grounding24k}
\begin{tabular}{@{}lcccccc@{}}
\toprule
Method & RefCOCO val & RefCOCO+ val & RefCOCOg val & RefCOCOg test & Ref-L4 val & Stress\\
\midrule
Proposal recall (ceiling) & 0.999 & 0.999 & 1.000 & 0.998 & 0.958 & 1.000\\
\midrule
Learned bridge (CLIP query) & 0.511\sd{.017} & 0.429\sd{.008} & 0.542\sd{.006} & 0.552\sd{.009} & 0.357\sd{.006} & 0.254\sd{.022}\\
Index scorer & 0.759\sd{.002} & 0.683\sd{.004} & 0.770\sd{.002} & 0.770\sd{.007} & 0.444\sd{.010} & 0.384\sd{.032}\\
Index scorer, audited$^\ast$ & 0.801\sd{.003} & 0.722\sd{.009} & 0.811\sd{.003} & 0.815\sd{.002} & 0.451\sd{.027} & 0.434\sd{.017}\\
Alignment hook & 0.789\sd{.007} & 0.707\sd{.010} & 0.787\sd{.004} & 0.793\sd{.007} & 0.464\sd{.010} & 0.384\sd{.035}\\
Relation-aware head, IoU-rank loss$^\ast$ & 0.849\sd{.003} & 0.772\sd{.003} & 0.846\sd{.002} & 0.849\sd{.001} & 0.522\sd{.007} & 0.388\sd{.014}\\
\midrule
Detector-native & 0.847 & 0.737 & 0.783 & 0.781 & 0.634 & 0.255\\
Coordinate-snap & 0.895 & 0.840 & 0.867 & 0.866 & 0.833 & 0.610\\
\bottomrule
\end{tabular}
\end{table*}

\paragraph{Reasoning-stress set by category.}
\cref{tab:app-stress} breaks the $231$-query stress set down by construction
category. Every interface built on the image-conditioned VLM query reaches
$0.57$--$0.62$ on negation, against $0.28$ for detector-native selection
and the CLIP-query bridge; affordance and world-knowledge requests remain hard
for every trained interface, and coordinate decoding leads every one of them by
at least $0.24$ on affordance. Because queries that share a template are not
independent, the table also gives intervals that resample templates
(\cref{sec:app-stress}).

\begin{table}[t]
\centering
\footnotesize
\setlength{\tabcolsep}{1.8pt}
\caption{Reasoning-stress set by category (selection accuracy at
$\mathrm{IoU}\ge0.5$, no abstention; trained rows mean$\pm$sample std over three
seeds; ${}^\dagger$single seed, rebuilt caches, \cref{sec:app-interface}). Coordinate decoding is scored on the raw VLM
box. ``Template interval'': $95\%$ bootstrap interval over the $25$ query
templates (clusters; $10{,}000$ resamples), for the hook computed on per-query
correctness averaged over seeds.}
\label{tab:app-stress}
\begin{tabular}{@{}lcccc@{}}
\toprule
Method & Afford. & Negation & World kn. & All\\
& ($n{=}125$) & ($n{=}71$) & ($n{=}35$) & ($n{=}231$)\\
\midrule
Learned bridge (CLIP) & 0.251\sd{.033} & 0.282\sd{.028} & 0.210\sd{.033} & 0.254\sd{.022}\\
Detector-native & 0.256 & 0.282 & 0.200 & 0.255\\
\quad template interval & \ci{.129}{.386} & \ci{.184}{.359} & \ci{.029}{.481} & \ci{.172}{.337}\\
Index scorer & 0.296\sd{.037} & 0.615\sd{.064} & 0.229\sd{.049} & 0.384\sd{.032}\\
Alignment hook & 0.309\sd{.038} & 0.568\sd{.080} & 0.276\sd{.066} & 0.384\sd{.035}\\
\quad template interval & \ci{.216}{.403} & \ci{.469}{.681} & \ci{.135}{.588} & \ci{.297}{.470}\\
Hook, no verifier loss & 0.360\sd{.050} & 0.592\sd{.042} & 0.305\sd{.016} & 0.423\sd{.017}\\
Relation-aware head & 0.325\sd{.012} & 0.582\sd{.008} & 0.219\sd{.044} & 0.388\sd{.014}\\
Discrete phrase${}^\dagger$ & 0.392 & 0.606 & 0.343 & 0.450\\
Coordinate decoding & \textbf{0.640} & \textbf{0.718} & \textbf{0.371} & \textbf{0.623}\\
\quad template interval & \ci{.536}{.745} & \ci{.636}{.817} & \ci{.211}{.679} & \ci{.534}{.706}\\
\bottomrule
\end{tabular}
\end{table}

\paragraph{POPE existence probes.}
\cref{tab:app-pope} isolates the two gates of the default veto at the fixed
thresholds and compares them with the VLM's own answer. The area in the last
column is taken under the hallucination--recall curve (recall = answer rate on
present objects, risk = hallucination rate); it is not the selective AURC used
elsewhere.

\begin{table}[t]
\centering
\footnotesize
\setlength{\tabcolsep}{2.5pt}
\caption{POPE at $(\gamma_v,\gamma_s)=(0.3,24)$. Hook rows: means over
three seeds; ``verifier only'' and ``affinity only'' replay the trained hook
with one gate disabled; ``correctness label'': verifier trained on
$y=\one[\text{present}\wedge\text{selection correct}]$. VLM rows are single runs.
Rec.: recall on present objects (identical across splits); Hall.\ and
area: adversarial split.}
\label{tab:app-pope}
\begin{tabular}{@{}lcccccc@{}}
\toprule
& \multicolumn{3}{c}{Accuracy} & & \multicolumn{2}{c}{Adversarial}\\
\cmidrule(lr){2-4}\cmidrule(lr){6-7}
System & Rand. & Pop. & Adv. & Rec. & Hall. & Area\\
\midrule
No verifier (affinity) & 0.573 & 0.586 & 0.589 & 0.521 & 0.343 & 0.341\\
Default veto & 0.599 & 0.600 & 0.594 & 0.216 & 0.028 & 0.265\\
Affinity only & 0.596 & 0.597 & 0.592 & 0.217 & 0.033 & 0.261\\
Verifier only & 0.806 & 0.802 & 0.763 & 0.666 & 0.140 & 0.135\\
Correctness label & 0.886 & 0.862 & 0.820 & -- & -- & --\\
\midrule
VLM, direct answer & 0.884 & 0.875 & 0.864 & 0.777 & 0.049 & --\\
VLM + contrastive dec. & \textbf{0.897} & \textbf{0.888} & \textbf{0.873} & 0.809 & 0.063 & --\\
\bottomrule
\end{tabular}
\end{table}

\begin{figure}[t]
\centering
\includegraphics[width=0.95\columnwidth]{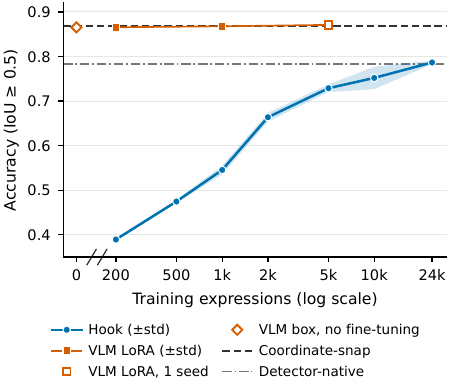}
\caption{Training budget on RefCOCOg val (accuracy at $\mathrm{IoU}\ge0.5$).
The hook (mean over three seeds) against low-rank fine-tuning of the
VLM in its native box format (three seeds at $200$ and $1{,}000$ expressions,
one seed at $5{,}000$) and the VLM's box without fine-tuning through the same
pipeline; horizontal lines are the coordinate-snap and detector-native
references, which receive no task-specific training here. Band and error bars span one sample standard deviation over
seeds.}
\label{fig:app-dataeff}
\end{figure}

\paragraph{Training budget.}
\cref{fig:app-dataeff} compares training the hook with low-rank fine-tuning of
the VLM in its native box format (protocol in \cref{sec:app-additional-discussion}).

\section{Extended Results}
\label{sec:app-extended}

All numbers in this section follow the metric contract of the main paper and,
for trained methods, are the mean$\pm$sample std over training seeds 42, 123 and 456 at
the fixed thresholds $(\gamma_v,\gamma_s)=(0.3,24.0)$.

\paragraph{Untrained hook.}
The hook with random projections, forced to emit its top proposal, measures the
modality gap rather than a threshold. Its RefCOCOg val accuracy is $0.007$,
$0.122$ and $0.115$ for the three random initializations ($0.082{\pm}0.065$;
intervals for every split in \cref{tab:app-intervals}). Under the operating
thresholds the untrained hook abstains on every query.

\paragraph{Proposal recall.}
Table~\ref{tab:app-coverage} reports the proposal ceiling used by the main
paper. At $M{=}300$ the pool covers $99.8$--$100\%$ of referents at IoU $0.5$
on the COCO-based splits and $95.8\%$ on Ref-L4, so the residual grounding
error is selection error rather than perception error. A hit need not be the
proposal of maximal IoU: on RefCOCOg val the hook picks that proposal for
$0.405$ of queries (three-seed mean) but hits at $\mathrm{IoU}\ge0.5$ for
$0.787$, so about half of its hits are other proposals that overlap the
referent (on RefCOCOg test $0.399$ against $0.793$).

\begin{table}[htbp]
\centering
\footnotesize
\setlength{\tabcolsep}{5pt}
\caption{Proposal recall at $M{=}300$ and the implied selection error of the
alignment hook. Proposal recall ($\mathrm{IoU}\ge0.5$) is the fraction of present referents covered by some
proposal at IoU $\ge0.5$ (the ceiling on selection accuracy); $\SeeErr$ is its
complement; $\SayErr(0.5)$ follows from Proposition~1 and the three-seed
selection accuracy.}
\label{tab:app-coverage}
\begin{tabular}{@{}lccc@{}}
\toprule
Split & Proposal recall & $\SeeErr$ & $\SayErr$ (hook)\\
\midrule
RefCOCO val      & 0.999 & 0.001 & 0.210\\
RefCOCO+ val     & 0.999 & 0.001 & 0.292\\
RefCOCOg val     & 1.000 & 0.000 & 0.213\\
RefCOCOg test    & 0.998 & 0.002 & 0.205\\
Ref-L4 val       & 0.958 & 0.042 & 0.516\\
Reasoning stress & 1.000 & 0.000 & 0.616\\
\bottomrule
\end{tabular}
\end{table}

\paragraph{Stricter localization.}
\Cref{tab:app-iou75} repeats the grounding comparison at $\mathrm{IoU}\ge0.75$.
Proposal-based selectors keep $0.88$--$0.93$ of their hits at the stricter
threshold on the standard splits, the VLM's raw box $0.87$--$0.89$, so the gap
to coordinate decoding narrows and on RefCOCOg the relation-aware head and the
token-state probe exceed the raw box; snapping that box to the proposals
($0.764$ on RefCOCOg val) recovers most of the difference.

\begin{table*}[t]
\centering
\footnotesize
\setlength{\tabcolsep}{4pt}
\caption{Accuracy at $\mathrm{IoU}\ge0.75$ on present queries, no abstention
(trained rows: mean$\pm$sample std over three seeds; rows without
task-specific training: one run). Last column: mean IoU of the returned box on RefCOCOg test (misses
included). --: boxes not stored for that run.}
\label{tab:app-iou75}
\begin{tabular}{@{}lccccccc@{}}
\toprule
Method & RefCOCO val & RefCOCO+ val & RefCOCOg val & RefCOCOg test & Ref-L4 val & Stress & mIoU test\\
\midrule
Learned bridge (CLIP query) & $0.460{\pm}0.017$ & $0.386{\pm}0.009$ & $0.478{\pm}0.008$ & $0.495{\pm}0.011$ & $0.316{\pm}0.008$ & $0.208{\pm}0.013$ & $0.544$\\
Index scorer & $0.699{\pm}0.002$ & $0.627{\pm}0.007$ & $0.692{\pm}0.004$ & $0.696{\pm}0.009$ & $0.390{\pm}0.007$ & $0.328{\pm}0.021$ & $0.735$\\
Alignment hook & $0.728{\pm}0.009$ & $0.655{\pm}0.014$ & $0.706{\pm}0.006$ & $0.720{\pm}0.009$ & $0.410{\pm}0.011$ & $0.323{\pm}0.039$ & $0.753$\\
Relation-aware head & $0.781{\pm}0.002$ & $0.712{\pm}0.005$ & $0.768{\pm}0.003$ & $0.779{\pm}0.002$ & $0.462{\pm}0.008$ & $0.320{\pm}0.011$ & $0.801$\\
Pooled query, cross-attention & -- & -- & $0.777{\pm}0.001$ & $0.785{\pm}0.002$ & $0.441{\pm}0.007$ & $0.271{\pm}0.002$ & $0.806$\\
Token states, cross-attention & -- & -- & $0.763{\pm}0.006$ & $0.769{\pm}0.003$ & $0.526{\pm}0.005$ & $0.328{\pm}0.028$ & $0.801$\\
\midrule
Detector-native & $0.792$ & $0.685$ & $0.715$ & $0.718$ & $0.567$ & $0.203$ & $0.745$\\
Coordinate-snap & -- & -- & $0.764$ & $0.770$ & -- & $0.532$ & $0.812$\\
Coordinate decoding (raw box) & $0.791$ & $0.742$ & $0.749$ & $0.757$ & $0.724$ & $0.468$ & $0.795$\\
\bottomrule
\end{tabular}
\end{table*}

\paragraph{Answer rate of the veto on positive-only splits.}
At the fixed thresholds the hook answers $0.818{\pm}0.054$ of RefCOCO val,
$0.758{\pm}0.045$ of RefCOCO+ val, $0.850{\pm}0.035$ of RefCOCOg val and
$0.856{\pm}0.028$ of RefCOCOg test queries, with selective accuracy $0.842$,
$0.778$, $0.824$ and $0.831$; the hook trained without the verifier, gated on
affinity only, answers $0.946$, $0.920$, $0.947$ and $0.951$. On Ref-L4 and the
stress set the answer rate falls to $0.337$ and $0.320$.

\paragraph{Coordinate-snap and detector-native on RefCOCO+ and Ref-L4.}
Coordinate-snap and detector-native are deterministic interfaces of the frozen
models and receive no task-specific training here. On every split of \cref{tab:app-grounding24k} they are computed on the
same rows and proposals as every other method; coordinate-snap is the VLM's box
snapped to the proposal of maximal IoU, detector-native the top-1 box of the
detector prompted with the expression (protocol in \cref{sec:app-detnative}). An
earlier, separate evaluation run of the same models gave raw-box accuracies of
$0.838$ (coordinate decoding) and $0.739$ (detector-native) on RefCOCO+ val and
$0.662$ (detector-native) on Ref-L4. The same-build values are $0.838$ for the
raw and $0.840$ for the snapped coordinate box and $0.737$ for detector-native
on RefCOCO+ val, and $0.634$ for detector-native on Ref-L4. On Ref-L4 the two
detector runs agree on $94.8\%$ of the queries, and almost all of the difference lies on
the COCO images ($0.649$ against $0.691$; $0.609$ against $0.610$ on the
Objects365 images). Every Ref-L4 detector-native number in this supplement
comes from the same-build run.

\paragraph{Standard test splits and Ref-L4 test.}
\label{sec:app-std-test}
\Cref{tab:app-eightsplit} completes the standard eight-split protocol with the
RefCOCO and RefCOCO+ testA (people) and testB (other objects) splits; RefCOCOg
uses the umd partition throughout. The test splits are built with the same
detector prompt, proposal budget and query readout as the other splits; the
first rows of RefCOCO val rebuilt this way match the original build exactly.
The hooks are the published checkpoints (they reproduce every published
per-query hit on RefCOCOg val and Ref-L4 val); the relation-aware heads were
retrained with the published recipe and seeds, reaching $0.848$ on RefCOCOg
val against the published $0.846$. The val ordering holds on all four test
splits. The hook stays below detector-native ($0.015$--$0.060$), the
relation-aware head is level with it on RefCOCO and above it on RefCOCO+, and
coordinate decoding leads every $24$k head and the full-set hook: by $0.10$--$0.13$ for the hook,
$0.05$--$0.06$ for the relation-aware head and $0.05$--$0.08$ for the hook
trained on all of RefCOCOg. The full-set relation-aware head trails it by
$0.030$ and $0.036$ on testA and is level on testB, and the combined selector is
level on testA and ahead on testB ($+0.013$ [$0.002$, $0.023$] and $+0.015$
[$0.003$, $0.026$]; \cref{sec:app-allsplits}). Every method is more accurate on people than on
other objects, by $0.05$--$0.08$ on RefCOCO and $0.11$--$0.18$ on RefCOCO+, but
the hook's distance to coordinate decoding is the same on both ($0.10$ on
RefCOCO, $0.12$--$0.13$ on RefCOCO+), so the selection deficit is not specific
to people. Averaged over $\mathrm{IoU}$ thresholds $0.50{:}0.05{:}0.95$ (mAcc),
the snapped coordinate box is best on all four test splits ($0.815$, $0.709$,
$0.783$, $0.628$), while the raw box falls below detector-native on RefCOCO
testA and testB ($0.778$ and $0.675$ against $0.795$ and $0.705$), as in the
stricter-localization comparison above.

\begin{table*}[t]
\centering
\scriptsize
\setlength{\tabcolsep}{3pt}
\caption{The standard eight-split protocol (Qwen2.5-VL): accuracy at
$\mathrm{IoU}\ge0.5$ on present queries, no abstention. Trained rows:
mean$\pm$sample std over three seeds (relation-aware head on testA/testB:
heads retrained with the published recipe and seeds; full-set relation-aware head on testA/testB and combined selector on RefCOCO/+: the evaluation of \cref{sec:app-allsplits}); rows without task-specific
training: one run. Val and RefCOCOg columns repeat \cref{tab:app-grounding24k}
($24$k rows, coordinate-snap) and Table 1 of the main paper (full-set rows,
detector-native, coordinate decoding); $95\%$ query
bootstrap half-widths on testA/testB are $0.007$--$0.013$.}
\label{tab:app-eightsplit}
\begin{tabular}{@{}lcccccccc@{}}
\toprule
& \multicolumn{3}{c}{RefCOCO} & \multicolumn{3}{c}{RefCOCO+} & \multicolumn{2}{c}{RefCOCOg (umd)}\\
\cmidrule(lr){2-4}\cmidrule(lr){5-7}\cmidrule(l){8-9}
Method & val & testA & testB & val & testA & testB & val & test\\
\midrule
Proposal recall (ceiling) & 0.999 & 0.999 & 0.998 & 0.999 & 0.999 & 0.999 & 1.000 & 0.998\\
\midrule
Alignment hook, $24$k & 0.789\sd{.007} & 0.827\sd{.006} & 0.755\sd{.011} & 0.707\sd{.010} & 0.770\sd{.007} & 0.625\sd{.009} & 0.787\sd{.004} & 0.793\sd{.007}\\
Relation-aware head, $24$k & 0.849\sd{.003} & 0.881\sd{.002} & 0.803\sd{.005} & 0.772\sd{.003} & 0.832\sd{.006} & 0.696\sd{.003} & 0.846\sd{.002} & 0.849\sd{.001}\\
Alignment hook, full set & 0.834\sd{.006} & 0.857\sd{.006} & 0.798\sd{.008} & 0.760\sd{.008} & 0.814\sd{.008} & 0.693\sd{.007} & 0.843\sd{.004} & 0.837\sd{.005}\\
Relation-aware head, full set & 0.872\sd{.003} & 0.898\sd{.002} & 0.846\sd{.003} & 0.802\sd{.006} & 0.855\sd{.001} & 0.748\sd{.003} & 0.877\sd{.002} & 0.878\sd{.003}\\
Combined selector, full set & 0.895\sd{.003} & 0.921\sd{.001} & 0.864\sd{.004} & 0.824\sd{.002} & 0.884\sd{.001} & 0.772\sd{.001} & 0.876\sd{.001} & 0.885\sd{.000}\\
\midrule
Detector-native & 0.847 & 0.887 & 0.806 & 0.737 & 0.822 & 0.640 & 0.783 & 0.781\\
Coordinate decoding (raw box) & 0.893 & \textbf{0.928} & 0.851 & 0.838 & \textbf{0.892} & 0.758 & 0.865 & 0.866\\
Coordinate-snap & \textbf{0.895} & \textbf{0.928} & \textbf{0.854} & \textbf{0.840} & 0.891 & \textbf{0.761} & \textbf{0.867} & \textbf{0.866}\\
\bottomrule
\end{tabular}
\end{table*}

\Cref{tab:app-refl4test} reports Ref-L4 test with its standard metrics. For
compute reasons it uses a fixed subset of $6{,}000$ of the $31{,}921$ official
test queries, drawn uniformly at random with a fixed seed within the two
vocabulary groups of the official annotation (COCO-80 category or other), in
proportion to their shares ($3{,}738$ and $2{,}262$ queries). The subset was
fixed before any method was run on it. Every method stays within $0.011$ of its
Ref-L4 val accuracy, so the val findings carry over to test: coordinate-snap
leads mAcc ($0.706$ against $0.468$ for the best trained head), and outside the
COCO categories every trained head stays near $0.20$ mAcc while
detector-native keeps $0.51$.

\begin{table*}[t]
\centering
\footnotesize
\setlength{\tabcolsep}{4pt}
\caption{Ref-L4 test, fixed stratified subset of $6{,}000$ queries
(Qwen2.5-VL): accuracy at $\mathrm{IoU}\ge0.5/0.75/0.9$ and mAcc (mean accuracy
over $\mathrm{IoU}$ $0.50{:}0.05{:}0.95$), overall and by whether the referent's
category is a COCO-80 class; last two columns: Ref-L4 val ($n{=}13{,}420$) for
comparison. Trained rows: mean over three seeds (sample std $\le0.012$); $95\%$
query bootstrap half-width of mAcc about $0.01$. --: not evaluated.}
\label{tab:app-refl4test}
\begin{tabular}{@{}lcccccc|cc@{}}
\toprule
& \multicolumn{6}{c|}{Ref-L4 test (subset, $n{=}6{,}000$)} & \multicolumn{2}{c}{Ref-L4 val}\\
Method & Acc@0.5 & Acc@0.75 & Acc@0.9 & mAcc & mAcc COCO cat. & mAcc other & Acc@0.5 & mAcc\\
\midrule
Proposal recall (ceiling) & 0.954 & 0.918 & 0.747 & 0.861 & 0.926 & 0.754 & 0.958 & 0.864\\
\midrule
Alignment hook, $24$k & 0.455 & 0.402 & 0.330 & 0.389 & 0.505 & 0.198 & 0.464 & 0.396\\
Relation-aware head, $24$k & 0.517 & 0.458 & 0.377 & 0.444 & 0.587 & 0.206 & 0.522 & 0.446\\
Alignment hook, full set & 0.542 & 0.483 & 0.401 & 0.468 & 0.623 & 0.213 & 0.549 & 0.473\\
Relation-aware head, full set & -- & -- & -- & -- & -- & -- & 0.595 & 0.516\\
\midrule
Detector-native & 0.644 & 0.575 & 0.450 & 0.546 & 0.566 & 0.511 & 0.634 & 0.537\\
Coordinate decoding (raw box) & \textbf{0.864} & 0.729 & 0.480 & 0.685 & 0.741 & \textbf{0.592} & \textbf{0.853} & 0.679\\
Coordinate-snap & 0.839 & \textbf{0.733} & \textbf{0.581} & \textbf{0.706} & \textbf{0.777} & 0.588 & 0.833 & \textbf{0.701}\\
\bottomrule
\end{tabular}
\end{table*}

\paragraph{Ref-L4 decomposition.}
Table~\ref{tab:app-refl4buckets} extends the main paper's vocabulary split of
Ref-L4 val with expression-length terciles, relative-box-area terciles and the
caption-origin flag of the benchmark (raw versus rewritten). The vocabulary
boundary dominates: the out-of-vocabulary bucket concentrates the collapse of
every trained interface ($0.24$--$0.27$) and also drives the apparent length
and size effects, since out-of-vocabulary referents are smaller and carry
shorter captions on average. Rewriting the caption changes the trained
heads by at most $0.03$. The expression-prompted detector also loses outside
the vocabulary and on small boxes ($0.649$ against $0.609$; $0.553$ on the
smallest tercile against $0.698$ on the largest), but far less than any trained
head. Gap shares are computed as the bucket's contribution $w_b\,\Delta_b$ to
the overall difference: the out-of-vocabulary bucket carries $79\%$ of the
detector-native$-$hook difference ($0.170$) and $50\%$ of the coordinate
decoding$-$hook difference ($0.389$), with $37\%$ of the queries. Every
referent the trained heads saw comes from RefCOCOg train and belongs to a COCO
category, so the prompt's vocabulary and the training vocabulary coincide; the
next paragraph separates vocabulary from image source and box size.

\begin{table}[t]
\centering
\footnotesize
\setlength{\tabcolsep}{1.9pt}
\caption{Ref-L4 val by bucket (selection accuracy at $\mathrm{IoU}\ge0.5$, no
abstention; trained rows are three-seed means). Cov.: proposal recall. Hook /
Rel. / Token: alignment hook, relation-aware head with IoU-rank loss,
token-state cross-attention. Det.\ / Coord.: detector-native and coordinate
decoding (the VLM's raw box).}
\label{tab:app-refl4buckets}
\begin{tabular}{@{}lrcccccc@{}}
\toprule
Bucket & $n$ & Cov. & Hook & Rel. & Token & Det. & Coord.\\
\midrule
COCO-80 category & 8483 & 0.999 & 0.592 & 0.680 & 0.784 & 0.649 & 0.900\\
Other category & 4937 & 0.886 & 0.243 & 0.249 & 0.266 & 0.609 & 0.774\\
\midrule
Caption $\le20$ words & 4726 & 0.918 & 0.367 & 0.404 & 0.439 & 0.678 & 0.813\\
Caption $21$--$27$ words & 4706 & 0.970 & 0.505 & 0.567 & 0.642 & 0.643 & 0.860\\
Caption $>27$ words & 3988 & 0.991 & 0.530 & 0.607 & 0.718 & 0.572 & 0.892\\
\midrule
Small box (lower tercile) & 4474 & 0.900 & 0.237 & 0.220 & 0.236 & 0.553 & 0.754\\
Medium box & 4473 & 0.981 & 0.559 & 0.693 & 0.755 & 0.652 & 0.875\\
Large box (upper tercile) & 4473 & 0.992 & 0.595 & 0.651 & 0.788 & 0.698 & 0.931\\
\midrule
Raw caption & 6769 & 0.957 & 0.479 & 0.537 & 0.598 & 0.662 & 0.861\\
Rewritten caption & 6651 & 0.958 & 0.448 & 0.506 & 0.588 & 0.606 & 0.845\\
\bottomrule
\end{tabular}
\end{table}

\paragraph{Vocabulary, image source and box size on Ref-L4.}
\label{sec:app-refl4-regression}
In Ref-L4 val the vocabulary split coincides with the image source: all
$8{,}483$ queries with a COCO-category referent are on COCO images and all
$4{,}937$ others on Objects365 images, whose referents are also small
($71$--$78\%$ fall in the smallest area tercile, against $9\%$ on COCO images).
\Cref{tab:app-refl4source} separates the factors. Within Objects365 images,
referents whose category maps to a COCO class (by a strict name mapping,
$1{,}664$ queries) are selected about twice as often by the trained heads as
the remaining $3{,}273$; the token-state probe's drop exceeds that of coordinate
decoding by $0.147$ [$0.110$, $0.184$]. In a fractional-logit regression of the
seed-averaged per-query hit on vocabulary, image source and box size (covered
queries, image-clustered intervals; \cref{tab:app-style}, bottom), the
vocabulary term stays large for the trained heads under every size
specification ($+0.84$ to $+0.94$ log-odds) and small for the references
without selection training ($+0.24$ to $+0.26$). The image-source term depends
on how size enters: with the standardized log relative area and its square it
is near zero for the trained heads ($-0.11$ for the token probe and $+0.01$ for
the hook, intervals including zero; $+0.16$ for the relation-aware head), with
the linear term alone or with size terciles it remains ($-0.27$ to $-0.55$ and
$-0.35$ to $-0.73$). Objects365 referents are mostly small, so image source and
box size are only weakly separable, while a category effect remains within one
image source under every specification. A class-agnostic proposer
(\cref{sec:app-upn}) and a caption-style control (\cref{sec:app-style}) show
that this effect is set by the selectors' training data, not by the proposal
prompt or the caption style. An image-overlap audit finds $746$ Ref-L4 COCO images ($1{,}341$
queries) among the hook's training images; the hook gains no more on them than
coordinate decoding does (difference in differences $+0.010$, interval
including zero). Objects365 images could not be audited against the detector's
training data.

\begin{table}[t]
\centering
\footnotesize
\setlength{\tabcolsep}{3pt}
\caption{Ref-L4 val by image source and referent vocabulary (selection
accuracy at $\mathrm{IoU}\ge0.5$, no abstention; trained rows are three-seed
means). COCO-mappable: an Objects365 category that maps to a COCO-80 class.}
\label{tab:app-refl4source}
\begin{tabular}{@{}llrcccc@{}}
\toprule
Images & Referent & $n$ & Hook & Token & Det. & Coord.\\
\midrule
COCO & COCO-80 & 8483 & 0.592 & 0.784 & 0.649 & 0.900\\
Objects365 & COCO-mappable & 1664 & 0.384 & 0.410 & 0.671 & 0.821\\
Objects365 & other & 3273 & 0.171 & 0.192 & 0.578 & 0.749\\
\bottomrule
\end{tabular}
\end{table}

\paragraph{Released contemporaneous models on our rows.}
\label{sec:app-e54}
\Cref{tab:app-e54} scores the released weights of Jag \cite{jag} and LOCUS
\cite{locus}, unchanged and not retrained, on the present rows of every column
of Table~1, with the same metric (a missing box counts as wrong) and
image-clustered bootstrap intervals. We used the authors' code at recorded
commits (Jag 57dcfa2, weights 8db6a87, PyTorch 2.10.0, Transformers 5.9.0; LOCUS
e8687aa, PyTorch 2.15.0, Transformers 5.22.1): Jag's predictor at its defaults, and
LOCUS's grounding call with exact verification, which also records the head's
one-pass draft. We disabled LOCUS's abstention so that every row gets a box, as in
Table~1. v1.0 never answers that the target is absent; v1.1 does so on at most
$5.5\%$ of these present rows, and counting those as wrong changes any verified
cell by at most $0.009$. FORUM \cite{forum}, which fuses several large frozen VLMs,
was not run.

LOCUS has two releases. Its paper describes v1.0 (weights 22ffbb1 for 2B, 2c6e041
for 0.8B), trained on the RefCOCO family; v1.1 (f393d11, 13cd584) adds no-target,
category-contrast, small-target and noun-swap rows, augmentation and $12$k steps.
Our first run loaded the hub's default, v1.1, and we noticed only because LOCUS-2B
fell outside the range we had predicted for it. We then ran v1.0 under the same
predictions, written before its run but with the v1.1 numbers known. v1.0 decides
the size statement in the main paper, which concerns VLMs fine-tuned on the RefCOCO
family, and v1.1 is reported beside it. The runs match the authors' numbers. On
Ref-L4 val targets under $1\%$ of the image we obtain $1.5$, $0.5$, $23.5$ and $12.2$
against their $1.4$, $0.5$, $22.8$ and $12.3$ (2B then 0.8B, v1.0 then v1.1). On
Objects365 images LOCUS-2B v1.0 scores $0.068$ and $0.848$ in the smallest and largest
size quarter, against their $0.062$ and $0.848$ on Ref-L4 test. Their figures for
targets under $2\%$ ($16.7$, $10.0$, $40.0$, $27.7$) come from a random $600$ of the
$2{,}127$ rows; on all of them we obtain $14.8$, $8.7$, $39.9$ and $27.5$.

On COCO images Jag and LOCUS v1.0 reach $0.897$--$0.917$, against $0.900$ and $0.918$
for decoding; $61\%$ of these rows are on images of the train union described below,
and outside it they reach $0.885$--$0.906$. On Objects365 images they reach $0.361$
[$0.342$, $0.380$] (Jag), $0.426$ [$0.407$, $0.445$] and $0.483$ [$0.464$, $0.502$]
(LOCUS-0.8B and 2B, verified): inside the ranges we predicted for Jag and LOCUS-2B,
and at least $0.29$ below decoding ($0.774$, $0.815$). They are above our combined
selectors on all Ref-L4 columns and on RB-PRO. On Ref-Adv-s, where we ran Qwen2.5-VL
heads only, LOCUS-2B v1.0 is clearly above ($0.375$ against $0.314$), while the gaps of
Jag and LOCUS-0.8B v1.0 ($0.038$, $0.035$) are about the half-width. The stress set
does not resolve the order (half-width $0.066$): Jag $0.338$, LOCUS v1.0 $0.385$ and
$0.442$, our combined selectors $0.381$ and $0.439$.

On the RefCOCO family we had planned to rank on images outside the train union
only. After the first run we found that LOCUS removes every image of these splits
from training, so all its cells there are held out like our all-three-sets row
($0.912$, $0.846$, $0.877$, $0.889$), and we read them on all rows. LOCUS-2B v1.0 is
above that row on RefCOCO and RefCOCO+ val ($0.923$, $0.874$), level with it on RefCOCOg
($0.881$, $0.893$; half-widths at most $0.013$), and below our Qwen3-VL-8B selector on
RefCOCOg ($0.898$, $0.895$). LOCUS-0.8B v1.0 is at or below the row. Outside the
union, which is a different set of rows, LOCUS-2B v1.0 is below the row on RefCOCO val
($0.906$). Base
models differ, so these gaps mix method and model size: Jag and LOCUS use $0.8$B
and $2$B VLMs, our rows $7$B and $8$B. The LOCUS paper reports $0.745$ on Objects365
images of Ref-L4 test for the untrained $2$B base of LOCUS-2B, against $0.478$ for
LOCUS-2B and $0.358$ for Jag; we did not run that base on our rows.

\Cref{tab:app-e54-size} refits the regression of \cref{tab:app-style} (bottom) on
the same covered rows. We fixed its reading before any run: (a) vocabulary remains
if the category term is at least $+0.50$ with an interval excluding zero under both
size specifications; (b) size accounts for the drop if under both the interval
includes zero or the term is at most $+0.26$, the top of our references without
selection training; (c) mixed otherwise. We predicted (a) for Jag and the LOCUS
drafts, and (a) or (c) for verified LOCUS. On v1.0 this held for verified LOCUS and
the 0.8B draft and missed for Jag and the 2B draft, which read (c). No model reads
(b). The large-target term ($+1.79$ to $+2.38$) exceeds the category term in every
tercile fit, in line with that paper's attribution to size. A category term remains
with size controlled: $+0.33$ to $+0.53$ on the Jag and verified v1.0 outputs, $+0.24$ to
$+0.64$ with the drafts, against $+0.84$ to $+0.94$ for our trained heads. The LOCUS-2B draft
is near the level of our references, and under continuous size its interval barely
excludes zero (lower bound $+0.0003$). In a paired comparison on the same rows with
images as clusters, which we added after these fits and ran for the hook only, the
hook's term exceeds each of these five under both specifications; the smallest gap
is $+0.243$ [$+0.010$, $+0.477$] (LOCUS-0.8B draft, continuous size). v1.1 raises
Objects365 accuracy by $0.09$ and $0.12$ at a cost of at most $0.010$ on the RefCOCO
family. Its large-target term falls ($+1.79$ to $+1.43$ for verified 2B), its category term
is larger, $+0.52$ to $+0.78$, reading (a) throughout, and the paired gap to the hook
stays above zero under terciles but includes zero under continuous size for the 0.8B
output and draft and the 2B draft.

Jag was trained on the raw union of the RefCOCO, RefCOCO+ and RefCOCOg train
splits, whose images recur in $57$--$59\%$ of the RefCOCO-family rows and $38\%$ of the
Ref-L4 rows, so its all-row cells there are not held out. \Cref{tab:app-e54} also
gives every model on images outside that union, but these cells and the
all-three-sets row cover different rows. We predicted seen above unseen images.
That holds on RefCOCO and RefCOCO+ val (Jag by $0.046$ and $0.051$), yet LOCUS v1.0
shows $0.022$--$0.030$ there without having trained on these images, so part of the gap
is row composition. The prediction fails on RefCOCOg test for every model (seen lower
by $0.01$ to $0.02$) and on RefCOCOg val for both LOCUS-2B releases (lower by $0.014$ and
$0.006$); we computed no intervals for these differences. We did not measure the overlap
of Ref-L4's COCO images with LOCUS's training images.

\begin{table*}[t]
\centering
\footnotesize
\setlength{\tabcolsep}{4pt}
\caption{Released contemporaneous models on the present rows of Table~1
(accuracy at $\mathrm{IoU}\ge0.5$, no retraining). Verified: the
LOCUS output; draft: its head's one-pass box. Reference rows are repeated from
Table~1. Image-clustered $95\%$ half-widths of the all-row Jag and verified cells are at
most $0.013$ on the RefCOCO family, $0.011$ on Ref-L4 all, $0.008$ on COCO, $0.019$ on
other, $0.066$ on Stress, $0.036$ on Ref-Adv-s and $0.017$ on RB-PRO; outside the train
union at most $0.020$ on the RefCOCO family, $0.015$ on Ref-L4 all and $0.013$ on COCO. Images
outside the train union: rows whose image is not in the RefCOCO, RefCOCO+ and
RefCOCOg train splits, the union Jag trained on. No Stress row and none of the $259$
Ref-Adv-s rows with a COCO image is in that union; RB-PRO rows carry no COCO id, so
its overlap is not measured.}
\label{tab:app-e54}
\begin{tabular}{@{}lcccccccccc@{}}
\toprule
& RefCOCO & RefCOCO+ & \multicolumn{2}{c}{RefCOCOg} & \multicolumn{3}{c}{Ref-L4 val} & & & \\
\cmidrule(lr){4-5}\cmidrule(lr){6-8}
& val & val & val & test & all & COCO cat. & other & Stress & Ref-Adv-s & RB-PRO\\
\midrule
\multicolumn{11}{@{}l}{\emph{References}}\\
Qwen2.5-VL-7B, coordinate decoding & 0.893 & 0.838 & 0.865 & 0.866 & 0.853 & 0.900 & 0.774 & 0.623 & 0.520 & 0.662\\
Qwen3-VL-8B, coordinate decoding & 0.918 & 0.868 & 0.886 & 0.890 & 0.880 & 0.918 & 0.815 & 0.671 & -- & --\\
Qwen2.5-VL-7B, combined selector & 0.895 & 0.824 & 0.876 & 0.885 & 0.641 & 0.852 & 0.280 & 0.381 & 0.314 & 0.313\\
\quad all three train sets & 0.912 & 0.846 & 0.877 & 0.889 & 0.640 & 0.853 & 0.274 & 0.374 & -- & --\\
Qwen3-VL-8B, combined selector & 0.910 & 0.855 & 0.898 & 0.895 & 0.637 & 0.861 & 0.252 & 0.439 & -- & --\\
\midrule
\multicolumn{11}{@{}l}{\emph{All rows}}\\
Jag & 0.912 & 0.856 & 0.872 & 0.876 & 0.707 & 0.909 & 0.361 & 0.338 & 0.352 & 0.386\\
LOCUS-0.8B v1.0, verified & 0.901 & 0.846 & 0.865 & 0.864 & 0.724 & 0.897 & 0.426 & 0.385 & 0.349 & 0.412\\
LOCUS-0.8B v1.0, draft & 0.910 & 0.856 & 0.869 & 0.869 & 0.719 & 0.907 & 0.397 & 0.377 & 0.352 & 0.404\\
LOCUS-2B v1.0, verified & 0.923 & 0.874 & 0.881 & 0.893 & 0.758 & 0.917 & 0.483 & 0.442 & 0.375 & 0.428\\
LOCUS-2B v1.0, draft & 0.931 & 0.879 & 0.886 & 0.899 & 0.734 & 0.925 & 0.406 & 0.433 & 0.350 & 0.401\\
LOCUS-0.8B v1.1, verified & 0.896 & 0.836 & 0.855 & 0.860 & 0.755 & 0.892 & 0.519 & 0.424 & 0.398 & 0.443\\
LOCUS-0.8B v1.1, draft & 0.907 & 0.847 & 0.863 & 0.869 & 0.744 & 0.902 & 0.473 & 0.398 & 0.356 & 0.418\\
LOCUS-2B v1.1, verified & 0.918 & 0.869 & 0.875 & 0.886 & 0.797 & 0.913 & 0.599 & 0.519 & 0.438 & 0.492\\
LOCUS-2B v1.1, draft & 0.927 & 0.872 & 0.887 & 0.890 & 0.783 & 0.920 & 0.547 & 0.519 & 0.418 & 0.457\\
\midrule
\multicolumn{11}{@{}l}{\emph{Images outside the train union (LOCUS: verified)}}\\
Jag & 0.886 & 0.826 & 0.854 & 0.882 & 0.574 & 0.891 & 0.361 & -- & -- & --\\
LOCUS-0.8B v1.0 & 0.888 & 0.828 & 0.856 & 0.878 & 0.611 & 0.885 & 0.426 & -- & -- & --\\
LOCUS-2B v1.0 & 0.906 & 0.857 & 0.889 & 0.899 & 0.653 & 0.906 & 0.483 & -- & -- & --\\
LOCUS-0.8B v1.1 & 0.880 & 0.815 & 0.850 & 0.870 & 0.663 & 0.878 & 0.519 & -- & -- & --\\
LOCUS-2B v1.1 & 0.898 & 0.853 & 0.878 & 0.896 & 0.720 & 0.901 & 0.599 & -- & -- & --\\
\bottomrule
\end{tabular}
\end{table*}

\begin{table*}[t]
\centering
\footnotesize
\caption{Category term against size on Ref-L4 val for the released models: the
regression of \cref{tab:app-style} (bottom), strict name mapping, covered rows
($n=12{,}854$), image-clustered $95\%$ intervals. Size: terciles, or the standardized log relative area and its square. Large
target: tercile term. Reading: the rule fixed before the runs (see text). Our rows
are from \cref{tab:app-style}, without selection training for the last two.}
\label{tab:app-e54-size}
\begin{tabular}{@{}lcccc@{}}
\toprule
& Category term, size terciles & Category term, continuous size & Large target & Reading\\
\midrule
Jag & $+0.450$ [$+0.276$, $+0.625$] & $+0.501$ [$+0.263$, $+0.739$] & $+2.38$ & (c)\\
LOCUS-0.8B v1.0, verified & $+0.454$ [$+0.290$, $+0.619$] & $+0.533$ [$+0.326$, $+0.740$] & $+1.90$ & (c)\\
LOCUS-0.8B v1.0, draft & $+0.511$ [$+0.345$, $+0.677$] & $+0.642$ [$+0.422$, $+0.862$] & $+2.26$ & (a)\\
LOCUS-2B v1.0, verified & $+0.335$ [$+0.171$, $+0.499$] & $+0.374$ [$+0.169$, $+0.579$] & $+1.79$ & (c)\\
LOCUS-2B v1.0, draft & $+0.294$ [$+0.123$, $+0.464$] & $+0.242$ [$+0.000$, $+0.483$] & $+2.32$ & (c)\\
LOCUS-0.8B v1.1, verified & $+0.662$ [$+0.501$, $+0.823$] & $+0.775$ [$+0.592$, $+0.958$] & $+1.59$ & (a)\\
LOCUS-0.8B v1.1, draft & $+0.634$ [$+0.470$, $+0.797$] & $+0.781$ [$+0.585$, $+0.977$] & $+2.15$ & (a)\\
LOCUS-2B v1.1, verified & $+0.519$ [$+0.354$, $+0.684$] & $+0.586$ [$+0.407$, $+0.764$] & $+1.43$ & (a)\\
LOCUS-2B v1.1, draft & $+0.584$ [$+0.423$, $+0.746$] & $+0.693$ [$+0.511$, $+0.875$] & $+1.94$ & (a)\\
\midrule
Ours: alignment hook & $+0.864$ & $+0.885$ & & \\
Ours: relation-aware head & $+0.875$ & $+0.941$ & & \\
Ours: token-state probe & $+0.841$ & $+0.930$ & & \\
Ours: detector-native & $+0.242$ & $+0.246$ & & \\
Ours: coordinate decoding & $+0.257$ & $+0.256$ & & \\
\bottomrule
\end{tabular}
\end{table*}

\paragraph{Extended proposal vocabulary on Ref-L4.}
Table~\ref{tab:app-extvocab} tests whether the out-of-vocabulary collapse is a
proposal failure. Only the detector is re-run on the Ref-L4 images: the
COCO-80 prompt is kept and the remaining Objects365 category names are added
as further prompts of at most $80$ names each; the detections of all prompts
are pooled and the $300$ highest-scoring boxes kept. The three hooks are those of the margin analysis below (retrained on
the earlier feature build (\cref{sec:app-builds})) and, like the query embeddings, are unchanged, and both proposal sets are scored on
the same $9{,}388$ present queries used in Table~\ref{tab:app-margin}. The
margin analysis below predicts the outcome, and we recorded that prediction
before the run: since the referent is already as separable outside the
vocabulary as inside it, proposal recall should rise but accuracy by no more than
the recall gain. That is what happens. Proposal recall of the out-of-vocabulary
bucket rises from $0.883$ to $0.957$; accuracy rises by $0.040$ (per seed
$+0.045$, $+0.036$, $+0.040$), mostly because more referents are covered, and
accuracy given a covering proposal moves only from $0.251$ to $0.273$. The bucket stays
far below both the in-vocabulary bucket ($0.62$) and detector-native
($0.609$ on the full bucket). Inside the vocabulary the extra prompts change nothing. One
caveat bounds this run: its hooks were trained on features fused with the
COCO-80 prompt. \Cref{tab:app-e31} retrains them on the extended-prompt features of the main build, which
lowers their accuracy there ($0.267$ to $0.241$).

\begin{table}[t]
\centering
\footnotesize
\setlength{\tabcolsep}{4pt}
\caption{Ref-L4 val with the fixed COCO-80 proposal prompt and with the
extended (COCO-80 plus Objects365) prompts; $9{,}388$ present queries, hook
accuracy as mean over three seeds. Acc$\mid$cov: accuracy among queries whose
referent is covered by a proposal.}
\label{tab:app-extvocab}
\begin{tabular}{@{}llrccc@{}}
\toprule
Referent & Prompt & $n$ & Coverage & Acc & Acc$\mid$cov\\
\midrule
COCO-80 & COCO-80 & 5863 & 0.999 & 0.621 & 0.622\\
 & extended & & 0.998 & 0.616 & 0.617\\
Other & COCO-80 & 3525 & 0.883 & 0.221 & 0.251\\
 & extended & & 0.957 & 0.261 & 0.273\\
\bottomrule
\end{tabular}
\end{table}

\paragraph{Prompt vocabulary versus training vocabulary.}
\label{sec:app-e31}
On Ref-L4 the prompt and training vocabularies coincide (both COCO-80), so the
category boundary could come from either. \Cref{tab:app-e31} separates them on
the main build. The extended prompts (COCO-80 plus all Objects365 names in
chunks, pooled, top $300$) are run over the training images, RefCOCOg val and
Ref-L4 val; hooks and relation-aware heads are trained on the $24$k rows with
fixed or extended features (three seeds). For the training vocabulary, Ref-L4's
Objects365 images are split into two image-disjoint halves ($2{,}492$ and
$2{,}445$ queries): a hook is trained on the $24$k rows plus the present rows of
one half and scored on the other half, and the two fold models are averaged on
the remaining rows. Extended prompts raise proposal recall on referents outside
the COCO categories from $0.834$ to $0.947$ and the published hook's accuracy
there by $0.10$, at a cost of $0.05$ on COCO-mappable Objects365 referents;
retraining on the extended features lowers it there ($0.267$ to $0.241$;
$+0.001$ overall). Adding non-COCO referents to the training data instead lifts the
other-category bucket from $0.17$ to $0.42$ with the original prompt and to
$0.50$ with both vocabularies, while COCO-image queries gain only
$0.02$--$0.04$ and RefCOCOg val moves by less than $0.01$: the boundary is set by
what the head was trained on, and the prompt matters only together with it.
The added rows also carry Ref-L4's caption style, but the same number of Ref-L4
rows from COCO images moves this bucket by only $-0.008$ [$-0.018$, $+0.002$]
(\cref{sec:app-style}); the gain comes from the Objects365 rows, partly through
their images (COCO-mappable referents gain $+0.159$) and partly through their
categories: the gain on other categories exceeds that on COCO-mappable ones by
$0.094$ [$0.065$, $0.125$] with the original prompt.

\begin{table*}[t]
\centering
\footnotesize
\setlength{\tabcolsep}{4pt}
\caption{Prompt vocabulary $\times$ training vocabulary on the main build
(Qwen2.5-VL, $24$k budget, three seeds; accuracy at $\mathrm{IoU}\ge0.5$).
``Ext.'': COCO-80 plus Objects365 prompts; ``+ Ref-L4 half'': training rows
extended by the present Ref-L4 queries on one image-disjoint half of its
Objects365 images, scored on the other half. O365: Ref-L4 queries on Objects365
images, split into referents whose category maps to a COCO class and the rest.
Hook rows with the COCO-80 prompt and $24$k rows: the published hooks;
relation-aware head with the COCO-80 prompt and $24$k rows: a retraining with
the published recipe and seeds (published run: $0.846$ on RefCOCOg val and
$0.522$ on Ref-L4 val).}
\label{tab:app-e31}
\begin{tabular}{@{}lllcccccc@{}}
\toprule
Head & Proposals & Training rows & RefCOCOg val & Ref-L4 val & COCO images & O365, COCO-mappable & O365, other\\
\midrule
\multicolumn{3}{@{}l}{Proposal recall, COCO-80 prompt} & 1.000 & 0.958 & 0.999 & 0.989 & 0.834\\
\multicolumn{3}{@{}l}{Proposal recall, extended prompts} & 0.998 & 0.984 & 0.998 & 0.984 & 0.947\\
\midrule
Hook & COCO-80 & $24$k & 0.787 & 0.464 & 0.592 & 0.384 & 0.171\\
Hook & ext. (not retrained) & $24$k & 0.776 & 0.470 & 0.576 & 0.331 & 0.267\\
Hook & ext. & $24$k, ext. features & 0.775 & 0.472 & 0.582 & 0.364 & 0.241\\
Hook & COCO-80 & $24$k + Ref-L4 half & 0.794 & 0.561 & 0.617 & 0.543 & 0.425\\
Hook & ext. & $24$k + Ref-L4 half & 0.778 & 0.586 & 0.625 & 0.544 & \textbf{0.504}\\
\midrule
Relation-aware head & COCO-80 & $24$k & 0.850 & 0.519 & 0.676 & 0.389 & 0.177\\
Relation-aware head & ext. (not retrained) & $24$k & 0.838 & 0.528 & 0.672 & 0.351 & 0.242\\
Relation-aware head & ext. & $24$k, ext. features & 0.842 & 0.529 & 0.672 & 0.366 & 0.243\\
\bottomrule
\end{tabular}
\end{table*}

\paragraph{A class-agnostic, prompt-free proposer.}
\label{sec:app-upn}
In every other Ref-L4 analysis the proposals and their region features come
from Grounding DINO prompted with the $80$ COCO names, whose decoder states are
fused with that prompt: the feature of a referent outside the COCO categories
is the detector's representation of the best-matching COCO phrase.
\Cref{tab:app-upn} removes the category prompt from perception. The universal
proposal network (UPN) released with ChatRex \cite{chatrex} is a class-agnostic
DETR-style proposer without text input; its queries are conditioned only on a
learned instance-level prompt embedding. Its boxes (input shorter side
$800$ pixels) are suppressed at $\mathrm{IoU}\,0.8$ in score order and the
$300$ highest-scoring ones are kept per image, the same number as for Grounding
DINO (at least $432$ boxes survive suppression on every image). The region
features come from one of two sources, neither of which has seen a category
prompt: DINOv2-L/14 \cite{dinov2}, a self-supervised backbone trained without text or
labels, whose final-layer patch tokens (input shorter side $644$ pixels) are
RoIAlign-pooled over $7{\times}7$ bins in each box and averaged ($1024$-d), or
UPN's own last decoder-layer query states ($256$-d).
The $24$k RefCOCOg training rows, RefCOCOg val and test, Ref-L4 val and the
stress set are re-detected; rows, ground-truth boxes and the frozen Qwen2.5-VL
queries and token states are unchanged, and the alignment hook, the
relation-aware head with IoU-rank loss, the token-state probe and the combined
selector (token states, two intermediate VLM depths, relations among proposals
and all-positive supervision with the IoU-rank loss) are retrained with the
recipes of \cref{tab:app-grounding24k} and Table 2 of the main paper, three
seeds each. The Grounding DINO rows are the
$24$k runs of \cref{tab:app-grounding24k} and of Table 2 of the main paper,
on the same rows; the combined selector, which the main paper trains only on
the full set, is trained at $24$k on all three feature sets.

The class-agnostic proposer covers more referents outside the COCO categories:
proposal recall on Ref-L4 ``other'' rises from $0.886$ to $0.966$ (from $0.834$
to $0.955$ on referents without a COCO counterpart), while RefCOCOg val and test, the RefCOCOg
training rows ($0.9953$ against $0.9985$) and the COCO-category referents lose
at most $0.003$, and the stress set loses $0.009$ ($0.991$ against $1.000$). Selection does not
follow. With UPN's query states every head is level with its Grounding DINO
counterpart on ``other'': $0.241$ against
$0.243$ for the hook, $0.261$ against $0.249$ for the relation-aware head and
$0.267$ against $0.266$ for the token-state probe (paired differences $-0.002$
[$-0.013$, $+0.009$], $+0.012$ [$+0.002$, $+0.022$] and $+0.001$ [$-0.008$,
$+0.010$]). Since selection accuracy is proposal recall times accuracy given a
covering proposal, the recall gain is canceled by a lower conditional accuracy
($0.276$ against $0.300$ for the token-state probe, $0.249$ against $0.274$ for
the hook): the referents that only UPN covers are selected less often than the
others, and on the referents both proposers cover the heads are level or
slightly worse ($-0.022$ to $-0.006$).

Within ``other'' the proposer only moves accuracy between the two Objects365
sub-buckets, $+0.040$ to $+0.056$ on referents without a COCO
counterpart, where its recall gain lies, and $-0.073$ to $-0.097$ on
COCO-mappable ones, whose category the COCO-80 prompt names; the gap
between COCO-category and other referents narrows by $0.02$--$0.04$ and stays at
$0.31$--$0.50$. On RefCOCOg the query states stay within $0.016$ of Grounding
DINO. DINOv2 features lose on every split and bucket ($-0.08$ to $-0.15$ on
RefCOCOg val, $-0.04$ to $-0.07$ on ``other''), so a region feature that never
saw a category label does not select better outside COCO than the prompt-fused
one. A proposer without a category prompt, with region features never fused
with one, thus leaves the category boundary where it was: the boundary is not
caused by the COCO-80 prompt of the proposer but lies in what the selectors
were trained on (\cref{sec:app-e31,sec:app-style}).

The combined selector, the head that shares most with VGent's design (token
states read at several depths, scored jointly over the proposal set), follows
the same pattern. Over UPN's query states it loses $0.016$ on RefCOCOg test
($0.845$ against $0.861$) and is level on Ref-L4 val ($0.599$ for both;
$0.000$ [$-0.006$, $+0.005$]); on ``other'' it gains $0.013$ [$0.003$, $0.023$],
again from $+0.050$ on referents without a COCO counterpart and $-0.060$ on
COCO-mappable ones, and with DINOv2 features it loses everywhere ($-0.081$ on
RefCOCOg test). Over the token-state probe it adds $0.008$--$0.009$ on RefCOCOg
test with either proposer and $0.014$ to $0.025$ on ``other''
(two-level intervals above zero), so the combination helps on both proposal
sets but leaves the category boundary where it was: $0.78$ against $0.29$ on
UPN, $0.79$ against $0.28$ on Grounding DINO.

\begin{table*}[t]
\centering
\footnotesize
\setlength{\tabcolsep}{3pt}
\caption{Class-agnostic, prompt-free proposals (Qwen2.5-VL, $24$k training
rows, three seeds; selection accuracy at $\mathrm{IoU}\ge0.5$, no abstention,
mean\,$\pm$\,sample std). Grounding DINO: COCO-80 prompt and prompt-fused
region features (the runs of \cref{tab:app-grounding24k}). UPN: the proposer
of ChatRex without text input, $300$ boxes per image, with DINOv2-L RoI
features or its own query states; both share one proposal set. Ref-L4 COCO
cat.\ / other: referents inside / outside the $80$ COCO categories (all of
the former are on COCO images, all of the latter on Objects365 images); O365
map.\ / O365 other: other referents whose Objects365 category does / does not
map to a COCO class by a strict name mapping. Combined selector: the main
paper's combined selector trained on the same $24$k rows (its Grounding DINO
row is a new run). $\Delta$: paired difference, UPN
minus Grounding DINO, of the seed-averaged per-query hits; ${}^\dagger$its
$95\%$ percentile-bootstrap interval over queries ($10{,}000$ resamples)
excludes zero (stress set: over queries, not templates).}
\label{tab:app-upn}
\begin{tabular}{@{}lcccccccc@{}}
\toprule
& \multicolumn{2}{c}{RefCOCOg} & \multicolumn{5}{c}{Ref-L4 val} & \\
\cmidrule(lr){2-3}\cmidrule(lr){4-8}
Proposals, region features & val & test & all & COCO cat. & other & O365 map. & O365 other & Stress\\
\midrule
\multicolumn{9}{@{}l}{\emph{Proposal recall}}\\
Grounding DINO, COCO-80 prompt & 1.000 & 0.998 & 0.958 & 0.999 & 0.886 & 0.989 & 0.834 & 1.000\\
UPN, no prompt & 0.997 & 0.996 & 0.986 & 0.998 & 0.966 & 0.986 & 0.955 & 0.991\\
\midrule
\multicolumn{9}{@{}l}{\emph{Alignment hook}}\\
Grounding DINO, fused features & 0.787\sd{.004} & 0.793\sd{.007} & 0.464\sd{.010} & 0.592\sd{.018} & 0.243\sd{.004} & 0.384\sd{.017} & 0.171\sd{.008} & 0.384\sd{.035}\\
UPN, DINOv2-L features & 0.640\sd{.003} & 0.639\sd{.003} & 0.393\sd{.015} & 0.522\sd{.023} & 0.172\sd{.003} & 0.270\sd{.006} & 0.121\sd{.005} & 0.241\sd{.018}\\
UPN, query states & 0.776\sd{.005} & 0.778\sd{.005} & 0.436\sd{.040} & 0.549\sd{.053} & 0.241\sd{.026} & 0.287\sd{.020} & 0.217\sd{.029} & 0.394\sd{.022}\\
\quad $\Delta$, DINOv2-L features & $-0.147^\dagger$ & $-0.154^\dagger$ & $-0.071^\dagger$ & $-0.070^\dagger$ & $-0.071^\dagger$ & $-0.113^\dagger$ & $-0.050^\dagger$ & $-0.143^\dagger$\\
\quad $\Delta$, query states & $-0.010^\dagger$ & $-0.015^\dagger$ & $-0.028^\dagger$ & $-0.044^\dagger$ & $-0.002$ & $-0.097^\dagger$ & $+0.046^\dagger$ & $+0.010$\\
\midrule
\multicolumn{9}{@{}l}{\emph{Relation-aware head, IoU-rank loss}}\\
Grounding DINO, fused features & 0.846\sd{.002} & 0.849\sd{.001} & 0.522\sd{.007} & 0.680\sd{.010} & 0.249\sd{.001} & 0.393\sd{.009} & 0.176\sd{.005} & 0.388\sd{.014}\\
UPN, DINOv2-L features & 0.729\sd{.003} & 0.735\sd{.003} & 0.461\sd{.002} & 0.608\sd{.004} & 0.208\sd{.001} & 0.315\sd{.002} & 0.153\sd{.003} & 0.291\sd{.015}\\
UPN, query states & 0.841\sd{.001} & 0.841\sd{.001} & 0.524\sd{.007} & 0.677\sd{.009} & 0.261\sd{.005} & 0.319\sd{.006} & 0.231\sd{.004} & 0.365\sd{.009}\\
\quad $\Delta$, DINOv2-L features & $-0.117^\dagger$ & $-0.114^\dagger$ & $-0.061^\dagger$ & $-0.072^\dagger$ & $-0.041^\dagger$ & $-0.078^\dagger$ & $-0.022^\dagger$ & $-0.097^\dagger$\\
\quad $\Delta$, query states & $-0.004$ & $-0.008^\dagger$ & $+0.003$ & $-0.003$ & $+0.012^\dagger$ & $-0.073^\dagger$ & $+0.056^\dagger$ & $-0.023$\\
\midrule
\multicolumn{9}{@{}l}{\emph{Token-state probe}}\\
Grounding DINO, fused features & 0.849\sd{.004} & 0.852\sd{.002} & 0.593\sd{.004} & 0.784\sd{.005} & 0.266\sd{.004} & 0.410\sd{.011} & 0.192\sd{.002} & 0.426\sd{.015}\\
UPN, DINOv2-L features & 0.765\sd{.003} & 0.766\sd{.001} & 0.537\sd{.005} & 0.719\sd{.005} & 0.223\sd{.004} & 0.335\sd{.006} & 0.167\sd{.005} & 0.322\sd{.009}\\
UPN, query states & 0.832\sd{.003} & 0.837\sd{.002} & 0.582\sd{.003} & 0.765\sd{.003} & 0.267\sd{.003} & 0.335\sd{.003} & 0.232\sd{.002} & 0.361\sd{.016}\\
\quad $\Delta$, DINOv2-L features & $-0.084^\dagger$ & $-0.086^\dagger$ & $-0.056^\dagger$ & $-0.064^\dagger$ & $-0.042^\dagger$ & $-0.076^\dagger$ & $-0.025^\dagger$ & $-0.104^\dagger$\\
\quad $\Delta$, query states & $-0.016^\dagger$ & $-0.015^\dagger$ & $-0.011^\dagger$ & $-0.018^\dagger$ & $+0.001$ & $-0.075^\dagger$ & $+0.040^\dagger$ & $-0.065^\dagger$\\
\midrule
\multicolumn{9}{@{}l}{\emph{Combined selector}}\\
Grounding DINO, fused features & 0.857\sd{.003} & 0.861\sd{.003} & 0.599\sd{.006} & 0.785\sd{.007} & 0.279\sd{.008} & 0.419\sd{.009} & 0.208\sd{.008} & 0.417\sd{.020}\\
UPN, DINOv2-L features & 0.772\sd{.004} & 0.780\sd{.001} & 0.557\sd{.005} & 0.735\sd{.006} & 0.252\sd{.007} & 0.361\sd{.013} & 0.196\sd{.004} & 0.317\sd{.017}\\
UPN, query states & 0.840\sd{.002} & 0.845\sd{.002} & 0.599\sd{.002} & 0.777\sd{.002} & 0.292\sd{.002} & 0.359\sd{.005} & 0.258\sd{.004} & 0.368\sd{.007}\\
\quad $\Delta$, DINOv2-L features & $-0.085^\dagger$ & $-0.081^\dagger$ & $-0.042^\dagger$ & $-0.050^\dagger$ & $-0.027^\dagger$ & $-0.058^\dagger$ & $-0.012^\dagger$ & $-0.100^\dagger$\\
\quad $\Delta$, query states & $-0.017^\dagger$ & $-0.016^\dagger$ & $-0.000$ & $-0.008^\dagger$ & $+0.013^\dagger$ & $-0.060^\dagger$ & $+0.050^\dagger$ & $-0.049^\dagger$\\
\bottomrule
\end{tabular}
\end{table*}
\paragraph{Caption style versus training data.}
\label{sec:app-style}
The Ref-L4 rows added in \cref{tab:app-e31} differ from RefCOCOg in three ways
at once: their caption style (long expressions, about half of them rewritten),
their image source (Objects365 photographs, whose referents are mostly small)
and their referent categories. A style control keeps the first and removes the
other two. Ref-L4's $4{,}802$ COCO images are split into two image-disjoint
halves ($2{,}401$ images each), and the $24$k RefCOCOg rows are extended by the
Ref-L4 queries of one COCO half, subsampled to exactly the number of queries of
the matching Objects365 half ($2{,}492$ and $2{,}445$): these rows carry
Ref-L4's caption style but only COCO images and COCO categories. Scoring is
cross-fitted as before: a query in a training half is scored by the model
trained on the other half, every other query by the mean of both. The hook and
the relation-aware head are trained on the COCO-80-prompt features, three
seeds per arm (\cref{tab:app-style}, top).

Caption style is a large effect where it comes with COCO images: the style arm
lifts the COCO-category referents by $+0.157$ [$+0.148$, $+0.165$] (hook) and
$+0.166$ [$+0.157$, $+0.175$] (relation-aware head), more than the Objects365
arm does ($+0.024$ and $+0.070$). On Objects365 referents it does almost
nothing: $-0.008$ [$-0.018$, $+0.002$] and $+0.015$ [$+0.007$, $+0.023$] on
referents without a COCO counterpart, $-0.010$ [$-0.026$, $+0.008$] and
$+0.009$ [$-0.003$, $+0.022$] on COCO-mappable ones. The gain of the Objects365
arm on referents without a COCO counterpart ($+0.253$ and $+0.332$ over the
$24$k rows) therefore lies beyond style: $+0.261$ [$+0.244$, $+0.279$] and
$+0.317$ [$+0.297$, $+0.337$]. Part of it is shared with the COCO-mappable
referents on the same Objects365 images ($+0.169$ [$+0.145$, $+0.194$] and
$+0.265$ [$+0.236$, $+0.293$]) and comes from the image source; the remainder,
specific to categories outside COCO, is $+0.092$ [$+0.063$, $+0.123$] and
$+0.052$ [$+0.015$, $+0.087$] (difference in differences). RefCOCOg val
changes by less than $0.01$ in every arm. Caption style thus explains almost
none of the transfer to Objects365 referents; what lifts them is training data from
their image source and, to a smaller but non-zero extent, from their
categories.

\Cref{tab:app-style} (bottom) reports the fractional-logit regression of
\cref{sec:app-refl4-regression} with size entered as relative-box-area terciles,
as a standardized log relative area with a quadratic term (the continuous
specification), and as the linear log-area term alone. The COCO-vocabulary
term is stable under all three: $+0.84$ to $+0.94$ log-odds for the trained
heads, against $+0.24$ to $+0.26$ for detector-native selection and coordinate
decoding. The image-source term is not. For the trained heads it is $-0.35$ to
$-0.73$ with terciles, $-0.11$ to $+0.16$ with the quadratic log area (the
intervals of the hook and the token-state probe include zero, the relation-aware
head's $+0.164$ [$+0.005$, $+0.323$] barely excludes it) and $-0.27$ to $-0.55$
with the linear term alone; detector-native selection has a positive
image-source term under each of the three ($+0.61$ to $+0.86$). Objects365
referents are concentrated at the small end ($71\%$ of the COCO-mappable and
$78\%$ of the other ones fall in the lowest area tercile, against $9\%$ on COCO
images), so image source and box size are only weakly separable. Letting the
terciles interact with the image source places the trained heads'
image-source deficit in the smallest tercile ($-0.91$ to $-1.83$, with
positive interactions for the larger ones), and a broader category mapping
gives the same pattern (COCO-vocabulary term $+1.01$ to $+1.11$ for the trained
heads). The category term is robust to the size specification; the
image-source term cannot be separated from box size and should not be read as
absent.

\begin{table*}[t]
\centering
\footnotesize
\setlength{\tabcolsep}{3pt}
\caption{Caption style versus training data on Ref-L4 val (Qwen2.5-VL,
COCO-80-prompt features, three seeds). \textbf{Top:} selection accuracy at
$\mathrm{IoU}\ge0.5$, no abstention (mean\,$\pm$\,sample std). Training rows:
the $24$k RefCOCOg rows, extended by the Ref-L4 queries on one image-disjoint
half of its COCO images (subsampled to the query count of the matching
Objects365 half) or on one half of its Objects365 images (the arm of
\cref{tab:app-e31}); cross-fitted scoring. Style term: COCO-image arm minus
$24$k rows; beyond style: Objects365 arm minus COCO-image arm; differences of
the seed-averaged per-query hits, with $95\%$ image-clustered bootstrap
intervals ($2{,}000$ resamples) on the line below. COCO cat.: referents on COCO
images, all of COCO categories; other: referents on Objects365 images, split
into O365 map.\ (category maps to a COCO class by a strict name mapping) and
O365 other. Relation-aware head: IoU-rank loss, trained as in
\cref{tab:app-e31}; its $24$k row is a retraining with the published recipe and
seeds (published run: $0.846$ and $0.522$). \textbf{Bottom:} fractional-logit regression of the
seed-averaged per-query hit on the $12{,}854$ covered Ref-L4 val queries
($7{,}005$ images); coefficients (log-odds) of the COCO-vocabulary indicator
(COCO category or strictly mapped to one) and the Objects365-image indicator,
with $95\%$ Wald intervals from the image-clustered sandwich covariance. Size
enters as relative-area terciles (two indicators), as the standardized log
relative area plus its square, or as the log relative area alone.}
\label{tab:app-style}
\begin{tabular}{@{}lcccccc@{}}
\toprule
& RefCOCOg & \multicolumn{5}{c}{Ref-L4 val}\\
\cmidrule(l){3-7}
Training rows & val & all & COCO cat. & other & O365 map. & O365 other\\
\midrule
\multicolumn{7}{@{}l}{\emph{Alignment hook}}\\
$24$k RefCOCOg rows & 0.787\sd{.004} & 0.464\sd{.010} & 0.592\sd{.018} & 0.243\sd{.004} & 0.384\sd{.017} & 0.171\sd{.008}\\
\quad + Ref-L4, COCO images & 0.792\sd{.008} & 0.560\sd{.008} & 0.749\sd{.016} & 0.234\sd{.006} & 0.374\sd{.009} & 0.163\sd{.005}\\
\quad + Ref-L4, Objects365 images & 0.794\sd{.003} & 0.561\sd{.016} & 0.617\sd{.022} & 0.465\sd{.008} & 0.543\sd{.002} & 0.425\sd{.013}\\
\quad Style term & & $+0.096$ & $+0.157$ & $-0.008$ & $-0.010$ & $-0.008$\\
& & \ci{.090}{.103} & \ci{.148}{.165} & \ci{$-$.017}{.001} & \ci{$-$.026}{.008} & \ci{$-$.018}{.002}\\
\quad Beyond style & & $+0.001$ & $-0.132$ & $+0.230$ & $+0.169$ & $+0.261$\\
& & \ci{$-$.008}{.009} & \ci{$-$.140}{$-$.125} & \ci{.216}{.246} & \ci{.145}{.194} & \ci{.244}{.279}\\
\midrule
\multicolumn{7}{@{}l}{\emph{Relation-aware head}}\\
$24$k RefCOCOg rows & 0.850\sd{.002} & 0.519\sd{.002} & 0.676\sd{.002} & 0.249\sd{.003} & 0.389\sd{.015} & 0.177\sd{.004}\\
\quad + Ref-L4, COCO images & 0.857\sd{.001} & 0.628\sd{.001} & 0.842\sd{.002} & 0.261\sd{.002} & 0.398\sd{.007} & 0.192\sd{.003}\\
\quad + Ref-L4, Objects365 images & 0.850\sd{.003} & 0.678\sd{.002} & 0.746\sd{.002} & 0.561\sd{.005} & 0.663\sd{.008} & 0.509\sd{.008}\\
\quad Style term & & $+0.110$ & $+0.166$ & $+0.013$ & $+0.009$ & $+0.015$\\
& & \ci{.103}{.116} & \ci{.157}{.175} & \ci{.006}{.020} & \ci{$-$.003}{.022} & \ci{.007}{.023}\\
\quad Beyond style & & $+0.050$ & $-0.095$ & $+0.300$ & $+0.265$ & $+0.317$\\
& & \ci{.041}{.059} & \ci{$-$.102}{$-$.089} & \ci{.282}{.317} & \ci{.236}{.293} & \ci{.297}{.337}\\
\bottomrule
\end{tabular}

\vspace{6pt}
\begin{tabular}{@{}llccc@{}}
\toprule
Method & Term & Size terciles & Log area and its square & Log area\\
\midrule
Alignment hook & COCO vocab. & $+0.864$\ci{.716}{1.012} & $+0.885$\ci{.732}{1.038} & $+0.879$\ci{.735}{1.024}\\
 & O365 image & $-0.346$\ci{$-$.479}{$-$.212} & $+0.006$\ci{$-$.136}{.148} & $-0.273$\ci{$-$.404}{$-$.141}\\
Relation-aware head & COCO vocab. & $+0.875$\ci{.718}{1.033} & $+0.941$\ci{.769}{1.113} & $+0.892$\ci{.739}{1.044}\\
 & O365 image & $-0.435$\ci{$-$.579}{$-$.292} & $+0.164$\ci{.005}{.323} & $-0.452$\ci{$-$.590}{$-$.313}\\
Token-state probe & COCO vocab. & $+0.841$\ci{.686}{.995} & $+0.930$\ci{.755}{1.105} & $+0.883$\ci{.730}{1.035}\\
 & O365 image & $-0.733$\ci{$-$.878}{$-$.589} & $-0.107$\ci{$-$.276}{.063} & $-0.553$\ci{$-$.696}{$-$.409}\\
\midrule
Detector-native & COCO vocab. & $+0.242$\ci{.085}{.399} & $+0.246$\ci{.084}{.408} & $+0.243$\ci{.086}{.401}\\
 & O365 image & $+0.605$\ci{.445}{.765} & $+0.859$\ci{.691}{1.028} & $+0.772$\ci{.613}{.932}\\
Coordinate decoding & COCO vocab. & $+0.257$\ci{.061}{.453} & $+0.256$\ci{.058}{.454} & $+0.257$\ci{.059}{.456}\\
 & O365 image & $-0.052$\ci{$-$.259}{.156} & $+0.232$\ci{.001}{.464} & $+0.237$\ci{.005}{.469}\\
\bottomrule
\end{tabular}
\end{table*}
\paragraph{Broad-vocabulary training phrases.}
\label{sec:app-broadvocab}
The Objects365 arm of \cref{tab:app-e31} adds referents from the target images
together with their categories. A second source asks whether categories outside
COCO, taken from other images and phrased differently, move the boundary as
well. From the training split of Flickr30k Entities \cite{flickr30kentities} we
keep phrases whose entity has exactly one box, and drop scene and non-visual
phrases, phrases whose last word names no object (a color, ``one'', a
pronoun), person phrases (the dataset's person type or a person noun) and
phrases whose head noun is a COCO class name or a synonym of one. One phrase is
kept per entity and images are visited in a seeded round-robin order, one
entity per image per round, until $30{,}000$ phrase--box pairs are reached. They
lie on $24{,}914$ images and are short (median two words); $40\%$ name clothing,
$10\%$ body parts, $46\%$ other objects and $4\%$ instruments, vehicles and
animals, with $3{,}228$ distinct head words. Their features are built exactly as
those of the training rows (COCO-80 proposal prompt, top $300$ proposals, same
VLM query); rebuilding $256$ RefCOCOg val rows in the same run reproduces the
cached features exactly, and a proposal covers $79\%$ of the phrases' referents.
The hook and the relation-aware head are trained on the $24$k RefCOCOg rows with
and without the phrases, three seeds each, with the same number of epochs (the
second arm thus passes over $2.25$ times as many rows). On these
caches the published hooks reproduce the main-table numbers, and the hook
trained without the phrases is identical to them. The relation-aware head is
retrained in the same run as its second arm and does not reproduce the earlier
run of \cref{tab:app-e31,tab:app-style} exactly although data, configuration,
code and seeds are the same: on RefCOCOg val it is $0.003$--$0.013$ lower per
seed ($0.842$ against $0.850$ for the mean), on Ref-L4 overall within $0.003$
per seed. Every
difference below is therefore paired within this one run. Before training, the Ref-L4 referents without a
COCO counterpart were split by whether the head noun of their category name is
among the phrases' head words ($2{,}461$ queries in $195$ categories, \emph{named})
or not ($812$ queries in $91$ categories).

The phrases raise accuracy on Objects365 images (\cref{tab:app-broadvocab}):
$+0.056$ [$+0.046$, $+0.066$] (hook) and $+0.081$ [$+0.071$, $+0.092$]
(relation-aware head; per seed $+0.03$ to $+0.08$ for the hook, $+0.08$ for the
relation-aware head, and the hook's gain on COCO-mappable referents ranges from
$-0.015$ to $+0.084$ across seeds), $+0.063$ and $+0.085$ on referents without a COCO
counterpart and $+0.041$ and $+0.074$ on COCO-mappable ones, while COCO-image
referents change by $-0.001$ and $+0.017$ and RefCOCOg by at most $+0.012$. The
gain is about a quarter of what Ref-L4's own Objects365 rows give ($+0.253$ and
$+0.332$ on referents without a COCO counterpart, \cref{tab:app-style}), and
most of it does not depend on naming the category: under the hook named and
other categories gain alike ($-0.006$ [$-0.032$, $+0.019$] between them); the
relation-aware head gains $+0.058$ even on categories the phrases never name
and $0.036$ [$0.008$, $0.065$] more on named ones. Referents outside COCO from another image source thus narrow the
boundary by a few points, mostly not through the category names, and leave
it far from closed.

\begin{table*}[t]
\centering
\footnotesize
\setlength{\tabcolsep}{2.5pt}
\caption{Broad-vocabulary training phrases on Ref-L4 val and RefCOCOg (Qwen2.5-VL,
COCO-80-prompt features, three seeds). Selection accuracy at
$\mathrm{IoU}\ge0.5$, no abstention (mean\,$\pm$\,sample std). Training rows:
the $24$k RefCOCOg rows, alone or with $30{,}000$ Flickr30k Entities phrases
whose referents are neither persons nor COCO categories
(paragraph ``Broad-vocabulary training phrases''). Difference: second arm minus
first, from the seed-averaged per-query hits, with $95\%$ bootstrap intervals
($2{,}000$ resamples; image-clustered on Ref-L4, over queries on RefCOCOg) on
the line below. Both arms come from one training run, so
each difference is paired within it; the intervals reflect query sampling only.
Retraining the relation-aware head with the same data and seeds in another run
moves its RefCOCOg val accuracy by up to $0.013$ per seed ($0.008$ for the
mean; its first row here: $0.842$, against $0.850$ in \cref{tab:app-style}) and
its overall Ref-L4 accuracy by at most $0.003$ per seed, so its RefCOCOg differences of about
$0.01$ are not resolved; the hook rows reproduce exactly across runs. Columns as in \cref{tab:app-style};
named / not named: the referents of the O365 other column whose category head
noun is / is not among the phrases' head words (split fixed before training);
named minus not named: difference of the two differences, bootstrapped within
each group.}
\label{tab:app-broadvocab}
\begin{tabular}{@{}lccccccccc@{}}
\toprule
& \multicolumn{2}{c}{RefCOCOg} & \multicolumn{7}{c}{Ref-L4 val}\\
\cmidrule(lr){2-3}\cmidrule(l){4-10}
Training rows & val & test & all & COCO cat. & other & O365 map. & O365 other & named & not named\\
\midrule
\multicolumn{10}{@{}l}{\emph{Alignment hook}}\\
$24$k RefCOCOg rows & 0.787\sd{.004} & 0.793\sd{.007} & 0.464\sd{.010} & 0.592\sd{.018} & 0.243\sd{.004} & 0.384\sd{.017} & 0.171\sd{.008} & 0.148 & 0.241\\
\quad + $30$k Flickr30k Entities phrases & 0.799\sd{.001} & 0.805\sd{.001} & 0.484\sd{.019} & 0.591\sd{.021} & 0.299\sd{.024} & 0.425\sd{.037} & 0.235\sd{.019} & 0.210 & 0.309\\
\quad Difference & $+0.012$ & $+0.012$ & $+0.020$ & $-0.001$ & $+0.056$ & $+0.041$ & $+0.063$ & $+0.062$ & $+0.068$\\
& \ci{.006}{.019} & \ci{.007}{.017} & \ci{.014}{.026} & \ci{$-$.009}{.006} & \ci{.046}{.066} & \ci{.022}{.060} & \ci{.052}{.075} & \ci{.050}{.074} & \ci{.044}{.090}\\
\quad Named minus not named & \multicolumn{7}{c}{} & \multicolumn{2}{c}{$-0.006$\ci{$-$.032}{.019}}\\
\midrule
\multicolumn{10}{@{}l}{\emph{Relation-aware head}}\\
$24$k RefCOCOg rows & 0.842\sd{.004} & 0.850\sd{.002} & 0.520\sd{.001} & 0.679\sd{.000} & 0.248\sd{.002} & 0.388\sd{.010} & 0.176\sd{.002} & 0.145 & 0.271\\
\quad + $30$k Flickr30k Entities phrases & 0.852\sd{.002} & 0.853\sd{.003} & 0.561\sd{.003} & 0.696\sd{.004} & 0.329\sd{.004} & 0.462\sd{.013} & 0.261\sd{.003} & 0.239 & 0.329\\
\quad Difference & $+0.009$ & $+0.003$ & $+0.041$ & $+0.017$ & $+0.081$ & $+0.074$ & $+0.085$ & $+0.093$ & $+0.058$\\
& \ci{.004}{.015} & \ci{$-$.001}{.007} & \ci{.036}{.047} & \ci{.011}{.024} & \ci{.071}{.092} & \ci{.056}{.092} & \ci{.072}{.097} & \ci{.080}{.108} & \ci{.034}{.084}\\
\quad Named minus not named & \multicolumn{7}{c}{} & \multicolumn{2}{c}{$+0.036$\ci{.008}{.065}}\\
\bottomrule
\end{tabular}
\end{table*}

\paragraph{Achievable versus realized selection margin.}
Table~\ref{tab:app-margin} measures, in the trained region space of each of
the three seeds, the margin $\mu/\kappa$ a best-possible query could attain at
the referent's proposal and the margin $\rho$ the emitted query does attain.
The achievable margin is nearly constant across regimes ($29.8$--$32.7$),
including on the reasoning-stress set and on the out-of-vocabulary bucket of
Ref-L4: the frozen features and the learned region projection leave as much
room for a correct selection there as on standard expressions. The realized
margin is what separates the regimes, from $+3.0$ on RefCOCOg val to $-8.7$
outside the proposal prompt's vocabulary. Since $\rho>0$ exactly when the
maximal-IoU proposal outranks every incorrect proposal, which implies a hit,
its sign largely restates accuracy. The informative comparison is with
$\mu/\kappa$ on the same rows, and its reading is limited: $\mu$ is computed
from request-independent features, so its flatness across regimes is expected,
and it is set by the nearest incorrect proposal, often an overlapping box of
the same object. It shows that per-instance separability does not differ
between regimes, not whether the deficit lies in the query, in the trained
projection or in what the region features encode. The median ratio of
$0.06$--$0.11$ on the four standard splits is likewise not evidence of waste:
a softmax-trained interface stops widening its margin once the loss saturates.
Two properties make the reading conservative. The achievable margin is taken at the proposal of maximal IoU,
although any proposal above the threshold counts as a hit, so it understates
the room a query could use. And the pattern is not an artifact of what the
projection learned: in the raw detector feature space, before any training,
the median achievable margin is likewise flat across regimes
($0.24$ on RefCOCO and RefCOCO+ val, $0.26$ on RefCOCOg val, $0.29$ on the stress
set, $0.25$ and $0.26$ on the two Ref-L4 buckets, on the unit-sphere scale), with the out-of-vocabulary
bucket no worse than the rest. Two details make the accuracies in
this table differ slightly from the main ones. The Ref-L4 rows are the
$9{,}388$ present queries of the evaluation set augmented with absent
referents, a content-independent $70\%$ sample of the split whose vocabulary
mix ($62.5$\% / $37.5$\%) matches the full set. And the hooks are retrained
for this analysis, on the earlier feature build (\cref{sec:app-builds}), with the same
detector, prompt and VLM, reaching $0.779\pm0.013$ on RefCOCOg val against
$0.787\pm0.004$ for the published hooks (\cref{tab:app-grounding24k}); the margins therefore
characterize the interface rather than one particular checkpoint.

\begin{table}[t]
\centering
\footnotesize
\setlength{\tabcolsep}{3pt}
\caption{Achievable ($\mu/\kappa$) and realized ($\rho$) selection margin at
the maximal-IoU proposal, medians over covered present queries, mean over
three seeds. Ratio: median of $\kappa\rho/\mu$, the share of the available
margin the trained interface attains. $n$: covered present queries; Acc:
selection accuracy of the same runs over these queries.}
\label{tab:app-margin}
\begin{tabular}{@{}lrcccc@{}}
\toprule
Split & $n$ & Acc & $\mu/\kappa$ & $\rho$ & Ratio\\
\midrule
RefCOCO val & 10815 & 0.780 & 30.0 & $+3.3$ & $0.11$\\
RefCOCO+ val & 10742 & 0.700 & 30.1 & $+1.7$ & $0.06$\\
RefCOCOg val & 4896 & 0.779 & 32.6 & $+3.0$ & $0.10$\\
RefCOCOg test & 9583 & 0.780 & 32.7 & $+3.1$ & $0.10$\\
Ref-L4, COCO-80 referent & 5858 & 0.622 & 31.0 & $-0.1$ & $-0.00$\\
Reasoning stress & 231 & 0.375 & 32.2 & $-6.5$ & $-0.20$\\
Ref-L4, other referent & 3112 & 0.251 & 29.8 & $-8.7$ & $-0.30$\\
\bottomrule
\end{tabular}
\end{table}

\paragraph{Query-source and head audit.}
Table~\ref{tab:app-controls} gives every variant of the two trained controls on
every split. Bridge variants other than the one used in the main grounding
table are trained with the hook's budget ($50$ epochs, uncovered positives
masked). Swept index-scorer variants choose their learning rate and number
of epochs on RefCOCOg val, so that column is optimistic for them; the other
splits are untouched by the choice. The bridge trained with the
image-conditioned VLM query reproduces the hook trained without the verifier on
every split and seed, to the last digit.

\begin{table*}[htbp]
\centering
\footnotesize
\setlength{\tabcolsep}{4pt}
\caption{Audit of the trained matched-perception controls: selection accuracy
at $\mathrm{IoU}\ge0.5$ (mean$\pm$sample std over three seeds).}
\label{tab:app-controls}
\begin{tabular}{@{}lcccccc@{}}
\toprule
Variant & RefCOCO val & RefCOCO+ val & RefCOCOg val & RefCOCOg test & Ref-L4 val & Stress\\
\midrule
Alignment hook (reference) & $0.789{\pm}0.007$ & $0.707{\pm}0.010$ & $0.787{\pm}0.004$ & $0.793{\pm}0.007$ & $0.464{\pm}0.010$ & $0.384{\pm}0.035$\\
\midrule
\multicolumn{7}{@{}l}{\emph{Index scorer (image-conditioned VLM query)}}\\
As proposed (index embedding, 30 epochs) & $0.759{\pm}0.002$ & $0.683{\pm}0.004$ & $0.770{\pm}0.002$ & $0.770{\pm}0.007$ & $0.444{\pm}0.010$ & $0.384{\pm}0.032$\\
No index embedding & $0.793{\pm}0.006$ & $0.714{\pm}0.005$ & $0.805{\pm}0.012$ & $0.808{\pm}0.007$ & $0.469{\pm}0.037$ & $0.420{\pm}0.004$\\
Bilinear scorer & $0.785{\pm}0.013$ & $0.711{\pm}0.012$ & $0.797{\pm}0.009$ & $0.795{\pm}0.008$ & $0.471{\pm}0.004$ & $0.404{\pm}0.011$\\
Uncovered positives masked & $0.754{\pm}0.003$ & $0.675{\pm}0.006$ & $0.763{\pm}0.005$ & $0.763{\pm}0.005$ & $0.428{\pm}0.008$ & $0.381{\pm}0.009$\\
Learning-rate and epoch sweep & $0.770{\pm}0.003$ & $0.692{\pm}0.004$ & $0.783{\pm}0.008$ & $0.786{\pm}0.003$ & $0.455{\pm}0.011$ & $0.397{\pm}0.026$\\
Audited (no index, masked, normalized, sweep) & $0.801{\pm}0.003$ & $0.722{\pm}0.009$ & $0.811{\pm}0.003$ & $0.815{\pm}0.002$ & $0.451{\pm}0.027$ & $0.434{\pm}0.017$\\
Audited, bilinear scorer & $0.796{\pm}0.002$ & $0.721{\pm}0.005$ & $0.807{\pm}0.002$ & $0.809{\pm}0.004$ & $0.468{\pm}0.015$ & $0.430{\pm}0.016$\\
\midrule
\multicolumn{7}{@{}l}{\emph{Learned bridge (query source varies)}}\\
CLIP text, 30 epochs (\cref{tab:app-grounding24k}) & $0.511{\pm}0.017$ & $0.429{\pm}0.008$ & $0.542{\pm}0.006$ & $0.552{\pm}0.009$ & $0.357{\pm}0.006$ & $0.254{\pm}0.022$\\
CLIP text & $0.504{\pm}0.012$ & $0.421{\pm}0.007$ & $0.541{\pm}0.005$ & $0.553{\pm}0.005$ & $0.359{\pm}0.005$ & $0.237{\pm}0.029$\\
Detector text tower & $0.593{\pm}0.006$ & $0.443{\pm}0.004$ & $0.596{\pm}0.005$ & $0.605{\pm}0.006$ & $0.388{\pm}0.004$ & $0.250{\pm}0.051$\\
Detector text tower, projected & $0.601{\pm}0.014$ & $0.454{\pm}0.011$ & $0.596{\pm}0.006$ & $0.611{\pm}0.010$ & $0.375{\pm}0.002$ & $0.224{\pm}0.027$\\
VLM, text only (no image) & $0.557{\pm}0.005$ & $0.444{\pm}0.012$ & $0.579{\pm}0.007$ & $0.593{\pm}0.007$ & $0.372{\pm}0.018$ & $0.294{\pm}0.033$\\
VLM, image-conditioned & $0.784{\pm}0.013$ & $0.705{\pm}0.012$ & $0.785{\pm}0.009$ & $0.791{\pm}0.013$ & $0.465{\pm}0.011$ & $0.423{\pm}0.017$\\
\bottomrule
\end{tabular}
\end{table*}

\paragraph{Hook capacity.}
Table~\ref{tab:app-capacity} varies the hook's shared dimension, width and
depth with the default single-positive supervision, and the positive set.
Capacity changes stay within $0.02$ of the default configuration; only the
positive-set definition moves accuracy beyond the seed variation.

\begin{table*}[htbp]
\centering
\footnotesize
\setlength{\tabcolsep}{5pt}
\caption{Hook capacity and positives on RefCOCOg (selection accuracy at
$\mathrm{IoU}\ge0.5$, instance-exact selection and recall@5 on test; three
seeds). $d_s$: shared dimension; $h$: hidden width; depth: hidden layers.}
\label{tab:app-capacity}
\begin{tabular}{@{}lcccc@{}}
\toprule
Configuration & Val & Test & Inst.-exact & R@5\\
\midrule
Default ($d_s{=}128$, $h{=}256$, depth 1) & $0.787{\pm}0.004$ & $0.793{\pm}0.007$ & $0.399{\pm}0.005$ & 0.958\\
$d_s{=}256$ & $0.792{\pm}0.002$ & $0.792{\pm}0.004$ & $0.394{\pm}0.007$ & 0.953\\
$d_s{=}512$ & $0.789{\pm}0.013$ & $0.794{\pm}0.006$ & $0.397{\pm}0.003$ & 0.957\\
$h{=}1024$ & $0.784{\pm}0.013$ & $0.790{\pm}0.004$ & $0.394{\pm}0.005$ & 0.957\\
Depth 2 & $0.789{\pm}0.004$ & $0.787{\pm}0.004$ & $0.391{\pm}0.013$ & 0.952\\
Depth 3 & $0.786{\pm}0.003$ & $0.787{\pm}0.004$ & $0.399{\pm}0.002$ & 0.951\\
Linear (depth 0) & $0.767{\pm}0.008$ & $0.775{\pm}0.006$ & $0.388{\pm}0.014$ & 0.953\\
\midrule
Multi-positive ($\mathrm{IoU}\ge0.5$) & $0.808{\pm}0.010$ & $0.805{\pm}0.014$ & $0.426{\pm}0.013$ & 0.965\\
Multi-positive, $d_s{=}256$, depth 2 & $0.821{\pm}0.006$ & $0.821{\pm}0.003$ & $0.433{\pm}0.005$ & 0.966\\
\bottomrule
\end{tabular}
\end{table*}

\paragraph{Relation-aware head.}
\label{sec:app-relation}
Each of the $M$ proposals is embedded from its region feature and its box
geometry $(x_1,y_1,x_2,y_2,w,h,\text{area})$, modulated feature-wise by the
projected query, and joined by one existence token. Two pre-normalized
transformer layers apply self-attention over the $M{+}1$ tokens with an
additive bias per head and layer computed from the relative geometry of every
pair (signed log offsets, log size ratios, center offsets and IoU),
cross-attention to eight tokens obtained by a linear expansion of the query,
and a feed-forward block. A proposal's logit is a scaled cosine between its
token and the query plus a small MLP over their concatenation and product; the
existence logit is read from the existence token. The losses are retrieval
cross-entropy with the single matched proposal, with every proposal at
$\mathrm{IoU}\ge0.5$ as positives, or both (hybrid), optionally an
IoU-weighted listwise term that favors tighter boxes, and existence
cross-entropy. The head has $4.29$M parameters; the capacity-matched hook
($4.29$M, $d_s{=}256$, $h{=}881$) and the default hook ($1.12$M) are trained in
the same runs. Removing the pairwise geometry bias from the hybrid-loss head
changes selection accuracy by at most $0.004$ on the four standard splits
($+0.002$, $+0.003$, $+0.003$ and $-0.001$ on RefCOCO, RefCOCO+ and RefCOCOg val
and RefCOCOg test), by $-0.002$ on Ref-L4 and by $-0.019$ on the stress set
(paired permutation $p=0.075$). On the stress set by category, the head reaches
$0.325{\pm}0.012$ (affordance), $0.582{\pm}0.008$ (negation) and
$0.219{\pm}0.044$ (world knowledge), against $0.309$, $0.568$ and $0.276$ for
the default hook.

\paragraph{Interface ablation: protocol, buckets and phrase examples.}
\label{sec:app-interface}
Both interface variants reuse the identical frozen detector ($M{=}300$), the
identical RefCOCOg training pool ($24$k rows, negative rate $0.3$), the
identical fixed thresholds and seed; only the query-embedding source differs.
The discrete variant's caches were rebuilt with the detector re-run on other
hardware; its covered-proposal flags agree with the main build on every
RefCOCOg val and stress query (\cref{sec:app-builds}).
For the discrete variant the frozen VLM is prompted with the image and the
instruction \emph{``In at most 6 words, name the object referred to by:
`\{expression\}'. Respond with only the short phrase,''} and the phrase
(truncated to six words) is re-embedded through the same readout used for the
continuous query. Table~\ref{tab:app-buckets} breaks RefCOCOg val down by
expression length and by the presence of relational and negation words. The
continuous query is ahead in every bucket; the gap is largest for
relational expressions and not significant for the $122$ expressions with a
negation word. The verbalization step also makes existence judgments of its
own: for present referents it sometimes answers \emph{``Not possible.''} or
\emph{``not visible''}, and for absent training referents \emph{``no zebra
present''} or \emph{``no such object''}. This is one reason why the discrete
variant answers fewer present queries at the fixed thresholds (answer rate
$0.539$ against $0.828$ for the continuous query, seed 42) and why it is not
simply a weaker version of the continuous one. On the stress set it compresses
affordance requests to object names (\emph{``something to ride''}
$\to$ \emph{``scooter''}; \emph{``something to eat with''} $\to$
\emph{``wooden spoon''}), which is where it gains: $0.450$ against $0.390$ for
the continuous query on the stress set (seed 42, $p=0.10$), against $0.681$ and
$0.787$ on RefCOCOg val. The continuous channel is thus worth about $0.1$ on
standard expressions but is not what limits the reasoning probes.

\begin{table}[htbp]
\centering
\footnotesize
\setlength{\tabcolsep}{3pt}
\caption{Continuous versus discrete interface on RefCOCOg val by expression
bucket (selection accuracy at $\mathrm{IoU}\ge0.5$, seed 42, same rows; discrete column on rebuilt caches, \cref{sec:app-interface}; mid-$p$
of the exact McNemar test, whose exact $p$ is in \cref{tab:app-tests}). Relational words: above, behind, below, beside,
between, bottom, closest, farthest, front, left, middle, near, next, on, right,
top, under. Negation words: empty, no, nobody, not, nothing, unoccupied,
without, and contractions ending in \emph{n't}.}
\label{tab:app-buckets}
\begin{tabular}{@{}lcccc@{}}
\toprule
Bucket & $n$ & Continuous & Discrete & mid-$p$\\
\midrule
All & 4896 & 0.787 & 0.681 & $1\times10^{-47}$\\
Length $\le4$ words & 693 & 0.797 & 0.729 & $2\times10^{-4}$\\
Length 5--8 words & 2079 & 0.783 & 0.667 & $1\times10^{-23}$\\
Length $\ge9$ words & 2124 & 0.789 & 0.678 & $3\times10^{-23}$\\
Relational word & 2161 & 0.768 & 0.634 & $9\times10^{-30}$\\
No relational word & 2735 & 0.803 & 0.717 & $4\times10^{-20}$\\
Negation word & 122 & 0.615 & 0.574 & 0.42\\
No negation word & 4774 & 0.792 & 0.683 & $4\times10^{-48}$\\
\bottomrule
\end{tabular}
\end{table}

\paragraph{Agreement between the hook and coordinate decoding.}
Table~\ref{tab:app-agreement} gives the agreement signal on every split with
coordinate-decoding boxes. Agreement means that the hook's selected proposal and
the coordinate-decoding box overlap at $\mathrm{IoU}\ge0.5$. The emitted box is
chosen on RefCOCOg val per seed among coordinate decoding, the hook and a
re-ranking of the proposals that overlap the coordinate box by hook affinity
(two seeds choose coordinate decoding, one the re-ranking), so the selective
accuracies average over these choices. On RefCOCOg val at least one of the two
is right on $0.92$ of queries, and answering only on agreement covers $0.79$
at $0.94$ selective accuracy; re-ranking by hook affinity does not beat
coordinate decoding, and the two agree on under half of the Ref-L4 and stress
queries.

\begin{table*}[htbp]
\centering
\footnotesize
\setlength{\tabcolsep}{4pt}
\caption{Hook--coordinate-decoding complementarity (present queries; mean over
three hook seeds). Oracle: at least one of the two is correct. Emit iff agree:
answer rate (Ans., over all present queries) and selective accuracy of the
operating point that answers only when the two agree. AURC of three confidence signals for the emitted box (lower is
better).}
\label{tab:app-agreement}
\begin{tabular}{@{}lccccccccc@{}}
\toprule
& & & & \multicolumn{2}{c}{Emit iff agree} & \multicolumn{3}{c}{AURC}\\
\cmidrule(lr){5-6}\cmidrule(lr){7-9}
Split & Hook & Coord.-snap & Oracle & Ans. & Sel.\ acc. & Agree, then affinity & Detector score & Verifier\\
\midrule
RefCOCOg val  & 0.787 & 0.867 & 0.922 & 0.785 & 0.936 & 0.051 & 0.060 & 0.091\\
RefCOCOg test & 0.793 & 0.866 & 0.922 & 0.791 & 0.934 & 0.048 & 0.057 & 0.097\\
RefCOCO val   & 0.789 & 0.895 & 0.935 & 0.785 & 0.958 & 0.032 & 0.035 & 0.085\\
Ref-L4 val    & 0.464 & 0.833 & 0.859 & 0.474 & 0.925 & 0.082 & -- & 0.101\\
Stress        & 0.384 & 0.610 & 0.668 & 0.440 & 0.748 & 0.263 & 0.387 & 0.288\\
\bottomrule
\end{tabular}
\end{table*}

\paragraph{Selecting inside the VLM: all variants and the prompts.}
\label{sec:app-invlm-text}
Table~\ref{tab:app-invlm} extends Table 3 of the main paper. The
frozen VLM sees the top-$K$ proposals of a ranker (the hook, seed 42, or the
detector prompted with the expression), with near-duplicates (IoU${>}0.7$ to a
higher-ranked box) removed, and may answer ``none''. Box-list coordinates are given in pixels of the smart-resized input frame in
which the VLM grounds. Its answer is taken
greedily; for the true/false variant each candidate is judged separately,
the candidate with the largest true-minus-false score is returned, and
``none'' means no candidate was judged true. The prompts are:
marks, \texttt{\{k\} numbered boxes are drawn on the image. Which box marks
"\{expression\}"? Answer with the number of the box only, or "none" if no box
marks it.}; box list, \texttt{Candidate boxes in this image, as [x1, y1,
x2, y2] pixel coordinates: \{one box per line\} Which box marks
"\{expression\}"? Answer with the number of the box only, or "none" if no box
marks it.}; true/false, \texttt{A red box is drawn on the image. Does the
box match the description "\{expression\}"? Answer true or false.} (prompts
in typewriter type are verbatim, including their straight quotes). Ten
candidates raise the ceiling but lower accuracy for every interface, and the
VLM rarely says ``none'' when the referent is present ($\le1.4\%$ of present
queries except true/false, $9.3\%$). On the absent rows ``none'' is most
reliable for the true/false variant ($0.953$), whose joint accuracy ($0.770$)
stays below the verifier attached to coordinate-snap ($0.888$,
\cref{tab:app-detach}). Its Acc column is a forced choice: when no candidate
is judged true it still returns the highest-scoring box, which is correct for
$210$ of the $454$ such present queries on RefCOCOg val; with abstention in
those cases, as in the Joint column, its present-query accuracy is $0.692$
($0.672$ on the RefCOCOg test subset). The result also depends on how per-box
judgments are combined. Returning instead the first candidate judged true in
the hook's ranking (with the same fallback) reaches $0.800$--$0.802$ on
RefCOCOg val, slightly above the hook ($0.787$); we found this rule after
seeing these results and have not evaluated it on held-out data, so we report
it as an observation, not as a result.

\begin{table}[t]
\centering
\footnotesize
\setlength{\tabcolsep}{1.6pt}
\caption{In-VLM selection over the same proposals, no fine-tuning. Acc: accuracy
at $\mathrm{IoU}\ge0.5$ on present queries (for true/false a forced choice,
see text); Ceil.: a correct box among the $K$
candidates; No-tgt.\ / Joint: ``none'' on the $30\%$ swapped absent rows and
joint accuracy over present and absent rows.}
\label{tab:app-invlm}
\begin{tabular}{@{}lccccccc@{}}
\toprule
& \multicolumn{4}{c}{RefCOCOg val} & \multicolumn{3}{c}{Stress}\\
\cmidrule(lr){2-5}\cmidrule(lr){6-8}
Variant & Acc & Ceil. & No-tgt. & Joint & Acc & Ceil. & Joint\\
\midrule
Marks, hook, $K{=}5$ & 0.587 & 0.973 & 0.743 & 0.632 & 0.411 & 0.736 & 0.376\\
Marks, hook, $K{=}10$ & 0.413 & 0.988 & 0.719 & 0.504 & 0.329 & 0.835 & 0.339\\
Marks, detector, $K{=}5$ & 0.479 & 0.980 & 0.740 & 0.555 & 0.429 & 0.719 & 0.415\\
Marks, detector, $K{=}10$ & 0.338 & 0.991 & 0.713 & 0.449 & 0.333 & 0.835 & 0.348\\
Box list, hook, $K{=}5$ & 0.760 & 0.973 & 0.791 & 0.767 & 0.524 & 0.736 & 0.500\\
Box list, hook, $K{=}10$ & 0.695 & 0.988 & 0.819 & 0.729 & 0.489 & 0.835 & 0.503\\
True/false, hook, $K{=}5$ & 0.735 & 0.973 & 0.953 & 0.770 & 0.489 & 0.736 & 0.467\\
\bottomrule
\end{tabular}
\end{table}

\paragraph{Other splits.}
\Cref{tab:app-invlm-splits} repeats the protocol on fixed $3{,}000$-query subsets
of RefCOCOg test, RefCOCO val and RefCOCO+ val, with $1{,}286$ swapped absent
rows added to RefCOCOg test by the generator used throughout
(\cref{sec:app-config-notes}), and on the
$2{,}121$ of the $3{,}000$ Ref-L4 queries of Table 3 of the main paper that are
present queries in that build, with five and ten candidates. These runs use
the earlier feature build (\cref{sec:app-builds}) and a hook of the same
recipe (seed 42) retrained on it; the hook and detector rows are therefore recomputed on the same subsets
and proposals. The pattern of Table 3 of the main paper holds on every split. The box
list stays within $0.04$ of the hook's first choice ($0.750$, $0.773$ and
$0.719$ against $0.790$, $0.779$ and $0.706$), although a correct box is among
its candidates for $0.94$--$0.97$ of the queries; true/false per box is close
behind, numbered marks lose $0.13$--$0.21$ to the hook, and marks over the
detector's ranking are lower still. Answering true/false first and falling
back to the marks is below true/false alone. On
Ref-L4 the box list gains $0.12$ over the hook, and ten candidates are again
worse than five for both interfaces. On RefCOCOg test with swapped absent rows
the best joint accuracy is $0.763$ with the VLM's own ``none'' and $0.784$ with
the thresholds chosen on RefCOCOg val, against $0.888$ for the verifier
attached to coordinate-snap on the full split (\cref{tab:app-detach}).

\begin{table*}[t]
\centering
\footnotesize
\setlength{\tabcolsep}{5pt}
\caption{In-VLM selection on the remaining splits ($3{,}000$-query subsets;
Ref-L4: the $2{,}121$ of the $3{,}000$ queries of Table 3 of the main paper that are
present queries in the earlier build), no fine-tuning. Acc: accuracy at $\mathrm{IoU}\ge0.5$ on present queries.
RefCOCOg test also has $1{,}286$ swapped absent rows: No-tgt.\ is the share
answered ``none'' (or no box judged true), Joint the accuracy over present and
absent rows. Hook and detector rows: their own first choice on the same
subset and proposals. ${}^\ast$True/false first, then the marks answer when
several or no boxes are judged true.}
\label{tab:app-invlm-splits}
\begin{tabular}{@{}lcccccc@{}}
\toprule
& \multicolumn{3}{c}{RefCOCOg test} & RefCOCO & RefCOCO+ & Ref-L4\\
\cmidrule(lr){2-4}
Variant & Acc & No-tgt. & Joint & val & val & val\\
\midrule
Hook, top-1 & \textbf{0.790} & -- & -- & 0.779 & 0.706 & 0.481\\
Detector, top-1 & 0.774 & -- & -- & \textbf{0.843} & \textbf{0.736} & --\\
Correct box in hook's top 5 & 0.970 & -- & -- & 0.960 & 0.943 & 0.767\\
\midrule
Marks, hook, $K{=}5$ & 0.578 & 0.767 & 0.632 & 0.614 & 0.578 & 0.518\\
Marks, hook, $K{=}10$ & -- & -- & -- & -- & -- & 0.398\\
Marks, detector, $K{=}5$ & 0.467 & 0.761 & 0.552 & 0.572 & 0.504 & --\\
Box list, hook, $K{=}5$ & 0.750 & 0.807 & \textbf{0.763} & 0.773 & 0.719 & \textbf{0.603}\\
Box list, hook, $K{=}10$ & -- & -- & -- & -- & -- & 0.589\\
True/false, hook, $K{=}5$ & 0.721 & \textbf{0.967} & 0.761 & 0.763 & 0.714 & --\\
True/false, then marks$^\ast$ & 0.611 & 0.767 & 0.656 & 0.651 & 0.623 & --\\
\bottomrule
\end{tabular}
\end{table*}

\paragraph{A VLM fine-tuned to select.}
\label{sec:app-e32}
The fine-tuned selector of Table 3 of the main paper answers exactly the
zero-shot box-list prompt, over the same candidates (the published seed-42
hook's top $20$, non-maximum suppression at IoU $0.7$, the five highest kept
and numbered left to right by box center), and is scored the same way. Its
training rows are the hook's $24{,}000$ training rows ranked by that hook:
present rows are labeled with the number of the highest-IoU correct candidate,
swapped rows with ``none'', and the $27$ present rows without a correct
candidate are dropped ($23{,}973$ rows; targets $1$--$5$: $3{,}121$, $3{,}884$,
$3{,}763$, $3{,}487$, $2{,}442$; ``none'': $7{,}276$). Low-rank adapters (rank
$16$, $\alpha=32$, dropout $0.05$) on the query, key, value and output
projections of Qwen2.5-VL-7B are trained for one epoch with the loss on the
answer tokens only: AdamW, learning rate $2{\times}10^{-4}$ (constant), weight
decay $0.01$, gradient clipping $1.0$, micro-batch $8$ with $8$ accumulation
steps, bfloat16, at most $1{,}003{,}520$ pixels, seed 42. Rerunning the
zero-shot box list in the same runtime gives $0.759$ against $0.760$, so the
gain is not an artifact of the inference setup. \Cref{tab:app-e32} gives the
details; on RefCOCOg val the trained selector is ahead of coordinate decoding
on $279$ present requests and behind on $215$ (exact McNemar $p=0.005$). The training candidates are easier than at evaluation, since the hook
ranks rows it was trained on (top-1 accuracy $0.967$ and top-five ceiling
$0.998$ there, against $0.787$ and $0.973$ on RefCOCOg val), yet given a correct
candidate the selector picks it for $0.90$ of RefCOCOg val and $0.92$ of Ref-L4
queries; on Ref-L4 its accuracy is bounded by the hook's top-five ceiling
($0.757$), and on the stress set it often answers ``none'' to present requests.
\Cref{sec:app-lora-matched} repeats this run with three seeds and on RefCOCOg
test, RefCOCO/+ val and gRefCOCO, against fine-tuning of the VLM's own box
format on the same rows and over the detector's own top five.

\begin{table*}[t]
\centering
\footnotesize
\setlength{\tabcolsep}{3pt}
\caption{VLM fine-tuned to select among the hook's top five (seed 42; all three seeds and the matched controls in \cref{tab:app-lora-matched,tab:app-lora-diffs}).
Accuracy at $\mathrm{IoU}\ge0.5$ on present requests; joint accuracy with the
swapped absent rows (``native'': abstain when ``none'' is the most likely
answer; ``tuned'': threshold on $1-P(\text{none})$ chosen on RefCOCOg val).
$\Delta$: paired difference on present requests with $95\%$ bootstrap interval.
Ref-L4: the fixed $3{,}000$-query subset of Table 3 of the main paper
(coordinate decoding $0.838$ there).}
\label{tab:app-e32}
\begin{tabular}{@{}lccc@{}}
\toprule
& RefCOCOg val & Ref-L4 & Stress\\
\midrule
Accuracy & 0.878 & 0.694 & 0.554\\
Top-five ceiling & 0.973 & 0.757 & 0.736\\
Accuracy given ceiling & 0.903 & 0.917 & 0.753\\
``None'' on present requests & 0.021 & 0.061 & 0.260\\
Joint accuracy, native / tuned & 0.899 / 0.901 & -- & 0.621 / 0.624\\
\midrule
$\Delta$ vs.\ zero-shot box list & $+0.118$\ci{.107}{.130} & $+0.118$\ci{.104}{.132} & $+0.030$\ci{$-$.026}{.091}\\
$\Delta$ vs.\ hook top-1 & $+0.091$\ci{.080}{.102} & $+0.243$\ci{.226}{.261} & $+0.165$\ci{.100}{.229}\\
$\Delta$ vs.\ coordinate decoding & $+0.013$\ci{.004}{.022} & $-0.144$\ci{$-$.161}{$-$.127} & $-0.069$\ci{$-$.126}{$-$.013}\\
\bottomrule
\end{tabular}
\end{table*}

\paragraph{A matched low-rank control for the fine-tuned selector.}
\label{sec:app-lora-matched}
\Cref{sec:app-e32} compares the fine-tuned selector with coordinate decoding
that received no task-specific training. To separate the answer interface from the
fine-tuning, we train three low-rank adaptations of Qwen2.5-VL-7B with exactly
that recipe (rank $16$ on the query, key, value and output projections, one
epoch, learning rate $2{\times}10^{-4}$, effective batch $64$, at most
$1{,}003{,}520$ pixels, loss on the answer tokens only) on the same $24$k
RefCOCOg training rows, and change only what the VLM is trained to answer.
(a)~\emph{Native box}: the coordinate-decoding prompt (\cref{sec:app-parse})
answered with the annotated box in the VLM's own format (absolute pixels of
its resized input frame). It uses the $16{,}697$ present rows of (b): the
$7{,}276$ swapped absent rows, which (b) answers with ``none'', have no answer
in this format, and the $27$ present rows that (b) drops for want of a correct
candidate are dropped here too. (a) and (b) thus see the same expressions and
images, and (a) takes $261$ optimizer steps against $374$ for (b). Its boxes are
decoded greedily with the parse rule of \cref{sec:app-parse} and scored raw
and snapped to the proposal of maximal IoU.
(b)~\emph{Index over the hook's top five}: the selector of
\cref{sec:app-e32}, unchanged ($23{,}973$ rows: $16{,}697$ present rows
answered with a candidate's number, $7{,}276$ swapped rows answered
``none''), now trained with seeds 42, 123 and 456 (seed 42 is the run of
\cref{tab:app-e32}).
(c)~\emph{Index over the detector's top five}, the candidate-region selection
setting \cite{crs2025} under matched perception. The detector is prompted with
the expression itself, as for detector-native selection, not with the fixed
category prompt that produces the proposals. Each of its $20$ highest-scoring
boxes (largest text logit) is replaced by the proposal of maximal IoU among the
same $300$ proposals and repeats are removed; suppression at IoU $0.7$, the cut
to five and the left-to-right numbering are those of (b). The training rows
are the same $24$k rows ranked in this way, without the $302$ present rows that
have no correct candidate ($23{,}698$ rows: $16{,}422$ present, $7{,}276$
``none''); one seed.
We evaluate (b) on every split of \cref{tab:app-lora-matched}, with $30\%$
swapped absent rows on RefCOCOg val and test and on the stress set, and with
the threshold on $1-P(\text{none})$ chosen per seed on RefCOCOg val and then
frozen; (a) on RefCOCOg val and test; (c) on RefCOCOg val and the Ref-L4
subset. On one RTX PRO 6000 shared with other jobs, so that wall-clock time
varies across seeds, a training run took $1.1$--$3.6$\,h for (a),
$2.2$--$3.2$\,h for (b) and $1.6$\,h for (c), with peak allocated memory of
$28.3$, $30.2$--$30.3$ and $30.2$\,GB.

\begin{table*}[t]
\centering
\footnotesize
\setlength{\tabcolsep}{4pt}
\caption{Matched low-rank control for the fine-tuned selector (Qwen2.5-VL-7B;
accuracy at $\mathrm{IoU}\ge0.5$ on present requests, a forced choice for the
selectors; mean$\pm$sample std over the number of seeds in column $S$). All
three fine-tuned variants train for one epoch on the same RefCOCOg rows with
the same recipe (\cref{sec:app-lora-matched}): (a) the coordinate-decoding
prompt answered with the annotated box, decoded raw or snapped to the proposal
of maximal IoU; (b) a candidate's number over the hook's top five; (c) the same
over the top five of the detector prompted with the expression. RefCOCO and
RefCOCO+ val and RefCOCOg test: all queries; Ref-L4: the fixed
$3{,}000$-query subset of Table 3 of the main paper. Hook, top-1: the ranking
hook's own choice. Ceiling rows: a correct box among the five candidates.
Coordinate decoding: the raw boxes of Table 1 of the main paper; same pipeline:
the decoding run of (a) without adapter. Zero-shot box lists are rerun in the
runtime of (b) and (c) (Table 3 of the main paper: $0.760$, $0.576$, $0.524$).
--: not evaluated.}
\label{tab:app-lora-matched}
\begin{tabular}{@{}lccccccc@{}}
\toprule
& & \multicolumn{2}{c}{RefCOCOg} & RefCOCO & RefCOCO+ & Ref-L4 & \\
\cmidrule(lr){3-4}
System & $S$ & val & test & val & val & subset & Stress\\
\midrule
Hook, top-1 & -- & 0.787 & 0.788 & 0.787 & 0.699 & 0.451 & 0.390\\
Correct box in hook's top 5 & -- & 0.973 & 0.973 & 0.961 & 0.943 & 0.757 & 0.736\\
Correct box in detector's top 5 & -- & 0.980 & -- & -- & -- & 0.898 & --\\
\midrule
Coordinate decoding, no fine-tuning & -- & 0.865 & 0.866 & 0.893 & 0.838 & 0.838 & 0.623\\
\quad same pipeline as (a) & -- & 0.865 & 0.868 & -- & -- & -- & --\\
Zero-shot box list, hook's top 5 & -- & 0.759 & 0.762 & -- & -- & 0.577 & 0.506\\
Zero-shot box list, detector's top 5 & -- & 0.730 & -- & -- & -- & 0.622 & --\\
\midrule
(a) Native box, fine-tuned, raw & 3 & 0.871\sd{.004} & 0.875\sd{.001} & -- & -- & -- & --\\
\quad snapped to a proposal & 3 & 0.871\sd{.005} & 0.875\sd{.001} & -- & -- & -- & --\\
(b) Index, hook's top 5, fine-tuned & 3 & 0.878\sd{.004} & 0.882\sd{.006} & 0.894\sd{.005} & 0.839\sd{.005} & 0.688\sd{.015} & 0.528\sd{.031}\\
(c) Index, detector's top 5, fine-tuned & 1 & 0.892 & -- & -- & -- & 0.763 & --\\
\bottomrule
\end{tabular}
\end{table*}

\begin{table*}[t]
\centering
\footnotesize
\setlength{\tabcolsep}{4pt}
\caption{Paired differences for \cref{tab:app-lora-matched} (accuracy at
$\mathrm{IoU}\ge0.5$ on present requests) with $95\%$ two-level
(seed $\times$ query) bootstrap intervals ($10{,}000$ resamples; queries drawn
once and shared, the seeds of each system drawn independently; a system with
one run contributes only its queries). Decoding: the raw boxes of Table 1 of
the main paper. $^\ast$(a) minus decoding in the same pipeline without
adapter. $^\dagger$The fine-tuned variant, (b) or (c), minus the zero-shot box list
over the same five candidates. Against the boxes of Table 1 instead, (a) gains
$+0.005$ and $+0.009$ and the difference of gains is $+0.114$ and $+0.112$.
On the stress set queries, not templates, are resampled. --: not evaluated.}
\label{tab:app-lora-diffs}
\begin{tabular}{@{}lcccc@{}}
\toprule
Difference & RefCOCOg val & RefCOCOg test & Ref-L4 subset & Stress\\
\midrule
(b) $-$ (a), raw & $+0.008$\ci{$-$.002}{.017} & $+0.007$\ci{$-$.001}{.015} & -- & --\\
(b) $-$ (a), snapped & $+0.007$\ci{$-$.002}{.017} & $+0.007$\ci{$-$.001}{.015} & -- & --\\
(b) $-$ decoding & $+0.013$\ci{.004}{.022} & $+0.016$\ci{.007}{.024} & $-0.150$\ci{$-$.172}{$-$.130} & $-0.095$\ci{$-$.157}{$-$.033}\\
Gain of (a)$^\ast$ & $+0.005$\ci{$-$.002}{.012} & $+0.006$\ci{.002}{.011} & -- & --\\
Gain of (b)$^\dagger$ & $+0.119$\ci{.107}{.131} & $+0.120$\ci{.111}{.130} & $+0.110$\ci{.090}{.129} & $+0.022$\ci{$-$.039}{.084}\\
Gain of (b) $-$ gain of (a) & $+0.114$\ci{.100}{.128} & $+0.114$\ci{.103}{.124} & -- & --\\
\midrule
(c) $-$ (b) & $+0.014$\ci{.006}{.022} & -- & $+0.076$\ci{.055}{.097} & --\\
(c) $-$ decoding & $+0.027$\ci{.018}{.035} & -- & $-0.075$\ci{$-$.089}{$-$.060} & --\\
Gain of (c)$^\dagger$ & $+0.162$\ci{.150}{.174} & -- & $+0.141$\ci{.124}{.158} & --\\
\bottomrule
\end{tabular}
\end{table*}

Fine-tuned on the same expressions, the native box and the index answer are
not distinguishable on RefCOCOg (\cref{tab:app-lora-diffs}): (b)$-$(a) is
$+0.008$ [$-0.002$, $+0.017$] on val and $+0.007$ [$-0.001$, $+0.015$] on
test, and matched seed pairs differ by $0.000$ to $0.013$ in favor of (b)
(exact McNemar $p\le0.03$ for two of the three pairs on each split).
Fine-tuning adds $0.119$ and $0.120$ to the box list but only $0.005$ and
$0.006$ to decoding run through the same pipeline; the gains differ by $0.114$
on both splits. Most of the gain from fine-tuning thus lies in learning to read
a candidate list, on which the VLM without fine-tuning falls $0.21$ short of its top-five ceiling,
whereas its own box format moves little. Over three seeds variant (b)
stays above coordinate decoding on RefCOCOg ($+0.013$ on val, $+0.016$ on test)
and is level with it on RefCOCO val ($+0.001$ [$-0.006$, $+0.008$]) and, in the
mean, on RefCOCO+ val ($0.839$ against $0.838$); native-box fine-tuning on the
same rows reaches $0.871$ and $0.875$, and the selector's lead over it is not
significant. On Ref-L4 and the stress set the selector stays below decoding ($-0.150$ and $-0.095$),
bounded by the hook's top five ($0.757$ and $0.736$): given a correct candidate
it picks it for $0.908$ of Ref-L4 and $0.718$ of stress queries, and it answers
``none'' to $0.19$ of present stress queries. Its seeds disagree most on these
two sets ($0.670$--$0.699$ and $0.494$--$0.554$).

Over the detector's own top five the same recipe reaches $0.892$ on RefCOCOg
val (one seed), $+0.014$ [$0.006$, $0.022$] over (b) and $+0.027$ [$0.018$,
$0.035$] over coordinate decoding, and $0.763$ on Ref-L4, $0.076$ above (b)
but $0.075$ below decoding. On Ref-L4 the lead over (b) comes from the
candidates: the expression-prompted detector has a correct box among its five
for $0.898$ of the queries against $0.757$ for the hook, while (c) picks a
correct candidate less often when one is present ($0.850$ against $0.908$).
The fine-tuned selector therefore does not need the hook's ranking; an
expression-prompted ranking serves at least as well. Without fine-tuning, the two
candidate sets order the other way on RefCOCOg ($0.730$ against $0.759$), and
fine-tuning adds $0.162$ over the detector's five against $0.119$ over the
hook's. A possible reason is that the hook ranks its own training rows almost
perfectly (top-1 $0.967$, top-five ceiling $0.998$; \cref{sec:app-e32}),
whereas the detector's ranking of the same rows (top-1 $0.807$, ceiling
$0.982$) is about as hard as at evaluation; we did not test this.

With its own ``none'' as the abstention, (b) reaches joint accuracy
$0.899$\sd{.003} on RefCOCOg val and $0.902$\sd{.004} on test with $30\%$
swapped absent rows ($0.900$ and $0.903$ with the tuned threshold), above the
$0.888$ of the verifier attached to coordinate-snap on the same rows
(\cref{tab:app-detach}), and $0.614$\sd{.015} on the stress set. The selector
was trained on swapped rows, as the verifier was. Of the swapped rows whose
donor image contains the referent's category ($524$ on val, $1{,}105$ on test)
it rejects $0.875$ and $0.892$ ($0.891$ and $0.903$ tuned), against $0.994$
and $0.991$ of the other swaps; joint accuracy over the present rows and this
hard subset is $0.871$ on val, against $0.857$ for the verifier and $0.796$ for
the VLM's yes/no answer on coordinate-snap (\cref{tab:app-hardneg}). On
gRefCOCO testB, with the threshold kept from RefCOCOg val, Pr@(F1$=$1) is
$0.447$\sd{.016} ($0.449$ tuned; N-acc $0.618$, T-acc $0.921$): above the
hook's veto ($0.412$) but below the verifier ($0.478$) and the VLM's yes/no
answer ($0.519$) on coordinate decoding (\cref{tab:app-grec-testb}). As for
the verifier, the advantage holds on the kind of negative it was trained on.

\section{Where the Hook's Selections Fail}
\label{sec:app-errors}

\cref{tab:app-errors} classifies the hook's wrong selections on RefCOCOg val and
test (three seeds pooled) by the overlap of the chosen proposal with the
referent and by the rank, under the hook's scores, of the first proposal that
covers the referent at $\mathrm{IoU}\ge0.5$. Near misses and selections of a
different object are about equally frequent; four in five errors keep a
covering proposal in the top five, and in $0.41$ ($0.42$ on test) of the errors
it is ranked second (median rank $3$). A covering proposal is missing for at
most $0.002$ of the queries, so retrieval is not what fails. Among the errors
that keep the referent in the top five, coordinate decoding is correct on
$0.67$--$0.68$ of the near misses and on $0.69$--$0.72$ of the other-object
selections.
The hook's error rate rises with the number of same-category distractors and
for small referents, but its accuracy gap to coordinate decoding stays within
$0.059$--$0.097$ across these buckets on RefCOCOg val ($0.079$ overall). Persons,
chairs and dining tables account for the most errors on both splits; among
categories with at least $50$ queries, dining tables have the highest error
rate on val ($0.435$) and the second highest on test.

\begin{table}[t]
\centering
\footnotesize
\setlength{\tabcolsep}{3pt}
\caption{Error taxonomy of the hook on RefCOCOg (present queries; shares of
all query--seed pairs over three seeds). Near miss: chosen proposal with
$0.1\le\mathrm{IoU}<0.5$ to the referent; other object: $\mathrm{IoU}<0.1$.
In top-5 / beyond top-5: rank, under the hook's scores, of the first proposal
covering the referent at $\mathrm{IoU}\ge0.5$. Coord.: coordinate decoding
(raw box). Distractors: other COCO instances of the referent's category in
the image; area: referent box relative to the image.}
\label{tab:app-errors}
\begin{tabular}{@{}lcccc@{}}
\toprule
& \multicolumn{2}{c}{RefCOCOg val} & \multicolumn{2}{c}{RefCOCOg test}\\
\cmidrule(lr){2-3}\cmidrule(l){4-5}
& Hook & Coord. & Hook & Coord.\\
\midrule
Error rate & 0.213 & 0.135 & 0.207 & 0.134\\
\quad near miss, in top-5 & 0.084 & -- & 0.082 & --\\
\quad other object, in top-5 & 0.087 & -- & 0.083 & --\\
\quad beyond top-5 & 0.042 & -- & 0.040 & --\\
\quad no covering proposal & 0.000 & -- & 0.002 & --\\
Ranked 2nd (share of errors) & 0.41 & -- & 0.42 & --\\
Ranked 2nd--5th (share) & 0.80 & -- & 0.80 & --\\
\midrule
\multicolumn{5}{@{}l}{\emph{Error rate by bucket}}\\
1 distractor & 0.193 & 0.124 & 0.187 & 0.122\\
2 distractors & 0.215 & 0.124 & 0.223 & 0.146\\
$\ge3$ distractors & 0.270 & 0.182 & 0.245 & 0.152\\
Area $2$--$10\%$ & 0.282 & 0.185 & 0.265 & 0.179\\
Area $10$--$30\%$ & 0.189 & 0.130 & 0.186 & 0.123\\
Area $\ge30\%$ & 0.159 & 0.067 & 0.166 & 0.090\\
\bottomrule
\end{tabular}
\end{table}

\section{The Verifier as a Detachable Abstention Layer}
\label{sec:app-abstention}

\paragraph{Protocol.}
Absent referents are created by pairing $30\%$ of the expressions of a split
with the image of another row (a fixed random draw) and re-encoding the query
with the VLM on that image; POPE carries its own absent objects. Target
accuracy is the fraction of present rows answered with a box at
$\mathrm{IoU}\ge0.5$, no-target accuracy the fraction of absent rows abstained
on, joint accuracy the accuracy over all rows and answer rate the fraction of
all rows (present and absent) answered. AURC is the area under the selective risk--coverage curve over all
rows, in which answering an absent row is an error; it is not comparable with
the hallucination--recall area of \cref{tab:app-pope}. ``+ verifier'' scores the
hook's verifier at the proposal returned by the named grounder, at the fixed
$\gamma_v$. ``VLM yes/no'' answers only if the VLM says that the referent is in
the image, asked \texttt{Is there "\{expression\}" in the image? Answer yes
or no.} (POPE: its own question followed by \texttt{Answer yes or no.});
``VLM yes-probability, tuned'' answers if the yes-minus-no score, the
log-probability of the yes tokens minus that of the no tokens at the first
answer position, is at least $-3.58$, and the detector score threshold
is $0.405$. These two thresholds maximize joint accuracy on RefCOCOg val with
absent referents and are frozen for every other split, so RefCOCOg test is
untouched by any choice (\cref{tab:app-tuning}).

\paragraph{Dependence on the absent rate.}
Joint accuracy weights target and no-target accuracy by the share of absent
referents. \Cref{tab:app-prior} reweights the RefCOCOg rows to other absent
rates (balanced accuracy equals the $50\%$ column) and gives target accuracy at
a fixed false-accept rate, with the threshold chosen on the absent rows of the
same split (an oracle operating point). When absences are rare the binary
yes/no gate costs more present referents than it saves: at $5\%$ it falls
below answering always. The verifier and the tuned yes-probability stay above
it at every rate, and at a $5\%$ false-accept rate the verifier keeps more
present referents than the VLM's yes-minus-no score ($0.860$ against
$0.834$).

\begin{table*}[t]
\centering
\footnotesize
\setlength{\tabcolsep}{2.4pt}
\caption{Joint accuracy of coordinate-snap gates and the hook on RefCOCOg val
(val) and test (test) when the absent rows are reweighted to an absent rate of
$5$--$50\%$; the actual rate is $30\%$. Last column: target accuracy at a
false-accept rate of at most $5\%$ on val (threshold chosen on the val absent
rows; the binary yes/no gate has a single point, false-accept rate $0.03$).
Hook rows: mean over three seeds.}
\label{tab:app-prior}
\begin{tabular}{@{}lccccc@{}}
\toprule
Absent rate & $5\%$ & $10\%$ & $30\%$ & $50\%$ & FAR${\le}5\%$\\
\midrule
Coordinate-snap, always answer (val / test) & 0.824 / 0.823 & 0.781 / 0.779 & 0.607 / 0.606 & 0.434 / 0.433 & --\\
\quad + VLM yes/no & 0.793 / 0.792 & 0.803 / 0.802 & 0.840 / 0.841 & 0.877 / 0.880 & 0.784\\
\quad + VLM yes-prob., tuned & 0.856 / 0.858 & 0.860 / 0.862 & 0.873 / 0.877 & 0.887 / 0.892 & 0.834\\
\quad + verifier & 0.861 / 0.861 & 0.866 / 0.866 & 0.888 / 0.888 & 0.909 / 0.910 & 0.860\\
\quad + verifier, hook's pick & 0.866 / 0.866 & 0.870 / 0.870 & 0.886 / 0.887 & 0.902 / 0.904 & 0.861\\
Hook, default veto & 0.714 / 0.724 & 0.727 / 0.737 & 0.780 / 0.788 & 0.833 / 0.840 & --\\
\bottomrule
\end{tabular}
\end{table*}

\paragraph{Transfer to an untouched split.}
\cref{tab:app-detach} gives both RefCOCOg splits. The thresholds chosen on val
transfer without loss: no system loses more than $0.004$ joint accuracy from
val to test (the detector threshold, $0.651\to0.648$), and on test the two
tuned thresholds are within $0.004$ of the best threshold chosen on test itself
(\cref{tab:app-tuning}).

\begin{table*}[t]
\centering
\footnotesize
\setlength{\tabcolsep}{3.4pt}
\caption{The verifier as a detachable abstention layer on RefCOCOg with $30\%$
swapped absent referents: the validation split, on which the two tuned
thresholds were chosen ($6{,}994$ rows), and the test split, untouched by any
choice ($13{,}717$ rows). Target: present rows answered with a box at
$\mathrm{IoU}\ge0.5$; No-tgt.: absent rows abstained on; Ans.: answer rate, the
fraction of all rows (present and absent) answered;
AURC over all rows (lower is better). Rows with the verifier or the hook: mean
over three seeds with the sample std of joint accuracy; the sample std of the
other columns is below $0.03$. Refusal gate: a fixed rule, so no AURC (--);
all other rows use Qwen2.5-VL.}
\label{tab:app-detach}
\begin{tabular}{@{}lcccccccccc@{}}
\toprule
& \multicolumn{5}{c}{RefCOCOg val + absent (tuning split)} & \multicolumn{5}{c}{RefCOCOg test + absent (untouched)}\\
\cmidrule(lr){2-6}\cmidrule(l){7-11}
System & Joint & Target & No-tgt. & Ans. & AURC & Joint & Target & No-tgt. & Ans. & AURC\\
\midrule
Coordinate-snap, always answer & 0.607 & 0.867 & 0.000 & 1.000 & 0.393 & 0.606 & 0.866 & 0.000 & 1.000 & 0.394\\
\quad + VLM yes/no & 0.840 & 0.784 & 0.970 & 0.626 & 0.134 & 0.841 & 0.782 & 0.978 & 0.624 & 0.135\\
\quad + VLM yes-probability, tuned & 0.873 & 0.853 & 0.921 & 0.708 & 0.134 & 0.877 & 0.854 & 0.929 & 0.708 & 0.135\\
\quad + verifier & 0.888\sd{.002} & 0.856 & 0.962 & 0.696 & 0.148 & 0.888\sd{.002} & 0.856 & 0.963 & 0.695 & 0.152\\
\quad + verifier at the hook's proposal & 0.886\sd{.001} & 0.862 & 0.942 & 0.712 & 0.161 & 0.887\sd{.001} & 0.861 & 0.947 & 0.711 & 0.168\\
\quad + VLM yes/no and verifier & 0.839\sd{.001} & 0.779 & 0.979 & 0.618 & 0.146 & 0.841\sd{.001} & 0.779 & 0.987 & 0.615 & 0.149\\
\midrule
Detector-native, score threshold, tuned & 0.651 & 0.731 & 0.467 & 0.790 & 0.236 & 0.648 & 0.734 & 0.446 & 0.798 & 0.239\\
Detector-native + verifier & 0.832\sd{.002} & 0.776 & 0.960 & 0.692 & 0.210 & 0.831\sd{.002} & 0.775 & 0.962 & 0.693 & 0.211\\
\midrule
Hook, default veto & 0.780\sd{.014} & 0.700 & 0.967 & 0.605 & 0.234 & 0.788\sd{.012} & 0.711 & 0.968 & 0.609 & 0.224\\
Hook, verifier only & 0.830\sd{.004} & 0.783 & 0.942 & 0.712 & 0.246 & 0.837\sd{.006} & 0.790 & 0.947 & 0.711 & 0.239\\
\midrule
\multicolumn{11}{@{}l}{\emph{Refusal gate: abstain when the answer has no parsable box}}\\
Qwen2.5-VL, coordinate-snap & 0.659 & 0.867 & 0.172 & 0.948 & -- & 0.658 & 0.866 & 0.174 & 0.948 & --\\
Qwen3-VL, raw box & 0.815 & 0.886 & 0.649 & 0.804 & -- & 0.819 & 0.890 & 0.655 & 0.802 & --\\
\bottomrule
\end{tabular}
\end{table*}

\begin{figure}[t]
\centering
\includegraphics[width=\columnwidth]{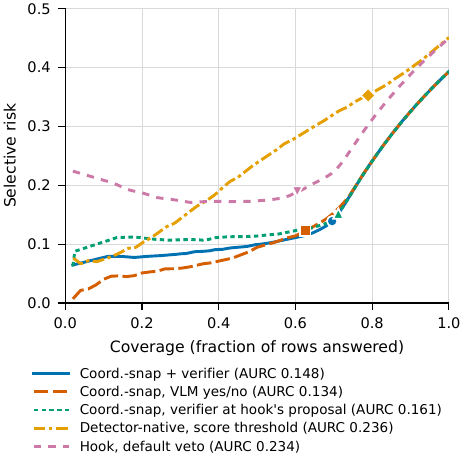}
\caption{Selective risk against coverage (the answer rate over all rows) on
RefCOCOg val with $30\%$ swapped absent referents (risk over answered rows; answering an absent row is an
error). Each curve orders the rows by the system's own confidence (verifier
probability, the VLM's yes-minus-no score, detector score); markers are the
operating points at the fixed or tuned thresholds. Curves involving the
verifier or the hook are averaged over three seeds; the legend gives the AURC.}
\label{fig:app-risk-coverage}
\end{figure}

\paragraph{The verifier against the VLM's own answer.}
Scoring the verifier at the coordinate-snap proposal gives the highest joint
accuracy on both RefCOCOg splits ($0.888$), but the VLM's own yes-probability,
thresholded once on val, trails it by only $0.015$ on val and $0.011$ on test
($0.873$ and $0.877$) and is far better on POPE ($0.932$, $0.897$ and $0.859$
against $0.697$, $0.695$ and $0.686$; \cref{tab:app-detach-ood}). Without a
threshold, the VLM's yes-minus-no score ranks rows better than the verifier probability:
AURC $0.134$ against $0.148$ on val and $0.135$ against $0.152$ on test, with a
lower risk at every coverage below about $0.55$ (\cref{fig:app-risk-coverage}).
The verifier's lead at the fixed operating points therefore comes from where it
operates: it keeps more present referents (target accuracy $0.856$ against
$0.784$ for the yes/no answer) at a slightly lower no-target accuracy ($0.962$
against $0.970$). Requiring both the yes/no answer and the verifier gives the
highest no-target accuracy of the coordinate-snap gates ($0.979$ on val,
$0.987$ on test) but also their lowest target accuracy, so its joint accuracy
stays at the level of the yes/no answer alone. On the stress set, gating on
the verifier at the hook's proposal gives the highest joint accuracy
($0.580$), ahead of the tuned yes-probability ($0.564$) and the yes/no answer
($0.545$). These stress-set differences are not established: paired tests do
not separate them (\cref{tab:app-tests}); moreover, a third of the stress set's absent labels are
questionable (see below).

\begin{table*}[t]
\centering
\footnotesize
\setlength{\tabcolsep}{4pt}
\caption{The systems of \cref{tab:app-detach} with the same frozen thresholds on
the reasoning-stress set with $30\%$ swapped absent referents ($330$ rows) and
on POPE, where joint accuracy is existence accuracy (a present object counts as
correct when the system answers). Rows with the verifier or the hook: mean over
three seeds with the sample std of joint accuracy; the sample std of the other
columns is below $0.05$. The POPE accuracy of the VLM yes/no gate differs from
the direct-answer run of \cref{tab:app-pope} by at most $0.001$.}
\label{tab:app-detach-ood}
\begin{tabular}{@{}lcccccc@{}}
\toprule
& \multicolumn{3}{c}{Stress + absent} & \multicolumn{3}{c}{POPE}\\
\cmidrule(lr){2-4}\cmidrule(l){5-7}
System & Joint & No-tgt. & AURC & Random & Popular & Adversarial\\
\midrule
Coordinate-snap, always answer & 0.427 & 0.000 & 0.573 & 0.500 & 0.500 & 0.500\\
\quad + VLM yes/no & 0.545 & 0.606 & 0.412 & 0.883 & 0.874 & 0.864\\
\quad + VLM yes-prob., tuned & 0.564 & 0.475 & 0.412 & 0.932 & 0.897 & 0.859\\
\quad + verifier & 0.528\sd{.042} & 0.923 & 0.377 & 0.697\sd{.033} & 0.695\sd{.034} & 0.686\sd{.030}\\
\quad + verifier at hook's proposal & 0.580\sd{.031} & 0.859 & 0.394 & 0.806\sd{.026} & 0.802\sd{.024} & 0.763\sd{.015}\\
\quad + VLM yes/no and verifier & 0.482\sd{.035} & 0.923 & 0.397 & 0.693\sd{.034} & 0.692\sd{.034} & 0.692\sd{.033}\\
\midrule
Det.-native, score threshold, tuned & 0.276 & 0.495 & 0.794 & 0.795 & 0.786 & 0.752\\
Det.-native + verifier & 0.390\sd{.014} & 0.916 & 0.699 & 0.715\sd{.035} & 0.711\sd{.036} & 0.696\sd{.027}\\
\midrule
Hook, default veto & 0.413\sd{.014} & 0.902 & 0.560 & 0.599\sd{.031} & 0.600\sd{.034} & 0.594\sd{.028}\\
Hook, verifier only & 0.478\sd{.031} & 0.859 & 0.586 & 0.806\sd{.026} & 0.802\sd{.024} & 0.763\sd{.015}\\
\bottomrule
\end{tabular}
\end{table*}

\paragraph{Hard negatives.}
A swapped expression can name an object whose category also occurs in the
donor image: $524$ of the $2{,}098$ swapped rows of RefCOCOg val ($0.250$) are
such category-present hard negatives. The VLM's yes/no answer accepts $0.092$
of them against $0.009$ of the category-absent swaps (Fisher exact
$p=1.7\times10^{-18}$); the detector score threshold accepts $0.807$ against
$0.442$. The VLM still rejects hard negatives more often than the verifier
($0.908$ against $0.869$ no-target accuracy; \cref{tab:app-hardneg}), while the
verifier keeps more present referents, so its joint accuracy remains higher on
the hard subset ($0.857$ against $0.796$). The highest-margin hard negatives
accepted by the VLM include several donor images that do contain an object
matching the expression, so part of this subset is label noise. On the stress
set the problem is larger: $35$ of its $99$ swapped rows ($25$ of the $33$
negation swaps) have a donor image that satisfies the query template under the
set's own construction rules, so their absent label is questionable; the VLM's
yes/no answer accepts $0.886$ of these rows and $0.125$ of the others.

\begin{table}[t]
\centering
\footnotesize
\setlength{\tabcolsep}{2.6pt}
\caption{Hard negatives among the $2{,}098$ swapped absent rows of RefCOCOg
val. Present: the donor image contains an object of the referent's COCO
category ($n{=}524$); absent: it does not ($n{=}1{,}574$). No-target: absent
rows abstained on. Joint: accuracy over the $4{,}896$ present rows and the
absent rows of the subset. Rows with the verifier or the hook: mean over three
seeds (sample std $\le0.019$).}
\label{tab:app-hardneg}
\begin{tabular}{@{}lccccc@{}}
\toprule
& \multicolumn{3}{c}{No-target} & \multicolumn{2}{c}{Joint}\\
\cmidrule(lr){2-4}\cmidrule(l){5-6}
System & All & Present & Absent & Present & Absent\\
\midrule
Coord.-snap + VLM yes/no & 0.970 & 0.908 & 0.991 & 0.796 & 0.834\\
Coord.-snap + yes-prob., tuned & 0.921 & 0.807 & 0.959 & 0.848 & 0.879\\
Coord.-snap + verifier & 0.962 & 0.869 & 0.993 & 0.857 & 0.889\\
Coord.-snap + yes/no + verifier & 0.979 & 0.929 & 0.996 & 0.794 & 0.832\\
Det.-native, score threshold & 0.467 & 0.193 & 0.558 & 0.679 & 0.689\\
Det.-native + verifier & 0.960 & 0.869 & 0.991 & 0.785 & 0.829\\
Hook, default veto & 0.967 & 0.889 & 0.993 & 0.718 & 0.771\\
\bottomrule
\end{tabular}
\end{table}

\paragraph{Verifier trained on harder negatives.}
\label{sec:app-hardneg-training}
Would harder negatives in training close the verifier's gap to the VLM? We
retrain the hook and its verifier with the recipe of the main paper
($24{,}000$ RefCOCOg training rows, $7{,}276$ of them swapped absent rows) and
change only the negatives, adding $7{,}276$ rows of one of two kinds to the
swaps (``+'') or using them instead of the swaps (``only''). A
\emph{same-image} negative scores a present expression at the gold proposal of
another expression of the same image whose referent it does not describe (that
gold box and the proposal each overlap the expression's own gold box with IoU
$<0.5$; $9{,}535$ such pairs over $1{,}311$ images); it supervises only the
verifier. A \emph{gRefCOCO} negative is a no-target expression of
gRefCOCO's training split with its own detector and VLM features, used as an
absent row; none of these images is a RefCOCOg val or test image. All
variants, including a baseline with the standard swaps, are trained with three
seeds in one run; their thresholds are chosen on RefCOCOg val with swaps and
frozen for RefCOCOg test and gRefCOCO testB (\cref{tab:app-hardneg-train}).

gRefCOCO negatives added to the swaps raise the verifier's no-target accuracy
on testB from $0.646$ to $0.821$, close to the VLM's yes/no answer ($0.840$),
and leave RefCOCOg joint accuracy unchanged ($0.887$ against $0.888$). They do
not raise the official score: the retrained verifier also rejects more
expressions that do have a target (target accuracy $0.728$ against $0.893$),
so Pr@(F1$=$1) stays at $0.475$, level with the tuned yes-probability and
below the yes/no answer ($0.519$). Using only gRefCOCO negatives moves further
along the same trade-off ($0.931$ no-target and $0.354$ target accuracy on
testB) and loses RefCOCOg's category-present swaps ($0.601$ rejected against
$0.895$). Same-image negatives help on neither RefCOCOg nor gRefCOCO; added to
the swaps they lower the rejection of category-present swaps from $0.895$ to
$0.754$. On these hard
negatives the standard verifier already rejects more than the tuned
yes-probability ($0.822$), and neither kind of harder negative raises that
rate. Matching the training negatives to gRefCOCO thus shifts the verifier's
operating point toward rejection rather than improving its ranking of present
and absent referents. The stress set ($231$ present queries, $99$ swapped
absent rows, a third of them questionable) orders the variants differently:
adding either kind of harder negative to the swaps raises its joint accuracy
($0.559$ and $0.544$ against $0.512$), and gRefCOCO negatives alone reach
$0.582$, level with the tuned yes-probability ($0.567$), whereas same-image
negatives alone do not help ($0.496$); on a set this small we do not read more into the difference.

\begin{table}[t]
\centering
\footnotesize
\setlength{\tabcolsep}{2.4pt}
\caption{Verifier retrained with harder negatives (coordinate-snap +
verifier; mean over three seeds, sample std $\le0.056$). RefCOCOg test with
$30\%$ swapped absent rows: joint accuracy, and no-target accuracy on the
$1{,}105$ swaps whose image contains an object of the referent's category
(``Hard''). Stress: joint accuracy with its swapped absent rows. gRefCOCO testB: Pr@(F1$=$1) (``Pr@1''), N-acc and T-acc. Thresholds chosen
on RefCOCOg val per variant. All rows use the earlier feature build (\cref{sec:app-builds}),
so the first row differs from
them by up to $0.02$. ${}^\ast$From \cref{tab:app-grec-testb}; its RefCOCOg
values are in \cref{tab:app-detach}.}
\label{tab:app-hardneg-train}
\begin{tabular}{@{}lcccccc@{}}
\toprule
& \multicolumn{2}{c}{RefCOCOg test} & & \multicolumn{3}{c}{gRefCOCO testB}\\
\cmidrule(lr){2-3}\cmidrule(l){5-7}
Training negatives & Joint & Hard & Stress & Pr@1 & N-acc & T-acc\\
\midrule
Swaps (main recipe) & \textbf{0.888} & \textbf{0.895} & 0.512 & 0.479 & 0.646 & 0.893\\
Swaps + same-image & 0.853 & 0.754 & 0.559 & 0.434 & 0.571 & 0.810\\
Same-image only & 0.759 & 0.580 & 0.496 & 0.390 & 0.542 & 0.657\\
Swaps + gRefCOCO & 0.887 & 0.870 & 0.544 & 0.475 & 0.821 & 0.728\\
gRefCOCO only & 0.853 & 0.601 & \textbf{0.582} & 0.375 & \textbf{0.931} & 0.354\\
\midrule
VLM yes-prob., tuned & 0.876 & 0.822 & 0.567 & 0.477 & 0.622 & \textbf{0.929}\\
VLM yes/no$^\ast$ & -- & -- & -- & \textbf{0.519} & 0.840 & 0.790\\
\bottomrule
\end{tabular}
\end{table}

\paragraph{Verifier supervision.}
\label{sec:app-labels}
Table~\ref{tab:app-labels} compares the existence label of the main paper
with the correctness label, each trained jointly with the hook or after the
fact on a frozen hook trained without the verifier, under both gates. The
existence label was fixed before this comparison; with the verifier gate the
correctness label is level on joint accuracy and better on AURC and POPE, so
the existence-label results of the main paper are a conservative estimate for
a verifier of this form. Attaching the verifier to another grounder requires
that grounder's box to be snapped to one of the detector's proposals, i.e.\ one
pass of the frozen detector per image in addition to the grounder.

\begin{table*}[htbp]
\centering
\footnotesize
\setlength{\tabcolsep}{5pt}
\caption{Verifier supervision (three seeds). Joint accuracy on RefCOCOg val
with $30\%$ swapped absent referents and its AURC (all rows), and POPE
accuracy. ``Post hoc'': verifier head trained on a frozen hook without
faithfulness training.}
\label{tab:app-labels}
\begin{tabular}{@{}llccccc@{}}
\toprule
Label & Gate & Val joint & Val AURC & Random & Popular & Advers.\\
\midrule
Existence & veto & $0.780$ & $0.234$ & $0.599$ & $0.600$ & $0.594$\\
Existence & verifier & $0.830$ & $0.246$ & $0.806$ & $0.802$ & $0.763$\\
Correctness & veto & $0.678$ & $0.173$ & $0.582$ & $0.581$ & $0.580$\\
Correctness & verifier & $0.832$ & $0.169$ & $0.886$ & $0.862$ & $0.820$\\
Existence, post hoc & veto & $0.807$ & $0.234$ & $0.700$ & $0.687$ & $0.676$\\
Existence, post hoc & verifier & $0.822$ & $0.233$ & $0.793$ & $0.766$ & $0.749$\\
Correctness, post hoc & veto & $0.808$ & $0.186$ & $0.686$ & $0.675$ & $0.667$\\
Correctness, post hoc & verifier & $0.824$ & $0.184$ & $0.756$ & $0.734$ & $0.723$\\
\bottomrule
\end{tabular}
\end{table*}

\paragraph{What the verifier reads.}
\label{sec:app-verifier-inputs}
The verifier sees the projected query, the projected selected region and
their affinity, and a swapped absent request's query is re-encoded with the
image it is asked about. \Cref{tab:app-verifier-inputs} retrains the
verifier head after the fact on each published hook (frozen, three seeds) with
one input at a time, on the same training rows and swaps, and adds a logistic
probe on the VLM's raw query state $u$ that never sees the hook. The query
alone carries the signal: with $\tilde u$ only, or with the raw hidden state,
the separation of present from absent requests equals that of the full
verifier on RefCOCOg val, on the untouched test split and on its same-category
hard negatives, and exceeds it on the stress set, whereas the region alone
loses $0.06$ AUROC on RefCOCOg and $0.12$ on the hard negatives, and the
affinity alone is in between. The verifier is therefore, in effect, an
existence probe on the VLM's image-conditioned hidden state, as in
hidden-state hallucination detection \cite{halp}; the region it is attached to
matters little, which is also why it transfers to proposals chosen by another
grounder. Probes on the final position after $3/4$ and $1/2$ of the layers,
trained and scored on image- and text-disjoint halves of the full training
split, reach held-out AUROC $0.996$ and $0.993$ against $0.995$ for the final
layer.

\begin{table}[t]
\centering
\footnotesize
\setlength{\tabcolsep}{2.2pt}
\caption{Verifier inputs (Qwen2.5-VL, $24$k hooks, three seeds). AUROC of
present against swapped absent requests ($30\%$) and joint accuracy at the
threshold tuned on RefCOCOg val (verifier only). Hard: same-category swaps.
``Probe on $u$'': logistic regression on the raw query state, no hook.}
\label{tab:app-verifier-inputs}
\begin{tabular}{@{}lccccc@{}}
\toprule
& \multicolumn{2}{c}{RefCOCOg val} & \multicolumn{2}{c}{RefCOCOg test} & Stress\\
\cmidrule(lr){2-3}\cmidrule(lr){4-5}
Input & AUROC & Joint & AUROC & Hard AUROC & AUROC\\
\midrule
Full $(\tilde u,\tilde e_\sigma,s_\sigma)$ & 0.993 & 0.833 & 0.994 & 0.982 & 0.860\\
Query $\tilde u$ only & 0.993 & 0.831 & 0.995 & 0.983 & 0.894\\
Region $\tilde e_\sigma$ only & 0.935 & 0.757 & 0.931 & 0.863 & 0.692\\
Affinity $s_\sigma$ only & 0.978 & 0.812 & 0.978 & 0.935 & 0.715\\
Probe on $u$ (no hook) & 0.993 & 0.833 & 0.994 & 0.982 & 0.916\\
\bottomrule
\end{tabular}
\end{table}

\paragraph{The query carries the referent's location.}
\label{sec:app-query-location}
The query is read at the position that would begin the VLM's answer to the
bare request, without a grounding instruction (\cref{tab:repro_config}).
A ridge regression from $u$ to the referent's box (center, log width and log
height), fitted on the present training rows and scored on held-out splits,
shows that this state nonetheless encodes where the referent is
(\cref{tab:app-query-location}): it explains about two thirds of the variance of the
box center on RefCOCOg, against about $0.06$ for the text-only query of the
same VLM, and snapping its box to the nearest proposal already selects the
referent for $0.56$ of the requests, against $0.79$ for the hook. Part of what
the image-conditioned query contributes to selection is thus the VLM's own
localization; the hook reads it out against the region features more
accurately than a linear map, and on Ref-L4 and the stress set both the
encoding and the hook are weaker.

\begin{table}[t]
\centering
\footnotesize
\setlength{\tabcolsep}{3pt}
\caption{Linear probe from the query to the referent's box (ridge, fitted on
the $24$k training rows). $R^2$ of the box center (mean over $x$ and $y$) and
of the log size (mean over width and height); ``Snap'': accuracy at
$\mathrm{IoU}\ge0.5$ of the proposal of maximal IoU with the probe's box;
hook: three-seed accuracy.}
\label{tab:app-query-location}
\begin{tabular}{@{}llcccc@{}}
\toprule
Split & Query & $R^2$ center & $R^2$ size & Snap & Hook\\
\midrule
RefCOCOg val & image-cond. & 0.685 & 0.577 & 0.564 & 0.787\\
 & text-only & 0.056 & 0.211 & 0.188 & \\
RefCOCOg test & image-cond. & 0.672 & 0.570 & 0.563 & 0.793\\
 & text-only & 0.055 & 0.211 & 0.189 & \\
Ref-L4 val & image-cond. & 0.413 & 0.268 & 0.247 & 0.464\\
Stress & image-cond. & 0.434 & $-0.072$ & 0.147 & 0.384\\
\bottomrule
\end{tabular}
\end{table}

\paragraph{Trained controls on existence probes.}
The learned bridge and the index scorer have no verifier and are gated at
their own confidence thresholds. On POPE the bridge reaches $0.560$--$0.582$
existence accuracy with hallucination $0.36$--$0.40$, and the index scorer
$0.490$--$0.494$ with hallucination $0.91$--$0.92$ (three seeds): their
retrieval or selection confidence carries little existence signal.

\section{Reasoning-Stress Set: Construction and Audit}
\label{sec:app-stress}

\paragraph{Construction and rejection protocol.}
From COCO val2017 instance annotations we semi-automatically generated
$1{,}323$ candidate queries across three categories: \emph{negation} (an image
with $\ge\!2$ people where exactly one has no spatial overlap with any instance
of a named accessory category), \emph{affordance} (an image containing exactly
one instance of an affordance-mapped category and no sibling category in the
same affordance group), and \emph{world-knowledge state} (a state-dependent
selection derivable from annotation geometry, e.g.\ an unoccupied chair among
images that also contain an occupied one). Each candidate records the image,
query, category, construction rule, ground-truth box and distractor count.
Rule-based filtering reduced this to $323$ candidates, which were manually
reviewed by annotators who are not authors of this paper and saw no model's
predictions, against three criteria---the ground-truth box matches the query, the
query has a unique correct referent, and the query reads as natural,
unambiguous language---rejecting $92$ and keeping $231$ approved queries
(affordance $125$, negation $71$, world-knowledge state $35$). All $231$ are
evaluated under the same thresholds and $M{=}300$ budget as every other split;
the query list and rejection log are in the accompanying code archive.

The query templates are fixed by the generation rules rather than chosen per
image: one negation template for each COCO accessory category whose wearing,
carrying or holding can be decided from box containment (tie, backpack,
handbag, umbrella, cell phone, skis, surfboard, tennis racket), one affordance
template for each of ten groups of COCO categories that share a function
(\cref{tab:app-stress-templates}), and one state template for each seat
category (chair, bench, couch) and each rideable category (horse, elephant,
bicycle, motorcycle). These are all $25$ templates the generator defines. Each
occurs among the $323$ candidates and keeps at least one approved query, so no
template was removed, by the manual review or at any later stage. The template
list was last changed (four accessory and two rideable categories added to
cover more categories) before the candidates were generated and reviewed, and before
any model was evaluated on the set.

\paragraph{Image overlap with training data.}
All $218$ stress images are COCO val2014 images (the val2017 images whose
annotations built the set are a subset of val2014); RefCOCO, RefCOCO+ and RefCOCOg
draw their images from COCO train2014. None of the stress images is among the
$9{,}948$ images of the $24$k training rows, the $21{,}899$ images of the full
training set, or any image of any RefCOCO-family split ($30{,}956$ images), so
neither the trained heads nor the detector's RefCOCO-family training data have
seen them. The pre-training data of the VLM and the detector beyond RefCOCO are
not documented and could not be audited.

\paragraph{Audit.}
The $231$ approved queries use $25$ distinct query strings, one per
construction template ($10$ affordance, $8$ negation and $7$ world-knowledge
templates), over $218$ images; $12$ images carry more than one query (at most
three). Manual verification rejected $15$ of $140$ affordance, $48$ of $119$
negation and $29$ of $64$ world-knowledge candidates. One negation template
contains a grammatical slip (``the person not wearing a skis'', three
queries). Because queries that share a template are not independent,
\cref{tab:app-stress} adds intervals that resample templates. They are wider
than query-level intervals (for the hook on all queries $[0.297, 0.470]$
against $[0.330, 0.439]$). On all queries and on affordance the template
intervals of the hook and coordinate decoding do not overlap; on negation they
overlap, and on world knowledge (seven templates, $35$ queries) every interval
is wide. Coordinate decoding is above the hook (three-seed mean) on $22$ of the
$25$ templates, level on one and below on two. Restricted to the $14$ templates
with at least ten approved queries (eight affordance, five negation and one
world-knowledge template; $189$ queries), coordinate decoding is above the hook
on all $14$; the template where the two are level and the two where the hook is
ahead have at most five queries each. \Cref{tab:app-stress-templates} gives every
template. Affordance templates whose object is small or held (``something to
eat with'', ``something to brush teeth with'') are the hardest for every
system, and the detector's own interface fails outright on several of them.

\begin{table*}[t]
\centering
\scriptsize
\setlength{\tabcolsep}{5pt}
\caption{Reasoning-stress set by query template: accuracy at
$\mathrm{IoU}\ge0.5$, no abstention, with $95\%$ percentile-bootstrap intervals
over queries ($10{,}000$ resamples). Hook: three-seed mean; Coord.: coordinate
decoding (raw box); Det.: detector-native. Templates with fewer than five
queries are listed for completeness; their intervals are uninformative.}
\label{tab:app-stress-templates}
\begin{tabular}{@{}lrccc@{}}
\toprule
Template & $n$ & Hook & Coord. & Det.\\
\midrule
\multicolumn{5}{@{}l}{\emph{Affordance}}\\
something that tells the time & 16 & 0.31\ci{.12}{.52} & 0.69\ci{.44}{.88} & 0.44\ci{.19}{.69}\\
something to cut with & 15 & 0.24\ci{.07}{.44} & 0.53\ci{.27}{.80} & 0.00\ci{.00}{.00}\\
something to eat with & 15 & 0.13\ci{.02}{.27} & 0.33\ci{.13}{.60} & 0.00\ci{.00}{.00}\\
something to ride & 15 & 0.42\ci{.24}{.62} & 0.60\ci{.33}{.87} & 0.40\ci{.13}{.67}\\
something to sit on & 15 & 0.47\ci{.24}{.69} & 0.60\ci{.33}{.87} & 0.27\ci{.07}{.47}\\
something to heat food with & 14 & 0.36\ci{.19}{.52} & 0.86\ci{.64}{1.00} & 0.21\ci{.00}{.43}\\
something to read & 13 & 0.18\ci{.00}{.38} & 0.77\ci{.54}{1.00} & 0.31\ci{.08}{.54}\\
something to drink from & 10 & 0.60\ci{.33}{.83} & 0.90\ci{.70}{1.00} & 0.70\ci{.40}{1.00}\\
something to brush teeth with & 8 & 0.00\ci{.00}{.00} & 0.50\ci{.12}{.88} & 0.00\ci{.00}{.00}\\
something to dry hair with & 4 & 0.33\ci{.00}{.75} & 0.75\ci{.25}{1.00} & 0.25\ci{.00}{.75}\\
\multicolumn{5}{@{}l}{\emph{Negation}}\\
the person not wearing a backpack & 15 & 0.49\ci{.29}{.69} & 0.60\ci{.33}{.87} & 0.40\ci{.13}{.67}\\
the person not carrying a handbag & 13 & 0.38\ci{.21}{.59} & 0.85\ci{.62}{1.00} & 0.31\ci{.08}{.54}\\
the person not holding a cell phone & 12 & 0.69\ci{.44}{.92} & 0.75\ci{.50}{1.00} & 0.25\ci{.00}{.50}\\
the person not holding a tennis racket & 12 & 0.53\ci{.28}{.75} & 0.58\ci{.33}{.83} & 0.08\ci{.00}{.25}\\
the person not wearing a tie & 12 & 0.72\ci{.50}{.89} & 0.75\ci{.50}{1.00} & 0.33\ci{.08}{.58}\\
the person not wearing a skis & 3 & 0.67\ci{.00}{1.00} & 1.00\ci{1.00}{1.00} & 0.33\ci{.00}{1.00}\\
the person not holding a surfboard & 2 & 0.33\ci{.00}{.67} & 0.50\ci{.00}{1.00} & 0.50\ci{.00}{1.00}\\
the person not holding an umbrella & 2 & 1.00\ci{1.00}{1.00} & 1.00\ci{1.00}{1.00} & 0.00\ci{.00}{.00}\\
\multicolumn{5}{@{}l}{\emph{World knowledge}}\\
the unoccupied chair & 12 & 0.14\ci{.00}{.33} & 0.25\ci{.00}{.50} & 0.17\ci{.00}{.42}\\
the motorcycle without a rider & 8 & 0.12\ci{.00}{.38} & 0.25\ci{.00}{.62} & 0.00\ci{.00}{.00}\\
the unoccupied bench & 5 & 0.13\ci{.00}{.40} & 0.00\ci{.00}{.00} & 0.00\ci{.00}{.00}\\
the unoccupied couch & 5 & 0.87\ci{.60}{1.00} & 0.80\ci{.40}{1.00} & 0.80\ci{.40}{1.00}\\
the elephant without a rider & 2 & 0.33\ci{.00}{.67} & 0.50\ci{.00}{1.00} & 0.00\ci{.00}{.00}\\
the horse without a rider & 2 & 0.67\ci{.33}{1.00} & 1.00\ci{1.00}{1.00} & 0.50\ci{.00}{1.00}\\
the bicycle without a rider & 1 & 0.00\ci{.00}{.00} & 1.00\ci{1.00}{1.00} & 0.00\ci{.00}{.00}\\
\bottomrule
\end{tabular}
\end{table*}

\begin{figure}[t]
\centering
\includegraphics[width=0.95\columnwidth]{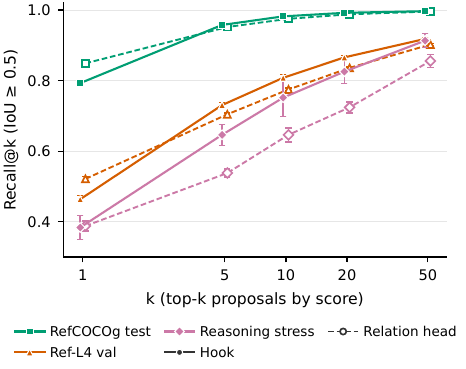}
\caption{\textbf{Recall@$k$ of a correct proposal} under the hook (solid) and the
relation-aware head (dashed), mean over three seeds with sample-std bars ($24$k
budget, Qwen2.5-VL). On RefCOCOg test the correct proposal is almost always in
the top five; on Ref-L4 and the stress set it often is not. RefCOCOg val is not
drawn: it differs from test by at most $0.006$ at every $k$ for both heads.
Full-budget recall@5 for both VLMs is in \cref{tab:app-full-iou75}.}
\label{fig:app-topk}
\end{figure}

\section{Public Reasoning and Rejection Benchmarks}
\label{sec:app-public}
The frozen pipeline, with every threshold kept as chosen on RefCOCOg val, is
evaluated on two public sets (\cref{tab:app-public}); nothing is tuned on them.
\emph{GroundingME} \cite{groundingme} ($1{,}005$ requests on high-resolution
images: $204$ discriminative, $300$ spatial, $300$ limited-view, $201$
rejection requests with no referent, correct only if the system abstains;
research-use license) is scored with its official Accuracy@0.5, which equals
our joint accuracy. Its gold boxes are partly seeded by Grounding DINO ($654$ of
the $804$ present referents, corrected by annotators; the other $150$ are
manual), which favors proposal-based systems: proposal recall at IoU $0.5$ is
$0.85$ on the seeded and $0.56$ on the manual boxes. \emph{RefCOCOg\_rej}
(released with GroundingME) edits an attribute of a RefCOCOg expression so that
it no longer describes any object; we use its $729$ negatives on RefCOCOg test
images together with their $729$ original expressions on the same images, so
that pair accuracy counts a pair as correct when the original is grounded and
the edit rejected. The detector, prompt and query extraction are those of every
other cache; the relation-aware head, token probe and combined selector, which
save no checkpoints, are retrained with the published recipe and reproduce the
published RefCOCOg val accuracies within $0.003$ per seed. GroundingME's own
prompt, which allows a null answer, reproduces the published Qwen2.5-VL-7B
score ($0.156$ at our input resolution, $0.178$ at the checkpoint default,
against $15.1\%$) and Qwen3-VL-8B ($0.290$ against $31.0\%$).

Two results carry over from the main paper. First, on GroundingME every
proposal-based selector fails although a proposal covers $79\%$ of present
referents: the trained heads reach target accuracy $0.05$--$0.07$ against
$0.24$ (Qwen2.5-VL) and $0.35$ (Qwen3-VL) for coordinate decoding, and $0.002$ on
the manually boxed subset; the default veto reaches its joint accuracy almost
only by abstaining. Second, on referent-level negatives the VLM's own yes/no
answer is the abstention that works: it rejects $0.87$--$0.91$ of the edited
expressions, the verifier scored at the decoded box $0.33$--$0.38$, and the
hook's veto $0.32$--$0.60$, consistent with the verifier being an existence
probe on the query (\cref{sec:app-verifier-inputs}): an attribute edit leaves
the object category present.

\paragraph{Where the GroundingME collapse comes from.}
The collapse is not a fault of the pipeline. All $879$ images match their
recorded sizes, every gold box lies inside its image, coordinate decoding in
the same frame reaches $0.243$ with a median center offset of zero, and the
proposals cover $0.792$ of present referents, as reported. The same retrained
heads, run through the same pipeline on COCO images (the RefCOCOg-based
negatives' present rows), reach $0.848$--$0.862$. On the $637$ covered
GroundingME referents, however, the selection error is $0.93$--$0.94$: the
first covering proposal has median rank $26$--$31$ of $300$ (rank $1$ on
RefCOCOg val), and it is among the top five for only $0.19$--$0.20$ of them.
The failure is worst on referents smaller than $1\%$ of the image (at most
$0.006$, against $0.16$--$0.23$ for coordinate decoding) and on the manually
boxed high-resolution subset ($0.002$), but it persists on boxes covering at
least $10\%$ of the image ($0.29$ against $0.43$) and on COCO-category
referents ($0.07$), so the category boundary does not explain it. Every
GroundingME image exceeds the VLM's pixel budget (median downscale $1.87$ per
side), whereas RefCOCOg images are read at their native size; reading them at
native resolution lifts no selector (\cref{sec:app-gme-diag}).

\begin{table*}[t]
\centering
\footnotesize
\setlength{\tabcolsep}{3.4pt}
\caption{Public benchmarks with frozen thresholds (accuracy at
$\mathrm{IoU}\ge0.5$; hook-based rows: full-budget hooks and heads, mean over
three seeds, sample std $\le0.017$). Target: present requests; No-tgt.: correct
abstention on requests without a referent; Joint: all requests (GroundingME's
official score); Pair: original grounded and edit rejected. ``+ verifier'':
the hook's verifier scored at the decoded box's proposal.}
\label{tab:app-public}
\begin{tabular}{@{}lcccccccccc@{}}
\toprule
& \multicolumn{6}{c}{GroundingME} & \multicolumn{4}{c}{RefCOCOg\_rej (test images)}\\
\cmidrule(lr){2-7}\cmidrule(lr){8-11}
System & Target & No-tgt. & Joint & Discr. & Spatial & Limited & Target & No-tgt. & Joint & Pair\\
\midrule
\multicolumn{11}{@{}l}{\emph{Qwen2.5-VL-7B}}\\
Coordinate decoding & \textbf{0.243} & 0.000 & \textbf{0.194} & 0.338 & 0.153 & 0.267 & 0.853 & 0.000 & 0.427 & 0.000\\
GroundingME prompt (null allowed) & 0.193 & 0.010 & 0.156 & 0.275 & 0.140 & 0.190 & 0.844 & 0.086 & 0.465 & 0.064\\
\quad at the checkpoint's default resolution & 0.223 & 0.000 & 0.178 & 0.358 & 0.160 & 0.193 & -- & -- & -- & --\\
Coordinate decoding + VLM yes/no & 0.111 & 0.224 & 0.133 & 0.206 & 0.090 & 0.067 & 0.771 & \textbf{0.870} & \textbf{0.820} & \textbf{0.657}\\
Coordinate decoding + tuned yes-prob. & 0.221 & 0.035 & 0.184 & 0.309 & 0.153 & 0.230 & 0.846 & 0.565 & 0.706 & 0.473\\
Coordinate decoding + verifier & 0.083 & 0.265 & 0.119 & 0.175 & 0.077 & 0.026 & 0.840 & 0.380 & 0.610 & 0.298\\
Detector-native & 0.056 & 0.000 & 0.045 & 0.127 & 0.033 & 0.030 & 0.783 & 0.000 & 0.392 & 0.000\\
Hook, forced & 0.051 & 0.000 & 0.041 & 0.088 & 0.049 & 0.028 & 0.815 & 0.000 & 0.408 & 0.000\\
Hook, default veto & 0.012 & \textbf{0.725} & 0.155 & 0.025 & 0.012 & 0.004 & 0.679 & 0.596 & 0.638 & 0.358\\
Relation-aware head & 0.055 & 0.000 & 0.044 & 0.096 & 0.057 & 0.024 & 0.848 & 0.000 & 0.424 & 0.000\\
Token-state probe & 0.046 & 0.000 & 0.036 & 0.077 & 0.053 & 0.017 & 0.850 & 0.000 & 0.425 & 0.000\\
Combined selector & 0.052 & 0.000 & 0.041 & 0.093 & 0.039 & 0.037 & \textbf{0.862} & 0.000 & 0.431 & 0.000\\
\midrule
\multicolumn{11}{@{}l}{\emph{Qwen3-VL-8B}}\\
Coordinate decoding & 0.351 & 0.000 & 0.281 & 0.515 & 0.223 & 0.367 & 0.881 & 0.000 & 0.440 & 0.000\\
GroundingME prompt (null allowed) & \textbf{0.362} & 0.000 & \textbf{0.290} & 0.500 & 0.280 & 0.350 & 0.881 & 0.141 & 0.511 & 0.118\\
Coordinate decoding + VLM yes/no & 0.269 & 0.169 & 0.249 & 0.392 & 0.203 & 0.250 & 0.805 & \textbf{0.912} & \textbf{0.859} & \textbf{0.727}\\
Coordinate decoding + verifier & 0.105 & 0.245 & 0.133 & 0.245 & 0.056 & 0.060 & 0.872 & 0.327 & 0.599 & 0.264\\
Hook, forced & 0.066 & 0.000 & 0.053 & 0.124 & 0.051 & 0.041 & 0.841 & 0.000 & 0.420 & 0.000\\
Hook, default veto & 0.034 & \textbf{0.405} & 0.108 & 0.077 & 0.019 & 0.021 & 0.818 & 0.321 & 0.570 & 0.244\\
\bottomrule
\end{tabular}
\end{table*}

\paragraph{Diagnosing the GroundingME collapse.}
\label{sec:app-gme-diag}
We traced the trained selectors' failure on GroundingME through the pipeline
(Qwen2.5-VL, $804$ present referents; \cref{tab:app-gme-diag}), and twelve
consistency checks pass. All $879$ images have their recorded size in the
detector pass. Proposals and gold boxes share one normalized frame: every
proposal lies in $[0,1]$, the $300$ proposals of an image reach its right and
bottom edges for landscape and portrait images alike (median largest right and
bottom coordinates $1.000$ and $0.998$, as on RefCOCOg val), the mean proposal
center $(0.496, 0.573)$ lies next to the mean gold-box center $(0.511,
0.586)$, and recomputing IoU recovers the stored covering proposal of every
referent. The queries and token states belong to their requests and images,
and re-scoring the hooks' stored selections reproduces the reported accuracies
exactly. The detector and the VLM see each image at the same scale (median
downscale per side $1.88$ and $1.87$; every image exceeds the VLM's
$1{,}280$-token budget), at which the square root of a referent's box area has
a median of $98$ pixels ($187$ in the original image), and coordinate decoding
from the same VLM input reaches $0.243$.

The failure sits in selection. Proposals cover $0.792$ of the present
referents, but on the $637$ covered ones the conditional selection error
$\SayErr$ of the two hooks and the three set-level heads is $0.930$--$0.942$,
against $0.160$--$0.212$ for the hooks on $2{,}000$ RefCOCOg val requests; the
first covering proposal has median rank $16$ and $19$ of $300$ for the two
hooks and $26$--$31$ for the three heads, deeper than in the detector's own
score order (rank $11$). The hooks' selections hardly depend on
the request: giving each request the query of another request lowers their
accuracy only from $0.055$ to $0.038$ ($24$k hook) and from $0.051$ to $0.037$
(full-budget hook), and the mean query of the split reaches $0.043$ and
$0.036$, whereas on RefCOCOg val the same replacement costs $0.50$ and $0.55$
($0.788$ to $0.284$ and $0.840$ to $0.294$). Selection thus falls back on a
prior over regions that suits COCO but not GroundingME, where the detector's
top proposal is the referent for $0.040$ of the referents against $0.300$ on
RefCOCOg val. The GroundingME queries also lie farther from the training
distribution (median cosine to the RefCOCOg val query centroid $0.913$ against
$0.973$), whereas description length is not the mechanism: every selector
scores at least as well beyond $40$ words as at $11$--$20$ words, and the $66$
descriptions truncated at $96$ tokens are not systematically worse ($0.040$--$0.091$ against
$0.046$--$0.052$). Small and manually boxed referents make the failure total
but do not create it. Below $32$ pixels at the detector input ($21\%$ of
present referents) no selector exceeds $0.003$ while decoding reaches
$0.14$--$0.19$; on the $150$ manually boxed referents, all on images of at
least $16$ megapixels, proposal recall falls to $0.560$ and every selector to
at most $0.002$, below the detector's top proposal ($0.053$). Yet on the
largest referents (at least $128$ pixels, recall $0.866$) the selectors reach
only $0.108$--$0.137$ against $0.294$ for decoding, with $\SayErr$ of
$0.842$--$0.875$, and on the discriminative subset only $0.077$--$0.098$
against $0.338$, while the same heads reach $0.848$--$0.862$ on the present
referents of RefCOCOg test images (\cref{tab:app-public}).

The VLM reads GroundingME images downscaled about twofold, unlike RefCOCOg
images, so we re-ran the frozen pipeline on $300$ present referents ($274$
images, drawn with a fixed seed and stratified by subset and box source) in
three arms that differ only in input resolution (\cref{tab:app-gme-res}): the
evaluation budget of $1{,}280$ visual tokens (re-extracted queries match the
original ones with cosine at least $0.9997$, and the hooks reproduce their
original accuracy within $0.001$); native resolution, the checkpoint's default
cap of $12.8$ megapixels, under which only the manually boxed images of at least
$16$ megapixels remain downscaled, with the proposals unchanged; and images
reduced to a longer side of $640$ pixels, the size of the COCO images every
head was trained on, with detector and VLM rerun. The set-level heads are the
saved full-budget heads of \cref{sec:app-allsplits}. At native resolution the
query changes little (median cosine $0.990$ to the budget query, minimum
$0.929$) and no selector improves significantly: all five stay at $0.049$--$0.062$ (paired
differences from $-0.014$ to $+0.009$, image-clustered intervals within
[$-0.027$, $+0.022$]) with proposal recall at $0.803$, while coordinate
decoding moves from $0.273$ to $0.237$ ($-0.037$ [$-0.088$, $+0.010$]); the
first covering proposal stays deep, at median rank $18$ for both hooks and
$23$--$26$ for the heads ($15$--$18$ and $24$--$26$ at the budget on the same
referents). At
$640$ pixels decoding loses $0.103$ ([$-0.157$, $-0.053$]) but the selectors
stay at $0.040$--$0.064$, and their $\SayErr$ lies between $0.92$ and $0.95$
in every arm. The numbers place the collapse on the selection side: the
pipeline is consistent, proposals cover four in five referents, and neither the
query's resolution nor the referent's size explains why the selectors miss
more than nine in ten covered referents. What fails is the mapping from a
GroundingME request to one of the covered proposals, which selectors trained on
RefCOCOg expressions over COCO-category referents do not learn.

\begin{table*}[t]
\centering
\footnotesize
\setlength{\tabcolsep}{3.5pt}
\caption{Where the GroundingME collapse sits (Qwen2.5-VL; present referents;
accuracy at $\mathrm{IoU}\ge0.5$, frozen pipeline, no abstention; trained rows:
mean over three seeds). Recall: proposal recall at $\mathrm{IoU}\ge0.5$.
Decoding: the VLM's raw coordinate box. Hooks: the alignment hooks trained on
$24$k rows and on the full set; set-level heads: the full-budget relation-aware
head, token-state probe and combined selector of \cref{tab:app-public}.
$\SayErr$: conditional selection error of the combined selector on covered
referents. Det.\ top: the detector's highest-scoring proposal. Size: square
root of the gold box's area in pixels at the detector input. Seeded: gold box
seeded by Grounding DINO and corrected by annotators; manual: drawn by
annotators, all on images of at least $16$ megapixels.}
\label{tab:app-gme-diag}
\begin{tabular}{@{}lcccccccccc@{}}
\toprule
& & & & \multicolumn{2}{c}{Hook} & \multicolumn{3}{c}{Set-level heads} & & \\
\cmidrule(lr){5-6}\cmidrule(lr){7-9}
Referents & $n$ & Recall & Decoding & $24$k & full & Relation & Token & Comb. & $\SayErr$ & Det.\ top\\
\midrule
All present & 804 & 0.792 & 0.243 & 0.055 & 0.051 & 0.055 & 0.046 & 0.052 & 0.935 & 0.040\\
\midrule
\multicolumn{11}{@{}l}{\emph{Size at the detector input}}\\
\quad $<16$ px & 108 & 0.519 & 0.139 & 0.000 & 0.000 & 0.003 & 0.000 & 0.003 & 0.994 & 0.056\\
\quad $16$--$32$ px & 64 & 0.672 & 0.188 & 0.000 & 0.000 & 0.000 & 0.000 & 0.000 & 1.000 & 0.031\\
\quad $32$--$64$ px & 112 & 0.875 & 0.179 & 0.003 & 0.003 & 0.012 & 0.012 & 0.024 & 0.973 & 0.036\\
\quad $64$--$128$ px & 221 & 0.819 & 0.271 & 0.033 & 0.023 & 0.006 & 0.014 & 0.014 & 0.983 & 0.009\\
\quad $\ge128$ px & 299 & 0.866 & 0.294 & 0.123 & 0.119 & 0.137 & 0.108 & 0.119 & 0.862 & 0.060\\
\midrule
\multicolumn{11}{@{}l}{\emph{Gold-box source}}\\
\quad Seeded, corrected & 654 & 0.846 & 0.258 & 0.068 & 0.063 & 0.067 & 0.056 & 0.063 & 0.925 & 0.037\\
\quad Manual & 150 & 0.560 & 0.173 & 0.000 & 0.000 & 0.002 & 0.000 & 0.002 & 0.996 & 0.053\\
\midrule
\multicolumn{11}{@{}l}{\emph{GroundingME subset}}\\
\quad Discriminative & 204 & 0.858 & 0.338 & 0.098 & 0.088 & 0.096 & 0.077 & 0.093 & 0.891 & 0.054\\
\quad Spatial & 300 & 0.803 & 0.153 & 0.054 & 0.049 & 0.057 & 0.053 & 0.039 & 0.952 & 0.033\\
\quad Limited view & 300 & 0.737 & 0.267 & 0.027 & 0.028 & 0.024 & 0.017 & 0.037 & 0.950 & 0.037\\
\bottomrule
\end{tabular}
\end{table*}

\begin{table}[t]
\centering
\footnotesize
\setlength{\tabcolsep}{3pt}
\caption{GroundingME at three input resolutions ($300$ present referents;
accuracy at $\mathrm{IoU}\ge0.5$, no abstention; trained rows: mean over three
seeds). Budget: $1{,}280$ visual tokens, as evaluated (coordinate decoding: the
original answers); native: the checkpoint's default cap of $12.8$ megapixels,
same proposals; $640$ px: image reduced to a longer side of $640$ pixels,
detector and VLM rerun. $\Delta$: native minus budget; ${}^\dagger$its
image-clustered bootstrap $95\%$ interval excludes zero. Set-level heads: the
saved full-budget heads of \cref{tab:app-allsplits}.}
\label{tab:app-gme-res}
\begin{tabular}{@{}lcccc@{}}
\toprule
& Budget & Native & $640$ px & $\Delta$\\
\midrule
Proposal recall & 0.803 & 0.803 & 0.813 & $+0.000$\\
Coordinate decoding & 0.273 & 0.237 & 0.170 & $-0.037$\\
Detector's top proposal & 0.040 & 0.040 & 0.037 & $+0.000$\\
\midrule
Hook, $24$k & 0.064 & 0.050 & 0.044 & $-0.014^\dagger$\\
Hook, full & 0.061 & 0.058 & 0.064 & $-0.003$\\
Relation-aware head & 0.051 & 0.060 & 0.053 & $+0.009$\\
Token-state probe & 0.058 & 0.062 & 0.040 & $+0.004$\\
Combined selector & 0.061 & 0.049 & 0.052 & $-0.012$\\
\bottomrule
\end{tabular}
\end{table}
\paragraph{Public reasoning benchmarks: Ref-Adv-s and RefBench-PRO.}
\label{sec:app-public-reasoning}
The stress set of the main paper is templated and small, so we also run the
frozen pipeline on two public referring benchmarks built around reasoning,
negation and rejection. Ref-Adv-s \cite{refadv} has $1{,}142$ expressions on
$614$ images, each with a referent among same-category distractors, which the
authors bin by number ($2$--$3$, $4$--$6$ and at least $7$); $457$ expressions
contain a negation and $421$ are marked as human-authored. Of its rows, $883$
are on OpenImages images and $259$ on COCO val2017 images; none of its images is
in a RefCOCO-family training split. RefBench-PRO \cite{refbenchpro}, whose data
are released under the name PR-Bench (CC BY-NC 4.0), has $6{,}000$ rows on
$3{,}102$ images: $1{,}000$ in each of five grounding tasks that test
attributes, positions, interactions, relations between objects and commonsense
knowledge, and $1{,}000$ in a reject task whose expression has no referent in
the image, where only an abstention is correct. Nothing is retrained or
re-tuned. The proposals come from the same detector pass with the
$80$-category prompt, the Qwen2.5-VL-7B query and token states are extracted
exactly as on every other split, the selectors are the alignment hook trained
on the $24$k rows of the probes and the full-budget heads of Table~1 of the
main paper (the alignment hook and the saved relation-aware head, token-state
probe and combined selector of \cref{sec:app-allsplits}), and every threshold
is the one chosen on RefCOCOg val (\cref{sec:app-abstention}). We report
accuracy at $\mathrm{IoU}\ge0.5$; on RefBench-PRO also the benchmark's mAcc,
the accuracy averaged over the IoU thresholds $0.50{:}0.05{:}0.90$, and the
authors' $\mathrm{Acc}_p$, the mean mAcc of the five grounding tasks (the
leaderboard score). Seeded systems are averaged over three seeds, and intervals
are image-clustered bootstrap $95\%$ intervals of the seed-averaged
correctness.

With each benchmark's own prompt the frozen VLM reproduces the published
numbers, which checks the data handling and scoring. On Ref-Adv-s the Ref-Adv
prompt at the checkpoint's default resolution, scored with the authors' rule,
reaches $0.395$ ($0.392$ at the pipeline's budget of $1{,}280$ visual tokens),
against $0.393$ for the Qwen2.5-VL-7B predictions released by the authors
($39.3$ in their paper, direct prompting); the two agree on $0.942$ of the rows
($34$ and $32$ discordant rows, exact McNemar $p=0.90$). The pipeline's own
coordinate-decoding prompt reaches $0.520$ on the same rows ($226$ rows correct
only for it, $81$ only for the released predictions, $p<10^{-16}$), so the decoding rows below are not weakened by
our prompt. On RefBench-PRO the benchmark's prompt, read as the authors read
it, gives Qwen2.5-VL-7B an $\mathrm{Acc}_p$ of $0.526$ (rejection accuracy
$0.038$), against $57.58$ ($3.1$) in the first version of the RefBench-PRO
paper, which was measured on an earlier version of the data, so this comparison
is approximate. Qwen3-VL-8B, which the release leaderboard lists, reaches
$0.605$ against $59.4$, with rejection accuracy $0.175$ against $16.2$.

\begin{table*}[t]
\centering
\footnotesize
\setlength{\tabcolsep}{3.5pt}
\caption{Ref-Adv-s through the frozen pipeline (Qwen2.5-VL-7B; every row has a
referent; accuracy at $\mathrm{IoU}\ge0.5$, no abstention; trained rows: mean
over three seeds; brackets: image-clustered bootstrap $95\%$ interval). Columns
split the rows by negation, by whether the expression is marked as
human-authored, by image source (COCO val2017 or OpenImages) and by the
authors' bins of the number of same-category distractors; the second header
row gives the rows per column. Coordinate decoding: the VLM's raw box under the
pipeline's prompt; coordinate-snap: that box snapped to the proposal of maximal
IoU; benchmark prompt: the Ref-Adv prompt at the checkpoint's default
resolution, scored with the authors' rule; detector-native: the detector
prompted with the expression, highest-scoring box. Alignment hooks: trained on
the $24$k rows of the probes and on the full set; the three set-level heads are
the saved full-budget heads of \cref{sec:app-allsplits}.}
\label{tab:app-refadv}
\begin{tabular}{@{}lcccccccccc@{}}
\toprule
& & \multicolumn{2}{c}{Negation} & \multicolumn{2}{c}{Human-authored} & \multicolumn{2}{c}{Image source} & \multicolumn{3}{c}{Distractors}\\
\cmidrule(lr){3-4}\cmidrule(lr){5-6}\cmidrule(lr){7-8}\cmidrule(l){9-11}
System & All & yes & no & yes & no & COCO & OpenImages & $2$--$3$ & $4$--$6$ & $\ge7$\\
\emph{Rows} & $1{,}142$ & $457$ & $685$ & $421$ & $721$ & $259$ & $883$ & $698$ & $315$ & $129$\\
\midrule
Proposal recall & 0.907 & 0.932 & 0.891 & 0.976 & 0.867 & 0.981 & 0.886 & 0.895 & 0.905 & 0.977\\
\midrule
\multicolumn{11}{@{}l}{\emph{No task training}}\\
Coordinate decoding & 0.520\ci{.485}{.555} & 0.554 & 0.498 & 0.475 & 0.546 & 0.409 & 0.553 & 0.557 & 0.479 & 0.419\\
Coordinate-snap & 0.499\ci{.461}{.536} & 0.532 & 0.477 & 0.466 & 0.519 & 0.402 & 0.528 & 0.539 & 0.444 & 0.419\\
Benchmark prompt & 0.395\ci{.360}{.429} & 0.387 & 0.400 & 0.297 & 0.452 & 0.197 & 0.453 & 0.441 & 0.346 & 0.264\\
Detector-native & 0.271\ci{.240}{.300} & 0.212 & 0.309 & 0.209 & 0.307 & 0.181 & 0.297 & 0.297 & 0.235 & 0.217\\
\midrule
\multicolumn{11}{@{}l}{\emph{Trained on RefCOCOg}}\\
Alignment hook, $24$k & 0.280\ci{.250}{.310} & 0.280 & 0.280 & 0.226 & 0.312 & 0.188 & 0.307 & 0.319 & 0.216 & 0.227\\
Alignment hook, full & 0.308\ci{.276}{.341} & 0.333 & 0.291 & 0.258 & 0.337 & 0.194 & 0.342 & 0.335 & 0.278 & 0.235\\
Relation-aware head & 0.317\ci{.282}{.353} & 0.333 & 0.307 & 0.249 & 0.357 & 0.156 & 0.364 & 0.349 & 0.293 & 0.202\\
Token-state probe & 0.326\ci{.291}{.362} & 0.346 & 0.313 & 0.272 & 0.358 & 0.192 & 0.365 & 0.358 & 0.292 & 0.235\\
Combined selector & 0.314\ci{.282}{.350} & 0.334 & 0.301 & 0.248 & 0.353 & 0.171 & 0.356 & 0.351 & 0.276 & 0.207\\
\bottomrule
\end{tabular}
\end{table*}

\begin{table*}[t]
\centering
\footnotesize
\setlength{\tabcolsep}{3pt}
\caption{RefBench-PRO grounding tasks through the frozen pipeline
(Qwen2.5-VL-7B; $1{,}000$ rows per task, every row with a referent; no
abstention; trained rows: mean over three seeds). @$0.5$: accuracy at
$\mathrm{IoU}\ge0.5$ (proposal recall in the first row); mAcc: the benchmark's
accuracy averaged over the IoU thresholds $0.50{:}0.05{:}0.90$. Five tasks:
mean accuracy at $\mathrm{IoU}\ge0.5$ over the $5{,}000$ rows with its
image-clustered bootstrap $95\%$ interval, and $\mathrm{Acc}_p$, the mean mAcc
of the five tasks (the leaderboard score). Benchmark prompt: the RefBench-PRO
prompt at the checkpoint's default resolution, read as by the authors. Other
rows as in \cref{tab:app-refadv}.}
\label{tab:app-prbench}
\begin{tabular}{@{}lcccccccccccc@{}}
\toprule
& \multicolumn{2}{c}{Attribute} & \multicolumn{2}{c}{Position} & \multicolumn{2}{c}{Interaction} & \multicolumn{2}{c}{Relation} & \multicolumn{2}{c}{Commonsense} & \multicolumn{2}{c}{Five tasks}\\
\cmidrule(lr){2-3}\cmidrule(lr){4-5}\cmidrule(lr){6-7}\cmidrule(lr){8-9}\cmidrule(lr){10-11}\cmidrule(l){12-13}
System & @$0.5$ & mAcc & @$0.5$ & mAcc & @$0.5$ & mAcc & @$0.5$ & mAcc & @$0.5$ & mAcc & @$0.5$ & $\mathrm{Acc}_p$\\
\midrule
Proposal recall & 0.927 & -- & 0.887 & -- & 0.908 & -- & 0.902 & -- & 0.794 & -- & -- & --\\
\midrule
\multicolumn{13}{@{}l}{\emph{No task training}}\\
Coordinate decoding & 0.725 & 0.595 & 0.693 & 0.566 & 0.622 & 0.524 & 0.607 & 0.510 & 0.662 & 0.494 & 0.662\ci{.645}{.679} & 0.538\\
Coordinate-snap & 0.702 & 0.589 & 0.658 & 0.546 & 0.597 & 0.513 & 0.577 & 0.499 & 0.562 & 0.444 & 0.619\ci{.601}{.636} & 0.518\\
Benchmark prompt & 0.724 & 0.591 & 0.676 & 0.547 & 0.628 & 0.524 & 0.574 & 0.482 & 0.647 & 0.486 & 0.650\ci{.633}{.668} & 0.526\\
Detector-native & 0.680 & 0.598 & 0.584 & 0.503 & 0.487 & 0.444 & 0.457 & 0.415 & 0.410 & 0.355 & 0.524\ci{.507}{.540} & 0.463\\
\midrule
\multicolumn{13}{@{}l}{\emph{Trained on RefCOCOg}}\\
Alignment hook, $24$k & 0.354 & 0.300 & 0.267 & 0.226 & 0.333 & 0.288 & 0.198 & 0.174 & 0.155 & 0.132 & 0.261\ci{.247}{.276} & 0.224\\
Alignment hook, full & 0.393 & 0.336 & 0.309 & 0.269 & 0.376 & 0.326 & 0.236 & 0.210 & 0.170 & 0.140 & 0.297\ci{.282}{.312} & 0.256\\
Relation-aware head & 0.401 & 0.339 & 0.288 & 0.250 & 0.372 & 0.327 & 0.235 & 0.211 & 0.155 & 0.131 & 0.290\ci{.274}{.306} & 0.252\\
Token-state probe & 0.427 & 0.362 & 0.329 & 0.277 & 0.390 & 0.329 & 0.244 & 0.209 & 0.176 & 0.146 & 0.313\ci{.297}{.329} & 0.265\\
Combined selector & 0.429 & 0.367 & 0.323 & 0.272 & 0.396 & 0.338 & 0.241 & 0.216 & 0.175 & 0.144 & 0.313\ci{.296}{.329} & 0.267\\
\bottomrule
\end{tabular}
\end{table*}

\begin{table*}[t]
\centering
\footnotesize
\setlength{\tabcolsep}{3.5pt}
\caption{Abstention on the public reasoning benchmarks with the thresholds
frozen on RefCOCOg val (Qwen2.5-VL-7B; accuracy at $\mathrm{IoU}\ge0.5$;
trained rows: mean over three seeds). Ref-Adv-s has a referent in every row, so
a gate can only lose referents. Answer rate (Ans.): fraction of present rows
answered (all rows, negated and other expressions on Ref-Adv-s); Target: accuracy on
present rows, an abstention counting as a miss; Reject: fraction of
RefBench-PRO's reject rows abstained on; Joint: accuracy over all $6{,}000$
RefBench-PRO rows, with its image-clustered bootstrap $95\%$ interval. Own
refusal: abstain when the answer contains no box. VLM yes/no: answer only if
the VLM says the referent is in the image; yes-probability, tuned: answer if
its yes-minus-no score reaches the threshold tuned on RefCOCOg val; verifier:
the $24$k hook's verifier scored at the snapped box's proposal; default veto:
verifier and affinity thresholds $(\gamma_v,\gamma_s)=(0.3,24)$; verifier
only: the verifier threshold alone; own gate: the head's existence probability
at $\gamma_v$; detector score threshold: the detector's snapped box if its score
reaches the threshold tuned on RefCOCOg val. Benchmark prompt: its first box
read in the model's own frame; read as by the authors: RefBench-PRO's reading,
under which an answer without a box is a rejection. --: not applicable.}
\label{tab:app-public-reject}
\begin{tabular}{@{}lcccccccc@{}}
\toprule
& \multicolumn{4}{c}{Ref-Adv-s ($1{,}142$ present)} & \multicolumn{4}{c}{RefBench-PRO ($5{,}000$ present, $1{,}000$ reject)}\\
\cmidrule(lr){2-5}\cmidrule(l){6-9}
& \multicolumn{3}{c}{Answer rate} & & & & & \\
\cmidrule(lr){2-4}
Gate & all & negated & other & Target & Reject & Target & Joint & Ans.\\
\midrule
\multicolumn{9}{@{}l}{\emph{Gates on the VLM's box}}\\
Coordinate decoding, own refusal & 0.999 & 1.000 & 0.999 & 0.520 & 0.052 & 0.662 & 0.560\ci{.545}{.575} & 0.996\\
Coordinate-snap + VLM yes/no & 0.616 & 0.503 & 0.691 & 0.346 & 0.565 & 0.516 & 0.524\ci{.507}{.540} & 0.755\\
Coordinate-snap + yes-probability, tuned & 0.957 & 0.932 & 0.974 & 0.482 & 0.209 & 0.613 & 0.545\ci{.529}{.561} & 0.976\\
Coordinate-snap + verifier & 0.815 & 0.810 & 0.818 & 0.450 & 0.271 & 0.517 & 0.476\ci{.461}{.492} & 0.776\\
Coordinate-snap + VLM yes/no + verifier & 0.534 & 0.430 & 0.604 & 0.318 & 0.635 & 0.450 & 0.481\ci{.465}{.497} & 0.636\\
Benchmark prompt, own refusal & 1.000 & 1.000 & 1.000 & 0.402 & 0.038 & 0.659 & 0.555\ci{.541}{.571} & 0.995\\
\quad read as by the authors & -- & -- & -- & -- & 0.038 & 0.650 & 0.548\ci{.532}{.563} & 0.995\\
\midrule
\multicolumn{9}{@{}l}{\emph{Gates on proposal-based selectors}}\\
Detector-native, score threshold & 0.765 & 0.799 & 0.743 & 0.223 & 0.128 & 0.437 & 0.385\ci{.371}{.401} & 0.770\\
Alignment hook $24$k, default veto & 0.424 & 0.410 & 0.433 & 0.180 & 0.794 & 0.150 & 0.258\ci{.245}{.270} & 0.310\\
Alignment hook $24$k, verifier only & 0.938 & 0.931 & 0.943 & 0.276 & 0.052 & 0.259 & 0.225\ci{.213}{.236} & 0.957\\
Alignment hook full, default veto & 0.418 & 0.413 & 0.421 & 0.200 & 0.778 & 0.194 & 0.292\ci{.278}{.305} & 0.338\\
Alignment hook full, verifier only & 0.893 & 0.902 & 0.887 & 0.299 & 0.158 & 0.292 & 0.270\ci{.257}{.283} & 0.903\\
Relation-aware head, own gate & 0.889 & 0.891 & 0.889 & 0.312 & 0.164 & 0.286 & 0.266\ci{.252}{.279} & 0.900\\
Token-state probe, own gate & 0.919 & 0.911 & 0.925 & 0.318 & 0.084 & 0.312 & 0.274\ci{.261}{.287} & 0.954\\
Combined selector, own gate & 0.859 & 0.880 & 0.845 & 0.304 & 0.153 & 0.309 & 0.283\ci{.270}{.296} & 0.886\\
\bottomrule
\end{tabular}
\end{table*}

Proposals are not the bottleneck on either benchmark
(\cref{tab:app-refadv,tab:app-prbench}): they cover $0.907$ of the Ref-Adv-s
referents and $0.794$--$0.927$ of the RefBench-PRO referents per task.
Selection is. On Ref-Adv-s the full-budget selectors reach $0.308$--$0.326$,
slightly above detector-native ($0.271$) but below the benchmark's own prompt
($0.395$) and far below coordinate decoding ($0.520$ [$0.485$, $0.555$], an
interval that lies above every selector's). On RefBench-PRO they reach $0.290$--$0.313$ over the five
grounding tasks ($\mathrm{Acc}_p$ $0.252$--$0.267$), against $0.662$ for
decoding and $0.524$ for detector-native, and they are weakest on relation
($0.235$--$0.244$) and commonsense ($0.155$--$0.176$), where decoding keeps
$0.607$ and $0.662$ and proposals still cover $0.902$ and $0.794$ of the
referents. Negation does not hurt the trained selectors: on negated Ref-Adv-s
expressions they reach $0.333$--$0.346$ against $0.291$--$0.313$ on the others,
as decoding does ($0.554$ against $0.498$) and with higher proposal recall
($0.932$ against $0.891$); detector-native loses clearly ($0.212$ against
$0.309$) and the benchmark prompt slightly ($0.387$ against $0.400$).
Human-authored expressions are harder for every system although their referents
are covered more often ($0.976$ against $0.867$): the selectors fall to
$0.248$--$0.272$ against $0.337$--$0.358$ and decoding to $0.475$ against
$0.546$. So are the rows on COCO val2017 images, the image domain the heads
were trained on: there the selectors reach only $0.156$--$0.194$ against
$0.342$--$0.365$ on OpenImages images, with $0.981$ of the referents covered,
while decoding drops from $0.553$ to $0.409$ and the benchmark prompt from
$0.453$ to $0.197$. Every system is also less accurate with at least seven
distractors than with two or three (decoding $0.419$ against $0.557$,
token-state probe $0.235$ against $0.358$), although proposals cover more of
those referents ($0.977$ against $0.895$).

Abstention repeats the pattern found on GroundingME's attribute-edited
negatives (\cref{tab:app-public,tab:app-public-reject}). On Ref-Adv-s the VLM's yes/no answer keeps
only $0.616$ of the referents, and $0.503$ of the negated ones against $0.691$
of the others; the tuned yes-probability keeps $0.957$ ($0.932$ of the negated
ones), the verifier at the decoded box $0.815$ with no difference by negation
($0.810$ against $0.818$), the heads' own gates $0.859$--$0.919$ and the hooks'
default vetoes only $0.418$--$0.424$. On RefBench-PRO's $1{,}000$ reject rows
the VLM's yes/no answer rejects $0.565$, with joint accuracy $0.524$ [$0.507$,
$0.540$] over all $6{,}000$ rows, whereas the verifier at the decoded box
rejects $0.271$ and reaches $0.476$ [$0.461$, $0.492$]. The tuned
yes-probability rejects only $0.209$ but keeps $0.976$ of the present queries
and gives the highest joint accuracy among the gates added to the decoded box
($0.545$). With five present queries per reject row, joint accuracy here
rewards answering: decoding's own refusal rejects only $0.052$ yet reaches the
highest joint accuracy of all gates ($0.560$ [$0.545$, $0.575$]) because it
keeps the raw box and its target accuracy of $0.662$, and the benchmark
prompt's refusal behaves alike ($0.038$ rejected, joint $0.555$). The trained
heads' own gates reject $0.084$--$0.164$ at joint accuracy $0.266$--$0.283$,
and the hooks' default vetoes reject $0.778$--$0.794$ only by answering on about
a third of the present queries (answer rate $0.310$--$0.338$, joint
$0.258$--$0.292$). On public reasoning data, too, proposals cover most
referents, but selectors trained on RefCOCOg expressions return a correct one
for less than a third of the queries, and the VLM's yes/no answer rejects more absent referents than the verifier, at the
price of half of Ref-Adv-s's negated expressions.

\section{Full Training Set and a Second VLM}
\label{sec:app-full}

\paragraph{Setup.}
The heads of Table 1 of the main paper are retrained on all of RefCOCOg (umd) train:
$80{,}512$ present expressions on $21{,}899$ images and $34{,}505$ additional swapped absent
rows ($30\%$ of all rows; \cref{sec:app-config-notes}), $115{,}017$ rows in total. The
diagnostic probes keep the $24$k budget, at which every variant, the
per-query statistics and the multiple-comparison tests were run; the full
budget repeats each comparison the main conclusions rest on. Proposals,
recipes, thresholds and seeds are those of the $24$k runs; the detector pass is
independent of the VLM, and its expression-prompted boxes coincide row by row
across the two VLM runs. The second VLM is Qwen3-VL-8B \cite{qwen3vl}. Its query
is read as for Qwen2.5-VL, as the final-layer state (after the final
normalization) at the position that would begin the answer, and both VLMs see
at most $1{,}280$ visual tokens per image. Qwen3-VL returns boxes in relative
$0$--$1000$ coordinates, which we rescale; decoding is greedy. Unparsable
answers, including refusals, fall back to a fixed box as for Qwen2.5-VL
(prompt and parse rule below); they
occur on at most $0.22\%$ of present queries, but Qwen3-VL gives no parsable
box for $66\%$ of the swapped absent requests on RefCOCOg test ($65\%$ on val), against $17\%$ for
Qwen2.5-VL. The bridge
trained on the image-conditioned query is the hook trained without the
verifier; their per-query outcomes coincide in every seed and split for both
VLMs, as at $24$k.

\paragraph{Coordinate answers: prompt and parse rule.}
\label{sec:app-parse}
Both VLMs receive the image and the prompt \emph{Locate
``$\langle$expression$\rangle$'' in this image and output its bounding box
coordinates in JSON format with the key ``bbox\_2d''.} They decode greedily
(Qwen2.5-VL with its checkpoint's sampling at temperature $10^{-6}$, which is
effectively greedy; \cref{sec:app-seeds}) for
at most $64$ new tokens and stop once the first complete box entry has been
generated. The box is read from the first entry under that key,
$[x_1,y_1,x_2,y_2]$, or, if there is none, from the
first group of four comma-separated numbers (in square brackets for Qwen3-VL).
Qwen2.5-VL values are absolute pixels of its resized input frame and are divided
by that frame's width and height (four values that are all at most $1.5$ are
taken as already normalized); Qwen3-VL values are divided by $1000$.
Coordinates are clipped to the image, and for Qwen3-VL reversed corners are
swapped. An answer without such a group of four numbers is unparsable; refusals
and statements that the object is not in the image fall in this class. Rows
that always answer (coordinate decoding and coordinate-snap) replace an
unparsable answer by the fixed box covering the top-left quarter of the image,
which almost always counts as a miss, and record the row as unparsable.

\paragraph{Refusal as an abstention gate.}
Unparsable answers are rare on present referents ($0.03\%$ of RefCOCOg test
rows for Qwen2.5-VL, $0.22\%$ for Qwen3-VL) but not on swapped absent ones
($17.4\%$ and $65.5\%$). Abstaining exactly on them gives coordinate decoding a
native gate. On RefCOCOg test with $30\%$ swapped absent referents it reaches
joint accuracy $0.658$ [$0.650$, $0.666$] for Qwen2.5-VL (snapped box; answer
rate $0.948$, no-target accuracy $0.174$, $0.087$ on same-category hard
negatives) and $0.819$ for Qwen3-VL (raw box; answer rate $0.802$, no-target
accuracy $0.655$, $0.506$ on hard negatives), against $0.606$ and $0.623$ for
always answering, $0.841$ and $0.860$ for the yes/no answer, $0.877$ and $0.891$
for the tuned yes-probability and $0.888$ and $0.907$ for the verifier. On the
stress set Qwen2.5-VL never refuses; Qwen3-VL's gate reaches $0.561$. On
gRefCOCO the Qwen2.5-VL coordinate-decoding row already abstains exactly on
answers without a box (no-target accuracy $0.030$ on testA, $0.071$ on testB).
Refusal is thus a real but weak gate, strongest for Qwen3-VL and below the
VLM's own yes/no answer for both models.

\paragraph{Accuracy.}
\Cref{tab:app-full-acc} lists every head with its sample std,
\cref{tab:app-full-iou75} accuracy at $\mathrm{IoU}\ge0.75$ and recall@5. Full
data raise every head on every split except the stress set, by $0.02$--$0.13$
(paired query-level intervals exclude zero); on the stress set the changes lie
between $-0.07$ and $+0.03$, and the token-state probe loses $0.071$
[$-0.107$, $-0.035$]. At $\mathrm{IoU}\ge0.75$ the relation-aware head, the token-state probe and
the combined selector are above the VLM's raw box on RefCOCOg for both VLMs
(the hook only with Qwen2.5-VL), and far below it on Ref-L4 and the stress set
(\cref{tab:app-full-iou75}). Qwen3-VL's raw box is more
accurate than Qwen2.5-VL's by $0.020$--$0.030$ on the standard splits and on
Ref-L4 and by $0.041$--$0.052$ at $\mathrm{IoU}\ge0.75$
(\cref{tab:app-full-dec}), yet on Ref-L4 its heads are not better than
Qwen2.5-VL's (hook $-0.018$, token states $-0.049$, combined $-0.005$), except
the relation-aware head ($0.605$ against $0.595$). Trained on subsets of the full split, the
hook reaches $0.734$ ($5$k rows), $0.772$ ($10$k) and $0.843$ (all) on RefCOCOg
val with Qwen2.5-VL and $0.788$, $0.797$ and $0.853$ with Qwen3-VL (random
subsets of the $115$k rows, so the $5$k and $10$k points differ by up to
$0.02$ from the curve over subsets of the $24$k rows in \cref{sec:app-dataeff-text}).

\begin{table*}[t]
\centering
\footnotesize
\setlength{\tabcolsep}{3.5pt}
\caption{Selection accuracy at $\mathrm{IoU}\ge0.5$ (present queries, no
abstention), mean$\pm$sample std over three seeds, at the $24$k budget
(Qwen2.5-VL, as in the main paper) and with the full RefCOCOg training set
($115$k rows) for both VLMs. Text-only query: the learned bridge on the request
read without the image. Matched hook: the hook at the relation-aware head's
parameter count with the same all-positive supervision. Pooled query and token
states: cross-attention probes of Table 2 of the main paper. Combined: token states, two
intermediate-depth states, relations among proposals and all-positive
supervision with the IoU-rank loss. Qwen2.5-VL cross-attention heads and combined selector on RefCOCO and RefCOCO+ val: the evaluation of \cref{sec:app-allsplits}. Qwen3-VL token states and combined selector on RefCOCO and RefCOCO+ val: the two heads retrained with the same recipe and seeds, which reproduce the other published cells within $0.002$ outside the stress set. --: not evaluated.}
\label{tab:app-full-acc}
\begin{tabular}{@{}llcccccc@{}}
\toprule
VLM, budget & Head & RefCOCO val & RefCOCO+ val & RefCOCOg val & RefCOCOg test & Ref-L4 val & Stress\\
\midrule
\multirow{6}{*}{Qwen2.5-VL, $24$k} & Text-only query & 0.557\sd{.005} & 0.444\sd{.012} & 0.579\sd{.007} & 0.593\sd{.007} & 0.372\sd{.018} & 0.294\sd{.033}\\
& Alignment hook & 0.789\sd{.007} & 0.707\sd{.010} & 0.787\sd{.004} & 0.793\sd{.007} & 0.464\sd{.010} & 0.384\sd{.035}\\
& Matched hook & 0.811\sd{.012} & 0.734\sd{.014} & 0.820\sd{.011} & 0.816\sd{.011} & 0.497\sd{.010} & 0.392\sd{.007}\\
& Relation-aware head & 0.849\sd{.003} & 0.772\sd{.003} & 0.846\sd{.002} & 0.849\sd{.001} & 0.522\sd{.007} & 0.388\sd{.014}\\
& Pooled query & -- & -- & 0.850\sd{.001} & 0.854\sd{.001} & 0.502\sd{.006} & 0.356\sd{.007}\\
& Token states & -- & -- & 0.849\sd{.004} & 0.852\sd{.002} & 0.593\sd{.004} & 0.426\sd{.015}\\
\midrule
\multirow{7}{*}{Qwen2.5-VL, full} & Text-only query & 0.683\sd{.001} & 0.517\sd{.006} & 0.672\sd{.005} & 0.676\sd{.002} & 0.442\sd{.008} & 0.312\sd{.020}\\
& Alignment hook & 0.834\sd{.006} & 0.760\sd{.008} & 0.843\sd{.004} & 0.837\sd{.005} & 0.549\sd{.007} & 0.413\sd{.005}\\
& Matched hook & 0.850\sd{.006} & 0.774\sd{.008} & 0.851\sd{.002} & 0.848\sd{.002} & 0.577\sd{.008} & 0.408\sd{.022}\\
& Relation-aware head & 0.872\sd{.003} & 0.802\sd{.006} & 0.877\sd{.002} & 0.878\sd{.003} & 0.595\sd{.004} & 0.358\sd{.009}\\
& Pooled query & 0.876\sd{.001} & 0.808\sd{.001} & 0.877\sd{.001} & 0.881\sd{.001} & 0.606\sd{.001} & 0.388\sd{.005}\\
& Token states & 0.886\sd{.002} & 0.815\sd{.003} & 0.871\sd{.003} & 0.878\sd{.002} & 0.627\sd{.005} & 0.355\sd{.016}\\
& Combined & 0.895\sd{.003} & 0.824\sd{.002} & 0.876\sd{.001} & 0.885\sd{.000} & 0.641\sd{.003} & 0.381\sd{.009}\\
\midrule
\multirow{6}{*}{Qwen3-VL, full} & Text-only query & 0.669\sd{.014} & 0.508\sd{.005} & 0.671\sd{.005} & 0.674\sd{.006} & 0.469\sd{.007} & 0.273\sd{.011}\\
& Alignment hook & 0.853\sd{.004} & 0.800\sd{.003} & 0.853\sd{.003} & 0.856\sd{.003} & 0.531\sd{.016} & 0.437\sd{.048}\\
& Matched hook & 0.872\sd{.004} & 0.818\sd{.003} & 0.873\sd{.005} & 0.877\sd{.002} & 0.572\sd{.034} & 0.444\sd{.010}\\
& Relation-aware head & 0.905\sd{.001} & 0.853\sd{.001} & 0.900\sd{.000} & 0.899\sd{.002} & 0.605\sd{.002} & 0.408\sd{.009}\\
& Token states & 0.909\sd{.001} & 0.858\sd{.003} & 0.897\sd{.002} & 0.894\sd{.001} & 0.579\sd{.006} & 0.434\sd{.010}\\
& Combined & 0.910\sd{.001} & 0.855\sd{.002} & 0.898\sd{.001} & 0.895\sd{.000} & 0.637\sd{.006} & 0.439\sd{.007}\\
\bottomrule
\end{tabular}
\end{table*}

\begin{table*}[t]
\centering
\footnotesize
\setlength{\tabcolsep}{2.6pt}
\caption{Full training set: accuracy at $\mathrm{IoU}\ge0.75$ (present queries,
mean over three seeds) and recall@5 at $\mathrm{IoU}\ge0.5$. Coordinate
decoding: the VLM's raw box; it ranks no proposals, hence no recall@5. Hook
values at $\mathrm{IoU}\ge0.75$ are computed from the stored per-query IoU of
the selected proposal (rounding bound $0.0003$). Token states and combined
selector on RefCOCO and RefCOCO+: the retrained heads of
\cref{tab:app-full-acc}.}
\label{tab:app-full-iou75}
\begin{tabular}{@{}llcccccccc@{}}
\toprule
& & \multicolumn{6}{c}{Accuracy at $\mathrm{IoU}\ge0.75$} & \multicolumn{2}{c}{Recall@5}\\
\cmidrule(lr){3-8}\cmidrule(l){9-10}
VLM & Head & RefCOCO & RefCOCO+ & RefCOCOg val & RefCOCOg test & Ref-L4 & Stress & RefCOCOg test & Ref-L4\\
\midrule
\multirow{5}{*}{Qwen2.5-VL} & Alignment hook & 0.770 & 0.702 & 0.760 & 0.761 & 0.490 & 0.355 & 0.971 & 0.761\\
& Relation-aware head & 0.810 & 0.745 & 0.802 & 0.808 & 0.536 & 0.278 & 0.960 & 0.725\\
& Token states & 0.819 & 0.757 & 0.791 & 0.806 & 0.560 & 0.281 & 0.955 & 0.762\\
& Combined & 0.829 & 0.765 & 0.797 & 0.814 & 0.573 & 0.300 & 0.953 & 0.752\\
& Coordinate decoding & 0.791 & 0.742 & 0.749 & 0.757 & 0.724 & 0.468 & -- & --\\
\midrule
\multirow{5}{*}{Qwen3-VL} & Alignment hook & 0.790 & 0.745 & 0.768 & 0.777 & 0.470 & 0.368 & 0.971 & 0.758\\
& Relation-aware head & 0.839 & 0.794 & 0.820 & 0.826 & 0.545 & 0.325 & 0.965 & 0.724\\
& Token states & 0.842 & 0.795 & 0.820 & 0.824 & 0.515 & 0.345 & 0.965 & 0.718\\
& Combined & 0.845 & 0.792 & 0.819 & 0.823 & 0.573 & 0.352 & 0.960 & 0.746\\
& Coordinate decoding & 0.832 & 0.789 & 0.801 & 0.807 & 0.774 & 0.554 & -- & --\\
\bottomrule
\end{tabular}
\end{table*}

\paragraph{Capacity, pooling and query source at the full budget.}
\Cref{tab:app-full-cap} reruns four capacity and supervision variants and the
pooling block with Qwen2.5-VL. Capacity settings again stay close: linear
projections and a width of $1{,}024$ are within $0.008$ of the hook on RefCOCOg
test, while three hidden layers lose $0.024$ and, at the fixed thresholds,
answer only about half of the queries. All-positive supervision and the
three-depth readout each add about $0.01$, and together $0.03$; averaging over
expression tokens still loses $0.05$ on RefCOCOg, gains on Ref-L4 ($0.582$
against $0.549$) and collapses on the stress set ($0.232$). The query-source gap
of the bridge, image-conditioned against text-only query, is $0.155$--$0.243$
on the RefCOCO family with Qwen2.5-VL and $0.184$--$0.302$ with Qwen3-VL, and
$0.08$--$0.18$ on Ref-L4 and the stress set (all paired intervals above
zero).

\begin{table}[t]
\centering
\footnotesize
\setlength{\tabcolsep}{2.6pt}
\caption{Capacity, supervision and readout of the hook at the $24$k budget and
on the full training set (Qwen2.5-VL; mean over three seeds; Ans.: answer rate,
the fraction of present RefCOCOg test queries answered at the fixed thresholds). Ref-L4 and stress were
not evaluated for these variants at $24$k (--).}
\label{tab:app-full-cap}
\begin{tabular}{@{}lcccccc@{}}
\toprule
& \multicolumn{2}{c}{$24$k} & \multicolumn{4}{c}{Full training set}\\
\cmidrule(lr){2-3}\cmidrule(l){4-7}
Variant & Test & Ans. & Test & Ans. & Ref-L4 & Stress\\
\midrule
Hook (final-token query) & 0.793 & 0.856 & 0.837 & 0.821 & 0.549 & 0.413\\
Linear projections & 0.775 & 0.970 & 0.844 & 0.957 & -- & --\\
Width $1{,}024$ & 0.790 & 0.749 & 0.836 & 0.952 & -- & --\\
Three hidden layers & 0.787 & 0.000 & 0.813 & 0.553 & -- & --\\
All-positive supervision & 0.805 & 0.636 & 0.849 & 0.767 & -- & --\\
Mean over expression & 0.721 & 0.746 & 0.790 & 0.737 & 0.582 & 0.232\\
Three VLM depths & 0.817 & 0.937 & 0.849 & 0.937 & 0.542 & 0.431\\
\quad + all positives & 0.842 & 0.757 & 0.866 & 0.896 & 0.595 & 0.403\\
\bottomrule
\end{tabular}
\end{table}

\paragraph{Coordinate decoding, abstention and in-VLM selection with Qwen3-VL.}
\Cref{tab:app-full-dec} compares the raw boxes of the two VLMs.
\Cref{tab:app-full-abst} repeats the abstention comparison of
\cref{tab:app-detach,tab:app-detach-ood} with Qwen3-VL and its full-budget hook and verifier: the ordering is unchanged,
with the verifier at the decoded proposal ahead of the tuned yes-probability
($0.907$ against $0.891$ joint accuracy on RefCOCOg test) and of the plain yes/no
answer ($0.860$); as an existence classifier the VLM's yes-minus-no score has
the same AUROC with both VLMs ($0.989$ on RefCOCOg test). The hard-negative
subset was not scored for Qwen3-VL. \Cref{tab:app-full-q3} repeats the box-list
prompt over the Qwen3-VL hook's five highest-ranked proposals: below the hook on
RefCOCOg val, above it on Ref-L4 and the stress set, and below the VLM's own
coordinate decoding everywhere, as with Qwen2.5-VL in the main paper.

\begin{table}[t]
\centering
\footnotesize
\setlength{\tabcolsep}{1.8pt}
\caption{Raw coordinate decoding of the two VLMs (present queries;
$95\%$ query-level intervals in parentheses for the Qwen3-VL minus Qwen2.5-VL
difference at $\mathrm{IoU}\ge0.5$). Snapped to the proposals, Qwen3-VL scores
$0.890$, $0.889$ and $0.684$ on RefCOCOg val, test and stress.}
\label{tab:app-full-dec}
\begin{tabular}{@{}lccccc@{}}
\toprule
& \multicolumn{2}{c}{$\mathrm{IoU}\ge0.5$} & \multicolumn{2}{c}{$\mathrm{IoU}\ge0.75$} & \\
\cmidrule(lr){2-3}\cmidrule(lr){4-5}
Split & Qwen2.5 & Qwen3 & Qwen2.5 & Qwen3 & Difference\\
\midrule
RefCOCO val & 0.893 & 0.918 & 0.791 & 0.832 & $+0.025$\ci{.020}{.030}\\
RefCOCO+ val & 0.838 & 0.868 & 0.742 & 0.789 & $+0.030$\ci{.023}{.037}\\
RefCOCOg val & 0.865 & 0.886 & 0.749 & 0.801 & $+0.020$\ci{.012}{.029}\\
RefCOCOg test & 0.866 & 0.890 & 0.757 & 0.807 & $+0.023$\ci{.017}{.030}\\
Ref-L4 val & 0.853 & 0.880 & 0.724 & 0.774 & $+0.027$\ci{.021}{.032}\\
Stress & 0.623 & 0.671 & 0.468 & 0.554 & $+0.048$\ci{-.013}{.104}\\
\bottomrule
\end{tabular}
\end{table}

\begin{table*}[t]
\centering
\footnotesize
\setlength{\tabcolsep}{4pt}
\caption{Abstention with $30\%$ swapped absent referents (joint accuracy) for
Qwen3-VL with its full-budget hook and verifier, next to the Qwen2.5-VL rows of
\cref{tab:app-detach,tab:app-detach-ood} ($24$k hook). Thresholds are tuned on RefCOCOg val where a row
tunes one; hook-based rows are means over three seeds.}
\label{tab:app-full-abst}
\begin{tabular}{@{}lcccccc@{}}
\toprule
& \multicolumn{3}{c}{Qwen3-VL, full training set} & \multicolumn{3}{c}{Qwen2.5-VL, $24$k}\\
\cmidrule(lr){2-4}\cmidrule(l){5-7}
System & RefCOCOg val & RefCOCOg test & Stress & RefCOCOg val & RefCOCOg test & Stress\\
\midrule
Coord.-snap, always & 0.623 & 0.623 & 0.479 & 0.607 & 0.606 & 0.427\\
\quad + VLM yes/no & 0.862 & 0.860 & 0.612 & 0.840 & 0.841 & 0.545\\
\quad + VLM yes-prob. & 0.890 & 0.891 & 0.615 & 0.873 & 0.877 & 0.564\\
\quad + verifier & 0.906 & 0.907 & 0.558 & 0.888 & 0.888 & 0.528\\
\quad + verifier, hook's pick & 0.906 & 0.906 & 0.638 & 0.886 & 0.887 & 0.580\\
Det.-native, own score & 0.651 & 0.648 & 0.276 & 0.651 & 0.648 & 0.276\\
Det.-native + verifier & 0.834 & 0.832 & 0.409 & 0.832 & 0.831 & 0.390\\
Hook, default veto & 0.863 & 0.867 & 0.495 & 0.780 & 0.788 & 0.413\\
Hook, verifier only & 0.880 & 0.884 & 0.513 & 0.830 & 0.837 & 0.478\\
\bottomrule
\end{tabular}
\end{table*}

\begin{table}[t]
\centering
\footnotesize
\setlength{\tabcolsep}{3pt}
\caption{Box-list selection by Qwen3-VL over its full-budget hook's five
highest-ranked proposals (the seed-$42$ hook ranks), on the same queries as the
hook's top-1 and the VLM's raw coordinate decoding (Ref-L4: the fixed
$3{,}000$-query subset of Table 3 of the main paper).}
\label{tab:app-full-q3}
\begin{tabular}{@{}lccc@{}}
\toprule
& RefCOCOg val & Ref-L4 & Stress\\
\midrule
Hook, top-1 & 0.856 & 0.530 & 0.446\\
Correct box in hook's top 5 & 0.982 & 0.775 & 0.714\\
VLM over hook's top 5: box list & 0.848 & 0.690 & 0.550\\
Coordinate decoding & \textbf{0.886} & \textbf{0.873} & \textbf{0.671}\\
\bottomrule
\end{tabular}
\end{table}

\paragraph{Text-only queries with the strongest heads.}
\label{sec:app-textq-heads}
The query-source comparison of the main paper holds the head fixed at the
learned bridge, the weakest full-budget head. \Cref{tab:app-textq-heads}
repeats it with the two strongest set-level heads, trained on the full RefCOCOg
training set with the proposals, recipes and three seeds of the
image-conditioned heads: the relation-aware head on the text-only query (the
same final-layer readout from a pass over the request without the image), and
the token-state probe on the final-layer states of the request tokens of that
image-free pass. Differences are paired over queries with two-level
(seed $\times$ query) intervals, clustered by template on the stress set. On
RefCOCOg the effect of the query does not depend on the head: on RefCOCOg test
the image-conditioned query leads the text-only one by $0.162$ [$0.151$,
$0.172$] for the bridge, $0.154$ [$0.143$, $0.164$] for the relation-aware head
and $0.167$ [$0.158$, $0.176$] for the token-state probe ($0.162$, $0.156$ and
$0.173$ on val), and on RefCOCO and RefCOCO+ val by $0.155$--$0.272$ for the
bridge and the relation-aware head (every interval above $0.14$; $0.179$ and
$0.282$ for the token-state probe, without interval). The two factors are close
to additive: on RefCOCOg test a set-level head adds $0.036$--$0.049$ over the
bridge with either query, a quarter to a third of the query effect, and with the
text-only query the three heads differ by at most $0.049$ on every
RefCOCO-family split. Ref-L4 behaves differently. The query effect shrinks to
$0.103$ [$0.088$, $0.119$], $0.130$ [$0.121$, $0.138$] and $0.080$ [$0.068$,
$0.092$] for the three heads, and with the text-only query the token-state
probe exceeds the bridge by $0.105$ [$0.092$, $0.118$], more than its own query
effect; the gain comes from COCO-category referents ($+0.155$ [$0.135$,
$0.177$], against $+0.018$ [$0.006$, $0.030$] on the rest). Outside the COCO
categories every combination of head and query stays at $0.19$--$0.27$, and the
image adds about $0.06$ to each head. On the stress set the image adds $0.105$
(template interval [$0.018$, $0.200$]) to the bridge and $0.095$ [$0.005$,
$0.203$] to the token-state probe, but only $0.048$ [$-0.024$, $0.125$] to the
relation-aware head. On the RefCOCO family, therefore, the query the VLM
computes with the image decides accuracy whatever the head; on the long
expressions of Ref-L4 the form of the head matters as much, and cross-attention
to the request tokens recovers part of the image-conditioned query's advantage
on COCO-category referents.

\paragraph{Set-level scoring at matched capacity, both VLMs.}
The main paper compares the relation-aware head with a hook of its parameter
count ($4.29$M) and all-positive supervision; with Qwen3-VL that hook is now
trained on the full set with the same three seeds
(\cref{tab:app-matched-hook-q3}). It is itself $0.018$--$0.021$ above the
default hook on the standard splits, and the relation-aware head leads it by
$0.033$ [$0.027$, $0.039$] on RefCOCO val, $0.035$ [$0.030$, $0.041$] on
RefCOCO+ val, $0.027$ [$0.019$, $0.035$] on RefCOCOg val and $0.023$ [$0.017$,
$0.028$] on RefCOCOg test, with every seed pair at least $+0.019$; with
Qwen2.5-VL the lead is $0.022$--$0.031$. On Ref-L4 the lead is $0.033$
[$0.008$, $0.070$] with Qwen3-VL (worst seed pair $+0.006$ [$-0.000$,
$+0.013$]) and $0.018$ [$0.008$, $0.028$] with Qwen2.5-VL, and it comes from
COCO-category referents ($+0.065$ and $+0.051$); outside the COCO categories it
reverses ($-0.022$ [$-0.034$, $-0.010$] and $-0.038$ [$-0.050$, $-0.026$]). On
the stress set the relation-aware head is behind the matched hook with both
VLMs: $-0.036$ with Qwen3-VL, not significant when clustered by template
([$-0.093$, $+0.021$]), and $-0.051$ [$-0.098$, $-0.007$] with Qwen2.5-VL.
Set-level scoring thus adds $0.02$--$0.04$ over a hook of the same size on every
standard split for both VLMs, but not beyond the COCO categories or on the
stress set.

\begin{table*}[t]
\centering
\footnotesize
\setlength{\tabcolsep}{3.4pt}
\caption{Query source crossed with head form (Qwen2.5-VL, full RefCOCOg
training set): accuracy at $\mathrm{IoU}\ge0.5$ on present queries, no
abstention, mean over three seeds (sample std $\le0.021$; two-level $95\%$
half-widths $0.006$--$0.021$, on Stress $0.054$--$0.061$). Image: the
image-conditioned query (bridge, relation-aware head) or the request-token
states read with the image (token-state probe); text: the same readouts from a
pass over the request without the image. The bridge on the image-conditioned
query is the hook trained without the verifier. Differences are paired over
queries; ${}^\dagger$their two-level (seed $\times$ query) $95\%$ interval
excludes zero (Stress: clustered by template). ${}^\ast$From the evaluation of
\cref{tab:app-allsplits}, whose token-state probe reproduces this probe's
per-query outcomes on every other split; the difference has no interval. Ref-L4
COCO cat.\ / other: referents inside / outside the $80$ COCO categories.}
\label{tab:app-textq-heads}
\begin{tabular}{@{}llcccccccc@{}}
\toprule
& & RefCOCO & RefCOCO+ & \multicolumn{2}{c}{RefCOCOg} & \multicolumn{3}{c}{Ref-L4 val} & \\
\cmidrule(lr){5-6}\cmidrule(lr){7-9}
Head & Query & val & val & val & test & all & COCO cat. & other & Stress\\
\midrule
\multirow{3}{*}{Learned bridge} & image & 0.839 & 0.760 & 0.834 & 0.838 & 0.546 & 0.714 & 0.256 & 0.417\\
& text & 0.683 & 0.517 & 0.672 & 0.676 & 0.442 & 0.587 & 0.194 & 0.312\\
& difference & $+0.155^\dagger$ & $+0.243^\dagger$ & $+0.162^\dagger$ & $+0.162^\dagger$ & $+0.103^\dagger$ & $+0.127^\dagger$ & $+0.062^\dagger$ & $+0.105^\dagger$\\
\midrule
\multirow{3}{*}{Relation-aware head} & image & 0.872 & 0.802 & 0.877 & 0.878 & 0.595 & 0.798 & 0.247 & 0.358\\
& text & 0.687 & 0.530 & 0.721 & 0.725 & 0.466 & 0.629 & 0.186 & 0.310\\
& difference & $+0.185^\dagger$ & $+0.272^\dagger$ & $+0.156^\dagger$ & $+0.154^\dagger$ & $+0.130^\dagger$ & $+0.169^\dagger$ & $+0.061^\dagger$ & $+0.048$\\
\midrule
\multirow{3}{*}{Token-state probe} & image & $0.886^\ast$ & $0.815^\ast$ & 0.871 & 0.878 & 0.627 & 0.834 & 0.272 & 0.355\\
& text & 0.707 & 0.533 & 0.698 & 0.711 & 0.547 & 0.742 & 0.212 & 0.260\\
& difference & $+0.179^\ast$ & $+0.282^\ast$ & $+0.173^\dagger$ & $+0.167^\dagger$ & $+0.080^\dagger$ & $+0.092^\dagger$ & $+0.060^\dagger$ & $+0.095^\dagger$\\
\midrule
\multicolumn{10}{@{}l}{\emph{Head contrasts at a fixed query}}\\
\multicolumn{2}{@{}l}{Relation $-$ bridge, image} & $+0.033^\dagger$ & $+0.042^\dagger$ & $+0.043^\dagger$ & $+0.041^\dagger$ & $+0.050^\dagger$ & $+0.084^\dagger$ & $-0.009$ & $-0.059^\dagger$\\
\multicolumn{2}{@{}l}{Relation $-$ bridge, text} & $+0.004$ & $+0.014^\dagger$ & $+0.049^\dagger$ & $+0.049^\dagger$ & $+0.023^\dagger$ & $+0.042^\dagger$ & $-0.008$ & $-0.001$\\
\multicolumn{2}{@{}l}{Token $-$ bridge, image} & -- & -- & $+0.037^\dagger$ & $+0.040^\dagger$ & $+0.082^\dagger$ & $+0.120^\dagger$ & $+0.016^\dagger$ & $-0.062^\dagger$\\
\multicolumn{2}{@{}l}{Token $-$ bridge, text} & $+0.023^\dagger$ & $+0.016^\dagger$ & $+0.026^\dagger$ & $+0.036^\dagger$ & $+0.105^\dagger$ & $+0.155^\dagger$ & $+0.018^\dagger$ & $-0.052^\dagger$\\
\bottomrule
\end{tabular}
\end{table*}

\begin{table*}[t]
\centering
\footnotesize
\setlength{\tabcolsep}{3.6pt}
\caption{Set-level scoring against a hook of the same size, both VLMs (full
RefCOCOg training set): accuracy at $\mathrm{IoU}\ge0.5$ on present queries, no
abstention, mean over three seeds. Hook: the default alignment hook ($1.12$M
parameters); matched: the capacity-matched hook ($4.29$M parameters, the
relation-aware head's count, with the same all-positive supervision); relation:
the relation-aware head. $\Delta$: relation minus matched with its two-level
(seed $\times$ query) $95\%$ interval, clustered by template on the stress set.
One seed of the Qwen3-VL matched hook is $0.05$--$0.06$ below the other two on
Ref-L4 (sample std $0.034$), which widens that interval.}
\label{tab:app-matched-hook-q3}
\begin{tabular}{@{}lcccccccc@{}}
\toprule
& \multicolumn{4}{c}{Qwen2.5-VL-7B} & \multicolumn{4}{c}{Qwen3-VL-8B}\\
\cmidrule(lr){2-5}\cmidrule(l){6-9}
Split & Hook & Matched & Relation & $\Delta$ & Hook & Matched & Relation & $\Delta$\\
\midrule
RefCOCO val & 0.834 & 0.850 & 0.872 & $+0.022$\ci{.014}{.030} & 0.853 & 0.872 & 0.905 & $+0.033$\ci{.027}{.039}\\
RefCOCO+ val & 0.760 & 0.774 & 0.802 & $+0.028$\ci{.018}{.039} & 0.800 & 0.818 & 0.853 & $+0.035$\ci{.030}{.041}\\
RefCOCOg val & 0.843 & 0.851 & 0.877 & $+0.026$\ci{.018}{.034} & 0.853 & 0.873 & 0.900 & $+0.027$\ci{.019}{.035}\\
RefCOCOg test & 0.837 & 0.848 & 0.878 & $+0.031$\ci{.025}{.036} & 0.856 & 0.877 & 0.899 & $+0.023$\ci{.017}{.028}\\
Ref-L4 val & 0.549 & 0.577 & 0.595 & $+0.018$\ci{.008}{.028} & 0.531 & 0.572 & 0.605 & $+0.033$\ci{.008}{.070}\\
\quad COCO cat. & 0.718 & 0.747 & 0.798 & $+0.051$\ci{.034}{.067} & 0.695 & 0.749 & 0.814 & $+0.065$\ci{.030}{.118}\\
\quad other & 0.259 & 0.285 & 0.247 & $-0.038$\ci{$-$.050}{$-$.026} & 0.250 & 0.269 & 0.247 & $-0.022$\ci{$-$.034}{$-$.010}\\
Stress & 0.413 & 0.408 & 0.358 & $-0.051$\ci{$-$.098}{$-$.007} & 0.437 & 0.444 & 0.408 & $-0.036$\ci{$-$.093}{.021}\\
\bottomrule
\end{tabular}
\end{table*}

\paragraph{Image overlap at the full budget.}
The full training set contains $918$ of the $1{,}500$ RefCOCO and RefCOCO+ val
images: subset A of \cref{tab:app-leakage} (images of the $24$k training rows)
and subset B (images first seen at the full budget). \Cref{tab:app-full-leak}
repeats the split. If the heads memorized images, their gain from full data
would be larger on B than on C (images outside RefCOCOg). For the Qwen2.5-VL
hook it is not: $+0.042$ on B against $+0.053$ on C on RefCOCO val (difference
$-0.011$ [$-0.024$, $+0.001$]) and $+0.003$ [$-0.010$, $+0.017$] on RefCOCO+ val.
The relation-aware head gains $0.013$ [$0.002$, $0.025$] more on B than on C, but
equally more on images it had already seen at $24$k ($+0.018$ [$0.007$, $0.029$]
for A against C), so the excess follows RefCOCOg's image selection rather than
newly seen images. As at $24$k, every trained head is $0.03$--$0.07$ lower on C
than on all rows and coordinate decoding $0.01$--$0.02$ lower; restricted to C,
the full-budget relation-aware head scores $0.822$ (Qwen2.5-VL) and $0.872$
(Qwen3-VL) on RefCOCO val against $0.875$ and $0.905$ for coordinate decoding.

\begin{table*}[t]
\centering
\footnotesize
\setlength{\tabcolsep}{4pt}
\caption{RefCOCO val accuracy by image subset (A: image of the $24$k training
rows; B: RefCOCOg train image first seen at the full budget; C: image not in
RefCOCOg), with $95\%$ query-level intervals. Trained heads: per-query
correctness averaged over three seeds; coordinate decoding: raw box.}
\label{tab:app-full-leak}
\begin{tabular}{@{}lcccccccc@{}}
\toprule
& & \multicolumn{3}{c}{Qwen2.5-VL} & \multicolumn{2}{c}{Qwen3-VL} & \multicolumn{2}{c}{Coordinate decoding}\\
\cmidrule(lr){3-5}\cmidrule(lr){6-7}\cmidrule(l){8-9}
Subset & Rows (images) & Hook, $24$k & Hook, full & Relation, full & Hook, full & Relation, full & Qwen2.5 & Qwen3\\
\midrule
A & $3{,}373$ ($481$) & 0.839\ci{.829}{.850} & 0.877\ci{.867}{.886} & 0.913\ci{.905}{.922} & 0.900\ci{.892}{.908} & 0.935\ci{.927}{.942} & 0.908\ci{.898}{.918} & 0.930\ci{.921}{.938}\\
B & $2{,}846$ ($437$) & 0.834\ci{.822}{.845} & 0.875\ci{.865}{.885} & 0.903\ci{.894}{.913} & 0.891\ci{.882}{.901} & 0.925\ci{.916}{.934} & 0.905\ci{.895}{.916} & 0.926\ci{.917}{.936}\\
C & $4{,}615$ ($582$) & 0.726\ci{.714}{.736} & 0.778\ci{.768}{.788} & 0.822\ci{.812}{.832} & 0.796\ci{.786}{.805} & 0.872\ci{.863}{.881} & 0.875\ci{.866}{.885} & 0.905\ci{.897}{.914}\\
All & $10{,}834$ ($1{,}500$) & 0.789\ci{.783}{.796} & 0.834\ci{.828}{.840} & 0.872\ci{.866}{.878} & 0.853\ci{.848}{.859} & 0.905\ci{.900}{.911} & 0.893\ci{.888}{.899} & 0.918\ci{.913}{.924}\\
\bottomrule
\end{tabular}
\end{table*}

\paragraph{RefCOCO+ and the test splits.}
The same split holds on RefCOCO+ val: on images outside RefCOCOg the hook
($24$k) falls from $0.707$ to $0.645$, the full-budget relation-aware head from
$0.802$ to $0.746$ and coordinate decoding from $0.838$ to $0.824$. On these
images the published full-budget relation-aware head trails decoding by $0.053$
[$0.040$, $0.065$] on RefCOCO val and $0.078$ [$0.063$, $0.093$] on RefCOCO+
val (two-level intervals; its retraining in \cref{sec:app-allsplits}: $0.048$
and $0.077$); with Qwen3-VL by $0.034$ and $0.036$, and the
Qwen3-VL combined selector by $0.024$ [$0.016$, $0.032$] and $0.027$ [$0.018$,
$0.036$]. RefCOCO/+ testA
and testB share the pattern: $60.5$--$65.1\%$ of their images are RefCOCOg
training images, and on the rest the $24$k relation-aware head trails decoding
by $0.06$--$0.08$, the retrained full-budget relation-aware head by $0.049$ and $0.056$ on testA and by $0.014$ and $0.018$ (intervals including zero) on testB, and the full-budget combined selector by at most $0.016$ (\cref{sec:app-allsplits}). The pooled RefCOCO/+ columns of the main paper therefore favor
the trained heads; they measure held-out expressions, not held-out images.

\paragraph{The full-budget heads on every RefCOCO and RefCOCO+ split.}
\label{sec:app-allsplits}
The published runs of cross-attention to the pooled query, the token-state
probe and the combined selector were evaluated on RefCOCOg, Ref-L4 and the
stress set only; their RefCOCO and RefCOCO+ val cells in the main paper's
two-VLM table come from the Qwen2.5-VL retraining described here and from a
same-seed retraining of the Qwen3-VL token-state probe and combined selector
(\cref{tab:app-full-acc}). \Cref{tab:app-allsplits} evaluates the full-budget Qwen2.5-VL heads
on the six RefCOCO and RefCOCO+ splits, overall and by image subset, and
\cref{tab:app-allsplits-diffs} gives the paired comparisons. The heads were
retrained with the published recipe and seeds and their weights kept:
cross-attention to the pooled query and the token-state probe reproduce every
per-query outcome of the published heads on RefCOCOg, Ref-L4 and the stress
set, and the combined selector and the relation-aware head agree with them on
$0.935$--$0.965$ of per-query outcomes and within $0.002$ in mean accuracy
outside the stress set (combined selector $-0.010$ on it). On all rows the
combined selector is the most accurate head on every split, ahead of the
relation-aware head by $0.018$--$0.028$, of cross-attention to the pooled query
by $0.016$--$0.028$ and of the token-state probe by $0.006$--$0.012$ (every
two-level interval above zero except the last on RefCOCO+ testB). Unlike on
RefCOCOg, token states help on these splits: the token-state probe leads the
relation-aware head by $0.012$--$0.020$ on the val and testA splits. Against
coordinate decoding the combined selector is level on RefCOCO val ($+0.001$
[$-0.005$, $+0.007$]) and on both testA splits ($-0.006$ and $-0.008$,
intervals reaching zero), ahead on both testB splits ($+0.013$ [$0.002$,
$0.023$] and $+0.015$ [$0.003$, $0.026$]) and behind on RefCOCO+ val ($-0.014$
[$-0.021$, $-0.007$]), whereas the relation-aware head trails decoding by
$0.020$--$0.036$ on the val and testA splits and is level with it on testB.

Subset C, the images outside RefCOCOg train, holds $39$--$43\%$ of the rows of
each split and was seen by no head. Every system is less accurate there and the
trained heads lose more than decoding, as in \cref{sec:app-leakage}, but the
combined selector's lead over the relation-aware head grows to
$0.022$--$0.041$, and its deficit to decoding on the val and testA splits
($0.014$--$0.041$) is a quarter to a half of the relation-aware head's
($0.048$--$0.077$); on both testB splits it is level with decoding ($+0.008$
[$-0.011$, $+0.027$] and $+0.005$ [$-0.016$, $+0.026$]). On the training images
(subsets A and B, both seen at this budget) the combined selector is never more
than $0.005$ below decoding. Its remaining deficit on RefCOCO and RefCOCO+ is
therefore confined to images outside RefCOCOg's image selection and is largest
on RefCOCO+ val ($-0.041$ [$-0.052$, $-0.029$]); the pooled RefCOCO/+ columns
of the main paper overstate held-out-image accuracy for every trained head.

\begin{table*}[t]
\centering
\footnotesize
\setlength{\tabcolsep}{3.5pt}
\caption{Full-budget heads (Qwen2.5-VL) on the RefCOCO and RefCOCO+ splits by
image subset: accuracy at $\mathrm{IoU}\ge0.5$ on present queries, no
abstention. Trained rows: mean over three seeds (all rows: $\pm$sample std);
two-level (seed $\times$ query) $95\%$ half-widths $0.006$--$0.013$ on all rows
and $0.009$--$0.021$ on the subsets. Coordinate decoding: the VLM's raw box,
one run (query-bootstrap half-widths $0.006$--$0.012$ and $0.010$--$0.021$).
Block headers give rows (images). A: image of the $24$k training rows; B: other
RefCOCOg train image; C: image outside RefCOCOg train, seen by no head. The
relation-aware head is the retraining paired with the other heads (published
run: $0.872$ and $0.802$ on RefCOCO and RefCOCO+ val).}
\label{tab:app-allsplits}
\begin{tabular}{@{}lcccccc@{}}
\toprule
& \multicolumn{3}{c}{RefCOCO} & \multicolumn{3}{c}{RefCOCO+}\\
\cmidrule(lr){2-4}\cmidrule(l){5-7}
& val & testA & testB & val & testA & testB\\
\midrule
\emph{All rows} & $10{,}834$ ($1{,}500$) & $5{,}657$ ($750$) & $5{,}095$ ($750$) & $10{,}758$ ($1{,}500$) & $5{,}726$ ($750$) & $4{,}889$ ($750$)\\
\quad Pooled query, cross-attention & 0.876\sd{.001} & 0.902\sd{.001} & 0.845\sd{.002} & 0.808\sd{.001} & 0.867\sd{.001} & 0.744\sd{.004}\\
\quad Token-state probe & 0.886\sd{.002} & 0.915\sd{.001} & 0.852\sd{.005} & 0.815\sd{.003} & 0.875\sd{.003} & 0.764\sd{.004}\\
\quad Combined selector & 0.895\sd{.003} & 0.921\sd{.001} & 0.864\sd{.004} & 0.824\sd{.002} & 0.884\sd{.001} & 0.772\sd{.001}\\
\quad Relation-aware head & 0.874\sd{.001} & 0.898\sd{.002} & 0.846\sd{.003} & 0.802\sd{.002} & 0.855\sd{.001} & 0.748\sd{.003}\\
\quad Coordinate decoding & 0.893 & 0.928 & 0.851 & 0.838 & 0.892 & 0.758\\
\midrule
\emph{A: $24$k training images} & $3{,}373$ ($481$) & $1{,}864$ ($249$) & $1{,}438$ ($228$) & $3{,}344$ ($481$) & $1{,}888$ ($249$) & $1{,}428$ ($228$)\\
\quad Pooled query, cross-attention & 0.903 & 0.900 & 0.886 & 0.830 & 0.872 & 0.782\\
\quad Token-state probe & 0.915 & 0.921 & 0.898 & 0.843 & 0.883 & 0.812\\
\quad Combined selector & 0.924 & 0.922 & 0.908 & 0.857 & 0.887 & 0.817\\
\quad Relation-aware head & 0.913 & 0.904 & 0.891 & 0.841 & 0.874 & 0.787\\
\quad Coordinate decoding & 0.908 & 0.918 & 0.902 & 0.848 & 0.887 & 0.796\\
\midrule
\emph{B: other RefCOCOg train} & $2{,}846$ ($437$) & $1{,}404$ ($205$) & $1{,}597$ ($260$) & $2{,}871$ ($437$) & $1{,}417$ ($205$) & $1{,}562$ ($260$)\\
\quad Pooled query, cross-attention & 0.901 & 0.922 & 0.883 & 0.842 & 0.892 & 0.802\\
\quad Token-state probe & 0.905 & 0.933 & 0.889 & 0.839 & 0.899 & 0.816\\
\quad Combined selector & 0.914 & 0.938 & 0.904 & 0.850 & 0.912 & 0.833\\
\quad Relation-aware head & 0.903 & 0.924 & 0.891 & 0.843 & 0.884 & 0.813\\
\quad Coordinate decoding & 0.905 & 0.942 & 0.879 & 0.849 & 0.917 & 0.813\\
\midrule
\emph{C: outside RefCOCOg train} & $4{,}615$ ($582$) & $2{,}389$ ($296$) & $2{,}060$ ($262$) & $4{,}543$ ($582$) & $2{,}421$ ($296$) & $1{,}899$ ($262$)\\
\quad Pooled query, cross-attention & 0.842 & 0.893 & 0.786 & 0.769 & 0.849 & 0.668\\
\quad Token-state probe & 0.852 & 0.899 & 0.791 & 0.779 & 0.856 & 0.685\\
\quad Combined selector & 0.861 & 0.910 & 0.802 & 0.784 & 0.865 & 0.688\\
\quad Relation-aware head & 0.827 & 0.878 & 0.779 & 0.748 & 0.824 & 0.665\\
\quad Coordinate decoding & 0.875 & 0.926 & 0.794 & 0.824 & 0.880 & 0.683\\
\bottomrule
\end{tabular}
\end{table*}

\begin{table*}[t]
\centering
\footnotesize
\setlength{\tabcolsep}{4pt}
\caption{Paired differences on the RefCOCO and RefCOCO+ splits for the
full-budget Qwen2.5-VL heads of \cref{tab:app-allsplits}, on all rows and on
subset C (images outside RefCOCOg train), with two-level (seed $\times$ query)
$95\%$ bootstrap intervals ($10{,}000$ resamples; rows drawn once and shared by
both systems). Coordinate decoding: the VLM's raw box, one run.}
\label{tab:app-allsplits-diffs}
\begin{tabular}{@{}llcccc@{}}
\toprule
& & \multicolumn{3}{c}{Combined selector minus} & Relation-aware head\\
\cmidrule(lr){3-5}
Split & Rows & relation-aware head & token-state probe & decoding & minus decoding\\
\midrule
RefCOCO val & all & $+0.021$\ci{.015}{.026} & $+0.009$\ci{.004}{.014} & $+0.001$\ci{$-$.005}{.007} & $-0.020$\ci{$-$.026}{$-$.014}\\
& C & $+0.034$\ci{.025}{.044} & $+0.009$\ci{.001}{.016} & $-0.014$\ci{$-$.024}{$-$.004} & $-0.048$\ci{$-$.060}{$-$.038}\\
RefCOCO testA & all & $+0.023$\ci{.017}{.030} & $+0.006$\ci{.001}{.012} & $-0.006$\ci{$-$.013}{.000} & $-0.030$\ci{$-$.037}{$-$.022}\\
& C & $+0.033$\ci{.022}{.044} & $+0.011$\ci{.002}{.020} & $-0.016$\ci{$-$.027}{$-$.005} & $-0.049$\ci{$-$.062}{$-$.036}\\
RefCOCO testB & all & $+0.018$\ci{.009}{.026} & $+0.012$\ci{.004}{.020} & $+0.013$\ci{.002}{.023} & $-0.005$\ci{$-$.016}{.006}\\
& C & $+0.022$\ci{.007}{.038} & $+0.011$\ci{$-$.003}{.025} & $+0.008$\ci{$-$.011}{.027} & $-0.014$\ci{$-$.033}{.005}\\
\midrule
RefCOCO+ val & all & $+0.022$\ci{.016}{.029} & $+0.009$\ci{.003}{.015} & $-0.014$\ci{$-$.021}{$-$.007} & $-0.036$\ci{$-$.043}{$-$.029}\\
& C & $+0.036$\ci{.025}{.048} & $+0.005$\ci{$-$.004}{.014} & $-0.041$\ci{$-$.052}{$-$.029} & $-0.077$\ci{$-$.089}{$-$.064}\\
RefCOCO+ testA & all & $+0.028$\ci{.021}{.036} & $+0.009$\ci{.002}{.015} & $-0.008$\ci{$-$.016}{.000} & $-0.036$\ci{$-$.045}{$-$.028}\\
& C & $+0.041$\ci{.029}{.054} & $+0.009$\ci{$-$.001}{.018} & $-0.015$\ci{$-$.029}{$-$.002} & $-0.056$\ci{$-$.072}{$-$.042}\\
RefCOCO+ testB & all & $+0.024$\ci{.014}{.035} & $+0.008$\ci{$-$.000}{.017} & $+0.015$\ci{.003}{.026} & $-0.010$\ci{$-$.022}{.003}\\
& C & $+0.023$\ci{.006}{.041} & $+0.003$\ci{$-$.012}{.017} & $+0.005$\ci{$-$.016}{.026} & $-0.018$\ci{$-$.040}{.004}\\
\bottomrule
\end{tabular}
\end{table*}
\paragraph{Training on RefCOCO, RefCOCO+ and RefCOCOg together.}
\label{sec:app-union}
Every trained head of the main paper sees RefCOCOg train only. To measure what
more referring data of the same family buys, we retrain the token-state probe
and the combined selector (only these two; neither the hook nor the
relation-aware head was retrained) with the full-budget recipe and seeds on the union of the RefCOCO and RefCOCO+
(UNC splits) and RefCOCOg (umd split) training sets, with the same frozen
proposals ($80$-category prompt, $M{=}300$) and the same frozen Qwen2.5-VL-7B
states. Every image of an evaluation split (RefCOCOg val and test, RefCOCO and
RefCOCO+ val, testA and testB) is removed from all three sources: $1{,}860$
RefCOCOg training images, which carry $8{,}143$ of its $80{,}512$ present
expressions, and the $1{,}891$ RefCOCOg val and test images that RefCOCO and
RefCOCO+ train each contain. RefCOCO and RefCOCO+ contribute at most one
expression per object and dataset, drawn at random after dropping texts
already taken for the same object by an earlier source (order RefCOCOg,
RefCOCO, RefCOCO+; $6{,}696$ and $10{,}245$ duplicates); without this cap the
union would hold $287{,}604$ present expressions. Each source receives $30\%$
swapped absent rows, an expression paired with another training image of the
same source; the RefCOCOg part keeps the full-budget swaps on its remaining
images and is topped up to $30\%$. The union has $210{,}836$ rows
($147{,}585$ present) on $24{,}407$ images: RefCOCOg $103{,}384$ rows
($72{,}369$ present) on $20{,}039$ images, RefCOCO $54{,}166$ ($37{,}916$) on
$15{,}101$ and RefCOCO+ $53{,}286$ ($37{,}300$) on $15{,}067$. As a control
for the smaller RefCOCOg part, both heads are also trained with the same recipe
and seeds on that part alone ($103{,}384$ rows). Of its images,
$4{,}368$ were new to the feature cache; re-detecting $16$ cached images
reproduced their boxes exactly (feature cosine $\ge0.99999$). Proposal recall
on the union's present rows is $0.999$ for every source. No union training
image is in an evaluation split or among the $4{,}802$ COCO images of Ref-L4
val, whereas RefCOCOg train contains $918$ of the $1{,}500$ RefCOCO/+ val
images and $1{,}452$ of those Ref-L4 images. Training took $40$--$66$\,min per
seed for the token-state probe and $62$--$67$\,min for the combined selector on
one RTX PRO 6000 shared with other jobs (peak $26.0$\,GB).

\begin{table*}[t]
\centering
\footnotesize
\setlength{\tabcolsep}{3.5pt}
\caption{Training on the union of RefCOCO, RefCOCO+ and RefCOCOg train
(Qwen2.5-VL-7B; accuracy at $\mathrm{IoU}\ge0.5$ on present queries, no
abstention; trained rows: mean over three seeds, sample std $\le0.009$ outside
the stress set and $\le0.016$ on it). Union: $210{,}836$ rows with every
evaluation image removed (\cref{sec:app-union}); RefCOCOg only: the full-budget
heads of Table 1 of the main paper; RefCOCOg part: the control trained on the
union's $103{,}384$ RefCOCOg rows alone. All: every query of the split; C: the
$582$ images outside RefCOCOg train ($4{,}615$ RefCOCO and $4{,}543$ RefCOCO+
queries; \cref{tab:app-full-leak}). Ref-L4 val: all $13{,}420$ queries, split
as in Table 1 of the main paper into COCO-category referents (COCO images) and
the rest (Objects365 images). Coordinate decoding: the VLM's raw box, no task
training. $^\dagger$The retraining of \cref{sec:app-allsplits}, since the
published runs were not evaluated on RefCOCO/+; it reproduces the published
token-state probe query for query and the published combined selector within
$0.002$ on RefCOCOg and Ref-L4.}
\label{tab:app-union}
\begin{tabular}{@{}lcccccccccc@{}}
\toprule
& \multicolumn{2}{c}{RefCOCOg} & \multicolumn{2}{c}{RefCOCO val} & \multicolumn{2}{c}{RefCOCO+ val} & \multicolumn{3}{c}{Ref-L4 val} & \\
\cmidrule(lr){2-3}\cmidrule(lr){4-5}\cmidrule(lr){6-7}\cmidrule(lr){8-10}
System & val & test & All & C & All & C & All & COCO cat. & other & Stress\\
\midrule
Token-state probe, union & 0.875 & 0.880 & 0.909 & 0.891 & 0.842 & 0.817 & 0.643 & 0.858 & 0.274 & 0.364\\
\quad RefCOCOg only & 0.871 & 0.878 & 0.886$^\dagger$ & 0.852$^\dagger$ & 0.815$^\dagger$ & 0.779$^\dagger$ & 0.627 & 0.834 & 0.272 & 0.355\\
\quad RefCOCOg part & 0.870 & 0.876 & 0.884 & 0.856 & 0.810 & 0.777 & 0.620 & 0.825 & 0.269 & 0.372\\
Combined selector, union & 0.877 & 0.889 & 0.912 & 0.893 & 0.846 & 0.820 & 0.640 & 0.853 & 0.274 & 0.374\\
\quad RefCOCOg only & 0.876 & 0.885 & 0.895$^\dagger$ & 0.861$^\dagger$ & 0.824$^\dagger$ & 0.784$^\dagger$ & 0.641 & 0.852 & 0.280 & 0.381\\
\quad RefCOCOg part & 0.875 & 0.882 & 0.887 & 0.858 & 0.816 & 0.782 & 0.643 & 0.851 & 0.285 & 0.382\\
\midrule
Coordinate decoding & 0.865 & 0.866 & 0.893 & 0.875 & 0.838 & 0.824 & 0.853 & 0.900 & 0.774 & 0.623\\
\bottomrule
\end{tabular}
\end{table*}

\begin{table*}[t]
\centering
\footnotesize
\setlength{\tabcolsep}{4pt}
\caption{Paired differences for \cref{tab:app-union} with $95\%$ two-level
(seed $\times$ query) bootstrap intervals ($10{,}000$ resamples; coordinate
decoding has one run and contributes only its queries; on the stress set
queries, not templates, are resampled). C: images outside RefCOCOg train.
$^\dagger$The RefCOCOg-only rows on RefCOCO/+ are the retraining of
\cref{sec:app-allsplits}. --: not computed.}
\label{tab:app-union-diffs}
\begin{tabular}{@{}lcccc@{}}
\toprule
& \multicolumn{2}{c}{Union $-$ RefCOCOg only} & \multicolumn{2}{c}{Combined selector $-$ decoding}\\
\cmidrule(lr){2-3}\cmidrule(l){4-5}
Split & Token-state probe & Combined selector & Union & RefCOCOg only\\
\midrule
RefCOCOg val & $+0.004$\ci{$-$.003}{.012} & $+0.001$\ci{$-$.005}{.007} & $+0.012$\ci{.003}{.021} & $+0.011$\ci{.001}{.020}\\
RefCOCOg test & $+0.002$\ci{$-$.003}{.007} & $+0.005$\ci{.000}{.009} & $+0.023$\ci{.017}{.030} & $+0.018$\ci{.012}{.025}\\
RefCOCO val$^\dagger$ & $+0.024$\ci{.019}{.028} & $+0.017$\ci{.012}{.022} & $+0.018$\ci{.013}{.024} & $+0.001$\ci{$-$.005}{.007}\\
\quad C$^\dagger$ & $+0.039$\ci{.031}{.046} & $+0.032$\ci{.025}{.040} & $+0.018$\ci{.009}{.027} & $-0.014$\ci{$-$.024}{$-$.004}\\
RefCOCO+ val$^\dagger$ & $+0.028$\ci{.022}{.034} & $+0.022$\ci{.017}{.028} & $+0.008$\ci{.001}{.015} & $-0.014$\ci{$-$.021}{$-$.007}\\
\quad C$^\dagger$ & $+0.038$\ci{.029}{.047} & $+0.036$\ci{.026}{.046} & $-0.004$\ci{$-$.016}{.007} & $-0.041$\ci{$-$.052}{$-$.029}\\
Ref-L4 val & $+0.015$\ci{.009}{.023} & $-0.001$\ci{$-$.007}{.005} & $-0.213$\ci{$-$.222}{$-$.205} & $-0.212$\ci{$-$.220}{$-$.204}\\
\quad COCO cat. & $+0.023$\ci{.016}{.032} & $+0.001$\ci{$-$.005}{.008} & $-0.046$\ci{$-$.054}{$-$.039} & --\\
\quad other & $+0.002$\ci{$-$.009}{.013} & $-0.006$\ci{$-$.017}{.005} & $-0.500$\ci{$-$.517}{$-$.483} & --\\
Stress & $+0.009$\ci{$-$.027}{.043} & $-0.007$\ci{$-$.042}{.026} & $-0.250$\ci{$-$.313}{$-$.186} & $-0.242$\ci{$-$.303}{$-$.182}\\
\bottomrule
\end{tabular}
\end{table*}

On RefCOCOg the added data change little
(\cref{tab:app-union,tab:app-union-diffs}): the combined selector moves by
$+0.001$ [$-0.005$, $+0.007$] on val and $+0.005$ [$0.000$, $0.009$] on test
and the token-state probe by $+0.004$ and $+0.002$, although the union holds
$10\%$ fewer RefCOCOg expressions. The control separates the two changes: on
that reduced RefCOCOg part alone both heads are level with full RefCOCOg
training on RefCOCOg (combined selector $-0.001$ [$-0.006$, $+0.004$] on val
and $-0.003$ [$-0.008$, $+0.002$] on test), so the smaller RefCOCOg part costs
nothing measurable there, and against the control the added RefCOCO/+
expressions are worth $+0.008$ [$0.003$, $0.013$] on RefCOCOg test for the
combined selector and $+0.005$ [$0.000$, $0.009$] for the token-state probe.
On RefCOCO and RefCOCO+ val both heads gain
$0.017$--$0.028$, and $0.032$--$0.039$ on the $582$ images outside RefCOCOg
train. The union heads saw none of these images, yet they also lead on the val
images that the RefCOCOg-only heads saw in training with other expressions
(combined selector $0.925$ against $0.919$ on RefCOCO val, $0.866$ against
$0.854$ on RefCOCO+ val), so the gain comes from the added expressions, not
from image exposure. With them the combined selector exceeds coordinate
decoding on RefCOCOg val and test and on RefCOCO and RefCOCO+ val; among the
RefCOCO-family columns it is level with decoding only on the RefCOCO+ images
outside RefCOCOg train ($-0.004$ [$-0.016$, $+0.007$], against $-0.041$ when
trained on RefCOCOg only). On Ref-L4 the combined selector does not change
($-0.001$ [$-0.007$, $+0.005$]); the token-state probe gains $0.015$, almost
all of it on COCO-category referents ($+0.023$). On Objects365 images both
heads stay at $0.27$ ($0.40$--$0.41$ for referents of a COCO-mappable
category, $0.21$ for the rest, bucketed as in \cref{tab:app-refl4source})
against $0.774$ for decoding, and the stress set does not move ($+0.009$ and
$-0.007$). More RefCOCO-family expressions thus lift the RefCOCO-family splits
whose expression style they add, but neither RefCOCOg nor the category
boundary of the main paper.

\paragraph{Position relative to VGent.}
VGent \cite{vgent} is the published design closest to the combined selector:
boxes from three proposers (UPN, SAM and GLEE) cross-attend to the hidden
states of every layer of a frozen, reinforcement-learning-tuned
Qwen2.5-VL-7B, and the head is trained on RefCOCO, RefCOCO+ and RefCOCOg
together with Objects365 and MaskGroups-2M. It reports $90.4$ and $90.1$ on
RefCOCOg val and test (\cref{tab:app-published}; not a matched-perception
result). Our combined selector is $2.8$ and $1.6$ points below when trained on
RefCOCOg alone and $2.7$ and $1.2$ points below when trained on the union; on
RefCOCO and RefCOCO+ val the union narrows the gap from $2.9$ to $1.2$ and from
$5.7$ to $3.5$ points. The RefCOCO-family part of VGent's training data thus
closes $1.7$ of $2.9$ and $2.2$ of $5.7$ points on RefCOCO and RefCOCO+, but
moves RefCOCOg little ($+0.005$ on test, $+0.008$ against the RefCOCOg part
alone) and Ref-L4 not at all ($-0.001$), and at the $24$k budget replacing the
category-prompted detector by a class-agnostic proposer (UPN \cite{chatrex};
\cref{sec:app-upn}) does not lift selection either, also not for the
combined selector ($-0.016$ on RefCOCOg test, level on Ref-L4).
What remains on RefCOCOg therefore lies in the components we held fixed:
VGent's tuned encoder, its readout of every layer rather than three depths,
its three proposers used together (we tested one class-agnostic proposer
alone) and its corpora beyond the RefCOCO family. Our experiments do not
apportion the remaining $1.2$--$2.7$ points among these factors. The
index-selecting VLM over the expression-prompted detector's top five comes
within $1.2$ points on RefCOCOg val ($0.892$, one seed;
\cref{sec:app-lora-matched}), but it adapts the VLM itself.
\paragraph{A third VLM.}
\label{sec:app-internvl3}
Both VLMs above are Qwen models. To test whether two findings depend on the Qwen models, we repeat the hook and coordinate decoding with InternVL3-8B
\cite{internvl3}, whose vision encoder (InternViT) and native multimodal
pre-training differ from Qwen's; its language model is initialized from
Qwen2.5 and shares its tokenizer, so the two are not fully independent. Proposals, region features, the $24$k training rows, the recipe
and the three seeds are those of the main build's hook; only the query
changes. It is read as for the Qwen VLMs (the final-layer state, after the
final normalization, at the position that would begin the answer), once with
the image and once from a pass over the request without it. Images are tiled as
in the model's own inference, with at most four $448$-pixel tiles and a
thumbnail ($\le1{,}280$ visual tokens, the cap of the Qwen runs). Decoding uses
the model's documented grounding prompt (``Please provide the bounding box
coordinates of the region this sentence describes'', with the expression
between reference tags), greedy, and rescales its relative $0$--$1000$
coordinates; an unparsable answer counts as wrong ($0.0\%$ of RefCOCOg val and
$0.2\%$ of Ref-L4 val queries). The reading was fixed before the runs: the
query finding transfers if the image-conditioned query leads the text-only one
by at least $0.10$ on RefCOCOg val with a paired interval excluding zero, and
the category finding transfers if decoding's lead over the hook is at least
$0.10$ larger on referents outside the COCO categories than inside them.

\Cref{tab:app-ivl3} gives the results. With InternVL3 the image-conditioned
query leads by $0.134$ [$0.122$, $0.146$] on RefCOCOg val (paired over queries,
on the three-seed mean) and by $0.042$ [$0.035$, $0.049$] on Ref-L4 val,
against $0.208$ and $0.092$ for Qwen2.5-VL at the same budget: the query
finding transfers, with a smaller effect. The hook again collapses outside the
COCO categories ($0.581$ inside against $0.213$ outside; $0.525$ against $0.196$
with the text-only query), but so does InternVL3's own decoding ($0.917$
against $0.561$, where Qwen2.5-VL keeps $0.900$ and $0.774$). Decoding's lead
over the hook is therefore $0.348$ outside and $0.336$ inside the COCO
categories, a difference of $0.012$ against $0.223$ for Qwen2.5-VL, and the
category finding does not transfer as registered: the trained head's boundary
appears with InternVL3 as with Qwen2.5-VL, but whether the VLM's own decoding crosses it
depends on the VLM. InternVL3's decoding still leads its hook on every column
($0.894$ against $0.735$ on RefCOCOg val).

\begin{table}[t]
\centering
\footnotesize
\setlength{\tabcolsep}{2pt}
\caption{A third VLM, InternVL3-8B (main build, $24$k training rows):
accuracy at $\mathrm{IoU}\ge0.5$ on present queries, no abstention; trained
rows are means$\pm$sample std over three seeds. Ref-L4 val: all queries and
referents inside / outside the $80$ COCO categories ($8{,}483$ / $4{,}937$
queries, proposal recall $0.999$ / $0.886$). Qwen2.5-VL rows from
\cref{tab:app-controls,tab:app-refl4buckets} and Table 1 of the main paper;
image / text-only: the query read with / without the image; Qwen2.5-VL's text-only row is the learned bridge, the hook trained without the verifier; decoding: the VLM's own box.
--: not evaluated.}
\label{tab:app-ivl3}
\begin{tabular}{@{}lcccc@{}}
\toprule
 & & \multicolumn{3}{c}{Ref-L4 val}\\
\cmidrule(l){3-5}
Selector & RefCOCOg val & all & COCO cat. & other\\
\midrule
\multicolumn{5}{@{}l}{\emph{InternVL3-8B}}\\
Hook, image & 0.735\sd{.005} & 0.446\sd{.007} & 0.581\sd{.007} & 0.213\sd{.007}\\
Hook, text-only & 0.601\sd{.007} & 0.404\sd{.008} & 0.525\sd{.011} & 0.196\sd{.008}\\
Decoding & 0.894 & 0.786 & 0.917 & 0.561\\
\midrule
\multicolumn{5}{@{}l}{\emph{Qwen2.5-VL-7B}}\\
Hook, image & 0.787\sd{.004} & 0.464\sd{.010} & 0.592 & 0.243\\
Bridge, text-only & 0.579\sd{.007} & 0.372\sd{.018} & -- & --\\
Decoding & 0.865 & 0.853 & 0.900 & 0.774\\
\bottomrule
\end{tabular}
\end{table}

\paragraph{A proposal source without RefCOCO data.}
\label{sec:app-second-detector}
The detector of every other table, Grounding DINO Swin-B, was trained with
RefCOCO-family data, so its region features may already encode referent
salience. \Cref{tab:app-second-detector} replaces it with MM-Grounding-DINO
Swin-B trained on Objects365 v1, GoldG and V3Det, whose documented training data
contain no RefCOCO/+/g annotations (its images are not documented as disjoint
from COCO). Everything else is held fixed: the COCO-80 prompt, the $300$
highest-scoring proposals with $256$-dimensional fused features, the full
RefCOCOg training rows and the Qwen2.5-VL queries, the hook (gates not
re-tuned) and the relation-aware head, three seeds; rows are paired with the
full-budget run on the original detector. Proposal recall is unchanged on
RefCOCO and RefCOCO+ val, within $0.004$ on RefCOCOg and $0.006$ on Ref-L4, and
falls from $1.000$ to $0.983$ on the stress set. The relation-aware head loses
at most $0.010$ on any standard split, the hook $0.026$--$0.034$ (and $0.037$
on Ref-L4), and the Ref-L4
collapse outside the COCO categories is the same with either detector
($0.25$--$0.26$), so neither the ranking finding nor the vocabulary boundary
depends on RefCOCO data in the detector. The detector itself, prompted with the
expression, does depend on it: without RefCOCO data its detector-native
accuracy falls by $0.15$--$0.32$ on the standard splits ($0.632$ against
$0.783$ on RefCOCOg val), $0.24$ below the relation-aware head over the same
detector's proposals.

\begin{table*}[t]
\centering
\footnotesize
\setlength{\tabcolsep}{4pt}
\caption{Second proposal source (Qwen2.5-VL, full training budget, three seeds;
accuracy at $\mathrm{IoU}\ge0.5$; detector-native rows are deterministic, one run).
``No RefC'': MM-Grounding-DINO Swin-B
without RefCOCO data; ``orig.'': Grounding DINO Swin-B. $\Delta$: paired
difference (no RefC minus orig.); ${}^\dagger$its $95\%$ bootstrap interval
excludes zero. Ref-L4 COCO cat.\ / other: referents inside / outside the $80$
COCO categories.}
\label{tab:app-second-detector}
\begin{tabular}{@{}lcccccccc@{}}
\toprule
& RefCOCO val & RefCOCO+ val & RefCOCOg val & RefCOCOg test & Ref-L4 val & \quad COCO cat. & \quad other & Stress\\
\midrule
Proposal recall, no RefC & 0.999 & 0.999 & 0.996 & 0.997 & 0.952 & 0.998 & 0.873 & 0.983\\
Proposal recall, orig. & 0.999 & 0.999 & 1.000 & 0.998 & 0.958 & 0.999 & 0.886 & 1.000\\
\midrule
Hook, no RefC & 0.806 & 0.733 & 0.809 & 0.811 & 0.512 & 0.657 & 0.264 & 0.398\\
Hook, orig. & 0.834 & 0.760 & 0.843 & 0.837 & 0.549 & 0.718 & 0.259 & 0.413\\
\quad $\Delta$ & $-0.028^\dagger$ & $-0.027^\dagger$ & $-0.034^\dagger$ & $-0.026^\dagger$ & $-0.037^\dagger$ & & & $-0.014$\\
Relation-aware head, no RefC & 0.870 & 0.792 & 0.870 & 0.873 & 0.590 & 0.790 & 0.247 & 0.397\\
Relation-aware head, orig. & 0.872 & 0.802 & 0.877 & 0.878 & 0.595 & 0.798 & 0.247 & 0.358\\
\quad $\Delta$ & $-0.002$ & $-0.010^\dagger$ & $-0.007^\dagger$ & $-0.006^\dagger$ & $-0.005^\dagger$ & & & $+0.039^\dagger$\\
\midrule
Detector-native, no RefC & 0.529 & 0.535 & 0.632 & 0.629 & 0.438 & & & 0.320\\
Detector-native, orig. & 0.847 & 0.737 & 0.783 & 0.781 & 0.634 & & & 0.255\\
\quad $\Delta$ & $-0.318^\dagger$ & $-0.202^\dagger$ & $-0.151^\dagger$ & $-0.151^\dagger$ & $-0.196^\dagger$ & & & $+0.065^\dagger$\\
\bottomrule
\end{tabular}
\end{table*}

\section{Image Overlap between Training and Evaluation Splits}
\label{sec:app-leakage}

The hook is trained on the first $24$k expressions of RefCOCOg train
($9{,}948$ images). RefCOCOg val and test share no image with RefCOCOg train,
but RefCOCO and RefCOCO+ val do: $481$ of their $1{,}500$ images are among the
hook's training images and $918$ are in RefCOCOg train. In turn, $0.488$ of the
RefCOCOg val images and $0.483$ of the test images appear in RefCOCO/RefCOCO+
train, the data family the detector checkpoint was trained with.

\cref{tab:app-leakage} splits RefCOCO val into images the hook was trained on
(A), RefCOCOg-train images it never saw (B) and images outside RefCOCOg (C).
The hook scores the same on A and B ($0.839$ against $0.834$; difference in
differences against coordinate decoding $+0.003$, $95\%$ interval
$[-0.013, +0.020]$), so there is no memorization effect. It scores $0.108$ lower
on C than on B, whereas coordinate decoding loses $0.030$ (difference in
differences $+0.078$, $[+0.062, +0.095]$): the hook's transfer accuracy depends
on how the images were selected, a domain effect of training on RefCOCOg's
image selection rather than memorization of seen images. The hook shows the
same pattern on RefCOCO+ val. On RefCOCOg val and test, rows whose images
appear in RefCOCO/RefCOCO+ train are not easier for the hook ($0.785$ against
$0.788$ on val, $0.780$ against $0.811$ on test) or for the detector's text
interface ($0.772$ against $0.800$, $0.759$ against $0.811$); no unexposed
detector or VLM is available, so these frozen-component comparisons are
descriptive.

\begin{table*}[t]
\centering
\footnotesize
\setlength{\tabcolsep}{4pt}
\caption{Accuracy by image subset on RefCOCO val (last column: the hook on
RefCOCO+ val, same images) with $95\%$ query-level bootstrap intervals. A:
image among the hook's training images; B: image in RefCOCOg train but never
seen by the hook; C: image not in RefCOCOg. The RefCOCO+ subsets hold
$3{,}344$, $2{,}871$ and $4{,}543$ rows. Hook: per-query correctness averaged
over three seeds; coordinate decoding and detector-native: raw boxes.
Differences are computed before rounding.}
\label{tab:app-leakage}
\begin{tabular}{@{}lccccc@{}}
\toprule
& & \multicolumn{3}{c}{RefCOCO val} & RefCOCO+ val\\
\cmidrule(lr){3-5}\cmidrule(l){6-6}
Image subset & Rows (images) & Hook & Coord.\ decoding & Detector-native & Hook\\
\midrule
A: among the hook's training images & $3{,}373$ ($481$) & 0.839\ci{.829}{.850} & 0.908\ci{.898}{.918} & 0.869\ci{.858}{.881} & 0.750\ci{.738}{.762}\\
B: in RefCOCOg train, never seen & $2{,}846$ ($437$) & 0.834\ci{.822}{.845} & 0.905\ci{.895}{.916} & 0.872\ci{.860}{.884} & 0.757\ci{.744}{.770}\\
C: not in RefCOCOg & $4{,}615$ ($582$) & 0.726\ci{.714}{.736} & 0.875\ci{.866}{.885} & 0.814\ci{.803}{.825} & 0.645\ci{.633}{.656}\\
\midrule
A $-$ B (memorization) & & +0.006 & +0.003 & $-$0.003 & $-$0.006\\
B $-$ C (image selection) & & +0.108 & +0.030 & +0.058 & +0.112\\
\bottomrule
\end{tabular}
\end{table*}

\section{Statistical Protocol}
\label{sec:app-stats}

\paragraph{Seed variation.}
Every trained result is the mean over three training seeds (42, 123, 456) with
the sample standard deviation ($n{-}1$ in the denominator), which for three
seeds is $\sqrt{3/2}\approx1.22$ times the population value. A $95\%$
$t$-interval over three seeds has a half-width of $2.48$ sample standard
deviations (\cref{tab:app-intervals}, upper block).

\paragraph{Query sampling.}
Detector-native selection, coordinate decoding and coordinate-snap receive no
task-specific training and are deterministic, so their uncertainty is query
sampling; \cref{tab:app-intervals} (middle block) gives percentile bootstrap
intervals over queries ($10{,}000$ resamples) for proposal recall, for every
cell of these three interfaces on the splits of \cref{tab:app-grounding24k} and for the hook's
three-seed mean. Half-widths are
$0.006$--$0.012$ on the RefCOCO-family splits and Ref-L4, and about $0.06$ on
the $231$-query stress set, where template-level intervals are wider still
(\cref{tab:app-stress}).

\paragraph{Seeds and queries together.}
Neither interval above carries both sources of variation. The two-level
bootstrap draws the three seeds with replacement and the queries with
replacement (one query draw shared by the drawn seeds, since the seeds share
the rows), $10{,}000$ times, and takes the mean accuracy over the drawn seeds
on the drawn queries (\cref{tab:app-intervals}, bottom block). We computed it
for $118$ cells: the hook and the hook without verifier at both budgets
(full budget with both VLMs), the relation-aware head with IoU-rank loss at
both budgets (full budget with both VLMs), the Qwen2.5-VL capacity-matched hook
at both budgets, pooled and token cross-attention at both budgets (token
cross-attention also with Qwen3-VL at the full budget), the combined selector with both VLMs, and
the image-conditioned and text-only learned bridges with both VLMs at the full
budget. In these cells, on the RefCOCO-family splits and Ref-L4, its half-width
is $0.005$--$0.016$, a median $1.2$ times the query-only half-width (at most
$2.2$ times, for the hooks and bridges whose seeds disagree most); on the
stress set it is $0.053$--$0.074$. The Qwen3-VL capacity-matched hook, whose
Ref-L4 seeds disagree more (sample std $0.034$, \cref{tab:app-full-acc}), is
not among them.

\paragraph{Paired differences with seeds and queries.}
\Cref{tab:app-twolevel-diffs} gives two-level intervals for the paired
differences the main paper's claims rest on: rows are drawn once per resample
and shared by both systems, the seeds of each system are drawn independently;
``worst pair'' is the seed pair closest to zero in the direction of the gap.
Set-level scoring (the relation-aware head with the IoU-rank loss) stays ahead
of a hook of the same size on every standard split
at both budgets and for both VLMs, and at the full budget falls behind it on the stress set (significantly only with Qwen2.5-VL) and outside the COCO categories of Ref-L4 (\cref{tab:app-matched-hook-q3}). The
combined selector is not distinguishable from the relation-aware head on
RefCOCOg val and leads it by $0.006$ on test with Qwen2.5-VL (not with
Qwen3-VL); both exceed the raw box on RefCOCOg test.

\begin{table*}[t]
\centering
\footnotesize
\setlength{\tabcolsep}{4pt}
\caption{Paired differences with $95\%$ two-level (seed $\times$ query)
bootstrap intervals ($10{,}000$ resamples; stress set additionally with
templates as the unit). Matched hook: the hook at the relation-aware head's
parameter count with the same all-positive supervision. Relation-aware head:
the variant with the IoU-rank loss, as in \cref{tab:app-grounding24k,tab:app-full-acc} (the variant without
it is compared in \cref{tab:app-tests}). Worst pair: the per-seed pair
closest to zero, with its query-level interval. Combined-selector rows on RefCOCO and RefCOCO+ val use the retraining of \cref{sec:app-allsplits}.}
\label{tab:app-twolevel-diffs}
\begin{tabular}{@{}llccc@{}}
\toprule
Comparison & Split & Difference & Two-level $95\%$ interval & Worst seed pair\\
\midrule
Relation-aware head (IoU-rank loss) $-$ matched hook, $24$k & RefCOCO val & $+0.038$ & [$+0.025$, $+0.049$] & $+0.021$ [$+0.015$, $+0.028$]\\
 & RefCOCO+ val & $+0.038$ & [$+0.023$, $+0.051$] & $+0.019$ [$+0.012$, $+0.027$]\\
 & RefCOCOg test & $+0.033$ & [$+0.021$, $+0.045$] & $+0.020$ [$+0.013$, $+0.027$]\\
 & Ref-L4 val & $+0.025$ & [$+0.012$, $+0.037$] & $+0.006$ [$-0.002$, $+0.014$]\\
 & Stress (templates) & $-0.004$ & [$-0.051$, $+0.038$] & --\\
Relation-aware head (IoU-rank loss) $-$ matched hook, full & RefCOCO val & $+0.022$ & [$+0.014$, $+0.030$] & $+0.014$ [$+0.008$, $+0.020$]\\
 & RefCOCO+ val & $+0.028$ & [$+0.018$, $+0.039$] & $+0.018$ [$+0.011$, $+0.026$]\\
 & RefCOCOg val & $+0.026$ & [$+0.018$, $+0.034$] & $+0.022$ [$+0.013$, $+0.031$]\\
 & RefCOCOg test & $+0.031$ & [$+0.025$, $+0.037$] & $+0.026$ [$+0.020$, $+0.033$]\\
 & Ref-L4 val & $+0.018$ & [$+0.008$, $+0.028$] & $+0.006$ [$-0.001$, $+0.013$]\\
 & Stress (templates) & $-0.051$ & [$-0.098$, $-0.007$] & --\\
\quad with Qwen3-VL & RefCOCO val & $+0.033$ & [$+0.027$, $+0.039$] & $+0.029$ [$+0.023$, $+0.034$]\\
 & RefCOCO+ val & $+0.035$ & [$+0.030$, $+0.041$] & $+0.032$ [$+0.026$, $+0.038$]\\
 & RefCOCOg val & $+0.027$ & [$+0.019$, $+0.035$] & $+0.021$ [$+0.013$, $+0.029$]\\
 & RefCOCOg test & $+0.023$ & [$+0.017$, $+0.028$] & $+0.019$ [$+0.013$, $+0.025$]\\
 & Ref-L4 val & $+0.033$ & [$+0.008$, $+0.070$] & $+0.006$ [$-0.000$, $+0.013$]\\
 & Stress (templates) & $-0.036$ & [$-0.093$, $+0.021$] & --\\
Combined $-$ relation-aware head, full & RefCOCOg val & $-0.001$ & [$-0.009$, $+0.007$] & $+0.002$ [$-0.007$, $+0.010$]\\
 & RefCOCOg test & $+0.006$ & [$+0.001$, $+0.012$] & $+0.004$ [$-0.001$, $+0.010$]\\
 & RefCOCO val & $+0.021$ & [$+0.015$, $+0.026$] & $+0.017$ [$+0.011$, $+0.023$]\\
 & RefCOCO+ val & $+0.022$ & [$+0.016$, $+0.029$] & $+0.019$ [$+0.012$, $+0.026$]\\
\quad with Qwen3-VL & RefCOCOg test & $-0.004$ & [$-0.008$, $+0.000$] & $-0.002$ [$-0.007$, $+0.003$]\\
Combined $-$ token-state probe, full & RefCOCOg test & $+0.007$ & [$+0.002$, $+0.012$] & --\\
Combined $-$ coordinate decoding (raw) & RefCOCOg test & $+0.018$ & [$+0.012$, $+0.025$] & $+0.018$ [$+0.011$, $+0.025$]\\
 & RefCOCO val & $+0.001$ & [$-0.005$, $+0.007$] & $-0.002$ [$-0.008$, $+0.004$]\\
 & RefCOCO+ val & $-0.014$ & [$-0.021$, $-0.007$] & $-0.012$ [$-0.019$, $-0.005$]\\
\quad with Qwen3-VL & RefCOCOg test & $+0.006$ & [$+0.000$, $+0.011$] & $+0.005$ [$-0.000$, $+0.011$]\\
Relation-aware head (full) $-$ coordinate decoding & RefCOCOg test & $+0.012$ & [$+0.005$, $+0.020$] & $+0.009$ [$+0.002$, $+0.016$]\\
 & RefCOCO val & $-0.021$ & [$-0.028$, $-0.015$] & $-0.019$ [$-0.025$, $-0.012$]\\
 & RefCOCO+ val & $-0.036$ & [$-0.045$, $-0.027$] & $-0.029$ [$-0.036$, $-0.021$]\\
 & Ref-L4 val & $-0.258$ & [$-0.267$, $-0.249$] & --\\
Token-state probe $-$ relation-aware head (IoU-rank loss) & Ref-L4 val ($24$k) & $+0.071$ & [$+0.062$, $+0.082$] & $+0.062$ [$+0.053$, $+0.070$]\\
 & Ref-L4 val (full) & $+0.032$ & [$+0.024$, $+0.041$] & $+0.024$ [$+0.017$, $+0.031$]\\
Verifier $-$ tuned yes-probability (joint) & RefCOCOg test & $+0.011$ & [$+0.008$, $+0.014$] & $+0.009$ [$+0.006$, $+0.012$]\\
\bottomrule
\end{tabular}
\end{table*}

\paragraph{Token states versus the pooled query.}
Fourteen further paired tests compare the token-state probe with pooled-query
interfaces (sign-flip permutation on seed-averaged per-query correctness,
$10{,}000$ permutations; on the stress set the resampling unit is the query
template). They are Holm-corrected jointly with the $44$ tests of
\cref{tab:app-tests}, a family of $58$; adding them changes the significance of
none of the $44$. On Ref-L4 the token probe exceeds the relation-aware head
trained in the same run ($0.517$) by $0.076$ ($0.108$ on COCO-category
referents, $0.021$ on the rest; $0.071$ over the variant with the IoU-rank
loss, $0.522$, in \cref{tab:app-twolevel-diffs}) and the hook by $0.129$, all with Holm-adjusted $p\le0.006$. On the
stress set it is level with the bridge on the image-conditioned query
($+0.003$ [$-0.044$, $+0.043$], $p=0.95$), ahead of the hook by $0.042$
($p=0.13$), and ahead of the pooled cross-attention control by $0.069$
[$0.025$, $0.109$] ($p=0.010$, Holm-adjusted $0.24$); the audited index
scorer ($0.434$) stored aggregates only and admits no paired test. With
templates as the resampling unit, the verifier's cost to the hook on the stress
set ($-0.039$, $p=0.039$) is not significant after correction (Holm $0.95$ in the
family below).

\paragraph{Remaining comparisons.}
Ten further tests complete the family, now $68$ tests; adding them changes the
significance of none of the $58$. Where one side stored only per-seed
aggregates, the test is the worst-case exact McNemar bound described below.
Multi-layer readout of the hook adds $+0.023$ on RefCOCOg val and $+0.024$ on
test (bound $p=0.047$ and $0.009$, Holm $1.0$ and $0.26$; seed $t$-intervals
$[0.011, 0.035]$ and $[0.008, 0.041]$); multi-positive supervision of the hook adds
$+0.022$ and $+0.012$ (bound $p=0.49$ and $0.06$, Holm $1.0$), and at matched
capacity, where both sides have per-query records, $+0.035$ and $+0.025$
(Holm $0.006$); averaging over expression tokens loses $0.077$ and $0.072$
(Holm $\le5{\times}10^{-10}$). On the untouched RefCOCOg test with swapped absent
rows the verifier at the coordinate-snap proposal leads the tuned
yes-probability by $0.011$ in joint accuracy ($0.888$ against $0.877$; paired
query-bootstrap interval $[0.009, 0.014]$ on the seed-averaged verifier, two-level
interval $[0.008, 0.014]$ in \cref{tab:app-twolevel-diffs}); with per-row verifier scores of all three seeds the
paired sign-flip test gives $p=10^{-4}$, the smallest attainable, and a
Holm-adjusted $p\le0.007$. On the stress set, with templates as the resampling unit, the
relation-aware head is level with the hook ($+0.004$ [$-0.036$, $+0.046$],
$p=0.90$), the continuous query is $0.061$ behind the discrete phrase
($p=0.13$), and the verifier at the hook's proposal is level with the tuned
yes-probability ($+0.016$ [$-0.031$, $+0.061$], $p=0.56$); none is significant
after correction.

\begin{table*}[t]
\centering
\footnotesize
\setlength{\tabcolsep}{3pt}
\caption{$95\%$ intervals for \cref{tab:app-grounding24k}, coordinate decoding and the untrained hook
(accuracy at $\mathrm{IoU}\ge0.5$, no abstention). Upper block: three-seed mean
with a $t$-interval over seeds (sample std in \cref{tab:app-grounding24k}). Middle block:
percentile bootstrap over queries ($10{,}000$ resamples); for the hook it
resamples queries with per-query correctness averaged over seeds. Bottom
block: two-level bootstrap over seeds and queries. Rows without task-specific training are computed on the same rows and
proposals as the trained rows; raw and snapped boxes differ in accuracy by at most $0.02$. Proposal recall, the first row of \cref{tab:app-grounding24k}, is the
ceiling for every snapped method.}
\label{tab:app-intervals}
\begin{tabular}{@{}lcccccc@{}}
\toprule
Method & RefCOCO val & RefCOCO+ val & RefCOCOg val & RefCOCOg test & Ref-L4 val & Stress\\
\midrule
\multicolumn{7}{@{}l}{\emph{Seed variation: $95\%$ $t$-interval over three training seeds}}\\
Naive cosine (untrained hook) & 0.078\ci{$-$.074}{.229} & 0.076\ci{$-$.074}{.227} & 0.082\ci{$-$.079}{.242} & 0.083\ci{$-$.081}{.248} & 0.065\ci{$-$.072}{.201} & 0.048\ci{$-$.072}{.167}\\
Learned bridge (CLIP query) & 0.511\ci{.468}{.553} & 0.429\ci{.409}{.449} & 0.542\ci{.527}{.558} & 0.552\ci{.531}{.574} & 0.357\ci{.341}{.373} & 0.254\ci{.200}{.308}\\
Index scorer & 0.759\ci{.754}{.765} & 0.683\ci{.672}{.694} & 0.770\ci{.766}{.775} & 0.770\ci{.753}{.787} & 0.444\ci{.420}{.468} & 0.384\ci{.303}{.465}\\
Index scorer, audited & 0.801\ci{.792}{.810} & 0.722\ci{.701}{.744} & 0.811\ci{.804}{.818} & 0.815\ci{.809}{.821} & 0.451\ci{.384}{.518} & 0.434\ci{.391}{.478}\\
Alignment hook & 0.789\ci{.772}{.807} & 0.707\ci{.682}{.733} & 0.787\ci{.777}{.797} & 0.793\ci{.776}{.810} & 0.464\ci{.439}{.489} & 0.384\ci{.297}{.471}\\
Relation-aware head (probe) & 0.849\ci{.842}{.856} & 0.772\ci{.763}{.780} & 0.846\ci{.841}{.850} & 0.849\ci{.846}{.852} & 0.522\ci{.505}{.538} & 0.388\ci{.354}{.423}\\
\midrule
\multicolumn{7}{@{}l}{\emph{Query sampling: $95\%$ bootstrap over queries}}\\
Proposal recall (ceiling) & 0.999\ci{.998}{.999} & 0.999\ci{.998}{1.000} & 1.000\ci{1.000}{1.000} & 0.998\ci{.997}{.999} & 0.958\ci{.954}{.961} & 1.000\ci{1.000}{1.000}\\
Alignment hook, three-seed mean & 0.789\ci{.783}{.796} & 0.707\ci{.700}{.715} & 0.787\ci{.777}{.796} & 0.793\ci{.786}{.800} & 0.464\ci{.457}{.471} & 0.384\ci{.330}{.439}\\
Detector-native & 0.847\ci{.840}{.853} & 0.737\ci{.729}{.745} & 0.783\ci{.772}{.795} & 0.781\ci{.772}{.789} & 0.634\ci{.626}{.642} & 0.255\ci{.199}{.312}\\
Coordinate-snap & 0.895\ci{.889}{.901} & 0.840\ci{.833}{.846} & 0.867\ci{.858}{.877} & 0.866\ci{.859}{.873} & 0.833\ci{.827}{.840} & 0.610\ci{.545}{.671}\\
Coordinate decoding & 0.893\ci{.888}{.899} & 0.838\ci{.831}{.845} & 0.865\ci{.856}{.875} & 0.866\ci{.860}{.873} & 0.853\ci{.847}{.859} & 0.623\ci{.563}{.684}\\
\midrule
\multicolumn{7}{@{}l}{\emph{Seeds and queries: $95\%$ two-level bootstrap (seeds and queries resampled together)}}\\
Alignment hook & 0.789\ci{.780}{.799} & 0.707\ci{.696}{.720} & 0.787\ci{.776}{.798} & 0.793\ci{.784}{.803} & 0.464\ci{.451}{.475} & 0.384\ci{.318}{.450}\\
Relation-aware head (probe) & 0.849\ci{.842}{.856} & 0.772\ci{.764}{.780} & 0.846\ci{.836}{.855} & 0.849\ci{.842}{.856} & 0.522\ci{.512}{.532} & 0.388\ci{.328}{.450}\\
\bottomrule
\end{tabular}
\end{table*}

\paragraph{Paired tests.}
\cref{tab:app-tests} lists the paired comparisons behind the main paper's
claims, with Holm correction over all $44$ tests. Whenever per-query outputs of
both systems exist, the test is a paired sign-flip permutation test on
per-query correctness averaged over seeds, whose statistic is the difference of
the reported three-seed means ($10{,}000$ permutations, so the smallest
attainable $p$ is $10^{-4}$ and the smallest Holm $p$ is $0.004$), or an exact
McNemar test when both sides are single runs. Only two comparisons lack
per-query outputs on one side and are reported as worst-case bounds, the
largest exact McNemar $p$-value compatible with the two systems' per-seed
accuracies: the audited index scorer, and the verifier at the
coordinate-snap proposal on POPE. For the audited index scorer the per-seed bounds
are $0.023$, $0.022$ and $0.0005$ on RefCOCOg val and $0.0001$, $0.019$ and
$0.00002$ on test; the table reports the largest. With Holm correction over
all $128$ per-seed values, the audited index scorer is ahead in two of three seeds
on RefCOCOg test and in one on RefCOCOg val, and in no seed on the other
splits, where the hook is ahead in one seed on Ref-L4. A bound is conservative, so a non-significant bound does not show that
a difference is absent.

After correction the hook is below coordinate decoding on every split tested,
below detector-native on RefCOCO val and above it on the stress set, and not
distinguishable from it on RefCOCOg val or test (Holm $p=1.000$ and $0.115$).
Set-level scoring (the relation-aware head with multi-positive supervision and
without the IoU-rank loss, against a hook with the same parameters and
supervision) is ahead on all four standard splits and on Ref-L4 (Holm
$p=0.004$ each) but not on the stress set. The verifier's joint-accuracy lead
over the VLM's yes/no answer on RefCOCOg val holds (Holm $p=0.004$), as does
the yes/no answer's lead on POPE adversarial. Its lead over the tuned
yes-probability on RefCOCOg val ($0.888$ against $0.873$) is also significant,
but that threshold was chosen on the same split; on the stress set the tuned
yes-probability is ahead ($0.564$ against $0.528$) without significance. The
stress-set cost of verifier training ($0.384$ against $0.423$) does not survive
correction (Holm $p=0.366$).

\begin{table*}[t]
\centering
\footnotesize
\setlength{\tabcolsep}{3pt}
\caption{Paired tests with Holm correction over all $44$ tests (selection
accuracy on present queries; ``Joint'': joint accuracy with absent referents,
existence accuracy on POPE, for the coordinate-snap box gated by the verifier
at its own or at the hook's proposal, against the same box gated by the VLM's
yes/no answer or its tuned yes-probability). $a$, $b$: accuracies of the first
and second system, three-seed means where trained. perm.: paired sign-flip
permutation test on per-query correctness averaged over seeds ($10{,}000$
permutations; smallest attainable $p=10^{-4}$); McNemar: exact test between
single runs; bound: largest exact McNemar $p$-value compatible with the
per-seed accuracies (per-seed values in the text). Relation head, no IoU-rank
loss: the relation-aware head with multi-positive supervision but without the
IoU-rank loss, against the capacity-matched hook with the same supervision (the
variant with the IoU-rank loss: \cref{tab:app-twolevel-diffs}).
${}^\dagger$Yes-probability
threshold chosen on this split.}
\label{tab:app-tests}
\begin{tabular}[t]{@{}lccccc@{}}
\toprule
Split & $a$ & $b$ & $a{-}b$ & $p$ & Holm $p$\\
\midrule
\multicolumn{6}{@{}l}{\emph{Hook vs index scorer} (perm.)}\\
\quad RefCOCO val & 0.789 & 0.759 & +0.030 & $1.0{\times}10^{-4}$ & 0.004\\
\quad RefCOCO+ val & 0.707 & 0.683 & +0.025 & $1.0{\times}10^{-4}$ & 0.004\\
\quad RefCOCOg val & 0.787 & 0.770 & +0.016 & $2.0{\times}10^{-4}$ & 0.005\\
\quad RefCOCOg test & 0.793 & 0.770 & +0.023 & $1.0{\times}10^{-4}$ & 0.004\\
\quad Ref-L4 val & 0.464 & 0.444 & +0.020 & $1.0{\times}10^{-4}$ & 0.004\\
\quad Stress & 0.384 & 0.384 & 0.000 & 1.000 & 1.000\\
\multicolumn{6}{@{}l}{\emph{Hook vs index scorer, audited} (bound)}\\
\quad RefCOCO val & 0.789 & 0.801 & $-$0.012 & 0.395 & 1.000\\
\quad RefCOCO+ val & 0.707 & 0.722 & $-$0.015 & 0.161 & 1.000\\
\quad RefCOCOg val & 0.787 & 0.811 & $-$0.024 & 0.023 & 0.394\\
\quad RefCOCOg test & 0.793 & 0.815 & $-$0.022 & 0.019 & 0.366\\
\quad Ref-L4 val & 0.464 & 0.451 & +0.013 & 0.168 & 1.000\\
\quad Stress & 0.384 & 0.434 & $-$0.051 & 0.714 & 1.000\\
\multicolumn{6}{@{}l}{\emph{Relation head, no IoU-rank loss, vs matched hook} (perm.)}\\
\quad RefCOCO val & 0.837 & 0.811 & +0.025 & $1.0{\times}10^{-4}$ & 0.004\\
\quad RefCOCO+ val & 0.759 & 0.734 & +0.025 & $1.0{\times}10^{-4}$ & 0.004\\
\quad RefCOCOg val & 0.843 & 0.820 & +0.023 & $1.0{\times}10^{-4}$ & 0.004\\
\quad RefCOCOg test & 0.842 & 0.816 & +0.026 & $1.0{\times}10^{-4}$ & 0.004\\
\quad Ref-L4 val & 0.518 & 0.497 & +0.021 & $1.0{\times}10^{-4}$ & 0.004\\
\quad Stress & 0.362 & 0.392 & $-$0.030 & 0.076 & 1.000\\
\multicolumn{6}{@{}l}{\emph{Relation-aware head (probe) vs hook} (perm.)}\\
\quad Ref-L4 val & 0.522 & 0.464 & +0.058 & $1.0{\times}10^{-4}$ & 0.004\\
\quad Stress & 0.388 & 0.384 & +0.004 & 0.894 & 1.000\\
\multicolumn{6}{@{}l}{\emph{Hook vs detector-native} (perm.)}\\
\quad RefCOCOg val & 0.787 & 0.783 & +0.003 & 0.617 & 1.000\\
\quad RefCOCOg test & 0.793 & 0.781 & +0.012 & 0.005 & 0.115\\
\quad RefCOCO val & 0.789 & 0.847 & $-$0.057 & $1.0{\times}10^{-4}$ & 0.004\\
\quad Stress & 0.384 & 0.255 & +0.128 & $2.0{\times}10^{-4}$ & 0.005\\
\bottomrule
\end{tabular}\hfill
\begin{tabular}[t]{@{}lccccc@{}}
\toprule
Split & $a$ & $b$ & $a{-}b$ & $p$ & Holm $p$\\
\midrule
\multicolumn{6}{@{}l}{\emph{Hook vs coordinate decoding} (perm.)}\\
\quad RefCOCOg val & 0.787 & 0.865 & $-$0.079 & $1.0{\times}10^{-4}$ & 0.004\\
\quad RefCOCOg test & 0.793 & 0.866 & $-$0.073 & $1.0{\times}10^{-4}$ & 0.004\\
\quad RefCOCO val & 0.789 & 0.893 & $-$0.104 & $1.0{\times}10^{-4}$ & 0.004\\
\quad Stress & 0.384 & 0.623 & $-$0.240 & $1.0{\times}10^{-4}$ & 0.004\\
\multicolumn{6}{@{}l}{\emph{Joint: coord.-snap + verifier vs + yes/no}}\\
\quad RefCOCOg val (perm.) & 0.888 & 0.840 & +0.048 & $1.0{\times}10^{-4}$ & 0.004\\
\quad Stress (perm.) & 0.528 & 0.545 & $-$0.017 & 0.572 & 1.000\\
\quad POPE adv.\ (bound) & 0.686 & 0.864 & $-$0.178 & $9.0{\times}10^{-41}$ & $3.9{\times}10^{-39}$\\
\multicolumn{6}{@{}l}{\emph{Joint: + verifier at hook's proposal vs + yes/no} (perm.)}\\
\quad RefCOCOg val & 0.886 & 0.840 & +0.046 & $1.0{\times}10^{-4}$ & 0.004\\
\quad Stress & 0.580 & 0.545 & +0.034 & 0.165 & 1.000\\
\quad POPE adv. & 0.763 & 0.864 & $-$0.101 & $1.0{\times}10^{-4}$ & 0.004\\
\multicolumn{6}{@{}l}{\emph{Joint: + verifier vs + yes-probability, tuned} (perm.)}\\
\quad RefCOCOg val$^\dagger$ & 0.888 & 0.873 & +0.014 & $1.0{\times}10^{-4}$ & 0.004\\
\quad Stress & 0.528 & 0.564 & $-$0.035 & 0.230 & 1.000\\
\multicolumn{6}{@{}l}{\emph{Continuous vs discrete query, seed 42} (McNemar)}\\
\quad RefCOCOg val & 0.787 & 0.681 & +0.107 & $1.4{\times}10^{-47}$ & $6.2{\times}10^{-46}$\\
\quad Stress & 0.390 & 0.450 & $-$0.061 & 0.103 & 1.000\\
\multicolumn{6}{@{}l}{\emph{Hook vs hook trained without verifier} (perm.)}\\
\quad RefCOCO val & 0.789 & 0.784 & +0.005 & 0.013 & 0.256\\
\quad RefCOCO+ val & 0.707 & 0.705 & +0.003 & 0.203 & 1.000\\
\quad RefCOCOg val & 0.787 & 0.785 & +0.001 & 0.717 & 1.000\\
\quad RefCOCOg test & 0.793 & 0.791 & +0.002 & 0.294 & 1.000\\
\quad Ref-L4 val & 0.464 & 0.465 & $-$0.001 & 0.637 & 1.000\\
\quad Stress & 0.384 & 0.423 & $-$0.039 & 0.020 & 0.366\\
\bottomrule
\end{tabular}
\end{table*}

\paragraph{Choices made on RefCOCOg val.}
\cref{tab:app-tuning} lists every choice made on RefCOCOg val and its measured
optimism on data untouched by the choice.

\begin{table*}[t]
\centering
\footnotesize
\setlength{\tabcolsep}{4pt}
\caption{Every choice made on RefCOCOg val and its measured optimism.
``Oracle'': joint accuracy at the best threshold chosen on the evaluated split
itself, an upper bound rather than a system.}
\label{tab:app-tuning}
\begin{tabular}{@{}>{\raggedright\arraybackslash}p{0.23\textwidth}>{\raggedright\arraybackslash}p{0.22\textwidth}>{\raggedright\arraybackslash}p{0.49\textwidth}@{}}
\toprule
Choice & Selected by & Untouched counterparts and measured optimism\\
\midrule
Veto thresholds $(\gamma_v,\gamma_s)=(0.3,24.0)$ & Faithfulness F1 over a grid on RefCOCOg val (present referents) with an earlier hook of the same recipe (\cref{sec:app-thresholds}); frozen since & Not re-tuned on the features reported here: with absent referents the default veto reaches $0.780$ joint accuracy on RefCOCOg val against a val oracle of $0.832$, and $0.788$ against $0.836$ on test. Selection accuracy without abstention does not depend on the thresholds.\\
Detector score threshold $0.405$ & Joint accuracy on RefCOCOg val with absent referents & Equal to the val oracle ($0.651$); on test $0.648$ against an oracle of $0.651$; on the stress set and POPE $0.011$--$0.032$ below the oracle.\\
VLM yes-probability threshold (yes-minus-no score $-3.58$) & Joint accuracy on RefCOCOg val with absent referents & On test $0.877$ against an oracle of $0.877$; on POPE $0.006$--$0.021$ and on the stress set $0.012$ below the oracle.\\
Agreement signal: re-ranking threshold and emitted box & RefCOCOg val, per seed & Tuned minus default choice: $-0.001$ on RefCOCOg test and RefCOCO val, $+0.001$ on Ref-L4 val, $-0.006$ on the stress set.\\
Index-scorer learning rate and epochs (three swept variants, six configurations each) & Selection accuracy on RefCOCOg val, per seed & Val-minus-test gap of the swept variants minus that of the four unswept variants: $-0.003$, i.e.\ no measurable selection optimism.\\
Reported relation-aware head (best of five variants) & Selection accuracy on RefCOCOg val & The same variant is best on five of the six splits and second on the stress set.\\
Oracle thresholds; union of hook and coordinate decoding & -- & Upper bounds by construction; labeled as oracles wherever they appear.\\
\bottomrule
\end{tabular}
\end{table*}

\paragraph{Intervals for the abstention systems.}
\Cref{tab:app-abst-intervals} gives query-level intervals for the abstention
table. On RefCOCOg val the joint-accuracy intervals of the verifier attached to
coordinate-snap ($[0.880,0.895]$) and of the tuned VLM yes-probability
($[0.865,0.881]$) barely overlap, and the VLM's yes-minus-no score has the lower AURC with
overlapping intervals; paired tests for these comparisons are in
\cref{tab:app-tests}. On the stress set every interval is wide and no two
abstention systems are separated.

\begin{table*}[t]
\centering
\footnotesize
\setlength{\tabcolsep}{5pt}
\caption{Query-level $95\%$ intervals for the abstention systems: joint accuracy
and AURC with $30\%$ swapped absent referents, percentile bootstrap over rows
($10{,}000$ resamples, present and absent rows together, the same resample for
the three seeds, so the interval covers the three-seed mean). Recomputed from
per-row records that reproduce the joint accuracies of \cref{tab:app-detach}
exactly and its AURCs within $0.005$ (the records store verifier probabilities
at reduced precision, which changes ties).}
\label{tab:app-abst-intervals}
\begin{tabular}{@{}lcccc@{}}
\toprule
& \multicolumn{2}{c}{RefCOCOg val + absent ($6{,}994$ rows)} & \multicolumn{2}{c}{Stress + absent ($330$ rows)}\\
\cmidrule(lr){2-3}\cmidrule(lr){4-5}
System & Joint & AURC & Joint & AURC\\
\midrule
Coordinate-snap, always answer & 0.607\ci{.596}{.618} & 0.393\ci{.382}{.404} & 0.427\ci{.376}{.482} & 0.573\ci{.518}{.624}\\
Coordinate-snap + VLM yes/no & 0.840\ci{.831}{.848} & 0.134\ci{.126}{.143} & 0.545\ci{.491}{.597} & 0.412\ci{.342}{.486}\\
Coordinate-snap + VLM yes-probability, tuned & 0.873\ci{.865}{.881} & 0.134\ci{.126}{.143} & 0.564\ci{.512}{.618} & 0.412\ci{.342}{.486}\\
Coordinate-snap + verifier & 0.888\ci{.880}{.895} & 0.146\ci{.137}{.156} & 0.528\ci{.479}{.578} & 0.377\ci{.309}{.447}\\
Coordinate-snap + verifier at the hook's proposal & 0.886\ci{.879}{.893} & 0.161\ci{.151}{.172} & 0.580\ci{.531}{.629} & 0.394\ci{.327}{.463}\\
Detector-native, score threshold & 0.651\ci{.640}{.663} & 0.236\ci{.225}{.249} & 0.276\ci{.227}{.324} & 0.794\ci{.732}{.850}\\
Detector-native + verifier & 0.832\ci{.823}{.840} & 0.209\ci{.198}{.221} & 0.390\ci{.340}{.441} & 0.699\ci{.627}{.769}\\
Hook, default veto & 0.780\ci{.772}{.788} & 0.233\ci{.223}{.244} & 0.413\ci{.365}{.463} & 0.560\ci{.494}{.626}\\
Hook, verifier only & 0.830\ci{.823}{.838} & 0.245\ci{.234}{.256} & 0.478\ci{.430}{.526} & 0.586\ci{.521}{.651}\\
\bottomrule
\end{tabular}
\end{table*}

\paragraph{Standard no-target evaluation (gRefCOCO).}
\Cref{tab:grec} evaluates the abstention layers on gRefCOCO testA, whose
no-target expressions were written to resemble the image content, with every
threshold kept as chosen on RefCOCOg val. Coordinate decoding almost never
abstains. Its own yes/no answer gives the best score and rejects $0.88$ of the
no-target expressions; the verifier and the tuned yes-probability keep more
single-target referents but reject fewer no-target ones.

\begin{table}[t]
\centering
\footnotesize
\setlength{\tabcolsep}{3pt}
\caption{\textbf{Standard no-target evaluation} on gRefCOCO testA
($19{,}200$ expressions: $4{,}448$ no-target, $5{,}917$ single-, $8{,}835$
multi-target) and testB (all expressions; breakdown in
\cref{tab:app-grec-testb}), with every threshold kept as chosen on RefCOCOg
val. Pr@(F1$=$1) is the official GREC score; on no-target expressions it
equals no-target accuracy. Hook-based rows (+ verifier, the hook, and marks
over its top five): mean over three seeds (sample std $\le0.034$), except the
marks row, which uses the seed-42 hook's ranking; on testB with hooks of the
same recipe retrained on the earlier feature build (\cref{sec:app-builds}).}
\label{tab:grec}
\begin{tabular}{@{}lccccc@{}}
\toprule
& \multicolumn{4}{c}{testA, Pr@(F1$=$1)} & testB\\
\cmidrule(lr){2-5}\cmidrule(l){6-6}
System & All & No-tgt. & Single & Multi & All\\
\midrule
Coordinate decoding & 0.336 & 0.030 & \textbf{0.907} & \textbf{0.107} & 0.326\\
\quad + VLM yes/no & \textbf{0.509} & \textbf{0.875} & 0.849 & 0.097 & \textbf{0.519}\\
\quad + VLM yes-prob. & 0.478 & 0.665 & 0.893 & 0.106 & 0.477\\
\quad + verifier & 0.484 & 0.744 & 0.860 & 0.102 & 0.478\\
Hook, default veto & 0.400 & 0.788 & 0.704 & 0.000 & 0.412\\
VLM over hook's top 5, marks & 0.334 & 0.649 & 0.595 & 0.000 & 0.362\\
\bottomrule
\end{tabular}
\end{table}

\paragraph{gRefCOCO testB and val.}
\Cref{tab:app-grec-testb} repeats the no-target evaluation on gRefCOCO testB.
The ranking of the abstention layers is the same as on testA: the VLM's own
yes/no answer is best overall, and the verifier and the tuned yes-probability
are level, with the verifier rejecting more no-target expressions. We do not
report val: in the released annotation file ($258{,}836$ expressions over all
splits) the val split holds $14{,}229$ expressions, $5{,}324$ multi-target and
$8{,}905$ no-target, and no single-target one, whereas testA and testB hold
$5{,}917$ and $5{,}646$ single-target expressions. The benchmark's
single-target expressions are inherited from RefCOCO, and val's are absent
from this release, so a val score here would not follow the published
protocol.

\begin{table}[t]
\centering
\footnotesize
\setlength{\tabcolsep}{1.4pt}
\caption{gRefCOCO testB ($16{,}063$ expressions: $4{,}673$ no-target, $5{,}646$
single-, $5{,}744$ multi-target), same systems and thresholds as \cref{tab:grec}
(every threshold chosen on RefCOCOg val). Pr@(F1$=$1)
overall and by number of targets; N-acc / T-acc: no-target / target
accuracy. ${}^\dagger$Hook-based rows use hooks of the same recipe retrained
on the earlier feature build (\cref{sec:app-builds}) (mean over three seeds;
sample std $\le0.063$; the marks row uses the seed-42 hook's ranking).}
\label{tab:app-grec-testb}
\begin{tabular}{@{}lcccccc@{}}
\toprule
& \multicolumn{4}{c}{Pr@(F1$=$1)} & & \\
\cmidrule(lr){2-5}
System & All & No-tgt. & Single & Multi & N-acc & T-acc\\
\midrule
Coordinate decoding & 0.326 & 0.071 & 0.773 & 0.095 & 0.071 & 0.988\\
\quad + VLM yes/no & 0.519 & 0.840 & 0.692 & 0.089 & 0.840 & 0.790\\
\quad + VLM yes-prob. & 0.477 & 0.622 & 0.746 & 0.095 & 0.622 & 0.929\\
\quad + verifier$^\dagger$ & 0.478 & 0.641 & 0.737 & 0.090 & 0.641 & 0.894\\
Hook, default veto$^\dagger$ & 0.412 & 0.681 & 0.608 & 0.000 & 0.681 & 0.739\\
Marks over hook's top 5$^\dagger$ & 0.362 & 0.584 & 0.547 & 0.000 & 0.584 & 0.877\\
\bottomrule
\end{tabular}
\end{table}

\section{Published Results for Context}
\label{sec:app-published}
\cref{tab:app-published} lists referring-expression accuracies as reported by
the original papers of the systems discussed in the related work. They are
given for orientation only: each system uses its own perception and training
data, most of which include the RefCOCO/+/g training splits, so they are not
matched-perception results and cannot be compared cell by cell with our
tables. The published values for the frozen VLM we use agree with our own
measurement of its coordinate decoding on the shared splits ($0.893$ on RefCOCO val and $0.865$ on RefCOCOg val, against $90.0$ and $87.2$ reported).
GROUNDHOG and F-LMM report mask metrics only, the Grounding DINO paper reports
no Swin-B referring result, and DINO-X \cite{dinox} and Grounding DINO 1.5
\cite{gdino15} report none, so they are not listed. Third-party evaluations
give Qwen3-VL-8B $88.2$ on RefCOCOg val \cite{groundingme}, close to our $0.886$.

\begin{table*}[t]
\centering
\footnotesize
\setlength{\tabcolsep}{4pt}
\caption{Published referring-expression results (accuracy at
$\mathrm{IoU}\ge0.5$, \%, as reported in the original papers), for context only;
these are not matched-perception results. RefC.\ data: whether RefCOCO-family
training data were used (undiscl.: not disclosed); proposals: the source of the
candidate boxes, if any. $^\dagger$Evaluated without RefCOCOg training, with
detections of the ground-truth category as box hints. $^\ddagger$The technical
report gives only the average over the RefCOCO splits. $^\S$F1 at IoU $0.5$
under a multi-target protocol, not accuracy.}
\label{tab:app-published}
\begin{tabular}{@{}lccccccccll@{}}
\toprule
& \multicolumn{3}{c}{RefCOCO} & \multicolumn{3}{c}{RefCOCO+} & \multicolumn{2}{c}{RefCOCOg} & & \\
\cmidrule(lr){2-4}\cmidrule(lr){5-7}\cmidrule(lr){8-9}
Model & val & testA & testB & val & testA & testB & val & test & RefC.\ data & Proposals\\
\midrule
Qwen2.5-VL-7B \cite{qwen25vl} & 90.0 & 92.5 & 85.4 & 84.2 & 89.1 & 76.9 & 87.2 & 87.2 & undiscl. & none\\
Qwen3-VL-8B$^\ddagger$ \cite{qwen3vl} & \multicolumn{8}{c}{average $89.1$; per split not published} & yes & none\\
Shikra-7B \cite{shikra} & 87.0 & 90.6 & 80.2 & 81.6 & 87.4 & 72.1 & 82.3 & 82.2 & yes & none\\
Ferret-7B \cite{ferret} & 87.5 & 91.4 & 82.5 & 80.8 & 87.4 & 73.1 & 83.9 & 84.8 & yes & none\\
Griffon-13B \cite{griffon} & 89.4 & 92.5 & 84.6 & 83.3 & 88.4 & 76.0 & 85.1 & 86.1 & yes & none\\
Florence-2-L, fine-tuned \cite{florence2} & 93.4 & 95.3 & 92.0 & 88.3 & 92.9 & 83.6 & 91.2 & 91.7 & yes & none\\
Rex-Omni$^\S$ \cite{rexomni} & -- & -- & -- & -- & -- & -- & 86.6 & 86.8 & yes & none\\
Groma-7B \cite{groma} & 89.5 & 92.1 & 86.3 & 83.9 & 88.9 & 78.1 & 86.4 & 87.0 & yes & own proposer\\
ChatRex-7B \cite{chatrex} & 91.0 & 94.1 & 87.0 & 89.8 & 91.9 & 79.3 & 89.8 & 90.0 & yes & UPN\\
VGent \cite{vgent} & 92.4 & 94.7 & 89.8 & 88.1 & 92.2 & 83.3 & 90.4 & 90.1 & yes & UPN, SAM, GLEE\\
Rex-Thinker-GRPO$^\dagger$ \cite{rexthinker} & -- & -- & -- & -- & -- & -- & 83.2 & 83.3 & no & detector hints\\
PropVG \cite{propvg} & 92.7 & 95.1 & 89.6 & 87.3 & 90.9 & 81.3 & 88.2 & 88.3 & yes & own proposer\\
RefFormer ViT-L \cite{refformer} & 90.9 & 93.7 & 86.6 & 83.3 & 89.0 & 75.8 & 85.0 & 84.9 & yes & none\\
Grounding DINO-T, fine-tuned \cite{groundingdino} & 89.2 & 91.9 & 86.0 & 81.1 & 87.4 & 74.7 & 84.2 & 84.9 & yes & none\\
\bottomrule
\end{tabular}
\end{table*}

\section{Real-Image Setup}
\paragraph{Configuration.}
Perception is a frozen Grounding DINO (Swin-B checkpoint) producing
request-independent top-$M$ proposals with $256$-dimensional region features
from a fixed prompt listing the COCO categories; queries are read from a frozen
Qwen2.5-VL-7B as one $3584$-dimensional embedding per request. The hook (two
projections and the verifier) and the other heads are trained on $24$k RefCOCOg
training rows ($16{,}724$ present expressions, $7{,}276$ swapped absent rows), and
on the full training split in \cref{sec:app-full}; the thresholds $(\gamma_v,\gamma_s)=(0.3,24.0)$ are selected once
on RefCOCOg val (\cref{sec:app-thresholds}) and held fixed across every
reported split. Evaluation covers
RefCOCO val ($10{,}834$ expressions), RefCOCO+ val ($10{,}758$), RefCOCOg val
($4{,}896$) and test ($9{,}602$), POPE random/popular/adversarial ($3{,}000$
each), Ref-L4 val ($13{,}420$) and the reasoning-stress set ($231$); for
abstention, RefCOCOg val, RefCOCOg test and the stress set are augmented with
$30\%$ swapped absent referents ($6{,}994$, $13{,}717$ and $330$ rows).

\paragraph{Data efficiency.}
\label{sec:app-dataeff-text}
Selection accuracy of the hook on RefCOCOg val rises from $0.389{\pm}0.002$ at
$200$ training expressions to $0.475{\pm}0.003$ ($500$), $0.545{\pm}0.009$
($1$k), $0.664{\pm}0.010$ ($2$k), $0.729{\pm}0.008$ ($5$k), $0.752{\pm}0.025$
($10$k) and $0.787{\pm}0.004$ at $24$k (three seeds); recall@5 rises from
$0.821$ to $0.958$ and instance-exact selection from $0.197$ to $0.405$ over the
same range.

\section{Training and Inference Algorithm}
\label{sec:app-algorithm}

Algorithm~\ref{alg:hook} gives the full training and inference procedure. For
clarity, $v_\psi$ denotes the verifier head parameterized by $\psi$.

\begin{algorithm}[htbp]
\caption{Alignment hook: training and inference}
\label{alg:hook}
\small
\begin{algorithmic}[1]
\Require image $I$, request $q$, frozen detector $D$, frozen LM $F$,
hook parameters $\theta,\phi,\psi$, proposal budget $M$
\State $\{(b_j,e_j)\}_{j=1}^{M} \leftarrow D(I)$ \Comment{proposals from the fixed prompt, request-independent}
\State $u \leftarrow F(I,q)$ \Comment{query embedding of the request}
\State $\tilde{e}_j \leftarrow f_\theta(e_j)$,\enspace $\tilde{u} \leftarrow g_\phi(u)$ \Comment{shared-space projections}
\State $s_{j} \leftarrow \langle\tilde{u},\tilde{e}_j\rangle/\kappa$ \Comment{affinities}
\State $j^\star \leftarrow \arg\max_j\; s_{j}$,\enspace $p \leftarrow v_\psi(\tilde{u},\tilde{e}_{j^\star},s_{j^\star})$
\If{training}
    \State $P \leftarrow$ matched proposal (maximal IoU, if $\ge0.5$; $\varnothing$ for absent requests)
    \State Compute $\mathcal{L}_{\mathrm{ret}}$ if $P\neq\varnothing$ and $\mathcal{L}_{\mathrm{faith}}$ from $p$ and the existence label
    \State Update $\theta,\phi,\psi$ only (perception and LM frozen)
\Else
    \If{$p\ge\gamma_v \;\wedge\; s_{j^\star}\ge\gamma_s$}
        \State emit box $\hat{b} \leftarrow b_{j^\star}$
    \Else\enspace abstain
    \EndIf
\EndIf
\end{algorithmic}
\end{algorithm}

\paragraph{Complexity and latency.}
Let $d_r$ and $d_q$ denote the detector-region and language-query dimensions,
$h$ the hidden width of the hook's projections and $d_s\le h$ the dimension of
its shared space. Projecting $M$ regions and $K$ queries
costs $\mathcal{O}(Md_rh + Kd_qh)$; the affinity matrix costs
$\mathcal{O}(MKd_s)$. With one query per request ($K{=}1$) selection is a
top-1 search and the dominant cost is the affinity computation over $M$
proposals, linear in $M$. Wall-clock hook latency measured on an NVIDIA H800
(200 timed calls after 50 warm-up calls, cache preloaded, $d_r{=}256$,
$d_q{=}3584$, $h{=}256$, $d_s{=}128$): $p_{50}=0.619$\,ms ($M{=}50$), $0.614$\,ms
($M{=}100$), $0.501$\,ms ($M{=}200$), $0.495$\,ms ($M{=}300$). At these scales
the measurement is dominated by a kernel-launch and timing floor rather than by
the $\mathcal{O}(M)$ arithmetic; the practical conclusion is that hook latency
is negligible at every tested budget.

\paragraph{End-to-end cost.}
\label{sec:app-cost}
\Cref{tab:app-cost} times every component of the compared pipelines on one
NVIDIA RTX PRO 6000 (Blackwell, 96\,GB) with nothing else running on the card.
The workload is $200$ timed requests (after $20$ untimed ones) drawn at random
from RefCOCOg val; the $220$ requests cover $201$ images of mean size
$584{\times}476$.
Inputs are the published ones: the detector at $M{=}300$, the VLM in bfloat16
at the cache's input resolution, coordinate decoding in float16 with the
effectively greedy decoding of \cref{sec:app-seeds} stopped at the first complete box, and the heads on the cached
features. The timed prefill reproduces the cached queries (cosine $1.000$). A pipeline's latency is the sum of its components' batch-1 means.
Proposal-based routes need the detector pass and one prefill but no
autoregressive decoding. The verifier needs the detector pass and a second
prefill of the bare request, while the yes/no gate needs only a second
prefill. The detector pass depends on the image alone, so it can be shared by
all requests on an image: $3.8$ per image on RefCOCOg val. The relation-aware
head and the token-state probe are timed with random weights of the published
size.

\begin{table*}[t]
\centering
\footnotesize
\setlength{\tabcolsep}{4pt}
\caption{End-to-end cost on one RTX PRO 6000. Top: components, batch-1
latency percentiles over $200$ requests, throughput at batch $1$ and at the
largest tested batch, and peak allocated GPU memory at batch $1$. Bottom:
per-request pipeline latency (sum of batch-1 means), with the detector pass
paid by every request or shared by the $3.8$ requests per image of RefCOCOg
val. $^\dagger$Left-padded batched greedy decoding, not the published per-request
protocol.}
\label{tab:app-cost}
\begin{tabular}{@{}lcccccc@{}}
\toprule
Component & Needs detector & $p_{50}$ (ms) & $p_{90}$ (ms) & Req/s, batch 1 & Req/s, batched (size) & Peak memory (GB)\\
\midrule
Detector pass (Swin-B, $M{=}300$) & yes & 143.1 & 156.3 & 7.1 & 5.4 (8) & 2.46\\
VLM prefill, query readout & -- & 92.4 & 112.0 & 10.3 & 6.1 (16) & 15.77\\
\quad with FlashAttention-2 & -- & 74.9 & 81.9 & 13.2 & 18.8 (16) & 15.77\\
Coordinate decoding & no & 585.9 & 617.0 & 1.7 & 4.9 (16)$^\dagger$ & 15.68\\
Yes/no answer & no & 96.6 & 111.8 & 10.1 & 6.1 (16) & 15.67\\
Alignment hook (selection and verifier) & yes & 0.37 & 0.38 & 2{,}688 & 355{,}919 (256) & 0.08\\
Relation-aware head & yes & 2.75 & 2.78 & 360 & 9{,}254 (256) & 0.11\\
Token-state probe & yes & 2.18 & 2.21 & 457 & 37{,}175 (256) & 0.21\\
\midrule
Pipeline & Needs detector & \multicolumn{2}{c}{Per request (ms)} & \multicolumn{2}{c}{Detector shared (ms)} & Frozen models\\
\midrule
Detector + prefill + hook & yes & \multicolumn{2}{c}{239.1} & \multicolumn{2}{c}{135.1} & both\\
Detector + prefill + relation-aware head & yes & \multicolumn{2}{c}{241.5} & \multicolumn{2}{c}{137.5} & both\\
Detector + prefill + token-state probe & yes & \multicolumn{2}{c}{240.9} & \multicolumn{2}{c}{136.9} & both\\
Coordinate decoding (raw box) & no & \multicolumn{2}{c}{587.8} & \multicolumn{2}{c}{587.8} & VLM\\
Coordinate decoding + yes/no gate & no & \multicolumn{2}{c}{686.7} & \multicolumn{2}{c}{686.7} & VLM\\
Coordinate-snap & yes & \multicolumn{2}{c}{729.5} & \multicolumn{2}{c}{625.4} & both\\
Coordinate-snap + verifier & yes & \multicolumn{2}{c}{827.0} & \multicolumn{2}{c}{722.9} & both\\
\bottomrule
\end{tabular}
\end{table*}

A proposal-based pipeline thus costs $0.41$ of coordinate decoding per request
($0.23$ with the detector pass shared), because one prefill replaces
autoregressive decoding; the trained head itself adds under $3$\,ms.
Abstention costs more with the verifier than with the yes/no gate ($827$ against
$687$\,ms per request) because the verifier needs the detector pass; with the
pass shared the difference falls to $36$\,ms. The two frozen models together
hold about $18$\,GB at batch $1$. Batching helps the prefill only with
FlashAttention-2 ($18.8$ requests/s at batch $16$), and the batched detector is
slower than batch $1$ at batch $8$ because of padding.

\section{Additional Discussion Details}
\label{sec:app-additional-discussion}

\paragraph{Low-rank fine-tuning protocol.}
Low-rank adaptation uses rank $16$ on the query, key, value and output
projections of Qwen2.5-VL-7B, learning rate $2{\times}10^{-4}$, three epochs,
micro-batch $8$ with $8$ accumulation steps (effective batch $64$) and a
maximum input of $401{,}408$ pixels. Each training example is the documented
grounding prompt with the expression, answered in the model's structured box
format with the box in absolute pixels of the resized input frame, followed by the end-of-turn token; the loss is applied to the answer
tokens only, so the image and prompt tokens do not dilute it. Evaluation parses
the answer on the full RefCOCOg val split ($n{=}4{,}896$; a parse failure
counts as a miss) and scores the unsnapped box. The point without fine-tuning runs
through the same pipeline: $0.865$ at $\mathrm{IoU}\ge0.5$ and $0.749$ at
$\mathrm{IoU}\ge0.75$. After fine-tuning, accuracy is $0.866{\pm}0.003$ /
$0.751{\pm}0.003$ ($200$ expressions), $0.868{\pm}0.002$ / $0.754{\pm}0.002$
($1{,}000$) and $0.871$ / $0.754$ ($5{,}000$, one seed), with no parse
failures. The pipeline runs at a lower input resolution than coordinate-snap,
so the two numbers without fine-tuning are different operating points of the
same model ($0.865$ unsnapped here, $0.867$ snapped for coordinate-snap). With the recipe of the fine-tuned selector instead (one epoch, $1{,}003{,}520$ pixels, its $16{,}697$ present expressions, three seeds), native-box fine-tuning reaches $0.871$ on RefCOCOg val and $0.875$ on test (\cref{sec:app-lora-matched}).

\paragraph{Seed protocol.}
Every hook-level result is the mean over three training seeds (42, 123, 456)
with the sample standard deviation, at one fixed threshold pair
(\cref{sec:app-stats}).

\paragraph{Broader impact and ethics.} No system we evaluate both grounds and rejects reliably on GroundingME (\cref{tab:app-public}), so none should drive compliance or enforcement decisions.
Faithful open-vocabulary grounding can be misused for surveillance, covert
person tracking and population monitoring. Abstention reduces false detections
but does not remove them: every gate studied here still answers on a share of
absent referents, and requests about people that cannot be verified visually
(sensitive attributes, social roles, ambiguous states) lie outside what the
verifier was trained to judge. We therefore recommend restricting downstream
deployment through access controls, usage auditing and explicit prohibition of
high-risk uses. The system inherits biases from both the frozen
detector and the language model: world-knowledge queries involving social
roles, safety-related attributes or cultural context may amplify stereotypical
correlations already present in COCO/RefCOCO-style training data. These
datasets do not uniformly represent all environments, occupations, body types
or cultural backgrounds, so benchmark accuracy should not be interpreted as
evidence of universal reliability or fairness. For production use, we
recommend a model card documenting training dependencies, known failure
modes, benchmark limitations and prohibited applications.

\paragraph{Ref-L4.}
Ref-L4~\citep{refl4} is a benchmark of long, GPT-4V-generated and
human-reviewed expressions (about $24$ words on average) over $365$
categories, on images of the RefCOCO-family evaluation splits and of
Objects365. We evaluate the RefCOCOg-trained hook zero-shot
on Ref-L4 val ($13{,}420$ queries): selection accuracy is $0.464{\pm}0.010$
(three seeds; $0.465$ without verifier training), instance-exact selection
$0.245$ and recall@5 $0.731$; the trained controls reach $0.444$ (index
selector) and $0.357$ (CLIP-query bridge) and the relation-aware head $0.522$.
Proposal recall is $0.958$, so $4\%$ of queries are lost to perception before selection. Detector-native selection reaches $0.634$ and coordinate-snap $0.833$
(raw coordinate decoding $0.853$, with images larger than $1{,}003{,}520$
pixels downscaled to that size). The drop from RefCOCOg ($0.787$) is much
larger for the hook than for coordinate-snap ($0.867\to0.833$); the vocabulary
decomposition (\cref{tab:app-refl4buckets,tab:app-refl4source,tab:app-extvocab,tab:app-upn,tab:app-style})
locates most of it in referents outside the COCO categories of the training
data, a category effect that persists within one image source, under every
box-size specification and with a class-agnostic proposer.

\paragraph{Coordinate-snap protocol.}
\label{sec:app-coordsnap}
Coordinate-snap prompts the frozen Qwen2.5-VL-7B for a box in its documented
absolute-pixel JSON format in the smart-resized input frame (prompt and parse
rule in \cref{sec:app-parse}), converts the box
to image coordinates, and snaps it to the proposal of maximal IoU. A
normalized-coordinate prompting convention does not match the model's
documented format, so the documented format is used. It gives $0.895$ (RefCOCO val), $0.840$ (RefCOCO+ val), $0.867$ (RefCOCOg val), $0.866$
(RefCOCOg test), $0.833$ (Ref-L4 val) and $0.610$ (stress set) after snapping;
unsnapped accuracies are within $0.02$ of the snapped ones, and $12$ of
$49{,}741$ present queries fail to parse; an unparsable answer is replaced by
the fixed box covering the top-left quarter of the image, which almost
always counts as a miss. These values are consistent with the
model's reported referring-expression accuracy and, together with the
detector-native interface, are the references of the main paper that receive
no task-specific training here; neither is zero-shot on these splits, since
the detector was trained with RefCOCO-family data and the VLM's grounding
corpora are undisclosed.

\paragraph{Detector-native protocol.}
\label{sec:app-detnative}
Detector-native selection runs the frozen detector of \cref{tab:repro_config}
once per request with the request itself as the text prompt: the full referring
expression, with surrounding white space removed and a period appended when it
does not end with one (on POPE, the object phrase of the question). No phrase
is extracted from the expression. The detector's processor resizes the image
exactly as in the proposal pass (shorter side $800$, longer side at most
$1{,}333$ pixels). Each of the $900$ decoder queries is scored by the sigmoid
of its largest logit over the token positions of the prompt, and the box of the
highest-scoring query is the answer. No box or text score threshold, no
non-maximum suppression and no phrase-level aggregation of token scores (for
example over the tokens of a parsed noun phrase) is applied, so the interface
always returns exactly one box; its score is the abstention signal of the
score-threshold rows. The pass does not involve the VLM, so one set of boxes
serves both VLM runs, and the rule is the same on every split. It is the rule
of the detector's official referring-expression evaluation, which also scores
each query by its largest probability over all prompt tokens, with the same
resizing and without threshold or suppression; a re-run reproduces every stored
box. Re-scoring each query by the mean probability over the expression's own
tokens (phrase-level scoring, as in the detector's given-phrase mode) changes
accuracy by $-0.007$ to $+0.002$ on RefCOCO/+ val and RefCOCOg val/test
($0.777$ on RefCOCOg val) and lowers Ref-L4 val by $0.034$, so the selection
rule does not explain the distance to the $84.2$ of the smaller Swin-T model
(\cref{tab:app-published}), which is a fine-tuned accuracy.

\section{Feature Builds}
\label{sec:app-builds}
Every result uses the same frozen detector checkpoint, prompt, proposal
budget and VLM, except the second proposal source (\cref{sec:app-second-detector})
and the class-agnostic proposer (\cref{sec:app-upn}), which replace the
detector, and the second and third VLMs (\cref{sec:app-full,sec:app-internvl3}), which replace the query. Unless listed in \cref{tab:app-builds}, a result comes from
the main build, the feature caches behind the main paper's tables, with the published
hooks. The earlier build was extracted with the same models on different
hardware and software before the main build. Its proposals cover the same
referents (recall at $\mathrm{IoU}\ge0.5$: $1.000$ against $1.000$ on RefCOCOg
val, $0.998$ against $0.998$ on test, $0.956$ against $0.958$ on the Ref-L4
queries both builds share, with per-query agreement $100\%$, $99.96\%$ and
$99.2\%$), but their scores and therefore their order differ: the main hook's
top-ranked proposal index points to a box of the same IoU in the earlier build
for only $9$--$28\%$ of queries, so hooks are retrained per build. Retrained
on the earlier build, the hook reaches $0.779\pm0.013$ on RefCOCOg val against
$0.787\pm0.004$; the VLM's boxes snapped to either build agree within
$0.004$ on RefCOCOg; detector-native selection on Ref-L4 reaches $0.662$
against $0.634$, the difference lying on COCO images ($0.691$ against
$0.649$; $0.610$ against $0.609$ elsewhere). We did not isolate the cause of
the score differences beyond the change of hardware and library versions.

\begin{table}[t]
\centering
\footnotesize
\setlength{\tabcolsep}{3pt}
\caption{Results not computed entirely on the main build.}
\label{tab:app-builds}
\begin{tabular}{@{}>{\raggedright\arraybackslash}p{0.40\columnwidth}>{\raggedright\arraybackslash}p{0.54\columnwidth}@{}}
\toprule
Result & Build \\
\midrule
Margin and separability analysis (\cref{tab:app-margin}) & Earlier build, hooks retrained on it \\
Extended-vocabulary proposals (\cref{tab:app-extvocab}) & Earlier build; Ref-L4 re-detected with the extended prompts; hooks of the margin analysis \\
In-VLM selection on the remaining splits (\cref{tab:app-invlm-splits}) & Earlier build, seed-42 hook retrained on it \\
Verifier with harder negatives (\cref{tab:app-hardneg-train}) & Earlier build, hooks retrained on it; gRefCOCO features extracted for this run \\
gRefCOCO testB (\cref{tab:grec,tab:app-grec-testb}) & testB features extracted with the earlier build; hooks retrained on it (testA: main build) \\
Discrete phrase interface (\cref{tab:app-stress,tab:app-buckets}) & Its caches rebuilt with the detector re-run; covered-proposal flags identical to the main build \\
Full-budget heads (Table 1 of the main paper, \cref{sec:app-full}) & Evaluation on the main build; the $21{,}899$ training images re-detected with the same models \\
Veto thresholds (\cref{sec:app-thresholds}) & Selected with a hook trained on the earlier build; re-selection on the main build in \cref{sec:app-thresholds} \\
Test splits and Ref-L4 test (\cref{tab:app-eightsplit,tab:app-refl4test}) & Extracted later with the main build's settings on its hardware (rebuilt RefCOCO val rows identical); relation-aware heads retrained with the published recipe \\
Training on all three RefCOCO-family sets (\cref{tab:app-union}) & Main build; the $4{,}368$ union training images new to the feature cache extracted later with the same models (re-detected cached images reproduce their boxes) \\
Class-agnostic proposals (\cref{tab:app-upn}) & Proposals and region features from UPN instead of the detector; the $24$k training rows, RefCOCOg val/test, Ref-L4 val and the stress set re-detected; queries and token states of the main build; heads retrained \\
\bottomrule
\end{tabular}
\end{table}

\section{Reproducibility Configuration}
\label{sec:app-repro}

\cref{tab:repro_config} lists the configuration of the real-image experiments;
the paragraph after it gives the details that do not fit the table (proposal
formation, the multi-layer readout, the control query encoders, the sampling of
absent rows and subsets, the VLM yes/no score and training time).

\begin{table*}[t]
\centering
\footnotesize
\setlength{\tabcolsep}{3pt}
\caption{Reproducibility configuration sheet for the real-image experiments.}
\label{tab:repro_config}
\begin{tabular}{@{}p{0.20\textwidth}p{0.77\textwidth}@{}}
\toprule
\textbf{Item} & \textbf{Value / notes} \\
\midrule
Frozen detector & Grounding DINO, Swin-B backbone, the authors' public base checkpoint; fixed request-independent prompt of the $80$ COCO category names; shorter image side $800$, longer side at most $1{,}333$ pixels; the top $M$ of its $900$ decoder queries by score, without non-maximum suppression or a score threshold (\cref{sec:app-config-notes}). \\
Region feature dimension & 256: the last decoder layer's state of each selected query, before the hook's projection. \\
Language model & Qwen2.5-VL-7B, frozen. Query: final-layer hidden state at the last position of the chat template whose user turn is the image followed by the bare request (no instruction), with the assistant prefix appended, i.e.\ the position that would begin the answer; no grounding prompt precedes the readout. Multi-layer readout: the same position after layers 28, 21 and 14 (of 28), concatenated and passed through the head's query projection (\cref{sec:app-config-notes}). Images resized by the model's processor to at most $1{,}003{,}520$ pixels. \\
Query feature dimension & 3584. \\
Proposal budget $M$ & 300. \\
Veto thresholds & $\gamma_v{=}0.3$, $\gamma_s{=}24.0$; selected once by faithfulness F1 on RefCOCOg val and fixed for every reported split (definition and provenance in \cref{sec:app-thresholds}). \\
Abstention baselines & Detector score threshold $0.405$ and VLM yes-minus-no score threshold $-3.58$, both maximizing joint accuracy on RefCOCOg val with $30\%$ swapped absent referents, frozen elsewhere. \\
Training split & RefCOCOg train, $24{,}000$ rows: $16{,}724$ present expressions and $7{,}276$ swapped absent rows (negative rate $0.3$). \\
Absent rows and subsets & Swapped absent rows in training and evaluation and the fixed $3{,}000$-request subsets are drawn with generator seed $0$ (rules in \cref{sec:app-config-notes}); Ref-L4 test uses a stratified $6{,}000$-request subset (\cref{sec:app-std-test}). \\
IoU threshold $\tau$ & 0.5 for selection accuracy and the positive-set definition. \\
Hook architecture & Query and region projections: MLPs $d{\to}256{\to}128$ (one GELU hidden layer), $\ell_2$-normalized outputs. Verifier: MLP $257{\to}256{\to}1$ (one GELU hidden layer, sigmoid output) on $[\tilde{u};\tilde{e}_\sigma;s_\sigma]$, the projected query, the projected top-ranked proposal and its affinity. No dropout; $1{,}115{,}650$ trainable parameters with Qwen2.5-VL queries. \\
Control query encoders & CLIP ViT-B/32 text encoder, $512$-dimensional pooled output; the detector's BERT-base text tower, mean-pooled ($768$-d), and the same after the detector's linear text projection ($256$-d, ``projected''); all frozen (\cref{sec:app-config-notes}). \\
VLM answers & Coordinates: greedy (Qwen2.5-VL: temperature $10^{-6}$, effectively greedy; \cref{sec:app-seeds}), at most $64$ new tokens; prompt, parse rule and fallback box in \cref{sec:app-parse}. Yes/no: prompt and score in \cref{sec:app-config-notes}. \\
Training time & Per seed on one RTX PRO 6000: hook about $1$\,min and the other heads $5$--$6.5$\,min at $24$k rows; heads $20$--$39$\,min at the full budget (\cref{sec:app-config-notes}). \\
Temperature $\kappa$ & Learned; initialized at 0.07, it converges to $0.017$ ($0.0166$--$0.0174$ over three seeds of the same recipe), so $\gamma_s{=}24$ is a cosine threshold of about $0.41$. \\
Loss weight & $\lambda_f{=}1.0$ (retrieval and verifier terms; no box regression). \\
Optimizer & AdamW, lr $1{\times}10^{-3}$, weight decay $10^{-4}$, batch 256, 50 epochs. \\
Seeds & 42, 123, 456 for every hook-level table (mean$\pm$sample std). \\
Hardware / software & NVIDIA H800 (80 GB), RTX PRO 6000 (96 GB) and an RTX 3090-class virtual GPU (48 GB); main build: PyTorch 2.8, Transformers 4.51.3 (4.57.6 for Qwen3-VL and InternVL3). \\
\bottomrule
\end{tabular}
\end{table*}

\paragraph{Configuration details.}
\label{sec:app-config-notes}
\emph{Proposals.} The detector processes each image once with the category
prompt. Each of its $900$ decoder queries is scored by the sigmoid of its
largest logit over the prompt tokens, and the $M{=}300$ highest-scoring queries
are kept with their boxes and their $256$-dimensional last-layer decoder
states. No non-maximum suppression or score threshold is applied, so
near-duplicate boxes can remain among the proposals; proposal recall is
measured on this set.
\emph{Multi-layer readout.} For Qwen2.5-VL the readout concatenates the query
state after layer $28$ (after the final normalization, i.e.\ the query $u$
itself) with the outputs of layers $21$ and $14$ at the same position, which
are taken without the final normalization; these are the last layer and
three-quarter and half depth (the same relative depths for Qwen3-VL). For the
multi-layer hook rows the resulting $10{,}752$-dimensional vector takes the
place of $u$, and the hook's query projection ($10{,}752{\to}256{\to}128$) is
the only fusion; there are no learned layer weights. The verifier reads the
projected query and therefore sees the three depths only through that
projection. All other hook rows and every verifier result use the last-layer
query alone. The combined selector instead receives the states after layers
$21$ and $14$ as two separate context tokens.
\emph{Control query encoders.} The CLIP query is the $512$-dimensional pooled
output of the CLIP ViT-B/32 text encoder at the end-of-text token, before
CLIP's text projection, with expressions truncated to $77$ tokens. The detector
text tower is the detector's own BERT-base text encoder, applied to the
lower-cased expression ending with a period and mean-pooled over its
non-padding tokens ($768$ dimensions). Its \emph{projected} variant passes that
mean through the detector's frozen linear text projection ($768{\to}256$),
which maps text features into the detector's $256$-dimensional model space
before its cross-modal fusion layers. All encoders are frozen; only the bridge
is trained.
\emph{Absent rows.} In the $24$k training split each row is made absent
independently with probability $0.3$ (generator seed $0$): its expression is
paired with the image of a uniformly drawn row of the same split, the VLM query
is computed on that image, and the row keeps that image's proposals. At the
full budget, $34{,}505$ expressions are drawn uniformly without replacement and
each is paired with a uniformly drawn different image (seed $0$). For
evaluation, present rows are drawn without replacement so that absent rows make
up $30\%$ of the augmented split, each paired with a uniformly drawn row that
shows a different image (seed $0$); every method sees the same absent rows.
\emph{Subsets.} The $3{,}000$-request subsets of the in-VLM selection tables
(Ref-L4 val, RefCOCO val, RefCOCO+ val and RefCOCOg test) are drawn uniformly
without replacement from the split's requests with generator seed $0$; where a
split has swapped absent rows, the same generator then draws the absent rows in
proportion ($1{,}286$ for RefCOCOg test).
\emph{VLM yes/no.} Both VLMs receive the image and the prompt \emph{Is there
``$\langle$expression$\rangle$'' in the image? Answer yes or no.} (on POPE the
benchmark question followed by \emph{Answer yes or no.}) at the budget of
$1{,}280$ visual tokens. One forward pass gives the score: the log of the total
next-token probability of the first tokens of \emph{Yes} and \emph{yes}, with
and without a leading space, minus the same quantity for \emph{No} and
\emph{no}, at the first answer position. The binary gate answers when the score
is positive; the tuned gate answers when it is at least $-3.58$.
\emph{Training time.} Wall-clock per seed as logged by the training runs on one
RTX PRO 6000, excluding feature extraction (\cref{tab:app-cost}): at $24$k
rows, $55$--$61$\,s for the hook, $6.1$--$6.4$\,min for the relation-aware head
and $4.9$\,min for the token-state probe; at the full budget ($115$k rows),
$28$--$29$\,min (Qwen2.5-VL) and $27$--$39$\,min (Qwen3-VL) for the
relation-aware head, $20$--$28$ and $20$--$27$\,min for the token-state probe,
and $30$--$32$ and $28$--$35$\,min for the combined selector. The training run
of the low-rank fine-tuned selector (one epoch) took $2.2$--$3.2$\,h per seed (\cref{sec:app-lora-matched}). Peak GPU
memory was $10$--$14$\,GB at $24$k rows and $22$--$26$\,GB at the full budget.

\paragraph{Provenance of the veto thresholds.}
\label{sec:app-thresholds}
$(\gamma_v,\gamma_s)$ were selected once, before the evaluation build used in
this paper, by maximizing faithfulness F1 over the grid
$\gamma_v\in\{0.3,0.4,0.5,0.6,0.7\}$ $\times$ $\gamma_s$ at the
$0.1,0.2,\dots,0.5$ quantiles of the per-query maximal affinity, on all
$4{,}896$ RefCOCOg val expressions (present referents only). Faithfulness F1
takes as its positive class an answered, instance-exact selection of a present
referent. A request is a true positive if the system answers, the referent is
present and the top-ranked proposal is the proposal matched to the annotated
box (the proposal of maximal IoU, provided that this IoU is at least $0.5$); a
false positive if the system answers and the request is not a true positive
(an answer to an absent referent, or an answered selection that is not
instance-exact); and a false negative if the system abstains on a present
referent, whether or not its selection would have been correct. Precision
$\mathrm{TP}/(\mathrm{TP}{+}\mathrm{FP})$, recall
$\mathrm{TP}/(\mathrm{TP}{+}\mathrm{FN})$ and their harmonic mean are computed
over all rows of the selection split. Because that split has no absent
referents, the criterion trades answering wrong selections against abstaining
on present referents and is not informed by absent referents. The hook used for
this selection had the recipe of \cref{tab:repro_config} but was trained on the
earlier feature build of \cref{sec:app-builds} with the configuration's default
seed. The selected
values were $(0.3,24.01)$, rounded to $(0.3,24.0)$. $\gamma_v=0.3$ is the
lowest value of the grid. Re-running the selection with hooks of seeds 42, 123
and 456 of that earlier build gave $\gamma_v=0.3$ and $\gamma_s=22.95$, $23.62$
and $23.36$; with the hooks of the main build it gives $\gamma_v=0.3$ and
$\gamma_s=21.94$, $22.00$ and $23.60$, the lowest quantile of its grid.
Extending the $\gamma_v$ grid down to $0.01$ on the main build selects
$\gamma_v=0.01$, $0.01$ and $0.1$ but changes the criterion by at most
$0.0009$, and extending the $\gamma_s$ grid as well runs to the corner without
an affinity gate: on rows without absent referents the criterion rewards
answering. On RefCOCOg test with swapped absent referents the hook's joint
accuracy is $0.788$ at the frozen pair, $0.786$ with the re-selected
$\gamma_v$, $0.808$ with the original grid re-selected on the main build and
$0.821$ with both grids extended; the coordinate-snap gate drops from $0.888$ to
$0.884$ at the re-selected $\gamma_v$. The veto rows are therefore reported at a
fixed operating point that is not optimal for joint accuracy (with absent
referents the default veto is $0.05$ below the RefCOCOg val oracle,
\cref{tab:app-tuning}); lowering $\gamma_v$ alone does not improve it. RefCOCOg val is also
an evaluation split, so the abstaining rows on it are tuned-split results
(\cref{tab:app-tuning}).

\paragraph{Relation-aware head.}
The query embedding passes through LayerNorm and an MLP
($3584{\to}256{\to}256$) and is expanded into eight slot tokens; a FiLM layer
(zero-initialized) modulates the proposal tokens with the pooled query. Each
proposal token is an MLP of its LayerNorm-ed region feature plus an MLP of its
seven-dimensional geometry $(x_1,y_1,x_2,y_2,w,h,\text{area})$; one existence
token is appended. Two pre-LayerNorm layers each apply self-attention over the
$M{+}1$ tokens with a pairwise bias, cross-attention to the slot tokens and a
GELU feed-forward block of width 1024 ($d=256$, 8 heads, dropout 0.1). The
pairwise bias is an MLP ($7{\to}16{\to}$ one scalar per head and layer) of
signed-log center offsets normalized by the first box's width and height,
log width and height ratios, raw center offsets and IoU. A proposal's logit is
the cosine of projected token and query divided by a learned temperature
(initialized at 0.07) plus an MLP of $[t;u;t\odot u]$; the existence logit is
linear in the existence token and the query, and the head answers when its
sigmoid is at least $0.3$. Loss: multi-positive InfoNCE (positives: proposals
with $\mathrm{IoU}\ge0.5$ and the matched proposal) $+\,1.0\times$
cross-entropy on the matched proposal $+\,0.5\times$ listwise cross-entropy with
targets proportional to $\mathrm{IoU}^4$ $+\,1.0\times$ binary cross-entropy on
existence. AdamW, learning rate $2{\times}10^{-4}$, weight decay $0.05$, batch
64, 20 epochs with 5\% linear warm-up and cosine decay, gradient clipping at
1.0, bfloat16 autocast ($7{,}500$ steps on the $24$k rows, $35{,}960$ on the
$115{,}017$ full-budget rows); $4.29$M parameters with Qwen2.5-VL queries,
$4.42$M with Qwen3-VL. This configuration was the best of five relation-head
variants on RefCOCOg val.

\paragraph{Token-state probe and combined selector.}
The context is the final-layer, final-norm states of the request tokens and of
the five chat-template tokens that close the user turn and open the answer (no
image positions); each state is LayerNorm-ed, projected $3584{\to}256$ and given
a type embedding and a position-from-end embedding. Proposal and existence
tokens are built as in the relation-aware head and pass through two layers of
cross-attention to the context (padding masked) and a feed-forward block; a
request summary for the logit head is pooled by attention with a learned query.
The combined selector adds proposal self-attention with the pairwise geometry
bias and two context tokens holding the final position after $3/4$ and $1/2$ of
the layers. Feed-forward widths (1956 for the probe, 1425--1427 for the
combined selector) are set so that both match the relation-aware head's
parameter count ($4.29$M / $4.42$M); loss and optimization are those of the
relation-aware head. The combined selector is trained only at the full budget, except in the
proposal-source comparison of \cref{sec:app-upn}, which retrains it at $24$k.

\paragraph{Index scorer.}
The query ($3584$-d) and each proposal's region feature with its box ($256+4$)
are projected by two-layer GELU MLPs to $256$ dimensions, and a learned
embedding of the proposal's index is added on the region side; an MLP over
$[q;r;q\odot r;|q-r|]$ ($1024{\to}256{\to}1$) gives one logit per proposal,
trained with cross-entropy on the matched proposal's index (present rows
without a covering proposal are trained toward index $0$ in the variant as
proposed). AdamW, learning rate $10^{-3}$, weight decay $10^{-4}$, batch $256$,
$30$ epochs, seeds 42, 123, 456; no verifier, emission by a maximum-softmax
threshold tuned on RefCOCOg val. The audited variant masks uncovered rows from
the loss, applies a parameter-free LayerNorm to both inputs and selects the
learning rate and number of epochs on RefCOCOg val from
$\{10^{-4},3{\times}10^{-4},10^{-3}\}\times\{50,100\}$ ($10^{-4}$ and $50$
selected); the other audit variants (no index embedding, bilinear scorer,
masking or the sweep alone) are in \cref{tab:app-controls}.

\paragraph{Ref-L4 vocabulary buckets.}
Each Ref-L4 referent carries the benchmark's source-category identifier, which
names either a COCO category or an Objects365 category. A query is in
vocabulary if that identifier is a COCO category ($8{,}483$ val queries) and
out of vocabulary otherwise ($4{,}937$). The COCO-mappable subset of
\cref{tab:app-refl4source} uses a strict manual name mapping of $79$
Objects365 categories, each to a distinct COCO-80 class (``sports ball'' has no
strict counterpart), which covers $1{,}664$ of the $4{,}937$ queries.

\paragraph{Achievable-margin computation.}
$\mu_p$ is computed per query by Frank--Wolfe over the convex hull of the
incorrect proposals' projected embeddings, started at the nearest incorrect
proposal, with exact line search and a fixed $300$ iterations (no early stop),
in batches of 512 queries in single precision. The hull iterate gives an upper
bound and the best separating direction a certified lower bound on $\mu_p$;
the ratio $\kappa\rho/\mu_p$ uses the lower bound. The medians of the two bounds
differ by at most $0.002$ on every split, and no query is left undecided at
tolerance $10^{-3}$.

\paragraph{Seeds and decoding.}
\label{sec:app-seeds}
Hook-level and trained-head rows are means over seeds 42, 123 and 456 unless
stated otherwise. Single-seed (seed 42) results: the hook boxes drawn in the
main paper's teaser; the hook rankings that define the candidates of every
in-VLM selection run (Qwen2.5-VL and Qwen3-VL), and the hook retrained on the
earlier feature build (\cref{tab:app-invlm-splits}); the discrete-phrase interface
(\cref{sec:app-interface}); the fine-tuned selector over the detector's top
five, variant (c) of \cref{sec:app-lora-matched}; and the $5{,}000$-expression
point of the native-box fine-tuning curve in \cref{fig:app-dataeff} (the
smaller points use three seeds). POPE rows of the VLM (\cref{tab:app-pope}) are single
deterministic runs. Rows without task-specific training need no seed: the detector is
deterministic, Qwen3-VL is decoded greedily, and Qwen2.5-VL keeps the
checkpoint's generation settings (sampling at temperature $10^{-6}$ with
repetition penalty $1.05$, i.e.\ effectively greedy).

\section{Availability and Reproducibility Notes}
All perception and language components are publicly available
\citep{groundingdino,qwen25vl}. The trainable parts (the hook's
projections and verifier, the other heads, and the adapters of the fine-tuned
selector) are fully specified above. No
human-subjects data is used. The detector checkpoint was trained with
RefCOCO-family data and the VLM's grounding training corpora are undisclosed,
as stated in the main paper.

An anonymized code archive accompanies this supplement. It holds the training,
evaluation, threshold-selection, statistical-test and figure scripts, the
reasoning-stress query list and rejection log, an environment lock for the main
build, and the stored per-sample prediction records and summary results. Datasets,
images, model weights, feature caches and trained checkpoints are not in the
archive, so several analyses cannot be re-run from it alone. Code and checkpoints
will be released publicly with the paper.

{\small\bibliographystyle{ieeenat_fullname}\bibliography{references}}

@article{chatrex,
  title={{ChatRex}: Taming Multimodal {LLM} for Joint Perception and Understanding},
  author={Jiang, Qing and Luo, Gen and Yang, Yuqin and Xiong, Yuda and Chen, Yihao and Zeng, Zhaoyang and Ren, Tianhe and Zhang, Lei},
  journal={arXiv preprint arXiv:2411.18363},
  year={2024}
}

@inproceedings{kosmos2,
  title={Grounding Multimodal Large Language Models to the World},
  author={Peng, Zhiliang and Wang, Wenhui and Dong, Li and Hao, Yaru and Huang, Shaohan and Ma, Shuming and Ye, Qixiang and Wei, Furu},
  booktitle={International Conference on Learning Representations (ICLR)},
  year={2024}
}

@article{shikra,
  title={{Shikra}: Unleashing Multimodal {LLM}'s Referential Dialogue Magic},
  author={Chen, Keqin and Zhang, Zhao and Zeng, Weili and Zhang, Richong and Zhu, Feng and Zhao, Rui},
  journal={arXiv preprint arXiv:2306.15195},
  year={2023}
}

@inproceedings{ferret,
  title={{Ferret}: Refer and Ground Anything Anywhere at Any Granularity},
  author={You, Haoxuan and Zhang, Haotian and Gan, Zhe and Du, Xianzhi and Zhang, Bowen and Wang, Zirui and Cao, Liangliang and Chang, Shih-Fu and Yang, Yinfei},
  booktitle={International Conference on Learning Representations (ICLR)},
  year={2024}
}

@inproceedings{groma,
  title={{Groma}: Localized Visual Tokenization for Grounding Multimodal Large Language Models},
  author={Ma, Chuofan and Jiang, Yi and Wu, Jiannan and Yuan, Zehuan and Qi, Xiaojuan},
  booktitle={European Conference on Computer Vision (ECCV)},
  year={2024}
}

@inproceedings{griffon,
  title={{Griffon}: Spelling out All Object Locations at Any Granularity with Large Language Models},
  author={Zhan, Yufei and Zhu, Yousong and Chen, Zhiyang and Yang, Fan and Tang, Ming and Wang, Jinqiao},
  booktitle={European Conference on Computer Vision (ECCV)},
  year={2024}
}

@inproceedings{mdetr,
  title={{MDETR} -- Modulated Detection for End-to-End Multi-Modal Understanding},
  author={Kamath, Aishwarya and Singh, Mannat and LeCun, Yann and Synnaeve, Gabriel and Misra, Ishan and Carion, Nicolas},
  booktitle={Proceedings of the IEEE/CVF International Conference on Computer Vision (ICCV)},
  pages={1780--1790},
  year={2021}
}

@inproceedings{transvg,
  title={{TransVG}: End-to-End Visual Grounding with Transformers},
  author={Deng, Jiajun and Yang, Zhengyuan and Chen, Tianlang and Zhou, Wengang and Li, Houqiang},
  booktitle={Proceedings of the IEEE/CVF International Conference on Computer Vision (ICCV)},
  pages={1769--1779},
  year={2021}
}

@inproceedings{groundingdino,
  title={{Grounding DINO}: Marrying {DINO} with Grounded Pre-Training for Open-Set Object Detection},
  author={Liu, Shilong and Zeng, Zhaoyang and Ren, Tianhe and Li, Feng and Zhang, Hao and Yang, Jie and Jiang, Qing and Li, Chunyuan and Yang, Jianwei and Su, Hang and Zhu, Jun and Zhang, Lei},
  booktitle={European Conference on Computer Vision (ECCV)},
  year={2024}
}

@inproceedings{glip,
  title={Grounded Language-Image Pre-training},
  author={Li, Liunian Harold and Zhang, Pengchuan and Zhang, Haotian and Yang, Jianwei and Li, Chunyuan and Zhong, Yiwu and Wang, Lijuan and Yuan, Lu and Zhang, Lei and Hwang, Jenq-Neng and Chang, Kai-Wei and Gao, Jianfeng},
  booktitle={Proceedings of the IEEE/CVF Conference on Computer Vision and Pattern Recognition (CVPR)},
  pages={10965--10975},
  year={2022}
}

@inproceedings{yoloworld,
  title={{YOLO-World}: Real-Time Open-Vocabulary Object Detection},
  author={Cheng, Tianheng and Song, Lin and Ge, Yixiao and Liu, Wenyu and Wang, Xinggang and Shan, Ying},
  booktitle={Proceedings of the IEEE/CVF Conference on Computer Vision and Pattern Recognition (CVPR)},
  pages={16901--16911},
  year={2024}
}

@inproceedings{owlvit,
  title={Simple Open-Vocabulary Object Detection with Vision Transformers},
  author={Minderer, Matthias and Gritsenko, Alexey and Stone, Austin and Neumann, Maxim and Weissenborn, Dirk and Dosovitskiy, Alexey and Mahendran, Aravindh and Arnab, Anurag and Dehghani, Mostafa and Shen, Zhuoran and Wang, Xiao and Zhai, Xiaohua and Kipf, Thomas and Houlsby, Neil},
  booktitle={European Conference on Computer Vision (ECCV)},
  year={2022}
}

@inproceedings{clip,
  title={Learning Transferable Visual Models from Natural Language Supervision},
  author={Radford, Alec and Kim, Jong Wook and Hallacy, Chris and Ramesh, Aditya and Goh, Gabriel and Agarwal, Sandhini and Sastry, Girish and Askell, Amanda and Mishkin, Pamela and Clark, Jack and Krueger, Gretchen and Sutskever, Ilya},
  booktitle={International Conference on Machine Learning (ICML)},
  series={Proceedings of Machine Learning Research},
  volume={139},
  pages={8748--8763},
  year={2021}
}

@article{qwenvl,
  title={{Qwen-VL}: A Versatile Vision-Language Model for Understanding, Localization, Text Reading, and Beyond},
  author={Bai, Jinze and Bai, Shuai and Yang, Shusheng and Wang, Shijie and Tan, Sinan and Wang, Peng and Lin, Junyang and Zhou, Chang and Zhou, Jingren},
  journal={arXiv preprint arXiv:2308.12966},
  year={2023}
}

@article{qwen2vl,
  title={{Qwen2-VL}: Enhancing Vision-Language Model's Perception of the World at Any Resolution},
  author={Wang, Peng and Bai, Shuai and Tan, Sinan and Wang, Shijie and Fan, Zhihao and Bai, Jinze and Chen, Keqin and Liu, Xuejing and Wang, Jialin and Ge, Wenbin and Fan, Yang and Dang, Kai and Du, Mengfei and Ren, Xuancheng and Men, Rui and Liu, Dayiheng and Zhou, Chang and Zhou, Jingren and Lin, Junyang},
  journal={arXiv preprint arXiv:2409.12191},
  year={2024}
}

@article{qwen25vl,
  title={{Qwen2.5-VL} Technical Report},
  author={Bai, Shuai and Chen, Keqin and Liu, Xuejing and Wang, Jialin and Ge, Wenbin and Song, Sibo and Dang, Kai and Wang, Peng and Wang, Shijie and Tang, Jun and others},
  journal={arXiv preprint arXiv:2502.13923},
  year={2025}
}

@article{qwen3vl,
  title={{Qwen3-VL} Technical Report},
  author={Bai, Shuai and Cai, Yuxuan and Chen, Ruizhe and Chen, Keqin and Chen, Xionghui and Cheng, Zesen and Deng, Lianghao and Ding, Wei and Gao, Chang and Ge, Chunjiang and others},
  journal={arXiv preprint arXiv:2511.21631},
  year={2025}
}

@article{internvl3,
  title={{InternVL3}: Exploring Advanced Training and Test-Time Recipes for Open-Source Multimodal Models},
  author={Zhu, Jinguo and Wang, Weiyun and Chen, Zhe and Liu, Zhaoyang and Ye, Shenglong and Gu, Lixin and Duan, Yuchen and Tian, Hao and Su, Weijie and Shao, Jie and others},
  journal={arXiv preprint arXiv:2504.10479},
  year={2025}
}

@inproceedings{pope,
  title={Evaluating Object Hallucination in Large Vision-Language Models},
  author={Li, Yifan and Du, Yifan and Zhou, Kun and Wang, Jinpeng and Zhao, Wayne Xin and Wen, Ji-Rong},
  booktitle={Proceedings of the Conference on Empirical Methods in Natural Language Processing (EMNLP)},
  pages={292--305},
  year={2023}
}

@inproceedings{rohrbach2018object,
  title={Object Hallucination in Image Captioning},
  author={Rohrbach, Anna and Hendricks, Lisa Anne and Burns, Kaylee and Darrell, Trevor and Saenko, Kate},
  booktitle={Proceedings of the Conference on Empirical Methods in Natural Language Processing (EMNLP)},
  pages={4035--4045},
  year={2018}
}

@inproceedings{objgndhalluc,
  title={Does Object Grounding Really Reduce Hallucination of Large Vision-Language Models?},
  author={Geigle, Gregor and Timofte, Radu and Glava{\v{s}}, Goran},
  booktitle={Proceedings of the Conference on Empirical Methods in Natural Language Processing (EMNLP)},
  pages={2728--2742},
  year={2024}
}

@inproceedings{refcoco,
  title={Modeling Context in Referring Expressions},
  author={Yu, Licheng and Poirson, Patrick and Yang, Shan and Berg, Alexander C. and Berg, Tamara L.},
  booktitle={European Conference on Computer Vision (ECCV)},
  pages={69--85},
  year={2016}
}

@inproceedings{coco,
  title={Microsoft {COCO}: Common Objects in Context},
  author={Lin, Tsung-Yi and Maire, Michael and Belongie, Serge and Hays, James and Perona, Pietro and Ramanan, Deva and Doll{\'a}r, Piotr and Zitnick, C. Lawrence},
  booktitle={European Conference on Computer Vision (ECCV)},
  pages={740--755},
  year={2014}
}

@inproceedings{flmm,
  title={{F-LMM}: Grounding Frozen Large Multimodal Models},
  author={Wu, Size and Jin, Sheng and Zhang, Wenwei and Xu, Lumin and Liu, Wentao and Li, Wei and Loy, Chen Change},
  booktitle={Proceedings of the IEEE/CVF Conference on Computer Vision and Pattern Recognition (CVPR)},
  pages={24710--24721},
  year={2025}
}

@inproceedings{rexomni,
  title={Detect Anything via Next Point Prediction},
  author={Jiang, Qing and Huo, Junan and Chen, Xingyu and Xiong, Yuda and Zeng, Zhaoyang and Chen, Yihao and Ren, Tianhe and Yu, Junzhi and Zhang, Lei},
  booktitle={Proceedings of the IEEE/CVF Conference on Computer Vision and Pattern Recognition (CVPR)},
  pages={25472--25483},
  year={2026}
}

@inproceedings{rexseek,
  title={Referring to Any Person},
  author={Jiang, Qing and Wu, Lin and Zeng, Zhaoyang and Ren, Tianhe and Xiong, Yuda and Chen, Yihao and Qin, Liu and Zhang, Lei},
  booktitle={Proceedings of the IEEE/CVF International Conference on Computer Vision (ICCV)},
  pages={21667--21678},
  year={2025}
}

@article{griffong,
  title={{Griffon-G}: Bridging Vision-Language and Vision-Centric Tasks via Large Multimodal Models},
  author={Zhan, Yufei and Zhao, Hongyin and Zhu, Yousong and Yang, Fan and Tang, Ming and Wang, Jinqiao},
  journal={arXiv preprint arXiv:2410.16163},
  year={2024}
}

@inproceedings{refl4,
  title={Revisiting Referring Expression Comprehension Evaluation in the Era of Large Multimodal Models},
  author={Chen, Jierun and Wei, Fangyun and Zhao, Jinjing and Song, Sizhe and Wu, Bohuai and Peng, Zhuoxuan and Chan, S.-H. Gary and Zhang, Hongyang},
  booktitle={Proceedings of the IEEE/CVF Conference on Computer Vision and Pattern Recognition (CVPR) Workshops},
  pages={513--524},
  year={2025}
}

@inproceedings{grefcoco,
  title={{GRES}: Generalized Referring Expression Segmentation},
  author={Liu, Chang and Ding, Henghui and Jiang, Xudong},
  booktitle={Proceedings of the IEEE/CVF Conference on Computer Vision and Pattern Recognition (CVPR)},
  pages={23592--23601},
  year={2023}
}

@inproceedings{refformer,
  title={Referencing Where to Focus: Improving Visual Grounding with Referential Query},
  author={Wang, Yabing and Tian, Zhuotao and Guo, Qingpei and Qin, Zheng and Zhou, Sanping and Yang, Ming and Wang, Le},
  booktitle={Advances in Neural Information Processing Systems (NeurIPS)},
  volume={37},
  pages={47378--47399},
  year={2024}
}

@inproceedings{counterfactualrec,
  title={Revisiting Counterfactual Problems in Referring Expression Comprehension},
  author={Yu, Zhihan and Li, Ruifan},
  booktitle={Proceedings of the IEEE/CVF Conference on Computer Vision and Pattern Recognition (CVPR)},
  pages={13438--13448},
  year={2024}
}

@inproceedings{propvg,
  title={{PropVG}: End-to-End Proposal-Driven Visual Grounding with Multi-Granularity Discrimination},
  author={Dai, Ming and Cheng, Wenxuan and Zhuang, Jiedong and Liu, Jiang-jiang and Zhao, Hongshen and Feng, Zhenhua and Yang, Wankou},
  booktitle={Proceedings of the IEEE/CVF International Conference on Computer Vision (ICCV)},
  pages={7058--7068},
  year={2025}
}

@inproceedings{vcd,
  title={Mitigating Object Hallucinations in Large Vision-Language Models through Visual Contrastive Decoding},
  author={Leng, Sicong and Zhang, Hang and Chen, Guanzheng and Li, Xin and Lu, Shijian and Miao, Chunyan and Bing, Lidong},
  booktitle={Proceedings of the IEEE/CVF Conference on Computer Vision and Pattern Recognition (CVPR)},
  pages={13872--13882},
  year={2024}
}

@inproceedings{geifman2017selective,
  title={Selective Classification for Deep Neural Networks},
  author={Geifman, Yonatan and El-Yaniv, Ran},
  booktitle={Advances in Neural Information Processing Systems (NeurIPS)},
  volume={30},
  year={2017}
}

@inproceedings{hu2022lora,
  title={{LoRA}: Low-Rank Adaptation of Large Language Models},
  author={Hu, Edward J. and Shen, Yelong and Wallis, Phillip and Allen-Zhu, Zeyuan and Li, Yuanzhi and Wang, Shean and Wang, Lu and Chen, Weizhu},
  booktitle={International Conference on Learning Representations (ICLR)},
  year={2022}
}

@inproceedings{groundhog,
  title={{GROUNDHOG}: Grounding Large Language Models to Holistic Segmentation},
  author={Zhang, Yichi and Ma, Ziqiao and Gao, Xiaofeng and Shakiah, Suhaila and Gao, Qiaozi and Chai, Joyce},
  booktitle={Proceedings of the IEEE/CVF Conference on Computer Vision and Pattern Recognition (CVPR)},
  pages={14227--14238},
  year={2024}
}

@inproceedings{vgent,
  title={{VGent}: Visual Grounding via Modular Design for Disentangling Reasoning and Prediction},
  author={Kang, Weitai and Kuen, Jason and Ren, Mengwei and Wei, Zijun and Yan, Yan and Liu, Kangning},
  booktitle={Proceedings of the IEEE/CVF Conference on Computer Vision and Pattern Recognition (CVPR)},
  pages={41160--41170},
  year={2026}
}

@inproceedings{rexthinker,
  title={{Rex-Thinker}: Grounded Object Referring via Chain-of-Thought Reasoning},
  author={Jiang, Qing and Chen, Xingyu and Zeng, Zhaoyang and Yu, Junzhi and Zhang, Lei},
  booktitle={International Conference on Learning Representations (ICLR)},
  year={2026}
}

@article{tfverify,
  title={Zero-Shot Referring Expression Comprehension via Vision-Language True/False Verification},
  author={Liu, Jeffrey and Hu, Rongbin},
  journal={arXiv preprint arXiv:2509.09958},
  year={2025}
}

@article{setofmark,
  title={{Set-of-Mark} Prompting Unleashes Extraordinary Visual Grounding in {GPT-4V}},
  author={Yang, Jianwei and Zhang, Hao and Li, Feng and Zou, Xueyan and Li, Chunyuan and Gao, Jianfeng},
  journal={arXiv preprint arXiv:2310.11441},
  year={2023}
}

@article{grec,
  title={{GREC}: Generalized Referring Expression Comprehension},
  author={He, Shuting and Ding, Henghui and Liu, Chang and Jiang, Xudong},
  journal={arXiv preprint arXiv:2308.16182},
  year={2023}
}

@article{grex,
  title={{GREx}: Generalized Referring Expression Segmentation, Comprehension, and Generation},
  author={Ding, Henghui and Liu, Chang and He, Shuting and Jiang, Xudong and Jiang, Yu-Gang},
  journal={International Journal of Computer Vision},
  volume={134},
  number={2},
  pages={79},
  year={2026},
  doi={10.1007/s11263-026-02729-y}
}

@inproceedings{groundingme,
  title={{GroundingME}: Exposing the Visual Grounding Gap in {MLLMs} through Multi-Dimensional Evaluation},
  author={Li, Rang and Li, Lei and Ren, Shuhuai and Tian, Hao and Gu, Shuhao and Li, Shicheng and Yue, Zihao and Wang, Yudong and Ma, Wenhan and Yang, Zhe and Ma, Jingyuan and Sui, Zhifang and Luo, Fuli},
  booktitle={Proceedings of the IEEE/CVF Conference on Computer Vision and Pattern Recognition (CVPR)},
  pages={2412--2422},
  year={2026}
}

@inproceedings{viro,
  title={{VIRO}: Robust and Efficient Neuro-Symbolic Reasoning with Verification for Referring Expression Comprehension},
  author={Park, Hyejin and Kwon, Junhyuk and Kwak, Suha and Ok, Jungseul},
  booktitle={Proceedings of the IEEE/CVF Conference on Computer Vision and Pattern Recognition (CVPR)},
  pages={33426--33435},
  year={2026}
}

@inproceedings{d3,
  title={Described Object Detection: Liberating Object Detection with Flexible Expressions},
  author={Xie, Chi and Zhang, Zhao and Wu, Yixuan and Zhu, Feng and Zhao, Rui and Liang, Shuang},
  booktitle={Advances in Neural Information Processing Systems (NeurIPS)},
  volume={36},
  pages={79095--79107},
  year={2023}
}

@article{rcgrpo,
  title={Teaching {MLLMs} to Say No: Generalized Referring Expression Comprehension via Refusal Calibrated {GRPO}},
  author={Yang, Xuzheng and Ling, Jun and Huang, Tao and Qin, Caiyan and Wang, Peng},
  journal={arXiv preprint arXiv:2608.04698},
  year={2026}
}

@inproceedings{mattnet,
  title={{MAttNet}: Modular Attention Network for Referring Expression Comprehension},
  author={Yu, Licheng and Lin, Zhe and Shen, Xiaohui and Yang, Jimei and Lu, Xin and Bansal, Mohit and Berg, Tamara L.},
  booktitle={Proceedings of the IEEE Conference on Computer Vision and Pattern Recognition (CVPR)},
  pages={1307--1315},
  year={2018}
}

@inproceedings{refnms,
  title={{Ref-NMS}: Breaking Proposal Bottlenecks in Two-Stage Referring Expression Grounding},
  author={Chen, Long and Ma, Wenbo and Xiao, Jun and Zhang, Hanwang and Chang, Shih-Fu},
  booktitle={Proceedings of the AAAI Conference on Artificial Intelligence},
  volume={35},
  number={2},
  pages={1036--1044},
  year={2021},
  doi={10.1609/aaai.v35i2.16188}
}

@inproceedings{reclip,
  title={{ReCLIP}: A Strong Zero-Shot Baseline for Referring Expression Comprehension},
  author={Subramanian, Sanjay and Merrill, William and Darrell, Trevor and Gardner, Matt and Singh, Sameer and Rohrbach, Anna},
  booktitle={Proceedings of the Annual Meeting of the Association for Computational Linguistics (ACL)},
  pages={5198--5215},
  year={2022},
  doi={10.18653/v1/2022.acl-long.357}
}

@inproceedings{onestagevg,
  title={A Fast and Accurate One-Stage Approach to Visual Grounding},
  author={Yang, Zhengyuan and Gong, Boqing and Wang, Liwei and Huang, Wenbing and Yu, Dong and Luo, Jiebo},
  booktitle={Proceedings of the IEEE/CVF International Conference on Computer Vision (ICCV)},
  pages={4683--4693},
  year={2019}
}

@inproceedings{refcocog,
  title={Generation and Comprehension of Unambiguous Object Descriptions},
  author={Mao, Junhua and Huang, Jonathan and Toshev, Alexander and Camburu, Oana and Yuille, Alan L. and Murphy, Kevin},
  booktitle={Proceedings of the IEEE Conference on Computer Vision and Pattern Recognition (CVPR)},
  pages={11--20},
  year={2016}
}

@inproceedings{refcocogumd,
  title={Modeling Context Between Objects for Referring Expression Understanding},
  author={Nagaraja, Varun K. and Morariu, Vlad I. and Davis, Larry S.},
  booktitle={European Conference on Computer Vision (ECCV)},
  pages={792--807},
  year={2016}
}

@article{vlmfo1,
  title={{VLM-FO1}: Bridging the Gap Between High-Level Reasoning and Fine-Grained Perception in {VLMs}},
  author={Liu, Peng and Shen, Haozhan and Fang, Chunxin and Sun, Zhicheng and Liao, Jiajia and Zhao, Tiancheng},
  journal={arXiv preprint arXiv:2509.25916},
  year={2025}
}

@inproceedings{llmwrapper,
  title={{LLM-wrapper}: Black-Box Semantic-Aware Adaptation of Vision-Language Models for Referring Expression Comprehension},
  author={Cardiel, Amaia and Zablocki, Eloi and Ramzi, Elias and Sim{\'e}oni, Oriane and Cord, Matthieu},
  booktitle={International Conference on Learning Representations (ICLR)},
  year={2025}
}

@inproceedings{rodmllm,
  title={{ROD-MLLM}: Towards More Reliable Object Detection in Multimodal Large Language Models},
  author={Yin, Heng and Ren, Yuqiang and Yan, Ke and Ding, Shouhong and Hao, Yongtao},
  booktitle={Proceedings of the IEEE/CVF Conference on Computer Vision and Pattern Recognition (CVPR)},
  pages={14358--14368},
  year={2025}
}

@inproceedings{groundingagent,
  title={Connecting the Dots: Training-Free Visual Grounding via Agentic Reasoning},
  author={Luo, Liqin and Chen, Guangyao and Zheng, Xiawu and Dai, Yongxing and Zou, Yixiong and Tian, Yonghong},
  booktitle={Proceedings of the AAAI Conference on Artificial Intelligence},
  volume={40},
  number={9},
  pages={7671--7679},
  year={2026},
  doi={10.1609/aaai.v40i9.37709}
}

@article{rrefcoco,
  title={Toward Robust Referring Image Segmentation},
  author={Wu, Jianzong and Li, Xiangtai and Li, Xia and Ding, Henghui and Tong, Yunhai and Tao, Dacheng},
  journal={IEEE Transactions on Image Processing},
  volume={33},
  pages={1782--1794},
  year={2024},
  doi={10.1109/TIP.2024.3371348}
}

@inproceedings{finecopsref,
  title={{FineCops-Ref}: A New Dataset and Task for Fine-Grained Compositional Referring Expression Comprehension},
  author={Liu, Junzhuo and Yang, Xuzheng and Li, Weiwei and Wang, Peng},
  booktitle={Proceedings of the Conference on Empirical Methods in Natural Language Processing (EMNLP)},
  pages={15440--15457},
  year={2024},
  doi={10.18653/v1/2024.emnlp-main.864}
}

@inproceedings{refzom,
  title={Beyond One-to-One: Rethinking the Referring Image Segmentation},
  author={Hu, Yutao and Wang, Qixiong and Shao, Wenqi and Xie, Enze and Li, Zhenguo and Han, Jungong and Luo, Ping},
  booktitle={Proceedings of the IEEE/CVF International Conference on Computer Vision (ICCV)},
  pages={4067--4077},
  year={2023}
}

@inproceedings{omnilabel,
  title={{OmniLabel}: A Challenging Benchmark for Language-Based Object Detection},
  author={Schulter, Samuel and G, Vijay Kumar B. and Suh, Yumin and Dafnis, Konstantinos M. and Zhang, Zhixing and Zhao, Shiyu and Metaxas, Dimitris},
  booktitle={Proceedings of the IEEE/CVF International Conference on Computer Vision (ICCV)},
  pages={11953--11962},
  year={2023}
}

@inproceedings{halp,
  title={{HALP}: Detecting Hallucinations in Vision-Language Models without Generating a Single Token},
  author={Kogilathota, Sai Akhil and G, Sripadha Vallabha E. and Sun, Luzhe and Zhou, Jiawei},
  booktitle={Proceedings of the Conference of the European Chapter of the Association for Computational Linguistics (EACL)},
  pages={6067--6085},
  year={2026},
  doi={10.18653/v1/2026.eacl-long.287}
}

@inproceedings{crg,
  title={Contrastive Region Guidance: Improving Grounding in Vision-Language Models Without Training},
  author={Wan, David and Cho, Jaemin and Stengel-Eskin, Elias and Bansal, Mohit},
  booktitle={European Conference on Computer Vision (ECCV)},
  pages={198--215},
  year={2024},
  doi={10.1007/978-3-031-72986-7_12}
}

@inproceedings{mllmsknow,
  title={{MLLMs} Know Where to Look: Training-free Perception of Small Visual Details with Multimodal {LLMs}},
  author={Zhang, Jiarui and Khayatkhoei, Mahyar and Chhikara, Prateek and Ilievski, Filip},
  booktitle={International Conference on Learning Representations (ICLR)},
  year={2025}
}

@inproceedings{lochead,
  title={Your Large Vision-Language Model Only Needs A Few Attention Heads For Visual Grounding},
  author={Kang, Seil and Kim, Jinyeong and Kim, Junhyeok and Hwang, Seong Jae},
  booktitle={Proceedings of the IEEE/CVF Conference on Computer Vision and Pattern Recognition (CVPR)},
  pages={9339--9350},
  year={2025}
}

@article{ovvg,
  title={{OV-VG}: A Benchmark for Open-Vocabulary Visual Grounding},
  author={Wang, Chunlei and Feng, Wenquan and Li, Xiangtai and Cheng, Guangliang and Lyu, Shuchang and Liu, Binghao and Chen, Lijiang and Zhao, Qi},
  journal={Neurocomputing},
  volume={591},
  pages={127738},
  year={2024},
  doi={10.1016/j.neucom.2024.127738}
}

@inproceedings{reliablevqa,
  title={Reliable Visual Question Answering: Abstain Rather Than Answer Incorrectly},
  author={Whitehead, Spencer and Petryk, Suzanne and Shakib, Vedaad and Gonzalez, Joseph and Darrell, Trevor and Rohrbach, Anna and Rohrbach, Marcus},
  booktitle={European Conference on Computer Vision (ECCV)},
  pages={148--166},
  year={2022},
  doi={10.1007/978-3-031-20059-5_9}
}

@article{crs2025,
  title={New Dataset and Methods for Fine-Grained Compositional Referring Expression Comprehension via Specialist-{MLLM} Collaboration},
  author={Yang, Xuzheng and Liu, Junzhuo and Wang, Peng and Wang, Guoqing and Yang, Yang and Shen, Heng Tao},
  journal={IEEE Transactions on Pattern Analysis and Machine Intelligence},
  volume={47},
  number={10},
  pages={8598--8612},
  year={2025},
  doi={10.1109/TPAMI.2025.3580090}
}

@inproceedings{deris2025,
  title={{DeRIS}: Decoupling Perception and Cognition for Enhanced Referring Image Segmentation through Loopback Synergy},
  author={Dai, Ming and Cheng, Wenxuan and Liu, Jiang-jiang and Yang, Sen and Cai, Wenxiao and Sun, Yanpeng and Yang, Wankou},
  booktitle={Proceedings of the IEEE/CVF International Conference on Computer Vision (ICCV)},
  pages={19936--19946},
  year={2025}
}

@inproceedings{somllava,
  title={List Items One by One: A New Data Source and Learning Paradigm for Multimodal {LLMs}},
  author={Yan, An and Yang, Zhengyuan and Wu, Junda and Zhu, Wanrong and Yang, Jianwei and Li, Linjie and Lin, Kevin and Wang, Jianfeng and McAuley, Julian and Gao, Jianfeng and Wang, Lijuan},
  booktitle={Conference on Language Modeling (COLM)},
  year={2024}
}

@article{llmoptic,
  title={{LLM-Optic}: Unveiling the Capabilities of Large Language Models for Universal Visual Grounding},
  author={Zhao, Haoyu and Ge, Wenhang and Chen, Ying-cong},
  journal={arXiv preprint arXiv:2405.17104},
  year={2024}
}

@inproceedings{refdetector,
  title={{RefDetector}: A Simple Yet Effective Matching-based Method for Referring Expression Comprehension},
  author={Wang, Yabing and Tian, Zhuotao and Qin, Zheng and Zhou, Sanping and Wang, Le},
  booktitle={Proceedings of the AAAI Conference on Artificial Intelligence},
  volume={39},
  number={8},
  pages={8033--8041},
  year={2025},
  doi={10.1609/aaai.v39i8.32866}
}

@inproceedings{hiea2g,
  title={Hierarchical Alignment-enhanced Adaptive Grounding Network for Generalized Referring Expression Comprehension},
  author={Wang, Yaxian and Ding, Henghui and He, Shuting and Jiang, Xudong and Wei, Bifan and Liu, Jun},
  booktitle={Proceedings of the AAAI Conference on Artificial Intelligence},
  volume={39},
  number={8},
  pages={8042--8050},
  year={2025},
  doi={10.1609/aaai.v39i8.32867}
}

@inproceedings{pin2024,
  title={{PIN}: Positional Insert Unlocks Object Localisation Abilities in {VLMs}},
  author={Dorkenwald, Michael and Barazani, Nimrod and Snoek, Cees G. M. and Asano, Yuki M.},
  booktitle={Proceedings of the IEEE/CVF Conference on Computer Vision and Pattern Recognition (CVPR)},
  pages={13548--13558},
  year={2024}
}

@inproceedings{vlmlocmech,
  title={Mechanisms of Object Localization in Vision-Language Models},
  author={Schauml{\"o}ffel, Timothy and Vilas, Martina G. and Roig, Gemma},
  booktitle={Proceedings of the IEEE/CVF Conference on Computer Vision and Pattern Recognition (CVPR)},
  pages={31356--31365},
  year={2026}
}

@inproceedings{interlayer2024,
  title={Unleashing Region Understanding in Intermediate Layers for {MLLM}-based Referring Expression Generation},
  author={Liang, Yaoyuan and Cai, Zhuojun and Xu, Jian and Huang, Guanbo and Wang, Yiran and Liang, Xiao and Liu, Jiahao and Li, Ziran and Wang, Jingang and Huang, Shao-Lun},
  booktitle={Advances in Neural Information Processing Systems (NeurIPS)},
  volume={37},
  pages={120578--120601},
  year={2024},
  doi={10.52202/079017-3833}
}

@inproceedings{realign2025,
  title={Generative Giants, Retrieval Weaklings: Why do Multimodal Large Language Models Fail at Multimodal Retrieval?},
  author={Feng, Hengyi and Sheng, Zeang and Qiang, Meiyi and Li, Yang and Zhang, Wentao},
  booktitle={Findings of the Association for Computational Linguistics: ACL 2026},
  pages={15917--15933},
  year={2026},
  doi={10.18653/v1/2026.findings-acl.781}
}

@inproceedings{refadv,
  title={{Ref-Adv}: Exploring {MLLM} Visual Reasoning in Referring Expression Tasks},
  author={Dong, Qihua and Yang, Kuo and Ju, Lin and Zhao, Handong and Zhang, Yitian and Wang, Yizhou and Zeng, Huimin and Lu, Jianglin and Fu, Yun},
  booktitle={International Conference on Learning Representations (ICLR)},
  year={2026}
}

@article{refbenchpro,
  title={{RefBench-PRO}: Perceptual and Reasoning Oriented Benchmark for Referring Expression Comprehension},
  author={Gao, Tianyi and Li, Hao and Fang, Han and Wei, Xin and Dong, Xiaodong and Sun, Hongbo and Yuan, Ye and He, Zhongjiang and Xu, Jinglin and Xin, Jingmin and Sun, Hao},
  journal={arXiv preprint arXiv:2512.06276},
  year={2025}
}

@inproceedings{knowdrrec,
  title={{KnowDR-REC}: Auditing Knowledge-Conditioned Visual Grounding in Referring Expression Comprehension},
  author={Jin, Guanghao and Wu, Jingpei and Guo, Tianpei and Niu, Yiyi and Zhou, Weidong and Yang, Linyi and Liu, Guoyang},
  booktitle={Findings of the Association for Computational Linguistics: ACL 2026},
  pages={38607--38629},
  year={2026},
  doi={10.18653/v1/2026.findings-acl.1923}
}

@article{openref,
  title={Towards Open-World Referring Expression Comprehension: A Benchmark with Training-free Multi-task Consistency Checker},
  author={Wu, Zongjian and Zhang, Lei},
  journal={arXiv preprint arXiv:2605.25706},
  year={2026}
}

@article{jag,
  title={Jag: Direct Box Prediction for Efficient Visual Grounding},
  author={Zhu, Xiuyuan and Lu, Ke and Wu, Hao and Du, Zijin and Zhang, Dongming and Xue, Jian},
  journal={arXiv preprint arXiv:2610.13239},
  year={2026}
}

@article{locus,
  title={{LOCUS}: Locate Once, Check Until Settled. Draft-and-Verify Visual Grounding with a Box Head and Coordinate-Token Verification},
  author={Talha, Muhammad},
  journal={arXiv preprint arXiv:2610.35053},
  year={2026}
}

@article{forum,
  title={{FORUM}: Frozen Outputs Reconciled Using Model Agreement for Visual Grounding},
  author={Sato, Taiyo and Sanda, Takamasa and Maeda, Keisuke and Ogawa, Takahiro and Haseyama, Miki and Nagashima, Shunya},
  journal={arXiv preprint arXiv:2609.37488},
  year={2026}
}

@article{prove,
  title={{PROVE}: Probabilistic Visual Grounding via Patch-Level Evidence for Vision-Language Models},
  author={Nautiyal, Mayank and Zhang, Tianru and Ju, Li and Vats, Ekta and Singh, Prashant},
  journal={arXiv preprint arXiv:2610.32440},
  year={2026}
}

@article{surrogatevg,
  title={Surrogate Uncertainty for Black-box Visual Grounding},
  author={Zhang, Hanbo and Hu, Tianrun and Bai, Cedar Site and Xiao, Anxing and Hsu, David},
  journal={arXiv preprint arXiv:2610.39798},
  year={2026}
}

@inproceedings{florence2,
  title={{Florence-2}: Advancing a Unified Representation for a Variety of Vision Tasks},
  author={Xiao, Bin and Wu, Haiping and Xu, Weijian and Dai, Xiyang and Hu, Houdong and Lu, Yumao and Zeng, Michael and Liu, Ce and Yuan, Lu},
  booktitle={Proceedings of the IEEE/CVF Conference on Computer Vision and Pattern Recognition (CVPR)},
  pages={4818--4829},
  year={2024}
}

@article{dinox,
  title={{DINO-X}: A Unified Vision Model for Open-World Object Detection and Understanding},
  author={Ren, Tianhe and Chen, Yihao and Jiang, Qing and Zeng, Zhaoyang and Xiong, Yuda and Liu, Wenlong and Ma, Zhengyu and Shen, Junyi and Gao, Yuan and Jiang, Xiaoke and Chen, Xingyu and Song, Zhuheng and Zhang, Yuhong and Huang, Hongjie and Gao, Han and Liu, Shilong and Zhang, Hao and Li, Feng and Yu, Kent and Zhang, Lei},
  journal={arXiv preprint arXiv:2411.14347},
  year={2024}
}

@article{gdino15,
  title={{Grounding DINO 1.5}: Advance the ``{Edge}'' of Open-Set Object Detection},
  author={Ren, Tianhe and Jiang, Qing and Liu, Shilong and Zeng, Zhaoyang and Liu, Wenlong and Gao, Han and Huang, Hongjie and Ma, Zhengyu and Jiang, Xiaoke and Chen, Yihao and Xiong, Yuda and Zhang, Hao and Li, Feng and Tang, Peijun and Yu, Kent and Zhang, Lei},
  journal={arXiv preprint arXiv:2405.10300},
  year={2024}
}

@article{hkvlm2026,
  title={{HKVLM}: Faithful Query--Region Binding for Frozen-Detector Visual Grounding},
  author={Ma, Bo},
  journal={arXiv preprint arXiv:2606.28862},
  year={2026}
}

@article{dinov2,
  title={{DINOv2}: Learning Robust Visual Features without Supervision},
  author={Oquab, Maxime and Darcet, Timoth{\'e}e and Moutakanni, Th{\'e}o and Vo, Huy and Szafraniec, Marc and Khalidov, Vasil and Fernandez, Pierre and Haziza, Daniel and Massa, Francisco and El-Nouby, Alaaeldin and others},
  journal={Transactions on Machine Learning Research},
  year={2024}
}

@inproceedings{flickr30kentities,
  title={{Flickr30k Entities}: Collecting Region-to-Phrase Correspondences for Richer Image-to-Sentence Models},
  author={Plummer, Bryan A. and Wang, Liwei and Cervantes, Chris M. and Caicedo, Juan C. and Hockenmaier, Julia and Lazebnik, Svetlana},
  booktitle={Proceedings of the IEEE International Conference on Computer Vision (ICCV)},
  pages={2641--2649},
  year={2015}
}
\end{document}